\documentclass[a4paper,9pt]{extarticle}
\usepackage[utf8]{inputenc}
\usepackage{float}
\usepackage{caption}
\usepackage{subcaption}
\usepackage[margin=0.6in]{geometry}
\usepackage{amsmath,amssymb,amsfonts}
\usepackage{amsthm}
\usepackage{graphicx}
\usepackage{textcomp}
\usepackage{tcolorbox}
\usepackage{booktabs}
\usepackage{multirow}
\usepackage{xspace}
\usepackage{enumitem}
\usepackage{navigator}
\usepackage[toc,page]{appendix}

\usepackage{tikz}
\usetikzlibrary{spy}
\usepackage{xcolor}
\definecolor{col1}{rgb}{0, 0.447, 0.7410}
\definecolor{col2}{rgb}{0.8500, 0.3250, 0.0980}
\definecolor{col3}{rgb}{0.9290, 0.6940, 0.1250}
\definecolor{col4}{rgb}{0.4940, 0.1840, 0.5560}
\definecolor{col5}{rgb}{0.4660, 0.6740, 0.1880}
\definecolor{col6}{rgb}{0.3020, 0.7450, 0.9330}
\definecolor{col7}{rgb}{0.6350, 0.0780, 0.1840}
\definecolor{col8}{rgb}{0.0000, 0.4470, 0.7410}
\usepackage[export]{adjustbox}

\usepackage{array}
\newcolumntype{P}[1]{>{\centering\arraybackslash}p{#1}}
\newcolumntype{M}[1]{>{\centering\arraybackslash}m{#1}}
\emergencystretch=\maxdimen
\newcommand{\eg}{{\it e.g.}}
\newcommand{\ie}{{\it i.e.}}

\newcommand{\Tr}{\mathop{\bf Tr}}
\newcommand{\prox}{\mathbf{prox}}
\newcommand{\proj}{\mathbf{proj}}
\newcommand{\diag}{\mathop{\bf diag}}
\newcommand{\argmin}{\mathop{\rm argmin}}
\newcommand{\argmax}{\mathop{\rm argmax}}
\newcommand{\norm}[1]{\left\lVert#1\right\rVert}
\newcommand{\R}{\mathbb{R}}

\newcommand{\vct}[1]{\boldsymbol{#1}}
\newcommand{\mtx}[1]{\boldsymbol{#1}}

\newcommand{\vf}{\vct{f}}

\newcommand{\vr}{\vct{r}}

\newcommand{\vt}{\vct{t}}

\newcommand{\vx}{\vct{x}}

\newcommand{\vz}{\vct{z}}
\newcommand{\vzero}{\vct{0}}
\newcommand{\vone}{\vct{1}}

\newcommand{\mA}{\mtx{A}}
\newcommand{\mB}{\mtx{B}}

\newcommand{\mF}{\mtx{F}}

\newcommand{\mI}{\mtx{I}}

\newcommand{\mP}{\mtx{P}}
\newcommand{\mQ}{\mtx{Q}}
\newcommand{\mR}{\mtx{R}}

\newcommand{\mT}{\mtx{T}}
\newcommand{\mU}{\mtx{U}}
\newcommand{\mV}{\mtx{V}}
\newcommand{\mW}{\mtx{W}}
\newcommand{\mX}{\mtx{X}}
\newcommand{\mY}{\mtx{Y}}
\newcommand{\mZ}{\mtx{Z}}

\newcommand{\mLambda}{\mtx{\Lambda}}

\newcommand{\mOmega}{\mtx{\Omega}}

\newcommand{\mzero}{{\bf 0}}
\newcommand{\mone}{{\bf 1}}

\newcommand{\set}[1]{\mathcal{#1}}

\newcommand{\setB}{\set{B}}

\newcommand{\setE}{\set{E}}

\newcommand{\setU}{\set{U}}
\newcommand{\setV}{\set{V}}

\newcommand{\setX}{\set{X}}
\newcommand{\setY}{\set{Y}}

\newtheorem{defn}{Definition}
\newtheorem{thm}{Theorem}
\newtheorem{lem}{Lemma}
\newtheorem{cor}{Corollary}

\title{\bf ADJUST: A Dictionary-Based Joint Reconstruction and Unmixing Method for Spectral Tomography}

\author{Math\'e T. Zeegers$^{1}$, Ajinkya Kadu$^{1,2}$, Tristan van Leeuwen$^{1,3}$, Kees Joost Batenburg$^{1,4}$ \\[2ex]
$^{1}$ Centrum Wiskunde \& Informatica,  Science Park 123, 1098 XG Amsterdam, The Netherlands \\
$^{2}$ University of Antwerp, Groenenborgerlaan 171, 2020 Antwerp, Belgium \\
$^{3}$ Mathematical Institute, Utrecht University, Budapestlaan 6, 3584 CD Utrecht, The Netherlands\\
$^{4}$ Leiden Institute of Advanced Computer Science, Niels Bohrweg 1, 2333 CA Leiden, The Netherlands\\
}
\date{}

\begin{document}
\maketitle

\begin{abstract}

Advances in multi-spectral detectors are causing a paradigm shift in X-ray Computed Tomography (CT). Spectral information acquired from these detectors can be used to extract volumetric material composition maps of the object of interest. If the materials and their spectral responses are known a priori, the image reconstruction step is rather straightforward. If they are not known, however, the maps as well as the responses need to be estimated jointly. A conventional workflow in spectral CT involves performing volume reconstruction followed by material decomposition, or vice versa. However, these methods inherently suffer from the ill-posedness of the joint reconstruction problem. To resolve this issue, we propose `A Dictionary-based Joint reconstruction and Unmixing method for Spectral Tomography' (ADJUST). Our formulation relies on forming a dictionary of spectral signatures of materials common in CT and prior knowledge of the number of materials present in an object. In particular, we decompose the spectral volume linearly in terms of spatial material maps, a spectral dictionary, and the indicator of materials for the dictionary elements. We propose a memory-efficient accelerated alternating proximal gradient method to find an approximate solution to the resulting bi-convex problem. From numerical demonstrations on several synthetic phantoms, we observe that ADJUST performs exceedingly well compared to other state-of-the-art methods. Additionally, we address the robustness of ADJUST against limited and noisy measurement patterns. The demonstration of the proposed approach on a spectral micro-CT dataset shows its potential for real-world applications. Code is available at \jumplink{https://github.com/mzeegers/ADJUST}{https://github.com/mzeegers/ADJUST}. 
\end{abstract}


\section{Introduction}
\label{sec:Intro}
X-ray Computed Tomography (CT) estimates the spatial attenuation map of the object of interest from its measured X-ray projections obtained from different angles. The conventional tomography acquisition setup consists of a polychromatic X-ray source and an X-ray detector that collects the transmitted X-rays. However, these conventional detectors do not discriminate between different incident photon energies and collect the attenuated X-rays solely in one energy bin (Figure~\ref{fig:Motivation:ConventionalVSSpectral}). Images of conventional detectors are reconstructed as grey scale volumes, representing the aggregate attenuation coefficients of the materials. Several materials may correspond to the same grey level, which makes it difficult to determine the material composition of an object. Nonetheless, the attenuations of materials are energy-dependent and their spectral attenuation curves are mutually different. Therefore, by probing multiple energy levels, additional information is obtained for discriminating the materials. A typical approach is dual-energy CT, which allows for more accurate \emph{material decomposition}~\cite{BhayanaParakh, TaguchiBlevis}, also known as material unmixing. It uses two polychromatic sources with different peak voltages, and correspondingly two sets of conventional detector panels to measure attenuated X-rays, each from one source. Dual-energy CT is commonly used in clinical settings for separating high-attenuating materials from low-attenuating materials, for example, determining the location of contrast agents~\cite{SiBar} such as iodine in the body~\cite{KarcaaltincabaAktas}. However, more accurate material decomposition and concentration determination requires multi-energy CT, enabled by multi-energy X-ray (photon counting) detectors~\cite{BhayanaParakh, TaguchiBlevis}.

\newcommand{\rulesep}{\unskip\ \vrule\ }
\begin{figure}[!htb]
    \centering
    \subcaptionbox{Data acquisition in conventional X-ray CT}[0.47\textwidth]{
		\includegraphics[width=0.47\textwidth]{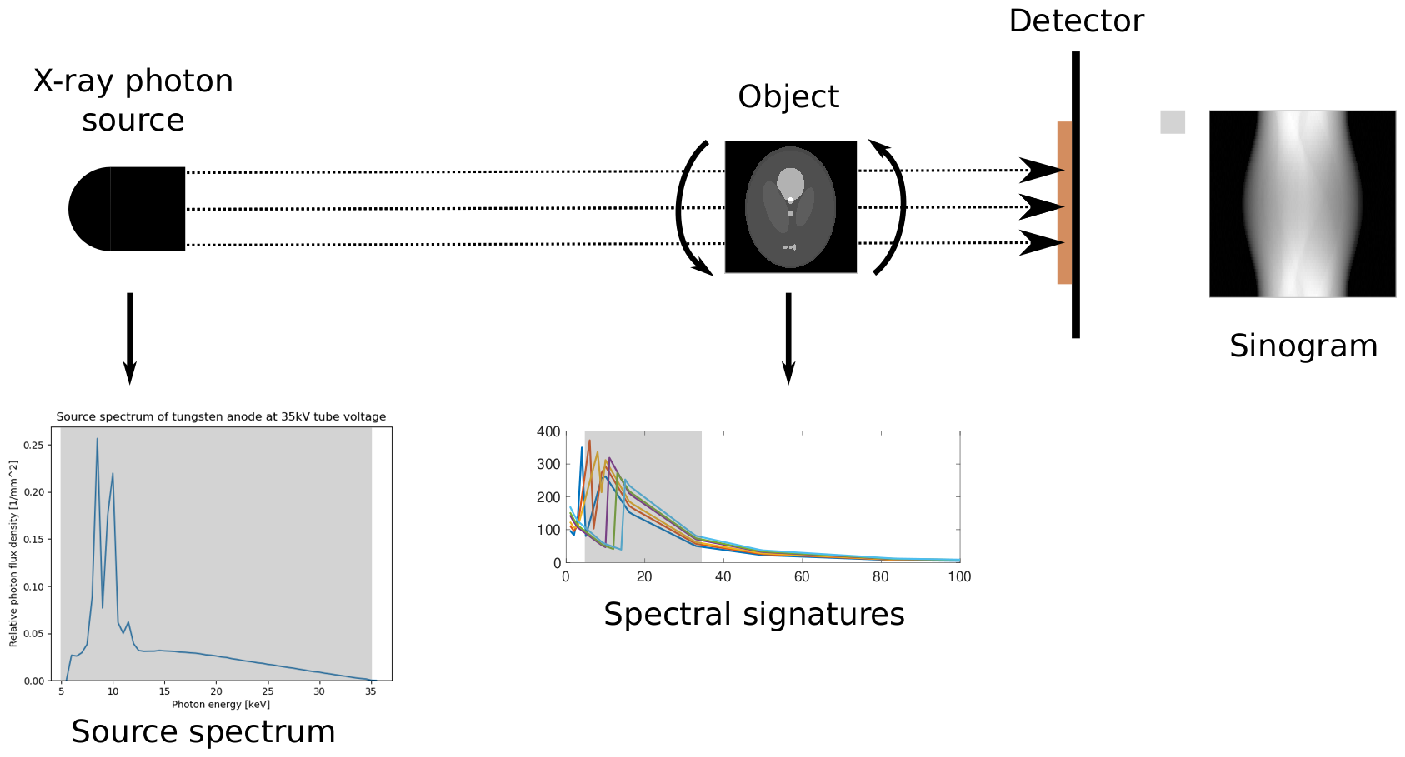} \qquad} \ \
    \rulesep
	\subcaptionbox{Data acquisition in spectral X-ray CT}[0.47\textwidth]{
		\includegraphics[width=0.47\textwidth]{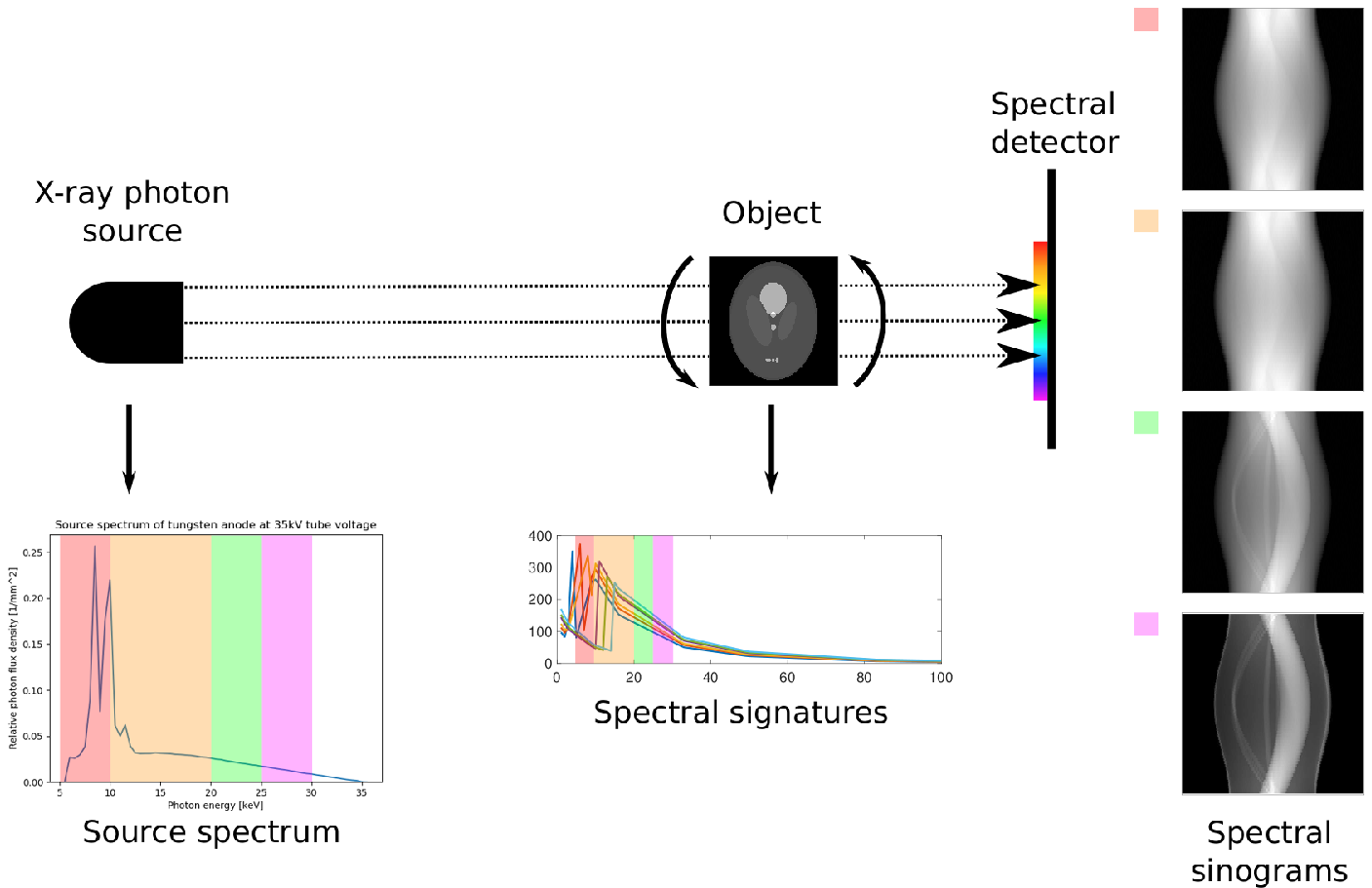} \qquad } \ \
    \caption{Comparison of data acquisitions in conventional X-ray CT (\textbf{a}) and spectral X-ray CT (\textbf{b}). In conventional X-ray CT, the energy window is dictated by the source spectrum (grey). None of the detected X-ray photons are distinguished by energy, leading to one sinogram (containing the projections ordered by angle). On the other hand, in spectral CT, the X-ray photons are distinguished according to their energy from multiple windows (colours), yielding multiple sinograms with different characteristics.}
    \label{fig:Motivation:ConventionalVSSpectral}
\end{figure}

Multi-energy X-ray detectors allow for collecting X-ray projection data in multiple energy bins with a very high spectral resolution. We can divide these detectors into two classes: (\emph{i})~detectors that measure all X-ray photons simultaneously and directly categorize these into various spectral channels, and (\emph{ii})~detectors that indirectly do this. Detectors from the first category are often used for hyperspectral imaging, and some examples include the Hexitec detector~\cite{EganJacques, JacquesEgan, VealeSeller}, Amptek X-123 CdTe detector~\cite{RedusHuber} and the SLcam \cite{OrdavoIhle,ScharfIhle}. On the other hand, the spectroscopic X-ray detectors from the second category indirectly measure many spectral channels by applying threshold scans. Detectors from the Medipix and Pixirad families are examples of this~\cite{BallabrigaAlozy, BallabrigaAlozy2}. The number of energy bins that can be recorded simultaneously is limited (usually up to 10), but different energy thresholds, with which X-ray photons with higher energies than that threshold can be detected, can be set between measurements. If the object of interest is static, hyperspectral images can, in principle, be easily obtained using these devices. \\

The measurements from these detectors, \ie\ the tomographic projections after their preprocessing and log-correction, are linearly related to the spectral characteristics of the materials present in the object. Although these detectors can now measure the X-ray projections in multiple energy bins (combined referred to as \textit{spectral projections}), it is not straightforward to optimally use the available spectral data for material decomposition. The conventional strategies for reconstruction and material decomposition are not designed to simultaneously process this type of spectral measurements and find accurate material maps (refer to Figure~\ref{fig:Motivation:comparison}). Hence, advanced computational techniques are required to infer the material composition from these spectral projections. 

\subsection{Motivation and problem description}

\begin{figure}[!htb]	
    \centering
	\includegraphics[width=0.8\textwidth]{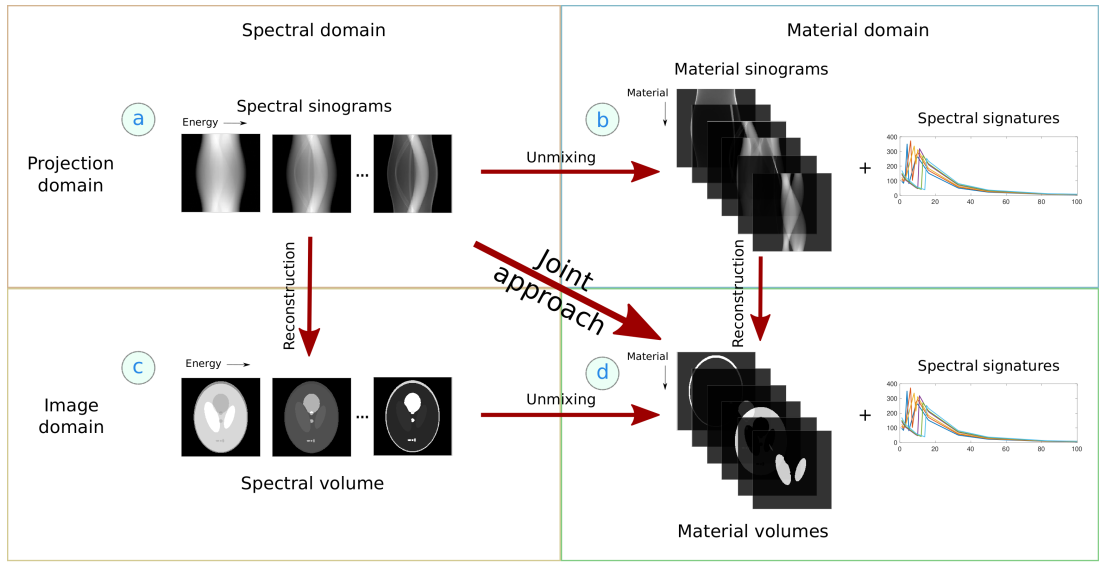}
    \caption{A schematic overview of various approaches to Spectral Computed Tomography:
The spectral sinograms (projections from various angles) obtained at different energy levels make up the input of these algorithms (\textbf{a}). The spectral sinograms can be decomposed first in the projection domain to obtain material sinograms and the spectral signatures (\textbf{b}). The spatial material maps (\textbf{d}) are then obtained by reconstructing each material sinogram separately. Alternatively, reconstruction can be done before the material decomposition by first making a CT reconstruction of every spectral channel using the associated spectral sinograms to obtain the spectral volumes (\textbf{c}). Then, in the image domain, these spectral volumes can be decomposed to obtain the spatial material volumes. These two-step approaches can also be combined into a one-step approach in which the decomposition and reconstruction are carried out as a joint approach.}
    \label{fig:Motivation:overview}
\end{figure}

Spectral tomographic projections are X-ray measurements at multiple energies that result from the spatial and spectral properties of the materials present in the object. We refer to \textit{Spectral Computed Tomography} as the process of estimating spatial maps and spectral signatures of materials present in the object from spectral tomographic projections. In general, the spectral CT workflow consists of two steps: (\emph{i})~ \textit{reconstruction}: computing spatial maps from their tomographic projections, (\emph{ii})~\textit{unmixing}: decomposition of the spectral volume into spatial material maps and their spectral signatures. Mathematically, reconstruction boils down to solving an inverse problem, while decomposition involves matrix factorization, an unsupervised learning approach. These two steps can be performed in a serial fashion to obtain the material maps from the spectral projections (refer to Figure~\ref{fig:Motivation:overview}). However, these two steps can also be combined into a one-step approach, which we refer to as \textit{joint}.  

To illustrate the performance of these two-step and one-step approaches, we consider a Shepp-Logan phantom, as described in Figure~\ref{fig:Motivation:comparison}. The numerical phantom consists of five materials (vanadium, chromium, manganese, iron and cobalt). For the tomographic projections, a full-view setting is chosen where 180 projections between 0 to $\pi$ are acquired. Figure~\ref{fig:Motivation:comparison} shows that the two-step approaches do not yield clear material decompositions or spectral signature reconstructions. Although the two-step methods are computationally efficient, they generally do not yield the same solution due to the ill-posedness of the problem. For the same reason, the joint approach is not expected to work well in all cases either. In particular, the current algorithms for joint methods suffer from ill-conditioning of the spectral profiles (refer to Section~\ref{sec:Illposedness}). The suboptimal performance of these methods motivates us to develop novel reconstruction methods to improve the spatial resolution and precise characterization of materials. To address the ill-posedness, we incorporate spatial and spectral prior information.

\newcommand{\centered}[1]{\begin{tabular}{l} #1 \end{tabular}}

\begin{figure}[!t]
    \renewcommand{\arraystretch}{1}
    \setlength{\tabcolsep}{3pt}
    \centering
    \begin{small}
    \begin{tabular}{ r c c c c c c}
    \centered{\bf GT} &
    \centered{\includegraphics[height=0.06\textheight,cfbox=col1 1pt 0pt]{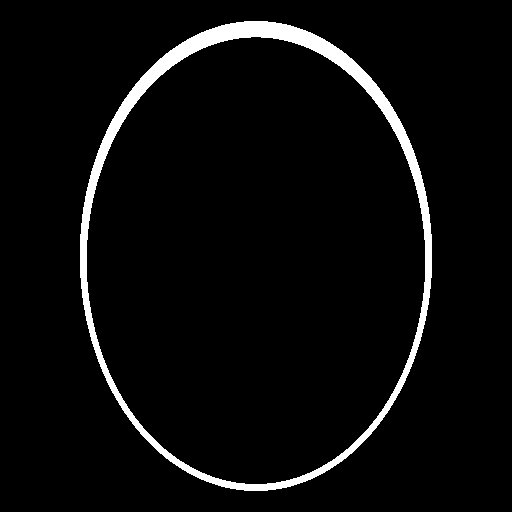}} & \centered{\includegraphics[height=0.06\textheight,cfbox=col2 1pt 0pt]{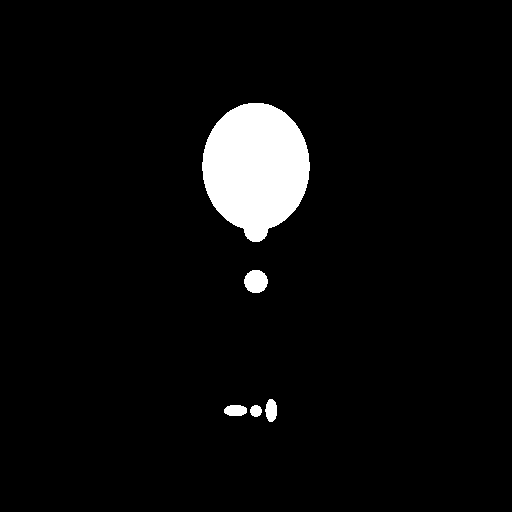}} & \centered{\includegraphics[height=0.06\textheight,cfbox=col3 1pt 0pt]{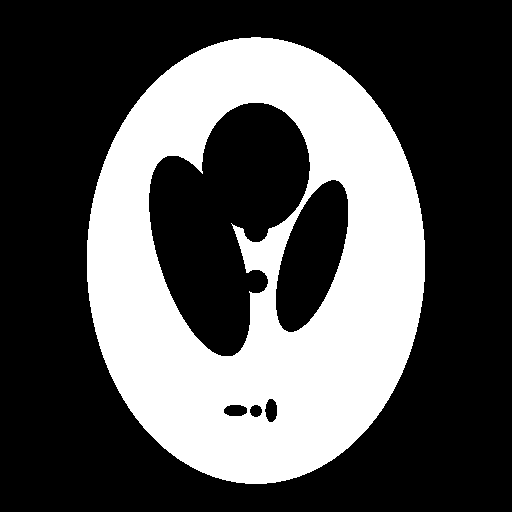}} & \centered{\includegraphics[height=0.06\textheight,cfbox=col4 1pt 0pt]{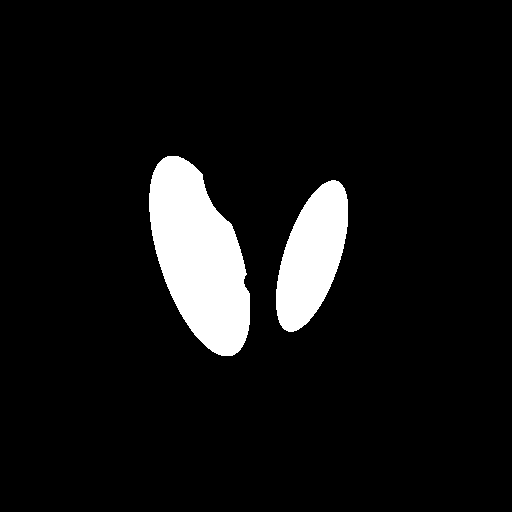}} &
    \centered{\includegraphics[height=0.06\textheight,cfbox=col5 1pt 0pt]{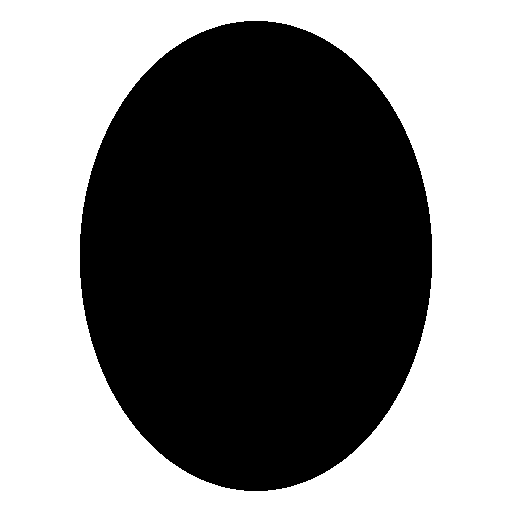}} &
    \centered{\includegraphics[height=0.06\textheight]{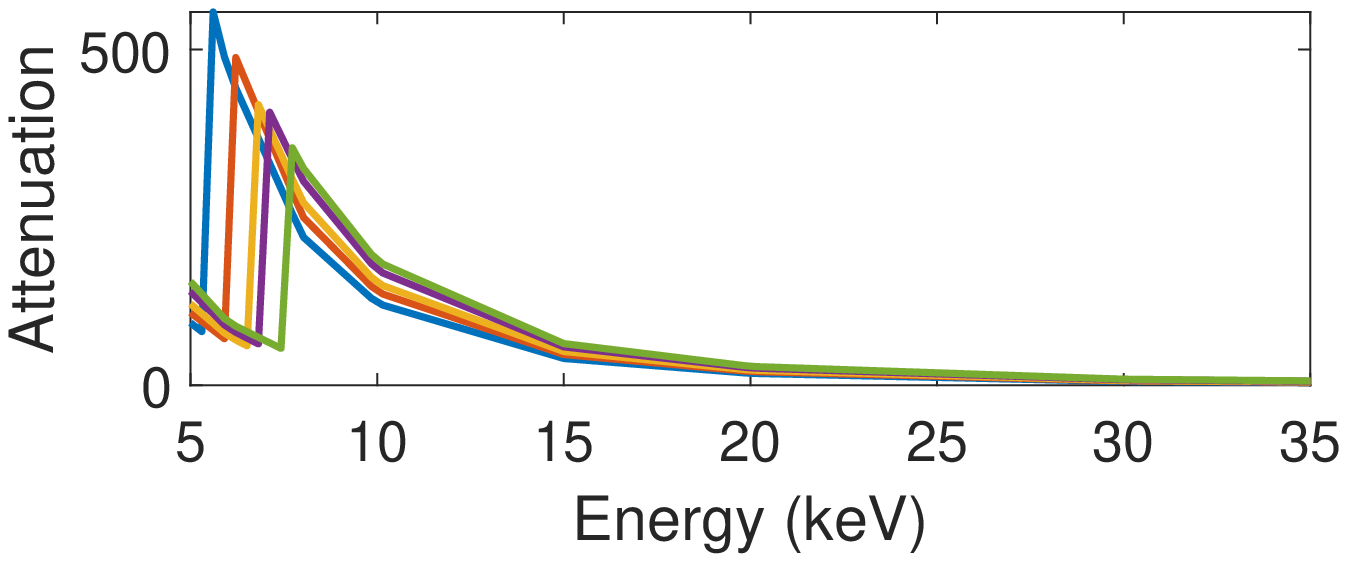}} \\ 
    \centered{\bf RU} &
    \centered{\includegraphics[height=0.06\textheight,cfbox=col1 1pt 0pt]{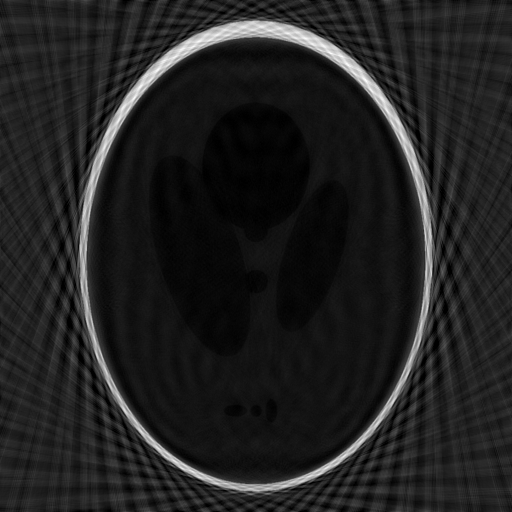}} & \centered{\includegraphics[height=0.06\textheight,cfbox=col2 1pt 0pt]{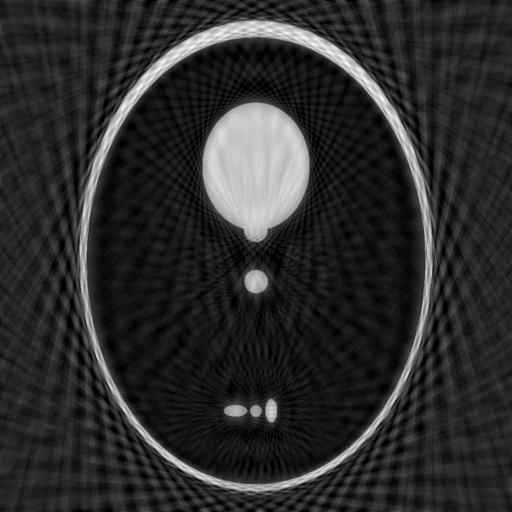}} & \centered{\includegraphics[height=0.06\textheight,cfbox=col3 1pt 0pt]{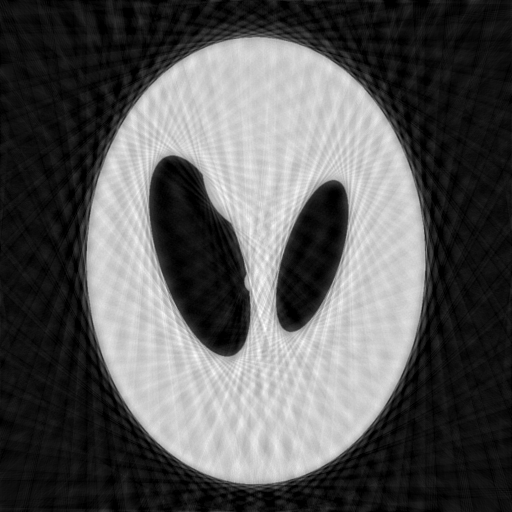}} & \centered{\includegraphics[height=0.06\textheight,cfbox=col4 1pt 0pt]{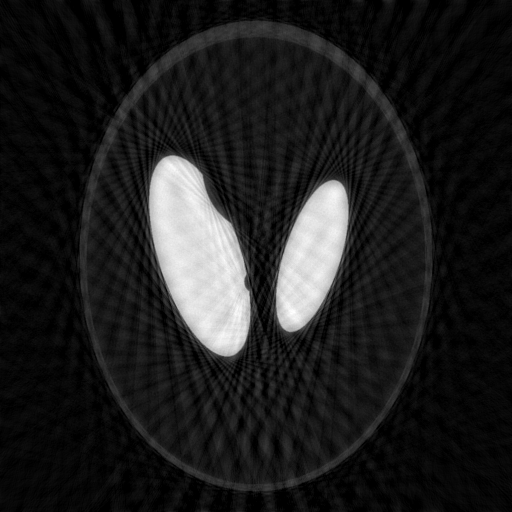}} &
    \centered{\includegraphics[height=0.06\textheight,cfbox=col5 1pt 0pt]{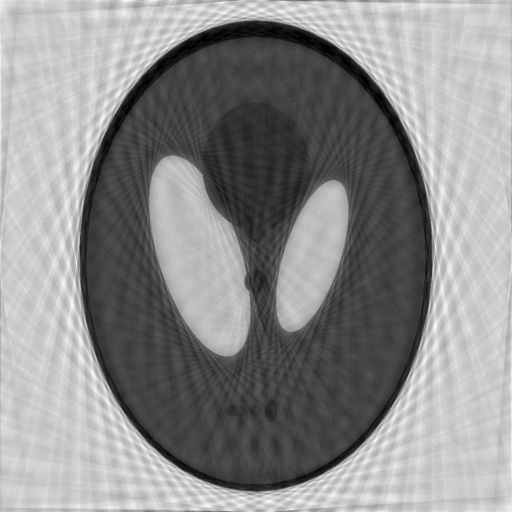}} &
    \centered{\includegraphics[height=0.06\textheight]{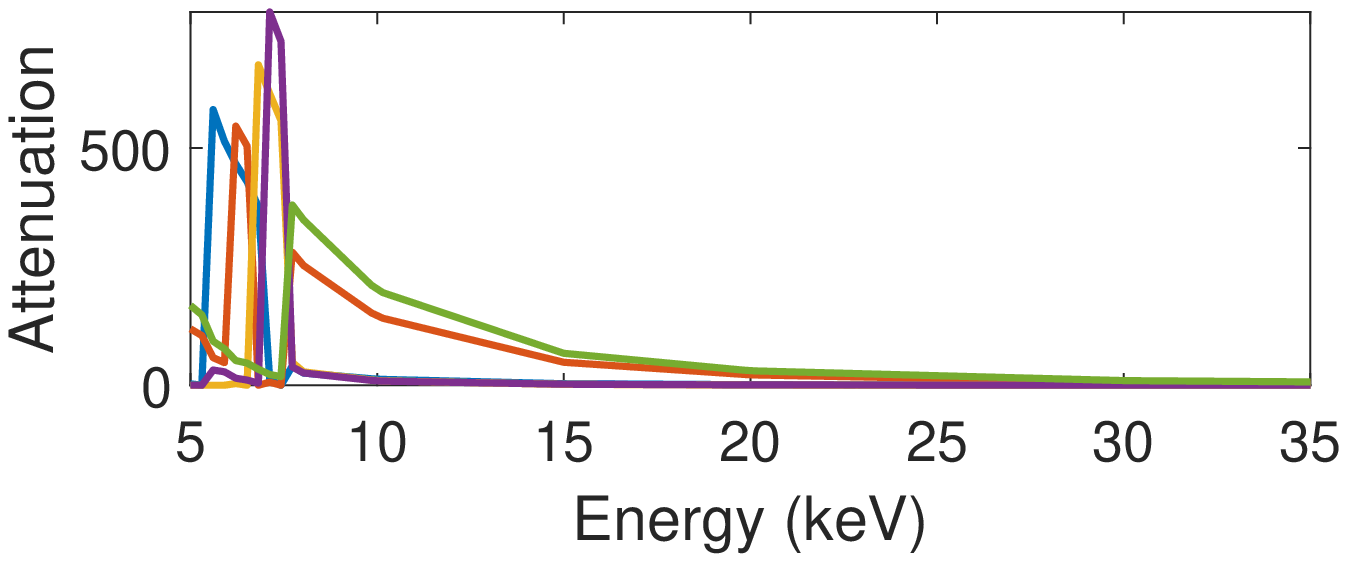}} \\ 
    \centered{\bf UR} &
    \centered{\includegraphics[height=0.06\textheight,cfbox=col1 1pt 0pt]{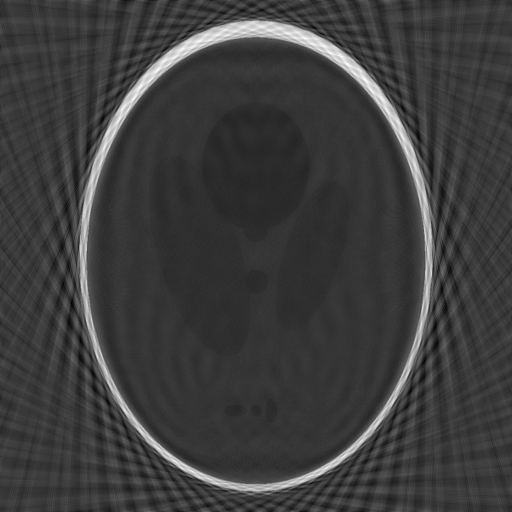}} & \centered{\includegraphics[height=0.06\textheight,cfbox=col2 1pt 0pt]{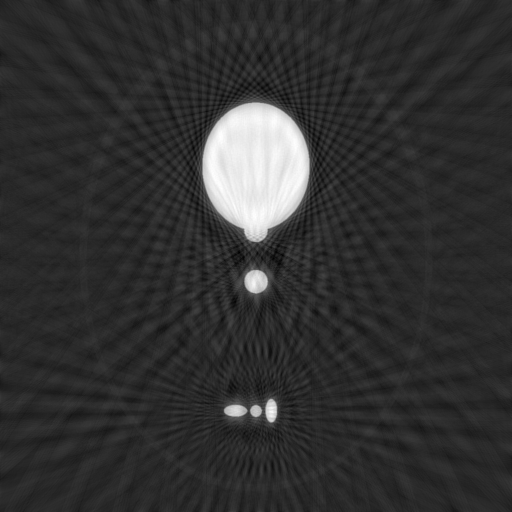}} & \centered{\includegraphics[height=0.06\textheight,cfbox=col3 1pt 0pt]{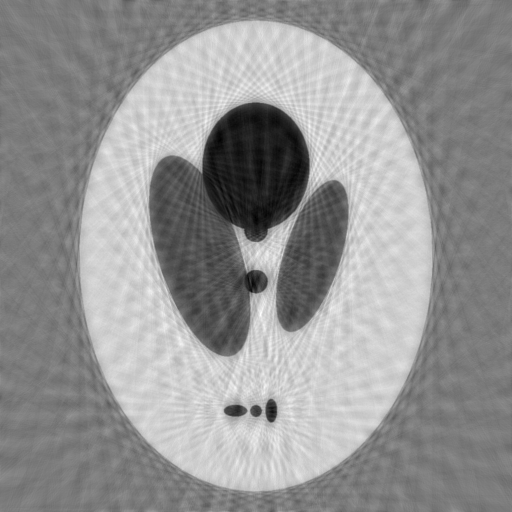}} & \centered{\includegraphics[height=0.06\textheight,cfbox=col4 1pt 0pt]{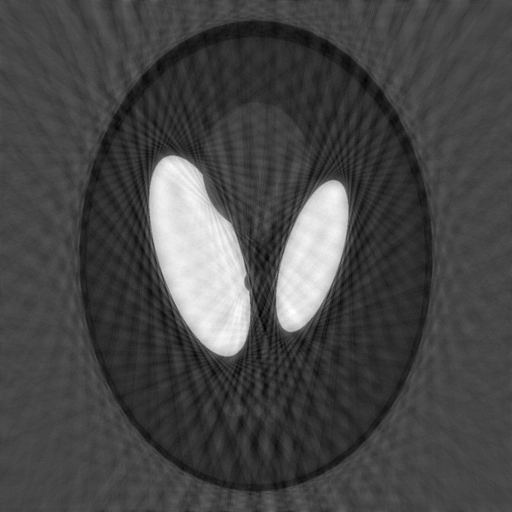}} &
    \centered{\includegraphics[height=0.06\textheight,cfbox=col5 1pt 0pt]{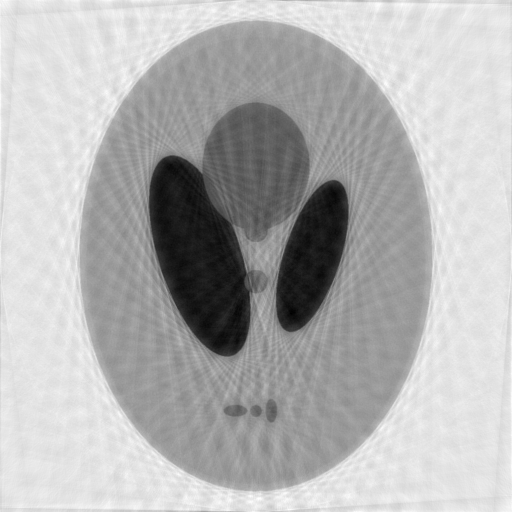}} &
    \centered{\includegraphics[height=0.06\textheight]{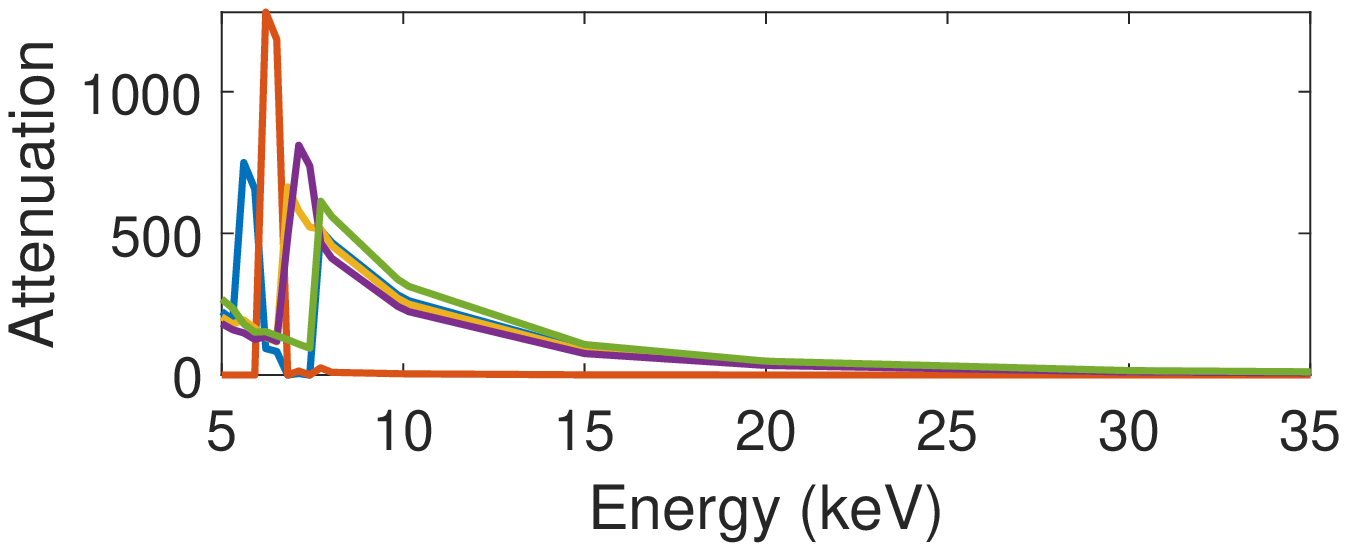}}  \\ 
    \centered{\bf cJoint} &
    \centered{\includegraphics[height=0.06\textheight,cfbox=col1 1pt 0pt]{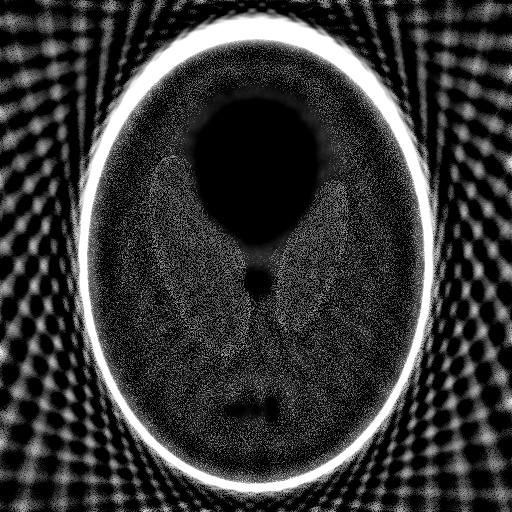}} & \centered{\includegraphics[height=0.06\textheight,cfbox=col2 1pt 0pt]{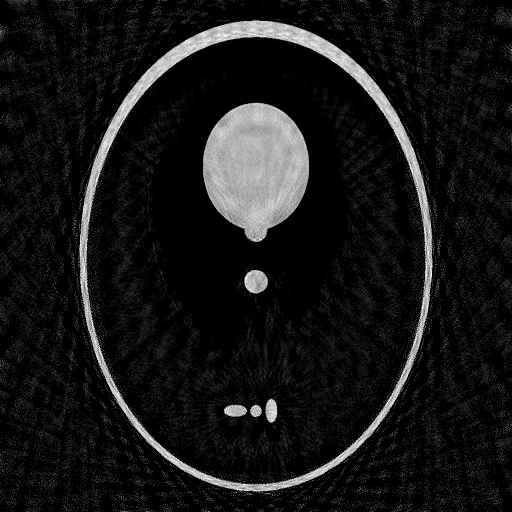}} & \centered{\includegraphics[height=0.06\textheight,cfbox=col3 1pt 0pt]{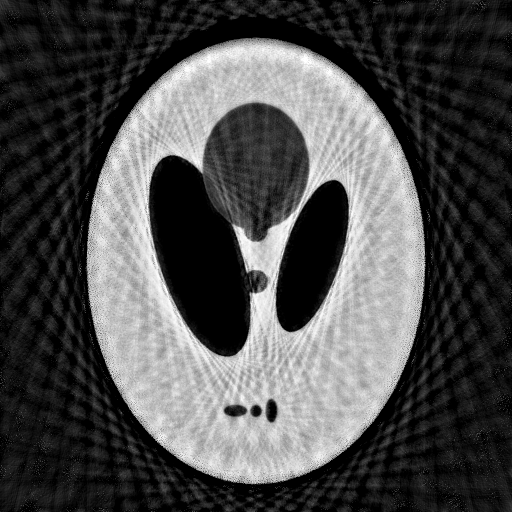}} & \centered{\includegraphics[height=0.06\textheight,cfbox=col4 1pt 0pt]{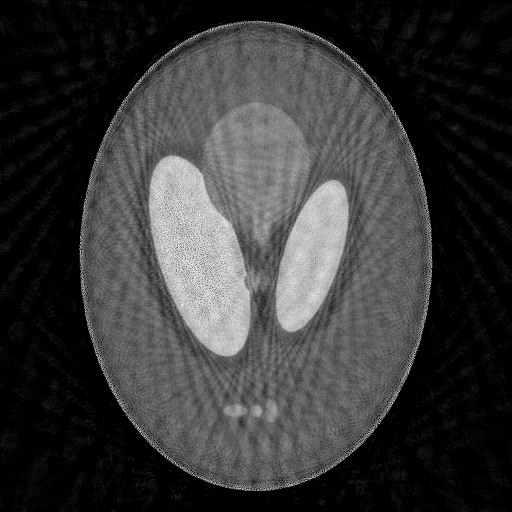}} &
    \centered{\includegraphics[height=0.06\textheight,cfbox=col5 1pt 0pt]{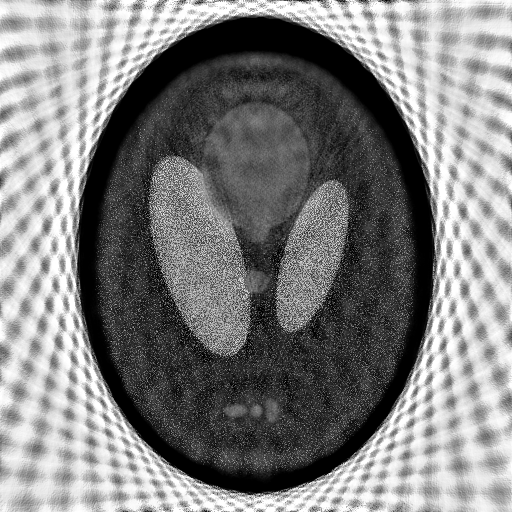}} &
    \centered{\includegraphics[height=0.06\textheight]{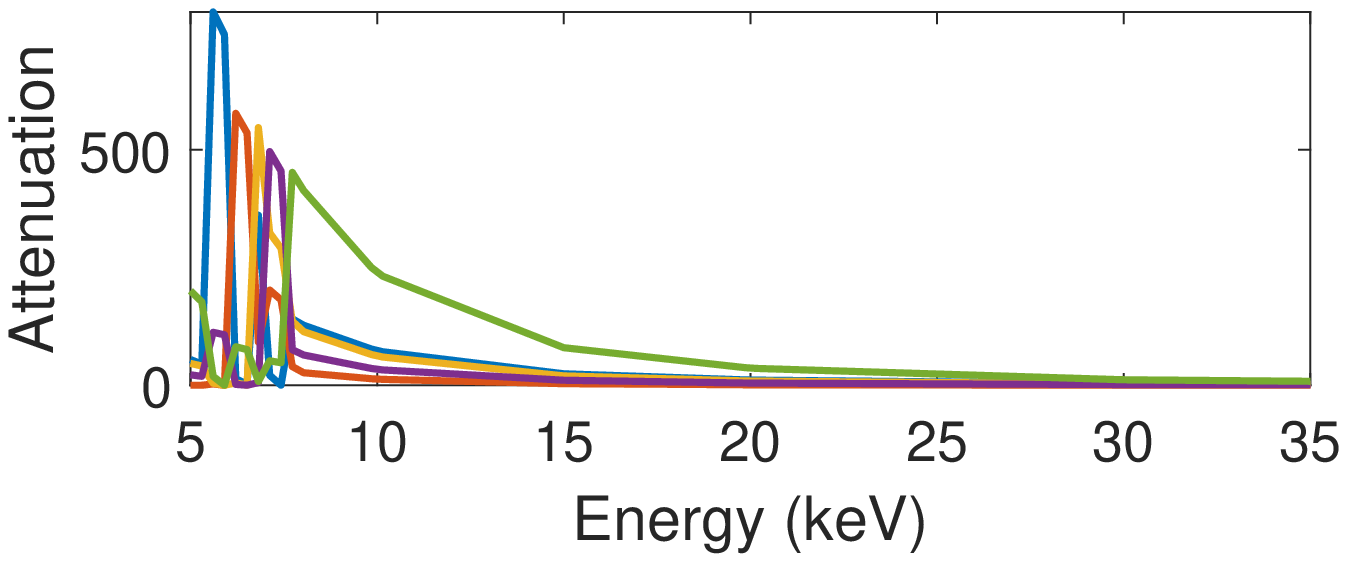}}  \\ 
    \centered{\bf ADJUST} &
    \centered{\includegraphics[height=0.06\textheight,cfbox=col1 1pt 0pt]{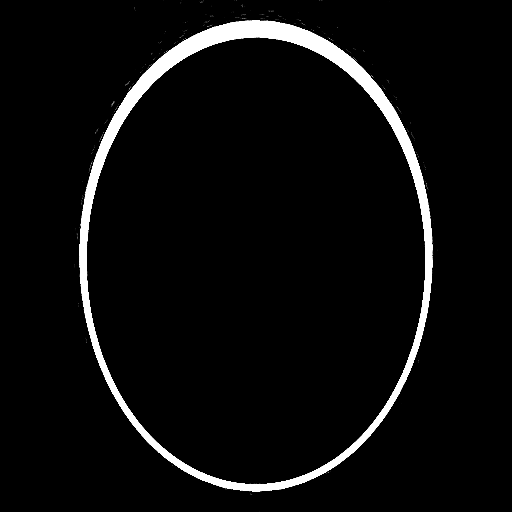}} & \centered{\includegraphics[height=0.06\textheight,cfbox=col2 1pt 0pt]{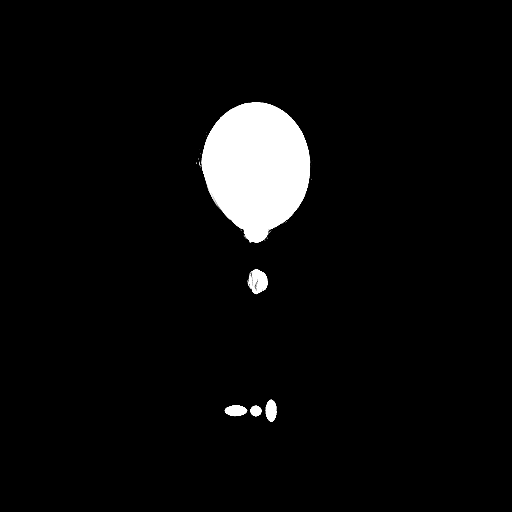}} & \centered{\includegraphics[height=0.06\textheight,cfbox=col3 1pt 0pt]{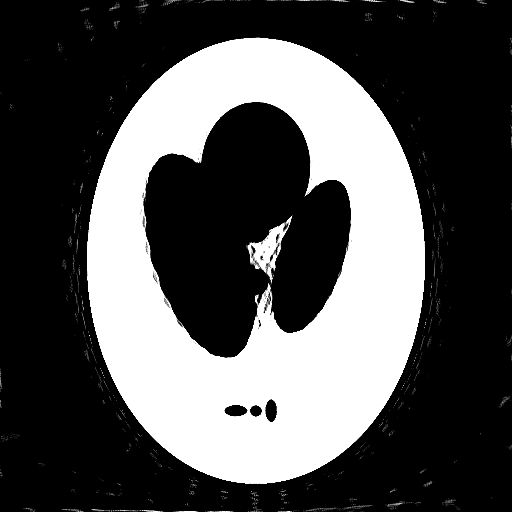}} & \centered{\includegraphics[height=0.06\textheight,cfbox=col4 1pt 0pt]{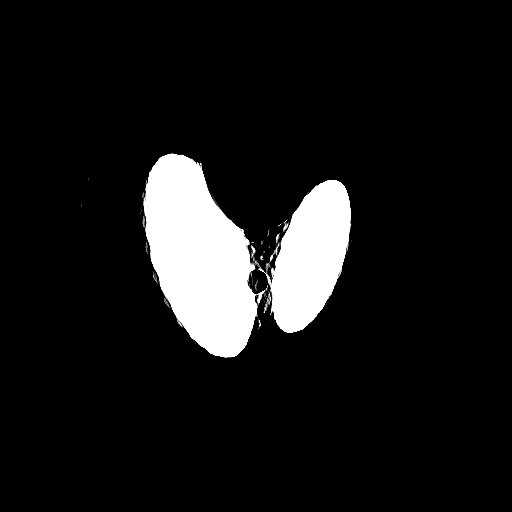}} &
    \centered{\includegraphics[height=0.06\textheight,cfbox=col5 1pt 0pt]{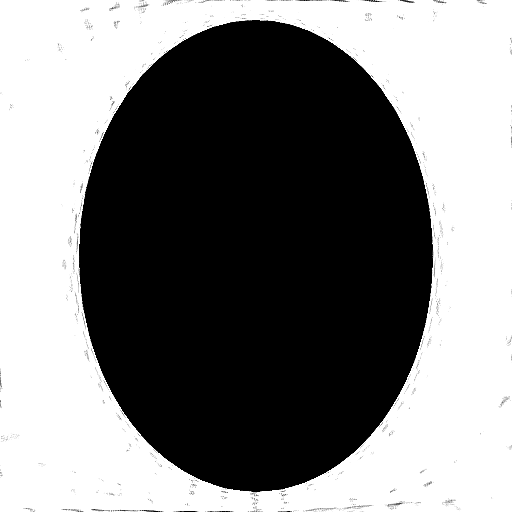}} &
    \centered{\includegraphics[height=0.06\textheight]{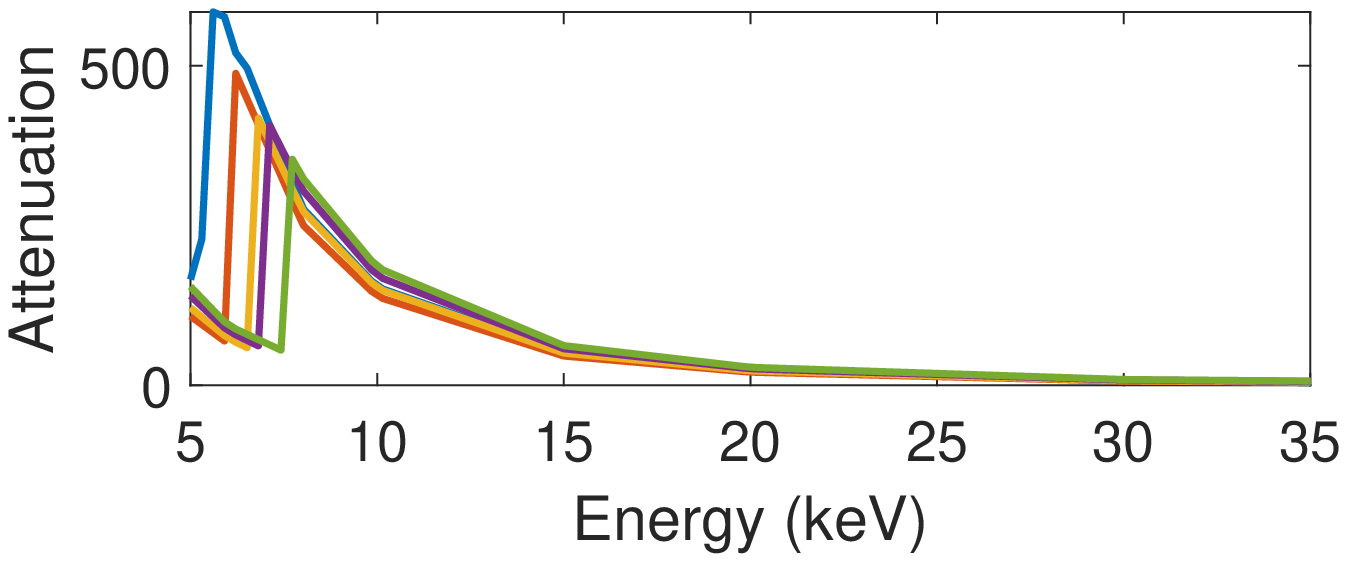}} \\
    \end{tabular}
    \end{small}
    \caption{Comparison of various methods for spectral CT on a five-material spectral Shepp-Logan phantom. The tomographic projections are gathered in 100 equally-sized bins between $5$ to $35$ keV. The top row shows the true material maps and the spectral signatures of each material. The second and third row show the material maps retrieved from two-step methods, reconstruction-then-unmixing (RU) and unmixing-then-reconstruction (UR), respectively. The fourth row shows the results of the classical joint method (see Section~(\ref{eq:joint:basic1})). In contrast, the fifth row shows results of the proposed method (see Section~\ref{sec:Proposed}).}
    \label{fig:Motivation:comparison}
\end{figure}

\subsection{Contributions and Outline}

In this paper, we propose a new technique called `\emph{A Dictionary-based Joint reconstruction and Unmixing method for Spectral Tomography} (ADJUST)' to reconstruct spatial material maps from their spectral tomographic measurements. ADJUST is a novel bi-convex optimization formulation that incorporates an effective \textit{spatiospectral} prior. This prior includes (\emph{i})~\textit{spatial}: the contribution of materials at each location should sum to 1, and (\emph{ii})~\textit{spectral}: the spectral signature of material should be a linear combination of the elements of a spectral dictionary. To solve the constrained optimization problem, we design a memory-efficient alternating proximal iterative scheme. In particular, we develop numerical methods to compute the required proximal operators quickly. In numerical experiments, we demonstrate that ADJUST performs better than the existing state-of-the-art methods. Furthermore, we show that ADJUST is also applicable to limited-angle problems found in industrial X-ray tomography, optical tomography and electron tomography.

The remainder of the paper is organized as follows. Section~\ref{sec:RelatedWork} discusses the existing work on spectral CT and material decomposition. Section~\ref{sec:ForwardModel} introduces the forward modelling of spectral X-ray tomography. In particular, we derive the linear map from the spatial map of materials to the spectral tomographic measurements. Then, we introduce the inverse problem in Section~\ref{sec:InverseProblem} that estimates the spatial material maps and spectral signatures from the spectral tomographic measurements. Here, we also discuss the ill-posedness involved in the inversion process. To reduce this ill-posedness, we introduce ADJUST in Section~\ref{sec:Proposed}. Moreover, we propose an iterative scheme that finds an approximate solution to the resulting biconvex formulation. In Section~\ref{sec:Experiments}, we numerically compare ADJUST with other methods on various synthetic phantoms. We also demonstrate the robustness of the method on limited measurement patterns such as sparse-angle tomography and limited view tomography, as well as in situations with limited spectral resolution and higher noise levels. In addition, we apply ADJUST to an experimental X-ray micro-CT dataset. Finally, we discuss the possibilities and limitations of the approach in Section~\ref{sec:Conclusions} and conclude the paper. 

\section{Related work}
\label{sec:RelatedWork}

For the sake of convenience, we categorize the previous work on spectral CT into (\emph{i})~(two-step) sequential approaches and (\emph{ii})~(one-step) joint approaches to reconstruction and material decomposition. Since we focus on multi-spectral CT, we do not discuss the advances in dual-energy CT. However, we refer the reader to the comprehensive review paper~\cite{SiBar} that covers dual-energy CT. \\
 
For spectral CT, sequential approaches (also known as two-step methods) are mainly (\emph{i})~reconstruction followed by decomposition~\cite{ClarkBadea, EganJacques2014, FredetteKavuri, WuChen2019, WuChen2020, ZhangMou}, and (\emph{ii})~decomposition followed by reconstruction~\cite{HohweillerDucros, MechlemSellerer, SchirraRoessl}. In the former category, material decomposition is carried out in the image domain, while in the latter category, it is carried out in the projection domain (see Figure~\ref{fig:Motivation:overview}). In both approaches, independent methods for material decomposition in the projection domain~\cite{DucrosAbascal}, (multi-channel) spectral reconstruction~\cite{ZeegersLucka} with various forms of structural or spectral regularization~\cite{KazantsevJorgensen, SalehjahromiZhang, SawatzkyXu}, and material decomposition in the image domain~\cite{FredetteKavuri} can be plugged in. Although these sequential two-step methods are computationally inexpensive, separating the reconstruction and decomposition steps causes information loss~\cite{MorySixou, WilleminkNoel}. \\

Joint methods (also known as one-step methods) that simultaneously reconstruct multi-channel spectral images and perform material decomposition have been developed to address the issues associated with sequential methods~\cite{JolivetLesaint, MorySixou}. All current one-step methods are iterative in nature and allow for the incorporation of prior knowledge through regularization. For example, (\emph{i})~structure on material maps is imposed through various penalties~\cite{HohweillerDucros, MorySixou}, (\emph{ii})~material maps are constrained using simplex constraints~\cite{FredetteKavuri, WuChen2019, WuChen2020}, and (\emph{iii})~structure on spectral signatures is enforced using a spectral dictionary~\cite{WuChen2020, ZhangZhao}. In particular, when the materials present in the object of interest are precisely known, various techniques improve the quality of the reconstructions. These tailored methods can image contrast agents, such as iodine~\cite{ZhangMou}, gold, gadolinium in angiography~\cite{MechlemSellerer}, in the presence of bone and tissue. Sometimes, knowledge about the materials can be used to choose the spectral bins effectively~\cite{FredetteKavuri}. Moreover, a deep learning approach has been proposed to perform joint decomposition and reconstruction task by generating a training set based on synthetic phantoms~\cite{AbascalDucros}. Most existing methods have been developed for, or demonstrated on, a limited number of materials (for example, two~\cite{JolivetLesaint, LiRavishankar, MechlemEhn}, three~\cite{BarberSidky, ZhangZhao}, and six~\cite{FredetteKavuri}), sometimes heavily relying on prior information about the materials, spectral signatures and energy bins. Some of these methods are extendable to more materials~\cite{AbascalDucros}, but this can be difficult for each additional parameter that may need to be estimated with each new material~\cite{BarberSidky}. However, these methods suffer when (\emph{i})~the number of materials present in the object is larger than 3, (\emph{ii})~the number of projections is smaller than the conventional criterion, (\emph{iii})~the measurements are corrupted with high noise~\cite{HeismannSchmidt}. \\

\section{Spectral Forward Model}
\label{sec:ForwardModel}

In this section, we provide the spectral X-ray forward model, by describing how we represent the objects and how spectral X-ray projections are obtained from this. The object is characterised by its attenuation coefficients $\mu(x,E) \in \mathbb{R}_{+}$, where $x \in \mathbb{R}^d$ is the location and $E > 0$ the X-ray photon energy, with $d\in\{2,3\}$ being the dimension of our space depending on whether we consider a slice-based or full 3D reconstruction. Given a polychromatic X-ray source with a source spectrum $I_0(E)$ at energy level $E$, and $C$ energy windows from the set $\setE = \{ \setE_c \}_{c \in \mathcal{C}} = \left\lbrace \left[ E_c^{\text{min}}, E_c^{\text{max}} \right] \right\rbrace_{c \in \mathcal{C}}$ (with $\mathcal{C}$ being its index set with $|\mathcal{C}| = C$) in which a spectral detector captures associated X-ray photons, we model the total X-ray photons captured by a detector pixel in energy bin $\setE_c$ as follows:
\begin{align}
    I(\setE_c)  &= \int_{E_{c}^{\min}}^{E_{c}^{\max}} I_0(E) \exp \left( -\int_{\ell}\mu(x,E)\, \mathrm{d} x \right) \, \mathrm{d} E. \label{eq:LambertBeer1} 
\end{align}
Here, the inner integral is taken over the line $\ell$ from the X-ray source to a detector pixel. The maximum and minimum energy range depend on detector specifications. We model the energy-dependent attenuation as a linear combination of energy-dependent material attenuations and their spatial contributions. We represent it mathematically as
\begin{align}
    \mu(x,E) &= \sum_{m \in \mathcal{M}} \mu_m(E) \alpha_m(x),
    \label{eq:Attenuation_Material_equation}
\end{align}
where the material attenuation coefficient $\mu_m$ is a function of energy. The material $m$ must be contained in a set of considered materials $\mathcal{M}$ of size $M$. The proportion of material $m$ at location $x$ is given by $\alpha_m$. From Equations~(\ref{eq:LambertBeer1}) and~(\ref{eq:Attenuation_Material_equation}), we arrive at the following continuous relationship of measured photons in terms of the material spatial distributions and their attenuations:
\begin{align}
    I(\setE_c)  &= \int_{E_{c_{\min}}}^{E_{c_{\max}}} I_0(E) \exp \left(-\sum_{m \in \mathcal{M}} \mu_m(E) \int_{\ell} \alpha_m(x) \, \mathrm{d} x \right) \, \mathrm{d} E.
\end{align}
If the spectral bins are sufficiently narrow then the source spectrum and the material-attenuation values can be approximated by their representative (average) values. These values are respectively $\overline{I}_0(\setE_c)$ and $\overline{\mu}_m(\setE_c)$, and form the following relation for the photon count in energy bin $\setE_c$ corresponding to the $c^{\text{th}}$ channel:
\begin{equation}
    \begin{split}
    I(\setE_c)  &\approx \overline{I}(\setE_c) \\ 
            &= \overline{I}_0(\setE_c) \exp \left( -\sum_{m \in \mathcal{M}} \overline{\mu}_m(\setE_c) \int_{\ell} \alpha_m(x) \, \mathrm{d} x \right).
    \end{split}
\end{equation}
In general, the photon count is perturbed by an energy dependent noise distribution. Moreover, the spectral X-ray detector is a photon counting detector, for which the noise in energy bin $\setE_c$ can be modelled using a Poisson distribution with parameter $I(\setE_c)$~\cite{PassmoreBates}. If the mean in each energy bin is sufficiently high, a realization $I^{\text{meas}}(\setE_c)$ of $I(\setE_c)$ measured by a spectral detector can be approximated using Gaussian distribution $\mathcal{N}( 0,\sigma^2)$, with variance $\sigma$ being inversely proportional to $I(\setE_c)$. In our experiments, we assume that the average photon count in each bin of detectors are sufficient for this approximation.

\section{Spectral Inverse Problem}
\label{sec:InverseProblem}
From the measured spectral X-ray projections, \textit{Spectral Computed Tomography} aims to retrieve the energy-dependent attenuation values $\mu_m(E_c)$ for each material $m$ and energy channel $E_c$ and the distribution $\alpha_m$ throughout the object for each material $m$. It boils down to solving the bilinear system, in terms of  $\overline{\mu}_m$ and $\alpha_m$,
\begin{align}
\sum_{m \in \mathcal{M}} \overline{\mu}_m(\setE_c) \int_{\ell_j} \alpha_m(x) dx &= -\ln \left(\frac{I_j^{\text{meas}}(\setE_c)}{\overline{I}_0(\setE_c)} \right), \quad \forall \, \ell_j \in L, \, \setE_c \in \setE,
\label{eq:bilinear}
\end{align}
with $L$ and $\setE$ denoting the set of rays (with size $|L| = J$) and the set of energy channels, respectively. In tomography, the material distributions are determined by discretizing the object space into the grid of either pixels (2D) or voxels (3D). For now, we consider a three-dimensional scene with $N$ voxels. The proportion of material $m$ in the $i^{\text{th}}$ voxel is then given by $a_{im}$. For each ray over line $\ell_j$ corresponding to the $j^{\text{th}}$ measurement, the quantity $w_{ji}$ determines the contribution of $i^{\text{th}}$ voxel to the $j^{\text{th}}$ measurement. Usually, the quantity $\overline{I}_0(\setE_c)$ can be determined accurately by performing a flatfield measurement (i.e. measurement without object). Therefore, the right-hand side of Equation~\eqref{eq:bilinear} is known. By expressing $-\ln\left(I_j^{\text{meas}}(\setE_c)/\overline{I}_0(\setE_c) \right) $ by $y_{jc}$ and $\overline{\mu}_m(\setE_c)$ by $f_{mc}$, we arrive at the following expression:
\begin{align}
    \sum_{m=1}^{M} \left(\sum_{i=1}^{N} w_{ji} a_{im} f_{mc} \right) = y_{jc}, \qquad j = 1, \dots, J, \mbox{ and } c = 1, \dots, C.
    \label{eq:compact}
\end{align}
Here, the total number of measurements for each channel is given by $J$, and $C$ denotes the total number of (energy) channels. Subsequently, we can write the expression in Equation~\eqref{eq:compact} in the following matrix notation
\begin{align}
    \mW \! \mA \mF = \mY,
    \label{eq:bilinearsystem}
\end{align}
where $\mY \in \R^{J \times C}$ represents tomographic measurements for $C$ number of channels, and $\mW \in \R^{J \times N}$ is a projection matrix containing the weights $w_{ji}$ described in Equation~\eqref{eq:compact}, $\mA \in \R^{N \times M}$ consists of $M$ columns of size $N$, with each column representing a spatial map corresponding to the material present in the object, while $\mF \in \R^{M \times C}$ consists of $M$ rows with each row denoting the channel attenuation information of the material. It is important to note that matrices $\mW$ and $\mY$ are known and matrices $\mA$ and $\mF$ are unknown. We formulate the joint spectral tomographic imaging and decomposition problem in a constrained least-squares form as
\begin{tcolorbox}[colframe=red!50!white,colback=red!10!white]
\begin{eqnarray}
    \begin{split}
    \underset{\mA, \mF }{\mbox{minimize}} \quad & \mathcal{J}(\mA, \mF) \triangleq \tfrac{1}{2} \norm{ \mY - \mW \! \mA \mF }_F^2 & \qquad \mbox{(least-squares misfit)}\\
    \mbox{subject to}  \quad & \mA \geq 0 & \mbox{(non-negativity of material maps)} \\
    & \mF \geq 0 & \qquad \mbox{(non-negativity of attenuation coefficients)}
    \end{split}
    \label{eq:joint:basic1}
\end{eqnarray}
\end{tcolorbox}
Here, we impose non-negativity constraints on both $\mA$ and $\mF$. 
The function $\mathcal{J}(\mA,\mF): \R^{N \times M} \times \R^{M \times C} \mapsto \R$ defines the misfit between the true measurements $\mY$ and the simulated measurements $\mW \! \mA \mF$ using the Frobenius norm. This norm is valid if the noise in the measurements is approximately Gaussian. We note that this joint formulation is a bi-convex optimization problem since the misfit function is bi-convex, and the constraints set is a bi-convex set~\cite{GorskiPfeuffer}. We denote the solution set of the joint formulation by 
\[
    \mathcal{B} = \bigg\lbrace \left( \mA_{\text{joint}}, \mF_{\text{joint}} \right) \bigg\rbrace = \argmin_{\mA, \mF} \bigg\lbrace \mathcal{J}(\mA, \mF) \ | \ \mA \geq 0, \mF \geq 0 \bigg\rbrace.
\]
This solution set $\mathcal{B}$ may contain more than one solution if the misfit function $\mathcal{J}$ is not strongly bi-convex. The solution set $\mathcal{B}$ cannot be determined trivially. To find the elements in set $\mathcal{B}$, we need to solve the optimization problem~\eqref{eq:joint:basic1} using an iterative scheme with an initial estimate of the solution~\cite{WendellHurter}.

\subsection{Practical Methods}
As outlined in Section~\ref{sec:RelatedWork}, joint methods aim to simultaneously estimate the spatial material distribution and the energy-dependent attenuation coefficients. However, these one-step methods are not practical due to their high computational cost. In practice, two-step methods are popular, where reconstruction and decomposition are performed separately, because of their modular nature. For each step, tailored solvers are readily available for different platforms. The first category of two-step methods, which we call \emph{RU} (short for Reconstruction-then-Unmixing), estimates a spectral volume from the spectral tomographic measurements, and then decomposes the resulting spectral volume to obtain the material maps and spectral signatures. That is, \emph{RU} solves the following problems in a serial fashion:
\begin{subequations}
\begin{align}
\mV_{\text{RU}} &= \argmin_{\mV \geq 0} \left\lbrace \frac{1}{2} \| \mW \mV - \mY \|_F^2 + \lambda \mathcal{R}_1(\mV) \right\rbrace\\
\left( \mA_{\text{RU}},\mF_{\text{RU}} \right) &= \argmin_{\mA \geq 0, \mF \geq 0} \left\lbrace \frac{1}{2} \| \mA \mF - \mV_{\text{RU}} \|_F^2 \right\rbrace
\end{align}
\label{eqn:RU}
\end{subequations}
\noindent where $\mV_{\text{RU}} \in \R^{N \times C}$ is a spectral volume, and $\mA_{\text{RU}}, \mF_{\text{RU}}$ are the material maps and spectral signatures respectively reconstructed by this method. $\mathcal{R}_1: \R^{N \times C} \mapsto \R$ is a regularization function that incorporates prior information about the spectral volumes and $\lambda \geq 0$ is a regularization parameter. Contrarily, \emph{UR} (short for Unmixing-then-Reconstruction), the other class of two-step methods, separates the spectral tomographic measurements into projections and spectral signatures. These projections then lead to the material maps. \emph{UR} mathematically reads
\begin{subequations}
\begin{align}
(\mP_{\text{UR}}, \mF_{\text{UR}}) &= \argmin_{\mP \geq 0,\mF \geq 0} \left\lbrace \frac{1}{2} \| \mP \mF - \mY \|_F^2 \right\rbrace\\
\mA_{\text{UR}} &= \argmin_{\mA \geq 0} \left\lbrace \frac{1}{2} \| \mW \mA - \mP_{\text{UR}} \|_F^2 + \gamma \mathcal{R}_2(\mA)\right\rbrace 
\end{align}
\label{eqn:UR}
\end{subequations}
where $\mP_{\text{UR}} \in \R^{J \times M}$ is a material volume, and $\mF_{\text{UR}}, \mA_{\text{UR}}$ are the material maps and spectral signatures respectively reconstructed by this method. Here, $\mathcal{R}_2: \R^{N \times M} \mapsto \R$ integrates prior information about the material maps with $\gamma \geq 0$ being the regularization parameter. These practical methods only work well when a complete projection series is available and the object is composed of materials whose spectra are clearly separable. However, the spectra of materials overlap in most industrial spectral tomography machines. Although these practical methods do not give accurate solutions, the results can be used as an initial guess for advanced reconstruction methods. 

\subsection{Ill-posedness}
\label{sec:Illposedness}
The Hadamard conditions to define a \textit{well-posed} problem consist of three criteria: (\emph{i})~\textit{existence}: There must be an $\mA^\star$ and $\mF^\star$ that satisfy $\mW \mA^\star \mF^\star = \mY$. (\emph{ii})~\textit{uniqueness}: The solution $\mA^\star$ and $\mF^\star$ must be unique. (\emph{iii})~\textit{stability}: small perturbations in the measurements $\mY$ should not lead to significant deviations in $\mA^\star$ and $\mF^\star$. If any of these conditions is violated, we call the problem \textit{ill-posed}. In general, we assume the existence of a solution to the least-squares problem since we use the Euclidean norm to measure the misfit in the discrete setting.  However, the uniqueness condition needs to be verified.  Moreover, the stability of the solution relies on the conditioning of projection matrix $\mW$ and the measurements $\mY$. \\

In general, the spectral inverse problem has multiple solutions if no prior information is incorporated. To see this, suppose $(\mA^{\star},\mF^{\star})$ is a solution to Equation~\eqref{eq:bilinearsystem}, then $\big( \alpha \mA^{\star},(1/\alpha) \mF^{\star} \big)$ is also a solution to Equation~\eqref{eq:bilinearsystem} for any $\alpha > 0$. Hence, the practical reconstruction methods and classical joint method solve an ill-posed problem. To reduce the non-uniqueness, and hence make the problem less ill-posed, we need to incorporate appropriate spatiospectral prior information. 

\section{Proposed Method - ADJUST}
\label{sec:Proposed}

Since the conventional spectral inverse problem remains ill-posed due to the non-uniqueness of solutions, we propose to incorporate spectral information through a spectral dictionary. The spectral profiles of many materials are readily available~\cite{NIST, HubbellSeltzer}, and the spectral responses in each spectral channel can easily be computed from these spectral profiles.  We model the spectral response for material $m$ as the binary combination of the dictionary elements. That is, 
\[
    \vf_{m} = \widehat{r}_{m1} \vt_1 + \widehat{r}_{m2} \vt_2 + \dots + \widehat{r}_{mD} \vt_D,
\]
where $\vt_1, \dots, \vt_D$ correspond to the spectral responses of $D$ distinct dictionary materials, and $\widehat{r}_{m1}, \widehat{r}_{m2}, \dots, \widehat{r}_{mD}$ are the coefficients of material $m$ that take the value of either $0$ or $1$. Suppose the $j^{\text{th}}$ material in the dictionary corresponds to the material $m$, then $\widehat{r}_{mj} = 1$, and the other coefficients will be zero. Hence, we can represent the spectral matrix $\mF \in \R^{M \times C}$ as 
\begin{align*}
    \underbrace{\left[\begin{array}{ccc}
    - & \vf_1 & - \\
     & \vdots &  \\
    - & \vf_M & - 
    \end{array} \right]}_{\mF}
    = 
    \underbrace{\left[\begin{array}{ccc}
    - & \widehat{\vr}_1 & - \\
    & \vdots & \\
    - & \widehat{\vr}_M & - 
    \end{array} \right]}_{\widehat{\mR}}
    \underbrace{\left[\begin{array}{ccc}
    | & & | \\
    \vt_1 & \dots & \vt_D \\
    | &  & |
    \end{array} \right]}_{\mT},
\end{align*}
where $\mT \in \R^{D \times C}$ is a dictionary of $D$ materials (with $D \geq M$) with spectral information for $C$ channels, and $\widehat{\mR} \in \{0,1\}^{M \times D}$ is a spectral coefficient matrix. \\  

Due to the binary constraints, finding such a matrix $\widehat{\mR}$ jointly with $\mA$ is a non-convex problem. To make it convex for fixed $\mA$, we relax the binary nature of the variable $\widehat{\mR}$. Moreover, we apply additional constraints on the material maps to ensure that the total contribution of materials at every voxel does not exceed 1. Thus, the resulting formulation, termed as \textit{A Dictionary-based Joint reconstruction and Unmixing method for Spectral Tomography} (ADJUST), is phrased as 
\begin{tcolorbox}[colframe=red!50!white,colback=red!10!white]
\begin{eqnarray}
    \begin{split}
    \underset{\mA, \mR }{\mbox{minimize}} \quad & \mathcal{J}(\mA, \mR) \triangleq \tfrac{1}{2} \norm{ \mY - \mW \! \mA \mR \mT }_F^2, & \qquad \mbox{(least-squares misfit)}\\
    \mbox{subject to} \quad & \mA \in \mathcal{C}_A, & \qquad \mbox{(constraints on spatial map)}\\
    & \mR \in \mathcal{C}_R, & \qquad \mbox{(constraints on dictionary coefficients)}
    \end{split}
    \label{eq:ADJUST}
\end{eqnarray}
\end{tcolorbox}
where $\mY \in \R^{J \times C}$ represents spectral tomographic measurements, $\mW \in \R^{J \times N}$ is a projection matrix containing the weights $w_{ji}$ described in Equation~(\ref{eq:compact}), $\mA \in \R^{N\times M}$ is the matrix that constitutes of the spatial contributions of the materials, $\mT \in \R^{D \times C}$ represents the fixed dictionary matrix containing attenuation spectra of many materials, and $\mR \in \R^{M \times D}$ is the dictionary coefficient matrix that represents the \emph{continuous} version of $\widehat{\mR}$. The constraint sets are
\begin{align*}
    \mathcal{C}_R &\triangleq \bigg\lbrace \mX \in \R^{M \times D} \, | \, \underbrace{\vphantom{\sum_{j}}x_{ij} \geq 0}_{(a)}, \, \, \underbrace{\sum_{j=1}^D x_{ij} \leq 1}_{(b)}, \, \underbrace{\sum_{i=1}^M x_{ij} \leq 1}_{(c)} \bigg\rbrace, \\
    \mathcal{C}_A &\triangleq \bigg\lbrace \mX \in \R^{N \times M} \, | \, \underbrace{\vphantom{\sum_{j}}x_{ij} \geq 0}_{(d)}, \, \underbrace{\sum_{j=1}^{M} x_{ij} \leq 1}_{(e)} \bigg\rbrace.
\end{align*}
We provide the details of each constraint below:
\begin{itemize}
    \item[(a)] \textit{Non-negativity of $\mR$}: Since $\mR$ is a dictionary-coefficient matrix that is a convex proxy for $\widehat{\mR}$, the values must be greater than or equal to 0.
    \item[(b)] \textit{Row-sum constraints for $\mR$}: In principle, we would like to impose that each material present in the object must be a part of the dictionary. The convex approximation of this condition is that the total contribution of the dictionary elements to represent material should not exceed 1. 
    \item[(c)] \textit{Column-sum constraints for $\mR$}: Each column of $\mR$ represents the contribution of the dictionary element to generate materials in the object. Since the number of materials in the object is smaller than the total number of dictionary elements, the contribution of many dictionary elements will be 0. Moreover, the materials present in the object must be distinct. Hence, the contribution of dictionary elements must not exceed 1. Hence, the column-sum constraints impose these conditions.
    \item[(d)] \textit{Non-negativity of $\mA$}: Each material should have a non-negative contribution to every voxel.
    \item[(e)] \textit{Row-sum constraints for $\mA$}: The total contribution of materials in each voxel must not exceed 1. 
\end{itemize}

We enumerate the benefits of ADJUST as follows. (\emph{i}) The ADJUST formulation is less ill-posed when compared with the two-step methods or the Joint formulation given in~\eqref{eq:joint:basic1}. This is due to the fact that the incorporation of the spectral dictionary resolves the scaling issue. (\emph{ii}) It is a \textit{parameter-free} approach since the constraints are simplex and do not involve any parameters that need to be estimated. The only parameter ADJUST requires is the number of materials present in the object. However, this prior knowledge is generally available to the user. (\emph{iii}) The optimization problem~\eqref{eq:ADJUST} is bi-convex (refer to Appendix~\ref{sec:Biconvexity} for the proof). Hence, it can be solved efficiently using the iterative minimization method.

\subsection{Numerical Optimization} \label{sec:NumericalOptimization}
\begin{figure}[!htb]
    \centering
    \includegraphics[width=0.5\textwidth]{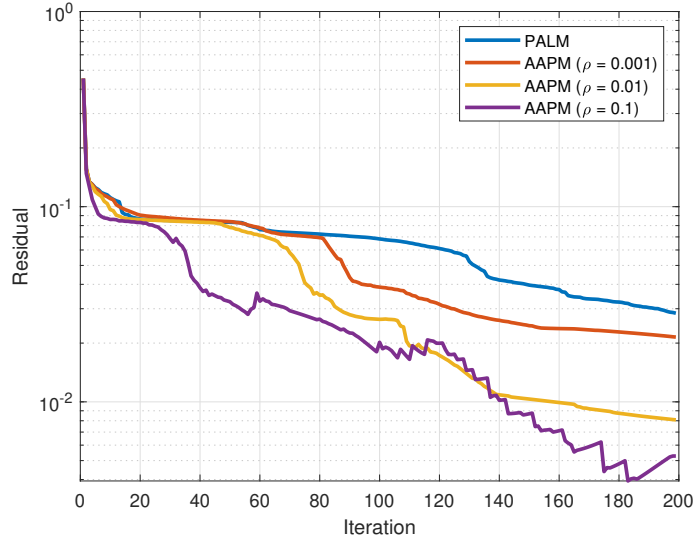}
    \caption{Numerical demonstration of proposed acceleration scheme on the Shepp-Logan phantom for various $\rho$ values.}
    \label{fig:residual_plot}
\end{figure}

To obtain an approximate solution to~\eqref{eq:ADJUST}, many alternating minimization schemes exist~\cite{AttouchBolte, BolteSabach, LeeuwenAravkin}. However, these schemes rely on complete minimization with respect to at least one variable in every step. Moreover, their convergence to a partially optimal solution is slow. Hence, such schemes might not be computationally feasible for large-scale problems. For practical applications, we propose an accelerated variant of Proximal Alternating Linearized Minimization (PALM)~\cite{BolteSabach}, by combining it with the acceleration strategy in alternating direction method of multipliers (ADMM) \cite{ParikhBoyd}. We term this variant as `Alternating Accelerated Proximal Minimization' (AAPM):
\begin{tcolorbox}[colframe=red!50!white,colback=red!10!white]
\begin{eqnarray}
\begin{split}
    \mbox{for} \quad k = 0, \dots &, K-1  \\
    \mR_{k+1} &= \proj_{\mathcal{C}_{R}} \left( \mR_{k} - \alpha \nabla_{\mR} \widetilde{\mathcal{J}}(\mA_k, \mR_k, \mU_k) \right) & \mbox{(update dictionary coefficients)} \\[1ex]
    \mA_{k+1} &= \proj_{\mathcal{C}_{A}} \left( \mA_{k} - \beta \nabla_{\mA} \widetilde{\mathcal{J}}(\mA_k, \mR_{k+1}, \mU_k) \right) & \mbox{(update material maps)} \\[1ex]
    \mU_{k+1} &= \mU_{k} + \rho \left( \mW \! \mA_{k+1} \mR_{k+1} \mT - \mY \right) & \mbox{(update running-sum-of-errors)}
\end{split}
\label{eq:ADJUST-optscheme}
\end{eqnarray}
\end{tcolorbox}
\noindent where $\alpha$ and $\beta$ are estimated using a line-search method (\eg, backtracking), $\rho$ is an acceleration parameter chosen from the range $[10^{-3},1)$, and $\widetilde{\mathcal{J}} ( \mA, \mR, \mU ) = \mathcal{J}(\mA,\mR) + \left\langle \mU,  \mY - \mW \! \mA \mR \mT \right\rangle$. The variable $U$ contains the running sum of errors (\ie, residuals):  
\[
    \mU_{k+1} = \mU_{0} + \rho \sum_{i=1}^{k} \mR_i \quad \mbox{where} \quad \mR_i = \mW \! \mA_i \mR_i \mT - \mY.
\]
For $\rho = 0$, AAPM is equivalent to PALM. When $\rho > 0$, the error signal $U$ is driven to zero by feeding back the integral of the error to its input. In Figure~\ref{fig:residual_plot}, we plot the residuals versus iterations for various values of $\rho$. From these results, we conclude that acceleration can be achieved by including the running-sum-of-errors into an alternative iteration scheme. As the $\rho$ values are increased, the residual decreases faster. However, for higher values of $\rho$, the monotonic decrease of the residual disappears. We consider $\rho$ as a hyperparameter for which a reasonable value can be determined heuristically. The derivation of the method is given in Appendix~\ref{sec:AAPM}.\\

\noindent
It is easy to compute the partial derivatives from basic linear algebra and calculus rules (refer to Appendix~\ref{sec:AppendixGradientComputations}). The partial derivatives are
\begin{align*}
    \nabla_{\mA} \widetilde{\mathcal{J}} ( \mA, \mR, \mU ) &= \mW^T \! \left( \mW \! \mA \mR \mT - \mY - \mU \right) \mT^T \! \mR^T, \\
    \nabla_{\mR} \widetilde{\mathcal{J}}(\mA, \mR, \mU) &= \mA^T \mW^T \! \left( \mW \! \mA \mR \mT - \mY -\mU \right) \mT^T.
\end{align*}
The proximal operators are derived in Section~\ref{sec:ProxOps}. We use the following parameters to determine the stopping criteria:
\begin{align*}
    \epsilon_k^{\text{abs}} &= \| \mY - \mW \mA_{k+1} \mR_{k+1} \mT \|_F / \| \mY \|_F \\
    \epsilon_k^{\text{rel}} &= \| \mA_{k+1} - \mA_{k} \|_F + \| \mR_{k+1} - \mR_{k} \|_F
\end{align*}
where $k$ is an iteration of the optimization scheme~\eqref{eq:ADJUST-optscheme}. The benefits of AAPM are the following:
\begin{itemize}[noitemsep]
    \item \textit{Simple gradient computations}: The gradients have explicit expressions and can be computed using simple matrix-matrix multiplications.
    \item \textit{Fast proximal operations}: The proximal operations are just orthogonal projections onto convex sets. These operations have either an explicit expression or a simple alternating routine to find the proximal point efficiently.
    \item \textit{Backtracking line-search}: The improved line-search (\ie\ finding $\alpha,\beta$) makes sure that the progress in the descent direction is appropriate for every iterate.
    \item \textit{Acceleration through running-sum-of-errors}: The regular update of variable $\mU$, which contains the running-sum-of-errors, helps in accelerating the convergence to the partial optimal solution. 
    \item \textit{Memory-efficiency}: The method relies only on forward and adjoint operations with the tomography operator $\mW$. Hence, it saves memory to explicitly store the tomography operator in either single or double precision.
\end{itemize}

\subsection{Orthogonal projections}
\label{sec:ProxOps}
In this section, we derive the projections onto the convex sets $\mathcal{C}_{R}$ and $\mathcal{C}_{A}$. For $\mathcal{C}_A = \left\lbrace \mX \in \R^{N \times M} \, | \, \mX \geq 0, \, \mX \vone \leq \vone \right\rbrace$, its orthogonal projection takes the following form:
\[
    \proj_{\mathcal{C}_A}(\mZ) = \max \left( \mZ - \mLambda \vone, \vzero \right), 
\]
where $\mLambda = \diag \left(\lambda_1, \dots, \lambda_N \right)$ is the diagonal matrix with weights computed from solving the inequalities
\[
    \left\langle \vone , \max \left( \vz_{i:}^T - \lambda_i \vone, \vzero \right) \right\rangle  \leq  1, \quad i=1, \dots, N,
\]
with $\vz_{i:}$ being the $i^{\text{th}}$ row of the matrix $\mZ$~\cite[Theorem~6.27]{Beck}. To find the optimal weight $\lambda_i$, we carry out bisection on $\lambda_i$ for which $\max \left(\left\langle \vone , \max \left( \vz_{i:}^T - \lambda_i \vone, \vzero \right) \right\rangle  -  1,0 \right) = 0$ holds, starting with the initial interval $\left[0, \max(\vz_{i:}) \right]$. The function $\left\langle \vone , \max \left( \vz_{i:}^T - \lambda_i \vone, \vzero \right) \right\rangle  =  1$ is piecewise linear, with breakpoints at the values $z_{i1}, \dots, z_{iM}$. Hence, once we have localized $\lambda_i$ to be between two adjacent
values, we can immediately compute the optimal value $\lambda_i^\star$. Furthermore, the following theorem entails the projection onto set $\mathcal{C}_{R}$:
\begin{thm} \label{thm:ProxR}
The convex set $\mathcal{C} = \left\lbrace \mX \in \R^{M \times D} \, | \, \mX \geq 0, \, \mX \vone \leq \vone, \mX^T \vone \leq \vone \right\rbrace$ is composed of convex sets $\mathcal{C}_1 = \left\lbrace \mX \in \R^{M \times D} \, | \, \mX \geq 0, \, \mX \vone \leq \vone \right\rbrace$ and $\mathcal{C}_2 = \left\lbrace \mX \in \R^{M \times D} \, | \, \mX \geq 0, \, \mX^T \vone \leq \vone \right\rbrace$. The projection of point $\mZ \in \R^{M \times D}$ onto set $\mathcal{C}$  is given by the fixed-point iteration scheme
\[
    \mX_{t+1} = \proj_{\mathcal{C}_1} \left(\proj_{\mathcal{C}_2} \left( \frac{\mX_t + \mZ}{2} \right)  \right), \quad t = 0, \dots, D
\]
with $\mX_{0} = \mZ$.
\end{thm}
\begin{proof}
    The proof is given in Appendix~\ref{sec:ProofsThm2}.
\end{proof}
For the set $\mathcal{C}_2 = \left\lbrace \mX \in \R^{M \times D} \, | \, \mX \geq 0, \, \mX^T \vone \leq \vone \right\rbrace$, the orthogonal projection, derived from~\cite[Theorem~6.27]{Beck}, takes the form
\[
    \proj_{\mathcal{C}_2}(\mZ) = \max \left( \mZ - \mone \mOmega, \mzero \right), 
\]
with $\mOmega = \diag \left(\omega_1, \dots, \omega_D \right)$ is the diagonal matrix with weights computed from solving the equation
\[
    \left\langle \vone , \max \left( \vz_{i} - \omega_i \vone, \vzero \right) \right\rangle  =  1, \quad i=1, \dots, D.
\]


\section{Experiments \& Results}
\label{sec:Experiments}

This section compares ADJUST with the sequential methods (RU, UR), the classical joint method (cJoint), and five state-of-the-art joint methods on a benchmark synthetic spectral phantom. After this, we compare ADJUST with the sequential methods and the classical joint method on more advanced spectral phantoms (in terms of number of materials and material shapes). Next, we examine the robustness of ADJUST against various limited measurement patterns. Finally, we show the application of ADJUST on spectral X-ray micro-CT dataset. Additional numerical experiments are presented in Appendices~\ref{sec:NumStudComparison},~\ref{sec:NumStudLimited},~\ref{sec:NumStudMixed}, and~\ref{sec:NumStud3D}.

\subsection{Experimental setup}
\label{sec:ExpSetup}
Before presenting the results, we first outline the experimental setup. We describe the phantoms that are used, the settings for the attenuation spectra and the source spectrum, the chosen materials for each phantom and a discussion on the implementation of the algorithms.

\subsubsection{Spectral phantoms}
\begin{figure}[H]
    \centering
    \begin{tabular}{cccc}
         \includegraphics[width=0.2\textwidth,clip]{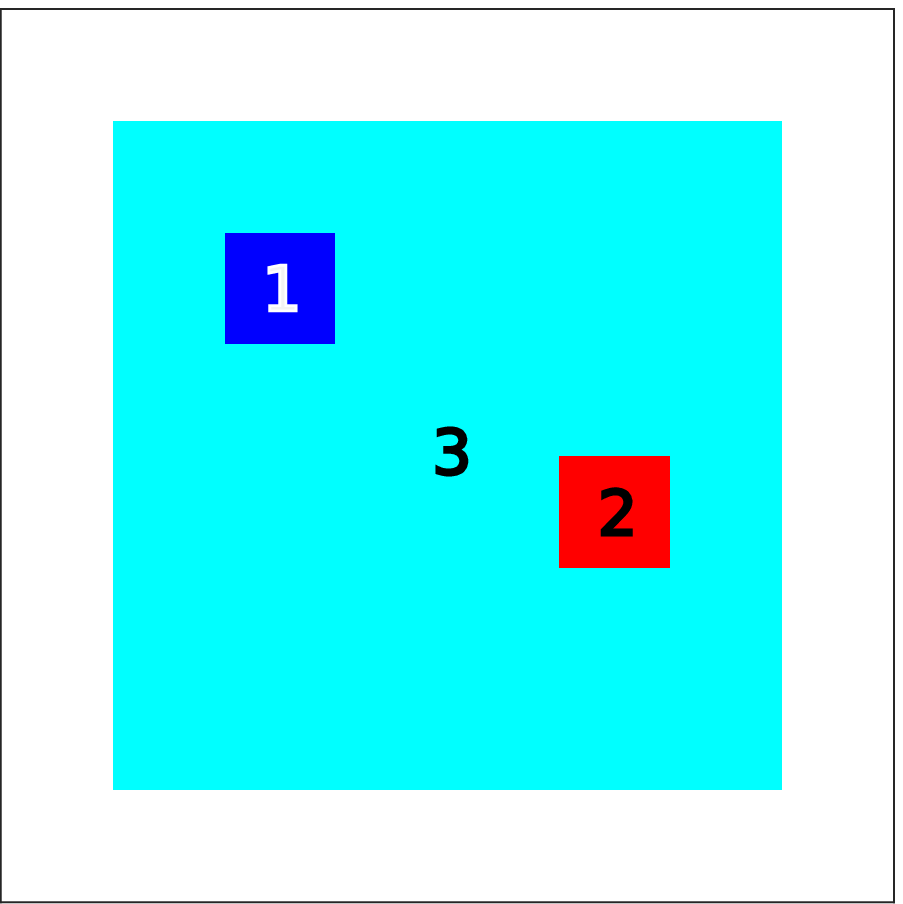} & \includegraphics[width=0.2\textwidth,clip]{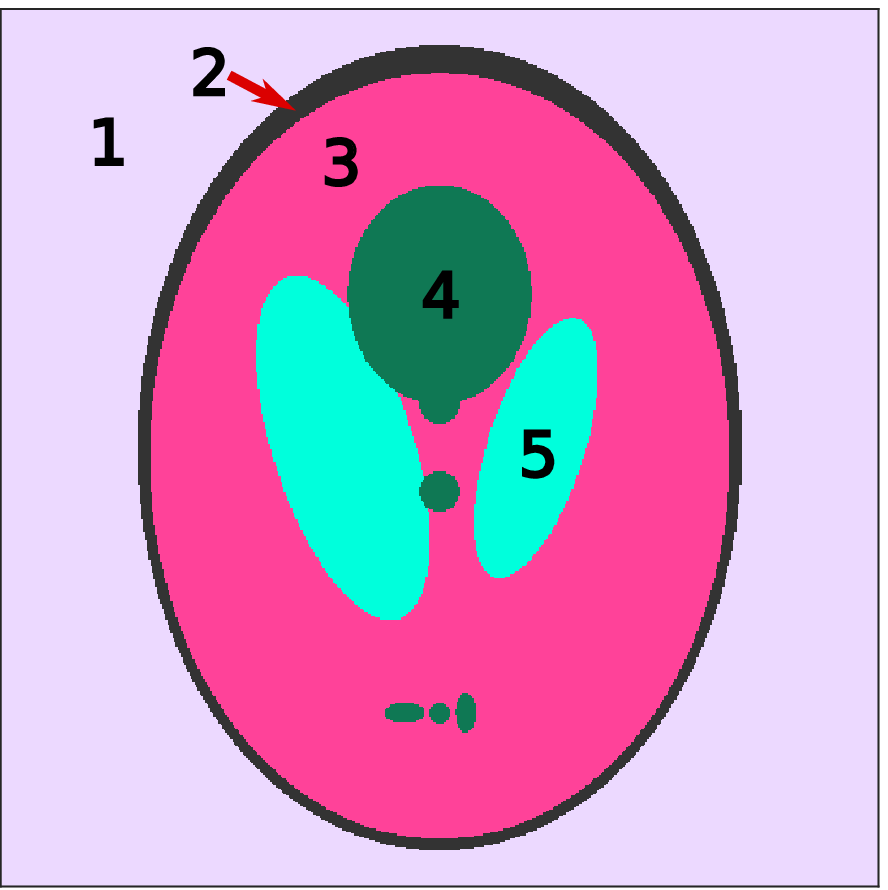} & \includegraphics[width=0.2\textwidth,clip]{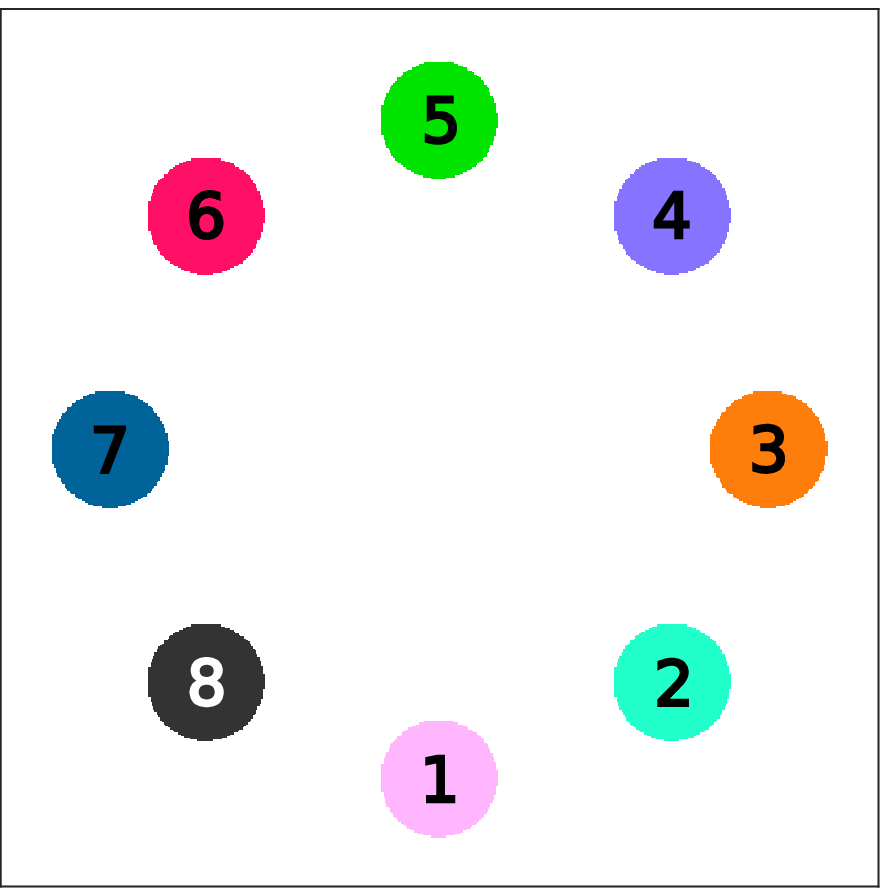} & \includegraphics[width=0.2\textwidth,clip]{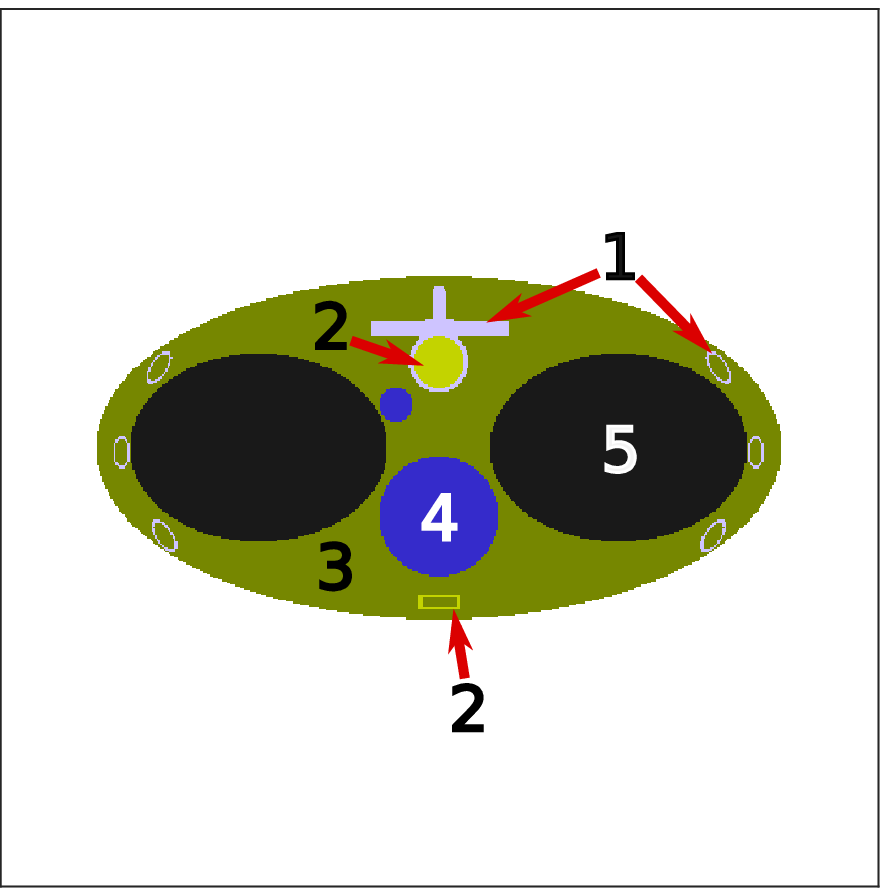} \\
         Mory phantom & Shepp-Logan Phantom & Disks phantom & Thorax phantom \\
    \end{tabular}
    \caption{Visualizations of the numerical phantoms used in the studies.}
    \label{fig:Exp:phantoms}
\end{figure}

A number of phantoms are used in our numerical studies, some of which are standardized while others are custom-made. All these phantoms are shown in Figure~\ref{fig:Exp:phantoms}.

\begin{description}
    \item[Mory phantom] We use a slightly modified version of the phantom provided in the work by Mory et al.~\cite{MorySixou} for comparing new one-step methods against the five one-step approaches addressed in their work. The phantom contains three different materials on a $128 \times 128$ grid. As opposed to the original phantom, each location contains only one material. 
    \item[Shepp-Logan phantom] The standard phantom is commonly used in tomography for benchmarking. We modify this phantom to have five unique grey values. We discretize it on a $512 \times 512$ uniform grid.
    \item[Disks phantom] In this custom phantom, several disks with different materials are placed on a circle. The phantom is created so that we can place up to $15$ different disks on this circle. However, for our numerics, we have taken eight disks and discretized the resulting phantom on a $512 \times 512$ pixel grid.
    \item[Thorax phantom] We use a modified thorax phantom provided in the CONRAD software framework~\cite{MaierHofmann}. We created a thorax phantom of $512^3$ voxels, took slice $z = 255$, and removed a few ribs. The resulting $512 \times 512$ phantom has eight different material candidates, on which we assign five different materials.
\end{description}
\textit{Inverse crime} refers to the process of using the same forward operator for the generation of synthetic measurements as for the subsequent reconstruction process~\cite{JariErkki}. To avoid inverse crime in all of our numerical experiments, we generate measurements by increasing the spatial resolution of the spectral phantoms by a factor of $2$. For example, the Shepp-Logan phantom is discretized on a grid of $1024 \times 1024$ pixels to generate the spectral tomographic measurements. These measurements are, however, acquired on $512$ (equally-spaced) detector pixels for $180$ projection angles in $[0,\pi)$. The spectral tomographic inversion is then performed on a grid of $512 \times 512$.

\subsubsection{Attenuation spectra, source spectrum and selected materials}

To generate the spectral sinograms and construct the dictionary matrix $\mT$, we use attenuation spectra provided by the National Institute for Standards and Technology (NIST)~\cite{NIST, HubbellSeltzer}. We perform linear interpolation to approximate the real spectra and discretize these such that the energetic centers are located at $100$ equidistant values ranging from $5$ to $35$~keV. We use the corresponding interpolated attenuation values as representative attenuation values in the bins. For experiments with the Thorax phantom, we use a spectral range of $[20,80]$~keV. Regarding the experiments with the Mory phantom, we use the attenuation spectra provided by Mory et al. in their implementation~\cite{MoryGithub}, providing 100 equidistant bins with energetic centers between $20$ and $119$~keV. In accordance with their data preprocessing procedure, we have scaled both the material maps and the dictionary entries for iodine and gadolinium by their densities and a value of $0.01$ to obtain a concentration of 10 mg/ml.\\ 

The materials chosen for each phantom (Figure~\ref{fig:Exp:phantoms}) are given in table~\ref{tab:ChosenMaterials}. For the up to $42$ chosen materials included in the dictionary matrix, we refer to Appendix~\ref{sec:Dictionary}. For the Thorax phantom, the dictionary consists of the materials that appear in the phantom. This also holds for the Mory phantom. \\

\begin{figure}[H]
    \centering
	\includegraphics[width=0.45\textwidth]{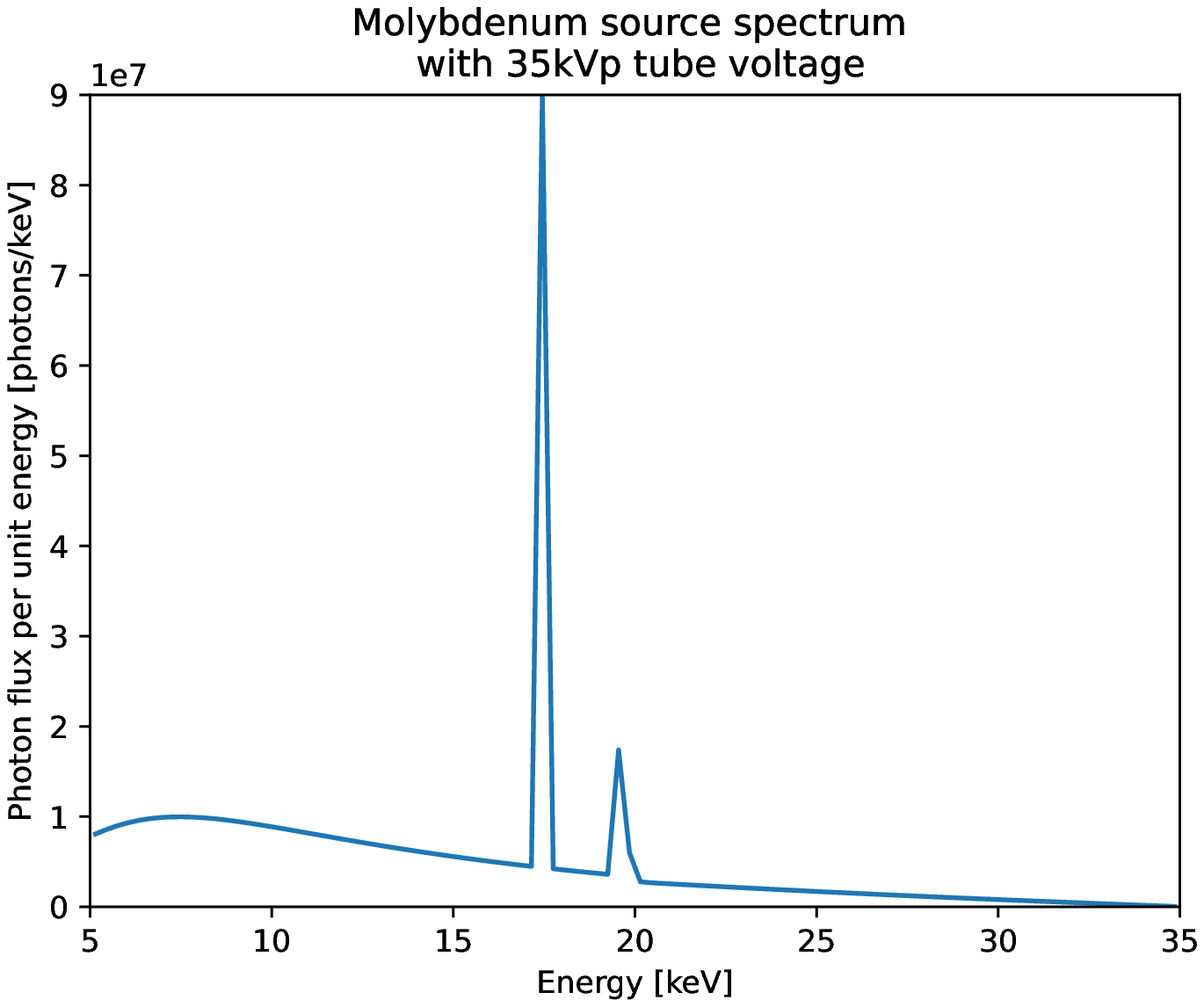}
	\includegraphics[width=0.45\textwidth]{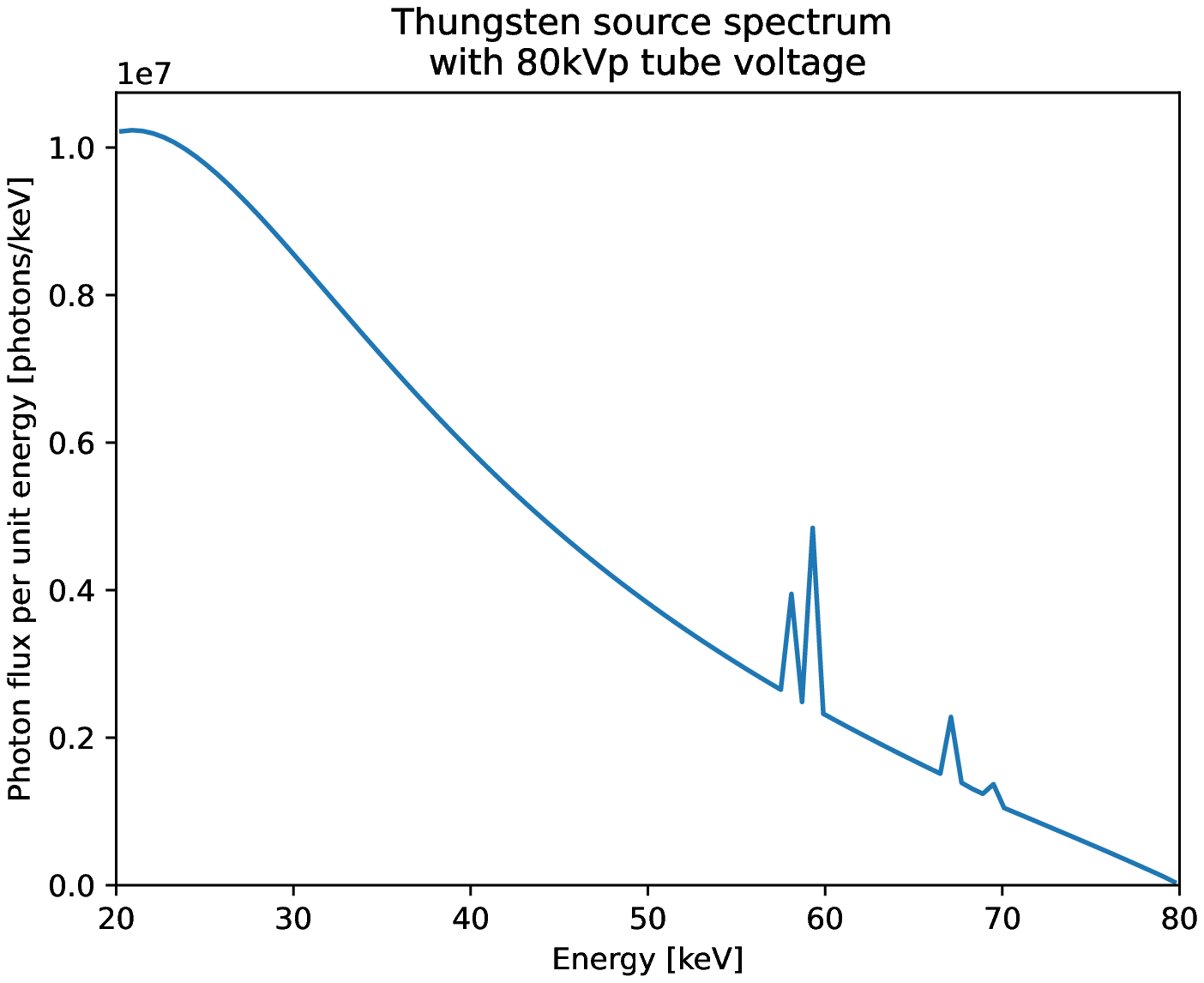}
    \caption{The simulated molybdenum-based and tungsten-based source spectra with respectively peak voltages of 35kVp and 80kVp used in the experiments.}
    \label{fig:Exp:SourceSpectrumoverview}
\end{figure}

To generate the source spectrum, we make use of the SpekPy software~\cite{BujilaOmar, PoludniowskiOmar}. With this, we have simulated an X-ray source spectrum from a molybdenum source with a peak voltage of $35$kV. This source material provides us with low energy bound of $5$~keV for a positive flux, enabling us to include more material absorption edges (mostly K-edges) in our simulations than with other source materials such as tungsten. However, for experiments with the Thorax phantom, we use a tungsten source with a peak voltage of $80$kV. Both source spectra are shown in Figure~\ref{fig:Exp:SourceSpectrumoverview}. For the Mory phantom, we use the source spectrum provided in their implementation~\cite{MoryGithub}, and have changed the matrix for detector response to an identity matrix. \\

\begin{table}[H]
    \centering
    \renewcommand{\arraystretch}{1.2}
    \begin{tabular}{c | cccc}
        \toprule
        & {\bf Shepp-Logan} & {\bf Disks} & {\bf Thorax} & {\bf Mory} \\ \midrule
        1 \quad & Vanadium & Arsenic & Bone & Iodine \\
        2 \quad & Chromium & Selenium & 90\% Blood + 10\% Iodine & Gadolinium \\ 
        3 \quad & Manganese & Bromine & Soft tissue & Water \\
        4 \quad & Iron & Krypton & Blood & - \\
        5 \quad & Cobalt & Rubidium & Lung tissue & - \\
        6 \quad & - & Strontium & - & - \\
        7 \quad & - & Yttrium & - & - \\
        8 \quad & - & Zirconium & - & - \\ \bottomrule
    \end{tabular}
    \caption{Selection of materials for each phantom}
    \label{tab:ChosenMaterials}
\end{table}

\subsubsection{Implementation of the numerical algorithms} \label{sec:NumericalImplementation}
We briefly describe the implementation of various algorithms used in our studies below. For the existing one-step methods, we keep a similar naming convention as in the work of Mory et al.~\cite{MorySixou} by using the last name of the first author of the associated paper. We use the ASTRA toolbox~\cite{AarlePalenstijn2,AarlePalenstijn} for the simulation of the X-ray projections and implementation of forward and the adjoint operator of tomography~\cite{BleichrodtLeeuwen}. For compatibility with the implementations of the existing methods in~\cite{MoryGithub}, we use the AIR Toolbox~\cite{HansenSaxild} for the computation of the forward operator and the adjoint instead of ASTRA whenever the Mory phantom is considered. For the last five listed existing one-step methods, we use the implementation and parameter values provided in~\cite{MoryGithub}. Table~\ref{tab:PriorInfo} summarizes the prior information used in these methods.

\begin{table}[H]
    \centering
    \renewcommand{\arraystretch}{1.2}
    \begin{tabular}{|c | c| c|}
        \toprule
        & {\bf Spatial prior information} & {\bf Spectral prior information}\\ \midrule
        RU \quad & Non-negativity  & Non-negativity \\
        UR \quad & Non-negativity  & Non-negativity  \\
        cJoint \quad & Non-negativity & Non-negativity \\
        Cai \quad & (Modified) TV & Spectral signatures of present materials\\
        Long \quad & Simplex and (modified) TV & Spectral signatures of present materials\\ 
        Weidinger \quad & (Weak) non-negativity and (modified) TV & Spectral signatures of present materials\\ 
        Mechlem \quad & (Modified) TV & Spectral signatures of present materials\\ 
        Barber \quad & Constrained TV & Spectral signatures of present materials \\ 
        ADJUST \quad & Simplex & Spectral dictionary\\ \bottomrule
    \end{tabular}
    \caption{Spatial and spectral information for each method. TV stands for Total-Variation regularization \cite{RudinOsher}. Note that each method has prior information on the number of materials. The spectral dictionary contains signatures of a large superset of present materials.}
    \label{tab:PriorInfo}
\end{table}

\begin{description}
    \item[RU] For the reconstruction, we solve the Tikhonov-regularized optimization problem with regularization parameter $\lambda$ set to $10^{-3}$. We perform a maximum of $20$ conjugate-gradient on the resulting normal equations with the tolerance of $10^{-6}$~\cite{HestenesStiefel}. We use non-negative matrix factorization (NMF) with an alternating least-squares algorithm~\cite{BerryBrowne} for $100$ maximum iterations in the decomposition step. Since NMF is a non-convex problem, we use ten different initializations to determine the solution. 
    \item[UR] We use the same settings for the decomposition and the reconstruction steps as described in RU.
    \item[cJoint] We solve the problem using an alternating minimization scheme. The maximum number of iterations is set to 2000 with tolerance, defined as the relative residual, of $10^{-4}$. In each iterate, we solve the minimization with a spectral projected gradient scheme~\cite{SchmidtBerg}.
    \item[ADJUST] We use the AAPM scheme described in~\eqref{eq:ADJUST-optscheme} to find the solution. For all the experiments, we choose $\rho$ value of $10^{-2}$ and set $\epsilon^{\text{abs}}_{k}, \epsilon^{\text{rel}}_k$ to $10^{-4}$ and $10^{-6}$, respectively. We run AAPM for maximum $1000$ iterations. 
    \item[Cai] This Bayesian reconstruction approach solves a minimization problem with a non-quadratic cost function using a monotone conjugate gradient algorithm with heuristic descent steps~\cite{CaiRodet}. We perform $5000$ iterations to find the solution.
    \item[Long] This is a regularized approach that uses Separable Quadratic Surrogates to minimize Kullback-Leibler cost function with edge-preserving regularization~\cite{LongFessler}. We run this method with $5000$ iterations.
    \item[Weidinger] This approach is similar to Long with modification in the approximation of regularization function by the Green potential, and leaving out Ordered Subsets that Long uses to speed up convergence~\cite{WeidingerBuzug}. We run the algorithm for $5000$ iterations.
    \item[Mechlem] This approach builds upon Weidinger while replacing the regularization using the Huber function. However, it uses Ordered Subsets and Nesterov acceleration to find the solution~\cite{MechlemEhn}. We run the algorithm for $200$ iterations.
    \item[Barber] This approach solves a constrained optimization problem using a primal-dual algorithm where constraints are composition of total-variation and simplex~\cite{BarberSidky}. We run a maximum of $10000$ primal-dual iterations. 
\end{description}


\subsection{Comparison of ADJUST with other methods}

\subsubsection{Results on the Mory phantom}

\begin{figure}[!htb]
    \centering
    \begin{small}
    \renewcommand{\arraystretch}{0}
    \begin{tabular}{r P{12em} P{12em} P{12em}}
        & {\bf Iodine} & {\bf Gadolinium} & {\bf Water} \\
        \centered{\bf GT} 
        & 
        \centered{\begin{tikzpicture}[spy using outlines={circle,red,magnification=2.5,size=1.2cm, connect spies}] \node {\includegraphics[width=0.115\textwidth,clip]{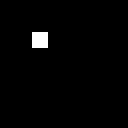}}; \spy on (-0.395,0.395) in node [] at (1.8,0); \end{tikzpicture}}
        & 
        \centered{ \begin{tikzpicture}[spy using outlines={circle,red,magnification=2.5,size=1.2cm, connect spies}] \node {\includegraphics[width=0.115\textwidth,clip]{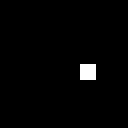}}; \spy on (0.38,-0.135) in node [] at (1.8,0); \end{tikzpicture}}
        &
        \centered{ \begin{tikzpicture}[spy using outlines={circle,red,magnification=2.5,size=1.2cm, connect spies}] \node {\includegraphics[width=0.115\textwidth,clip]{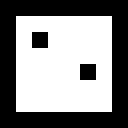}}; \spy on (-0.395,0.395) in node [] at (1.8,0); \end{tikzpicture}} \\
        \centered{\bf RU} 
        & 
        \centered{\begin{tikzpicture}[spy using outlines={circle,red,magnification=2.5,size=1.2cm, connect spies}] \node {\includegraphics[width=0.115\textwidth,clip]{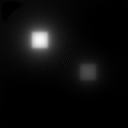}}; \spy on (-0.395,0.395) in node [] at (1.8,0); \end{tikzpicture}}
        & 
        \centered{ \begin{tikzpicture}[spy using outlines={circle,red,magnification=2.5,size=1.2cm, connect spies}] \node {\includegraphics[width=0.115\textwidth,clip]{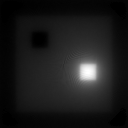}}; \spy on (0.38,-0.135) in node [] at (1.8,0); \end{tikzpicture}}
        &
        \centered{ \begin{tikzpicture}[spy using outlines={circle,red,magnification=2.5,size=1.2cm, connect spies}] \node {\includegraphics[width=0.115\textwidth,clip]{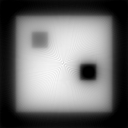}}; \spy on (-0.395,0.395) in node [] at (1.8,0); \end{tikzpicture}} \\
        \centered{\bf UR} 
        & 
        \centered{\begin{tikzpicture}[spy using outlines={circle,red,magnification=2.5,size=1.2cm, connect spies}] \node {\includegraphics[width=0.115\textwidth,clip]{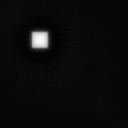}}; \spy on (-0.395,0.395) in node [] at (1.8,0); \end{tikzpicture}}
        & 
        \centered{ \begin{tikzpicture}[spy using outlines={circle,red,magnification=2.5,size=1.2cm, connect spies}] \node {\includegraphics[width=0.115\textwidth,clip]{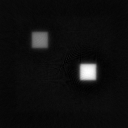}}; \spy on (0.38,-0.135) in node [] at (1.8,0); \end{tikzpicture}}
        &
        \centered{ \begin{tikzpicture}[spy using outlines={circle,red,magnification=2.5,size=1.2cm, connect spies}] \node {\includegraphics[width=0.115\textwidth,clip]{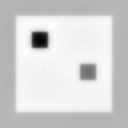}}; \spy on (-0.395,0.395) in node [] at (1.8,0); \end{tikzpicture}} \\
        \centered{\bf cJoint} 
        & 
        \centered{\begin{tikzpicture}[spy using outlines={circle,red,magnification=2.5,size=1.2cm, connect spies}] \node {\includegraphics[width=0.115\textwidth,clip]{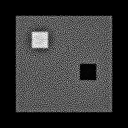}}; \spy on (-0.395,0.395) in node [] at (1.8,0); \end{tikzpicture}}
        & 
        \centered{ \begin{tikzpicture}[spy using outlines={circle,red,magnification=2.5,size=1.2cm, connect spies}] \node {\includegraphics[width=0.115\textwidth,clip]{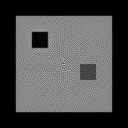}}; \spy on (0.38,-0.135) in node [] at (1.8,0); \end{tikzpicture}}
        &
        \centered{ \begin{tikzpicture}[spy using outlines={circle,red,magnification=2.5,size=1.2cm, connect spies}] \node {\includegraphics[width=0.12\textwidth,clip]{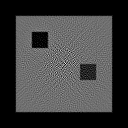}}; \spy on (-0.395,0.395) in node [] at (1.8,0); \end{tikzpicture}} \\
        \centered{\bf Cai} 
        & 
        \centered{\begin{tikzpicture}[spy using outlines={circle,red,magnification=2.5,size=1.2cm, connect spies}] \node {\includegraphics[width=0.115\textwidth,clip]{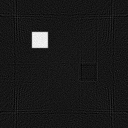}}; \spy on (-0.395,0.395) in node [] at (1.8,0); \end{tikzpicture}}
        & 
        \centered{ \begin{tikzpicture}[spy using outlines={circle,red,magnification=2.5,size=1.2cm, connect spies}] \node {\includegraphics[width=0.115\textwidth,clip]{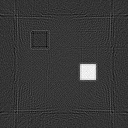}}; \spy on (0.38,-0.135) in node [] at (1.8,0); \end{tikzpicture}}
        &
        \centered{ \begin{tikzpicture}[spy using outlines={circle,red,magnification=2.5,size=1.2cm, connect spies}] \node {\includegraphics[width=0.115\textwidth,clip]{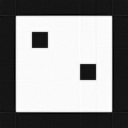}}; \spy on (-0.395,0.395) in node [] at (1.8,0); \end{tikzpicture}} \\
        \centered{\bf Weidinger} 
        & 
        \centered{\begin{tikzpicture}[spy using outlines={circle,red,magnification=2.5,size=1.2cm, connect spies}] \node {\includegraphics[width=0.115\textwidth,clip]{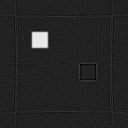}}; \spy on (-0.395,0.395) in node [] at (1.8,0); \end{tikzpicture}}
        & 
        \centered{ \begin{tikzpicture}[spy using outlines={circle,red,magnification=2.5,size=1.2cm, connect spies}] \node {\includegraphics[width=0.115\textwidth,clip]{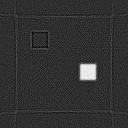}}; \spy on (0.38,-0.135) in node [] at (1.8,0); \end{tikzpicture}}
        &
        \centered{ \begin{tikzpicture}[spy using outlines={circle,red,magnification=2.5,size=1.2cm, connect spies}] \node {\includegraphics[width=0.115\textwidth,clip]{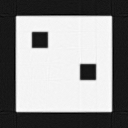}}; \spy on (-0.395,0.395) in node [] at (1.8,0); \end{tikzpicture}} \\
        \centered{\bf Long} 
        & 
        \centered{\begin{tikzpicture}[spy using outlines={circle,red,magnification=2.5,size=1.2cm, connect spies}] \node {\includegraphics[width=0.115\textwidth,clip]{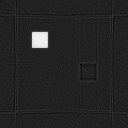}}; \spy on (-0.395,0.395) in node [] at (1.8,0); \end{tikzpicture}}
        & 
        \centered{ \begin{tikzpicture}[spy using outlines={circle,red,magnification=2.5,size=1.2cm, connect spies}] \node {\includegraphics[width=0.115\textwidth,clip]{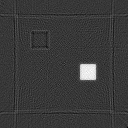}}; \spy on (0.38,-0.135) in node [] at (1.8,0); \end{tikzpicture}}
        &
        \centered{ \begin{tikzpicture}[spy using outlines={circle,red,magnification=2.5,size=1.2cm, connect spies}] \node {\includegraphics[width=0.115\textwidth,clip]{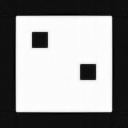}}; \spy on (-0.395,0.395) in node [] at (1.8,0); \end{tikzpicture}} \\
        \centered{\bf Mechlem} 
        & 
        \centered{\begin{tikzpicture}[spy using outlines={circle,red,magnification=2.5,size=1.2cm, connect spies}] \node {\includegraphics[width=0.115\textwidth,clip]{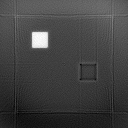}}; \spy on (-0.395,0.395) in node [] at (1.8,0); \end{tikzpicture}}
        & 
        \centered{ \begin{tikzpicture}[spy using outlines={circle,red,magnification=2.5,size=1.2cm, connect spies}] \node {\includegraphics[width=0.115\textwidth,clip]{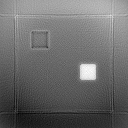}}; \spy on (0.38,-0.135) in node [] at (1.8,0); \end{tikzpicture}}
        &
        \centered{ \begin{tikzpicture}[spy using outlines={circle,red,magnification=2.5,size=1.2cm, connect spies}] \node {\includegraphics[width=0.115\textwidth,clip]{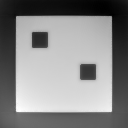}}; \spy on (-0.395,0.395) in node [] at (1.8,0); \end{tikzpicture}} \\
        \centered{\bf Barber} 
        & 
        \centered{\begin{tikzpicture}[spy using outlines={circle,red,magnification=2.5,size=1.2cm, connect spies}] \node {\includegraphics[width=0.115\textwidth,clip]{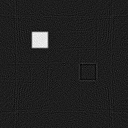}}; \spy on (-0.395,0.395) in node [] at (1.8,0); \end{tikzpicture}}
        & 
        \centered{ \begin{tikzpicture}[spy using outlines={circle,red,magnification=2.5,size=1.2cm, connect spies}] \node {\includegraphics[width=0.115\textwidth,clip]{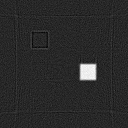}}; \spy on (0.38,-0.135) in node [] at (1.8,0); \end{tikzpicture}}
        &
        \centered{ \begin{tikzpicture}[spy using outlines={circle,red,magnification=2.5,size=1.2cm, connect spies}] \node {\includegraphics[width=0.115\textwidth,clip]{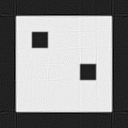}}; \spy on (-0.395,0.395) in node [] at (1.8,0); \end{tikzpicture}} \\
        \centered{\bf ADJUST} 
        & 
        \centered{\begin{tikzpicture}[spy using outlines={circle,red,magnification=2.5,size=1.2cm, connect spies}] \node {\includegraphics[width=0.115\textwidth,clip]{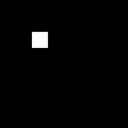}}; \spy on (-0.395,0.395) in node [] at (1.8,0); \end{tikzpicture}}
        & 
        \centered{ \begin{tikzpicture}[spy using outlines={circle,red,magnification=2.5,size=1.2cm, connect spies}] \node {\includegraphics[width=0.115\textwidth,clip]{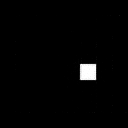}}; \spy on (0.38,-0.135) in node [] at (1.8,0); \end{tikzpicture}}
        &
        \centered{ \begin{tikzpicture}[spy using outlines={circle,red,magnification=2.5,size=1.2cm, connect spies}] \node {\includegraphics[width=0.115\textwidth,clip]{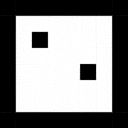}}; \spy on (-0.395,0.395) in node [] at (1.8,0); \end{tikzpicture}} \\
    \end{tabular}
    \end{small}
    \caption{Visual comparison of various methods on Mory phantom. GT refers to ground truth.}
    \label{fig:Exp:Mory}
\end{figure}

To compare the ADJUST method with the other approaches listed in Section~\ref{sec:NumericalImplementation}, we perform numerical studies on the Mory phantom. We take projections from $363$ equidistant angles in $[0, \pi)$ using $181$ detectors. Moreover, we apply Poisson noise to the resulting projections, with the incident photons being proportional to the source spectrum. We show the reconstruction results in Figure~\ref{fig:Exp:Mory}, and tabulate the performance measures in Table~\ref{tab:Exp:Mory}. We use Mean Squared Error (MSE), Peak Signal-to-Noise Ratio (PSNR), and Structural Similarity Index Measure (SSIM) to assess the reconstruction results with respect to the ground truth phantom (details of the measures are given in Appendix~\ref{sec:PerformanceMeasuresExt}). We list the results as the measures averaged over all material maps. We observe that ADJUST obtains the best values for PSNR, SSIM and MSE. 

The two-step methods (RU and UR) find the iodine and the gadolinium locations, but the results appear to be spatially smeared out, which is also reflected in the relatively low values of PSNR and MSE. However, the UR method still recovers the shapes reasonably well, as shown in the high value for the SSIM. The cJoint method finds the water location, but fails to recover the locations of iodine and gadolinium. The state-of-the-art one-step methods overall perform better than UR, RU, and cJoint. However, these methods gives rise to edge artefacts (e.g. water edges partially appearing in iodine or gadolinium maps, and blurring near the edges), resulting in suboptimal PSNR and SSIM values. These artefacts hardly appear in ADJUST reconstructions, resulting in significantly higher values of the PSNR and SSIM.  

RU, UR and cJoint method do not incorporate any prior information about the object other than non-negativity constraints. Hence, these methods suffer to find the optimal solution. However, one-step methods strongly assume the knowledge of materials present in the object. Although they perform better than RU, UR and cJoint, they can not be applied when the composition of object is not known. In contrast, ADJUST does not know about the material composition of the object. It solely relies on the number of materials present in the object (this number can be estimated through trial and error). 

\begin{table}[!htb]
    \centering
    \renewcommand{\arraystretch}{1.2}
    \begin{tabular}{r|c|c|c|c|c|c|c|c|c}
        \toprule
        & {\bf RU} & {\bf UR} & {\bf cJoint} & {\bf Cai} & {\bf Weidinger} & {\bf Long} & {\bf Mechlem} & {\bf Barber} & {\bf ADJUST}\\ \midrule
        MSE & 0.0267 & 0.0790 & 0.0860 & 0.0115 & 0.0117 & 0.0112 & 0.0406 & 0.0115 & {\bf 0.0103} \\
        PSNR & 20.35 & 16.44 & 16.03 & 19.54 & 19.33 & 19.62 & 14.45 & 19.41 & {\bf 21.94} \\
        SSIM & 0.4413 & 0.6720 & 0.4119 & 0.6034 & 0.6131 & 0.6187 & 0.2159 & 0.6460 & {\bf 0.9616} \\ 
        \bottomrule
    \end{tabular}
    \caption{Reconstruction error in terms of Mean Squared Error (MSE), Peak Signal to Noise Ratio (PSNR) and Structural Similarity Index Measure (SSIM) of various methods for the Mory phantom.}
    \label{tab:Exp:Mory}
\end{table}

\subsubsection{Results on the remaining phantoms}
\begin{figure}
    \centering
    \begin{small}
    \renewcommand{\arraystretch}{1.2}
    \begin{tabular}{c cccc}
        {\bf GT} & {\bf RU} & {\bf UR} & {\bf cJoint} & {\bf ADJUST} \\
        \includegraphics[width=0.17\textwidth,cfbox=col1 1pt 0pt]{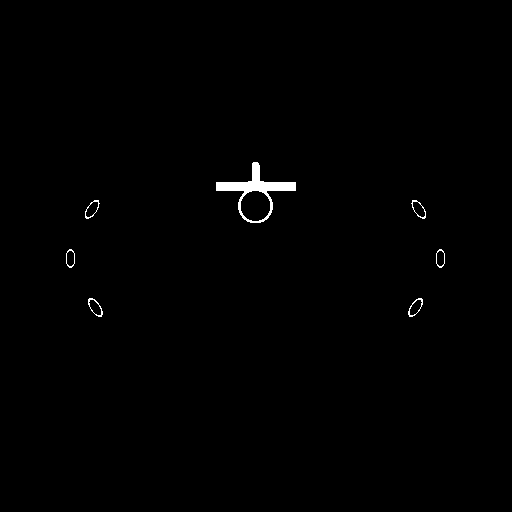} \quad & \includegraphics[width=0.17\textwidth,cfbox=col1 1pt 0pt]{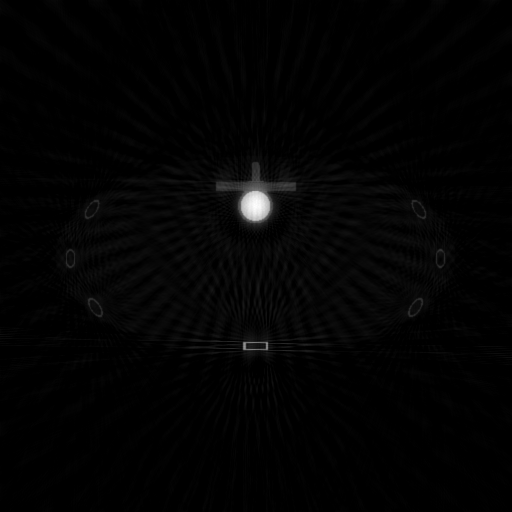} & \includegraphics[width=0.17\textwidth,cfbox=col1 1pt 0pt]{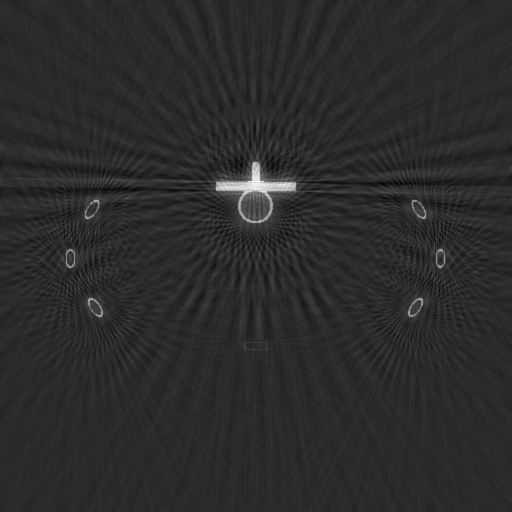} & \includegraphics[width=0.17\textwidth,cfbox=col1 1pt 0pt]{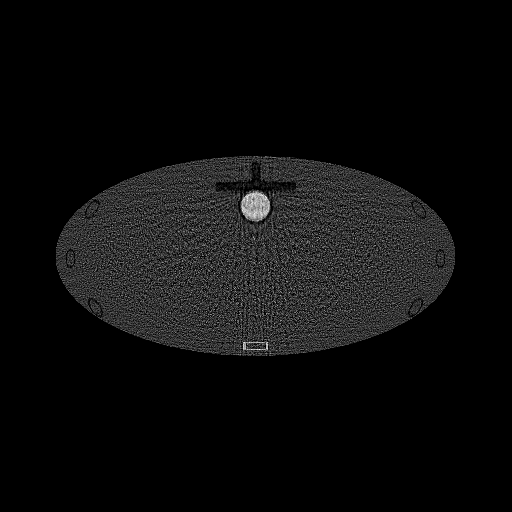}& \includegraphics[width=0.17\textwidth,cfbox=col1 1pt 0pt]{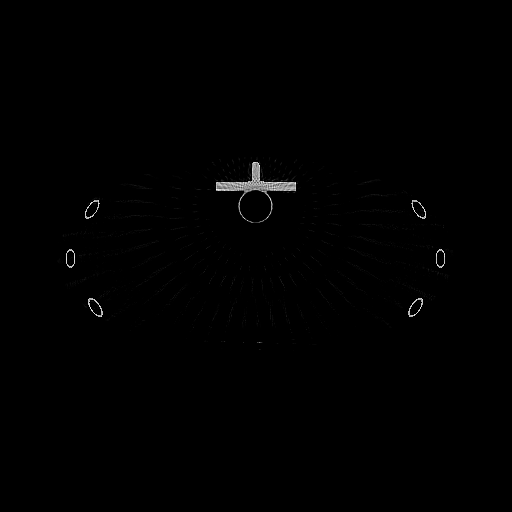} \\
        \includegraphics[width=0.17\textwidth,cfbox=col2 1pt 0pt]{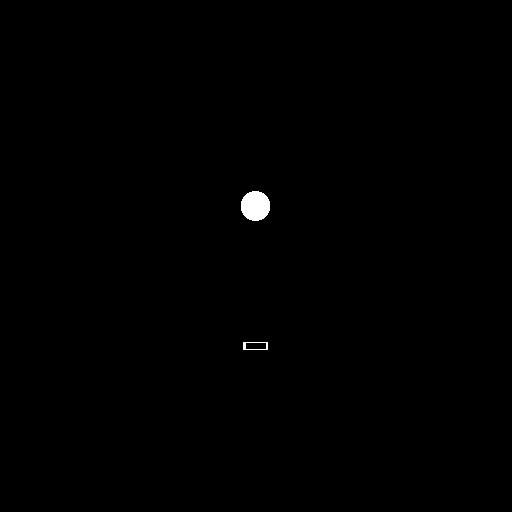} \quad & \includegraphics[width=0.17\textwidth,cfbox=col2 1pt 0pt]{Thorax_RU_2.png} & \includegraphics[width=0.17\textwidth,cfbox=col2 1pt 0pt]{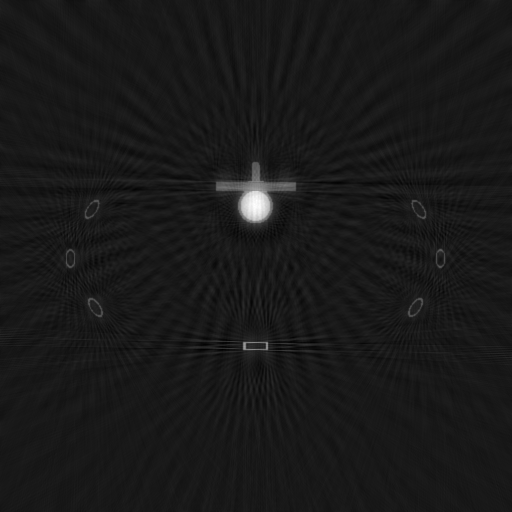} & \includegraphics[width=0.17\textwidth,cfbox=col2 1pt 0pt]{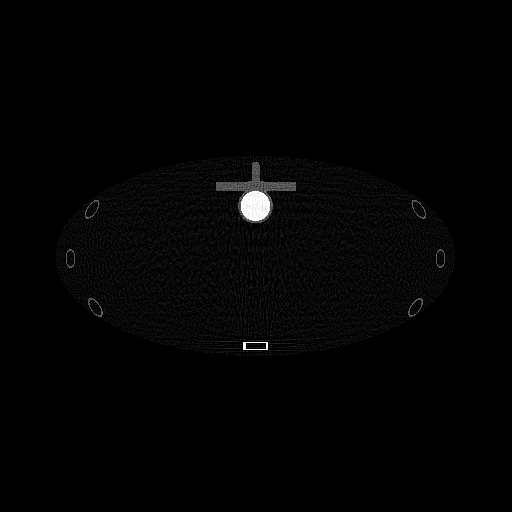} & \includegraphics[width=0.17\textwidth,cfbox=col2 1pt 0pt]{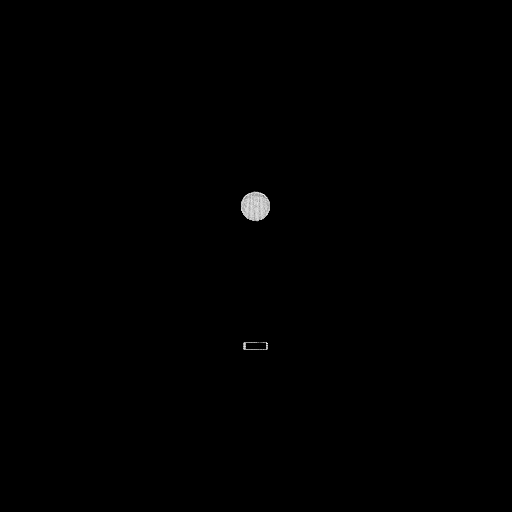} \\
        \includegraphics[width=0.17\textwidth,cfbox=col3 1pt 0pt]{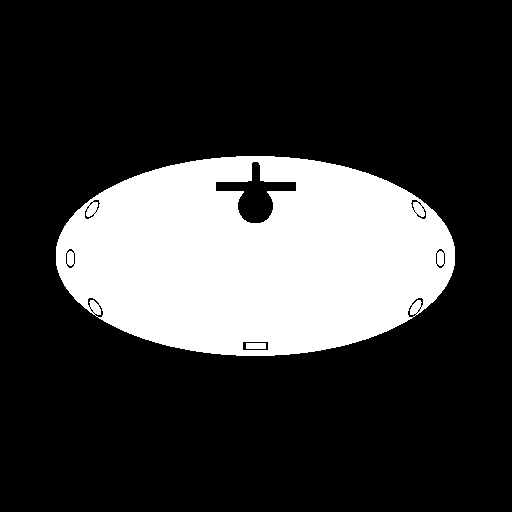} \quad & \includegraphics[width=0.17\textwidth,cfbox=col3 1pt 0pt]{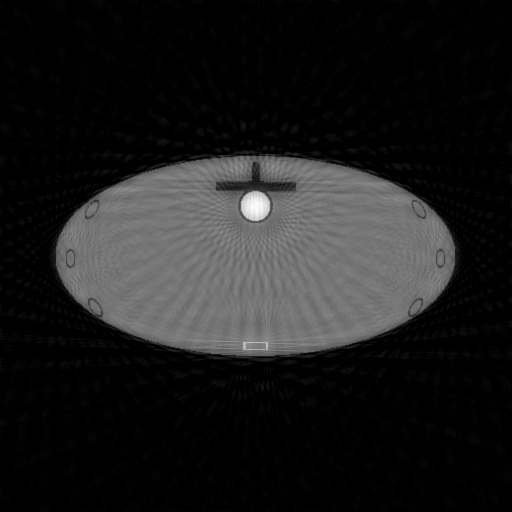} & \includegraphics[width=0.17\textwidth,cfbox=col3 1pt 0pt]{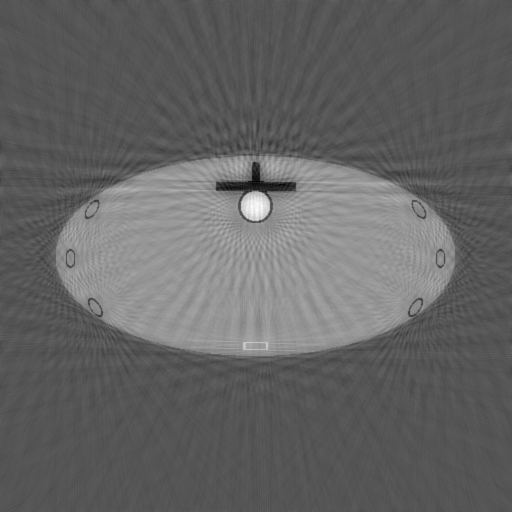} & \includegraphics[width=0.17\textwidth,cfbox=col3 1pt 0pt]{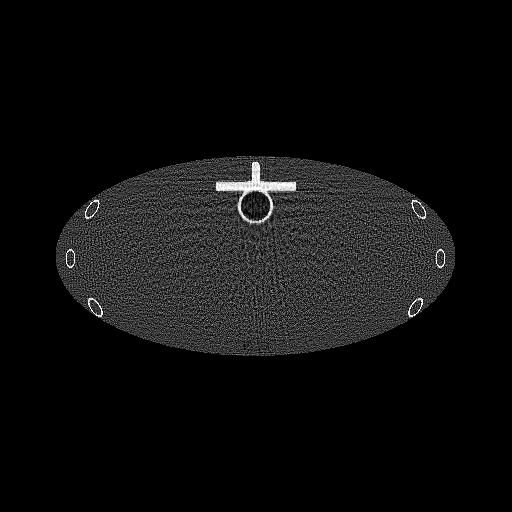} & \includegraphics[width=0.17\textwidth,cfbox=col3 1pt 0pt]{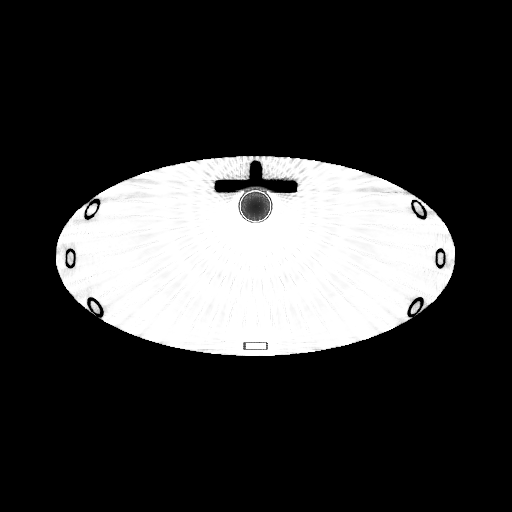} \\
        \includegraphics[width=0.17\textwidth,clip]{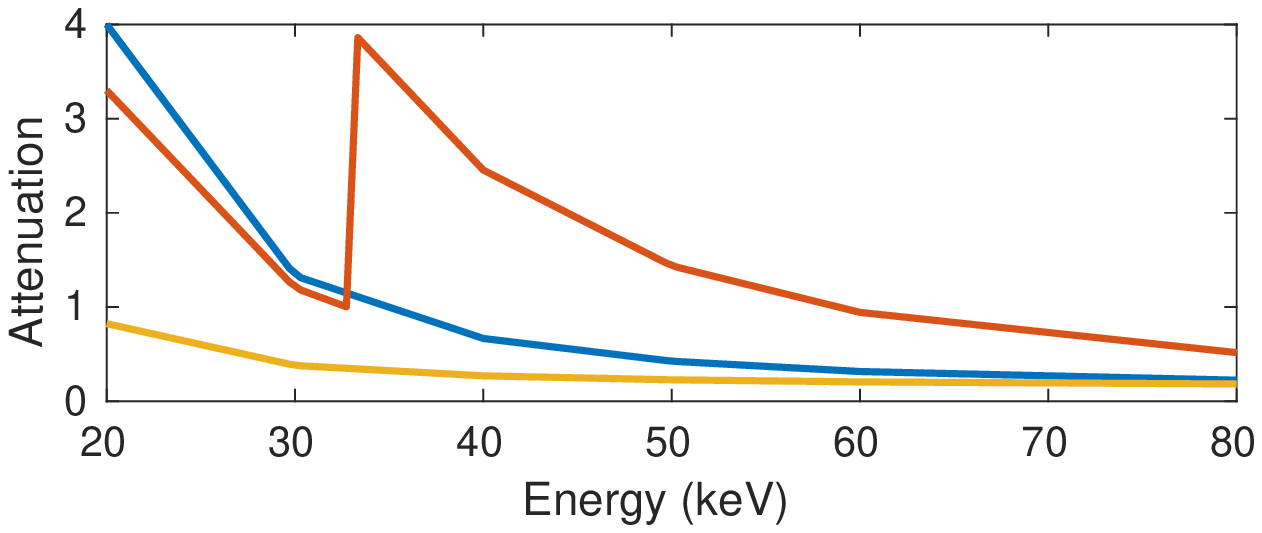} \quad & \includegraphics[width=0.17\textwidth,clip]{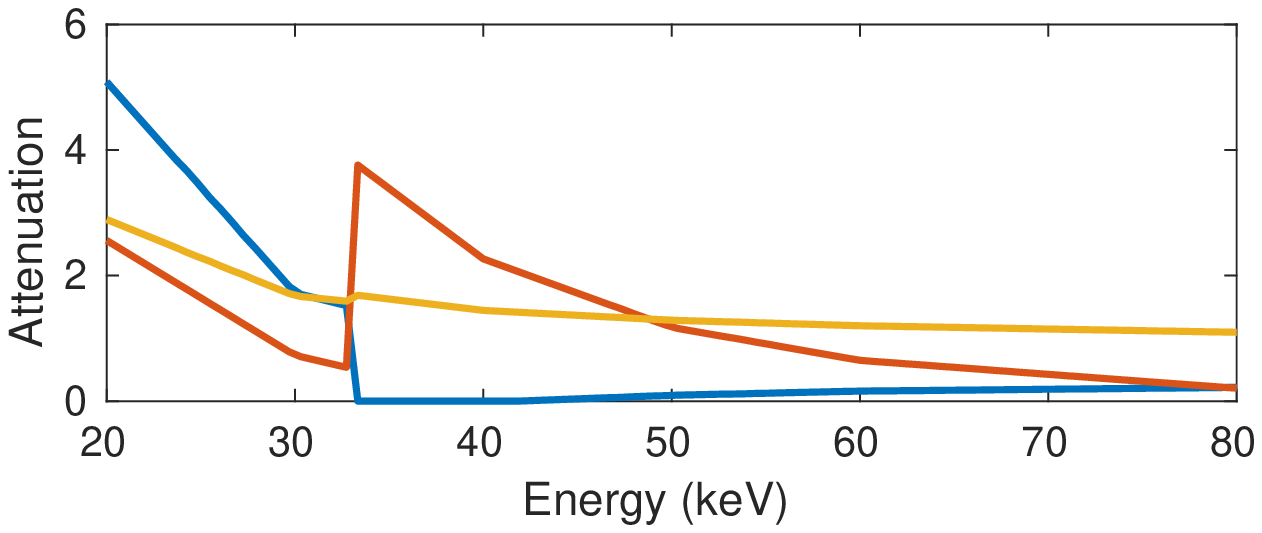} & \includegraphics[width=0.17\textwidth,clip]{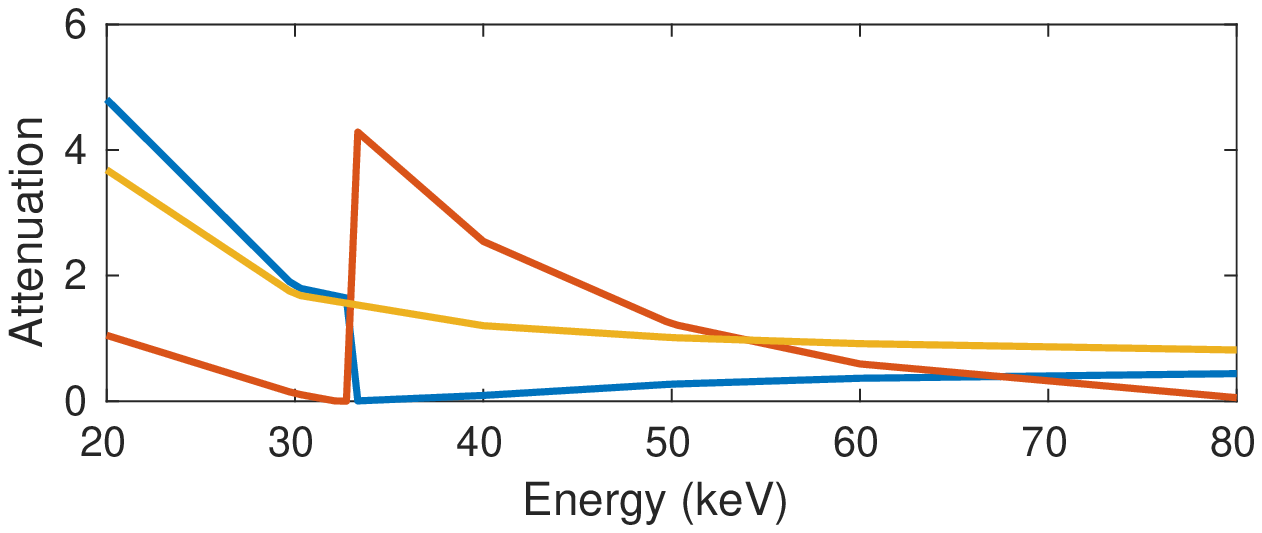} & \includegraphics[width=0.17\textwidth,clip]{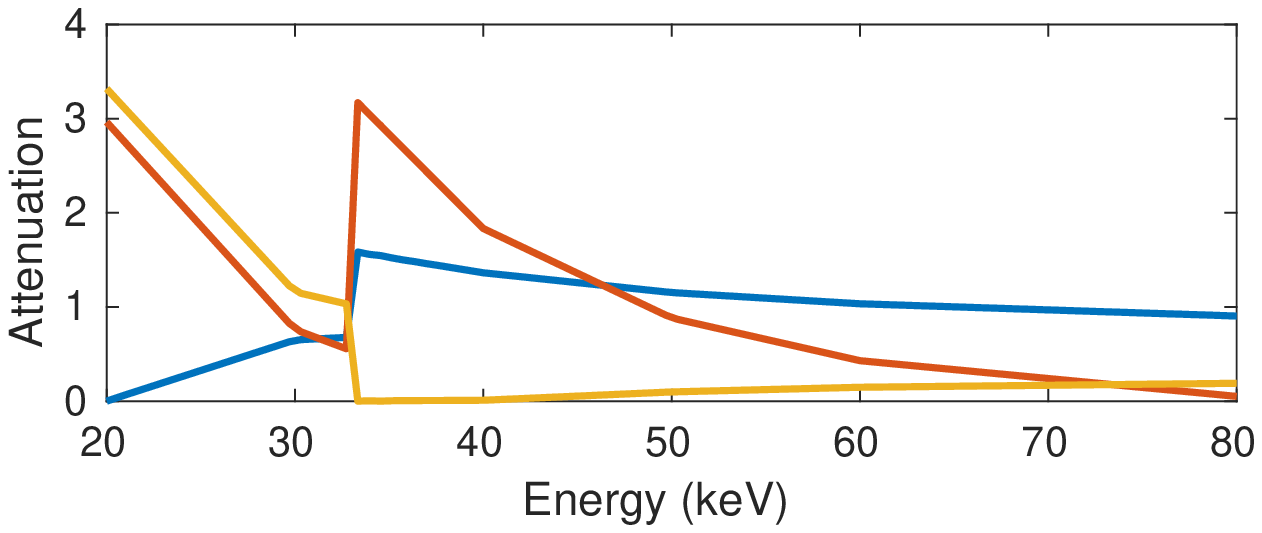} & \includegraphics[width=0.17\textwidth,clip]{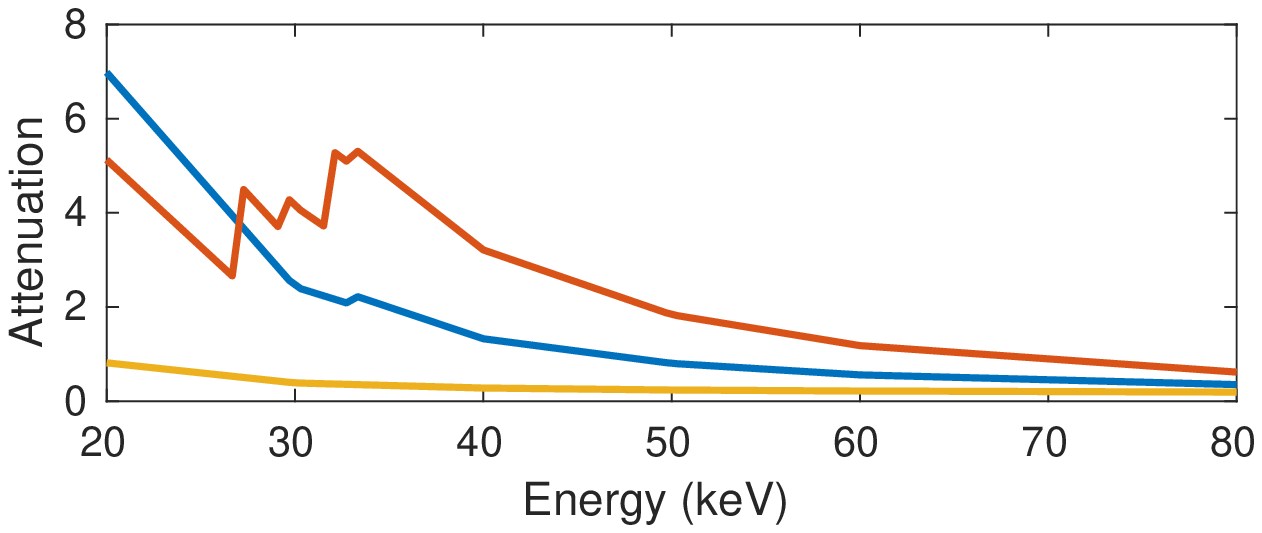} \\ 
    \end{tabular}
    \end{small}
    \caption{Visual comparison of ADJUST with RU, UR, and cJoint method on the Thorax phantom. The top row shows the ground truth and reconstructed material maps and spectral signatures for bone, the second row shows these for iodine and the third row shows (averages of) these for soft materials. We match the colors of the bounding box for material maps with the (recovered) spectral signatures shown on the bottom row.}
    \label{fig:Exp:HardSoft}
\end{figure}

We compare the proposed method ADJUST with RU, UR, and Joint method on the Shepp-Logan, Disks and Thorax phantoms. Comparison with the other methods is not possible as they are designed for handling only a limited number of materials. The first two phantoms consist of only hard materials, and hence K-edges are present in the spectra. All these phantoms are more advanced compared to the Mory phantom. The Shepp-Logan and Thorax phantoms are structurally more complicated. On other hand, the Disks phantom contains up to eight hard materials. For the Thorax phantom, we also include soft materials. Nevertheless, we aim to reconstruct the bone, the iodine-blood mixture and the remaining soft materials into three separate classes. For all three phantoms, we measure tomographic projections for $180$ equidistant angles between $[0,\pi)$. These measurements consist of Poisson noise that is proportional to the incoming photons on the detector. We tabulate the measures on the solutions produced by the RU, UR, cJoint and ADJUST algorithms for the three phantoms in Table~\ref{tab:Exp:comparison}. Moreover, for the Thorax phantom, we show the reconstructed material maps and the recovered spectra in Figure~\ref{fig:Exp:HardSoft}.

We see that for all measures, ADJUST outperforms the other three methods. The results for the other methods are similar to each other. In general, the Disks phantom leads to very good similarity measures for ADJUST, indicating the capability of dealing with eight materials. For the other two phantoms, where the proportions of materials are different, the measures are slightly worse, but we observed that the reconstructions for the Shepp-Logan and Disks phantoms are visually satisfactory (the visual results for the Disks phantom are given in Appendix~\ref{sec:NumStudComparison}, while visual results for the Shepp-Logan phantom are given in Figure~\ref{fig:Motivation:comparison}). However, for the Thorax phantom we see a striped pattern in the material map of the soft tissues. A possible reason is that the spectral signature of the combined tissue materials may not be present in the dictionary, and therefore produces visually suboptimal results. On the other hand, it can also be observed that the spectrum of bone is not fully correctly recovered. This may be because it does not have a discontinuity in the chosen spectral range, and is therefore too similar in shape to the tissue spectra. Despite this, ADJUST outperforms the other three methods on all phantoms.

\begin{table}
    \centering
    \renewcommand{\arraystretch}{1.2}
    \begin{tabular}{c|c|P{5em}|P{5em}|P{5em}|P{5em}}
        \toprule
        {\bf Phantom} & & {\bf RU} & {\bf UR} & {\bf cJoint} & {\bf ADJUST} \\ \midrule
        \multirow{3}*{\bf Shepp-Logan} & MSE & 0.0711 & 0.0598 & 0.0548 & {\bf 0.0061} \\
        & PSNR & 16.41 & 16.66 & 13.74 & {\bf 23.12} \\
        & SSIM & 0.2433 & 0.4497 & 0.1077 & {\bf 0.9599} \\ \midrule
        \multirow{3}*{\bf Disks} & MSE & 0.0125 & 0.0082 & 0.0063 & {\bf 0.0030} \\ 
        & PSNR & 19.50 & 21.22 & 23.72 & {\bf 33.32} \\
        & SSIM & 0.8970 & 0.8975 & 0.8882 & {\bf 0.9925} \\ \midrule
        \multirow{3}*{\bf Thorax} & MSE & 0.0587 & 0.0622 & 0.0525 & {\bf 0.0020} \\ 
        & PSNR & 19.92 & 20.96 & 19.11 & {\bf 36.68} \\
        & SSIM & 0.6902 & 0.6094 & 0.7628 & {\bf 0.9198} \\ \bottomrule
    \end{tabular}
    \caption{Reconstruction error in terms of mean-squared error (MSE), peak signal-to-noise ratio (PSNR) and structural similarity index measure (SSIM) of various methods for various phantoms.}
    \label{tab:Exp:comparison}
\end{table}

\subsection{Limited Measurement Patterns}
Through the numerical experiments in this section, we demonstrate that ADJUST is robust. We consider three scenarios: (\emph{i})~\textit{Sparse-angle tomography}, where the number of measurements is reduced by sampling fewer projections angles, (\emph{ii})~\textit{Limited-view tomography}, where measurements from a particular range of angles are missing (representing the case of hardware limitations), and (\emph{iii})~\textit{Sparse channels}, where the number of spectral bins of the detector is limited. We apply these settings to the Shepp-Logan phantom and the Disks phantom, and report the results in Table~\ref{tab:Exp:comparison_corrupteddata}.\\

For the sparse-angle tomography setup, we consider tomographic projections from $10$ equidistant angles in the range of $0$ to $\pi$. For the Shepp-Logan phantom, the spectral signatures are determined correctly and we observed very minor artefacts. For the Disks phantom, the material maps and the spectral signatures are precisely reconstructed, as reflected in the very low MSE and very high PSNR and SSIM values. Therefore, for these phantoms and the selected angles, ADJUST performs well.\\

For the limited-view tomography setup, we limit the projection angle range from $0$ to $2\pi/3$. Restricting the angle range for the projections results in a well-known missing-wedge artefact. We take projections for 60 equidistant projection angles and add Poisson noise. For both phantoms, we observed no missing wedge artefacts when reconstructed with ADJUST. The PSNR and SSIM measures remain very high and the MSE measure remains low. We conclude that for these phantoms, ADJUST can deal well limited-view measurements. \\

In the sparse channel setting, we reduce the number of spectral bins based on the spectral dictionary. Since we consider a dictionary of 42 hard materials, we reduce the spectral channels from 100 to 42. These 42 spectral channels are chosen based on the independent columns of the spectral dictionary. For tomography, we choose 60 equidistant angles between $0$ and $\pi$. We observed that the Shepp-Logan phantom has been reconstructed precisely, but a few artefacts are visible on the edges of the disks for the Disks phantom. These artefacts are reflected in the slightly higher MSE and slightly lower PSNR and SSIM. For the Shepp-Logan phantom there is no obvious decrease of quality in terms of the measurements. So ADJUST appears to be capable of dealing with a sparse channel setting with the given phantoms and the spectral setup. \\

\begin{table}[!htb]
    \centering
    \renewcommand{\arraystretch}{1.2}
    \begin{tabular}{c|c|P{5em}|P{5em}|P{5em}|P{5em}}
        \toprule
        {\bf Phantom} & & {\bf Full sampling} & {\bf Sparse-angle} & {\bf Limited-view} & {\bf Sparse channels} \\ \midrule 
        \multirow{3}*{\bf Shepp-Logan} & MSE & 0.0061 & 0.0113 & 0.0330 & 0.0066 \\ 
        & PSNR & 23.12 & 20.27 & 18.41 & 23.04 \\ 
        & SSIM & 0.9599 & 0.9435 & 0.9112 & 0.9670 \\ 
        \midrule
        \multirow{3}*{\bf Disks} & MSE & 0.0030 & 0.0028 & 0.0057 & 0.0002 \\ 
        & PSNR & 33.32 & 33.05 & 26.31 & 36.76\\ 
        & SSIM & 0.9925 & 0.9924 & 0.9807 & 0.9929 \\ 
        \bottomrule
    \end{tabular}
    \caption{Reconstruction error in terms of MSE, PSNR, SSIM with ADJUST for various phantoms with limited measurement pattern experiments.}
    \label{tab:Exp:comparison_corrupteddata}
\end{table}

\begin{table}
    \centering
    \renewcommand{\arraystretch}{1.2}
    \begin{tabular}{c|c|P{5em}|P{5em}|P{5em}|P{5em}}
        \toprule
        {\bf Phantom} & & {\bf RU} & {\bf UR} & {\bf cJoint} & {\bf ADJUST} \\ \midrule
        \multirow{3}*{\bf Shepp-Logan} & MSE & 0.1446 & 0.1856 & 0.1980 & {\bf 0.0976} \\
        & PSNR & 9.042 & 9.392 & 7.259 & {\bf 10.94} \\
        & SSIM & 0.0213 & 0.3773 & 0.0209 & {\bf 0.6228} \\ \midrule
        \multirow{3}*{\bf Thorax} & MSE & 0.0543 & 0.0715 & 0.0533 & {\bf 0.0010} \\ 
        & PSNR & 21.22 & 19.64 & 17.95 & {\bf 33.05} \\
        & SSIM & 0.6820 & 0.6579 & 0.7608 & {\bf 0.9707} \\ \bottomrule
    \end{tabular}
    \caption{Reconstruction error in terms of MSE, PSNR, SSIM for various phantoms with 10 times smaller spectral resolution (10 spectral bins in total).}
    \label{tab:Exp:comparisonLimSpecRes}
\end{table}

\subsection{Limited spectral resolution}

\begin{figure}[!t]
    \centering
    \renewcommand{\arraystretch}{1}
    \begin{tabular}{c c}
        \includegraphics[width=0.45\textwidth,clip]{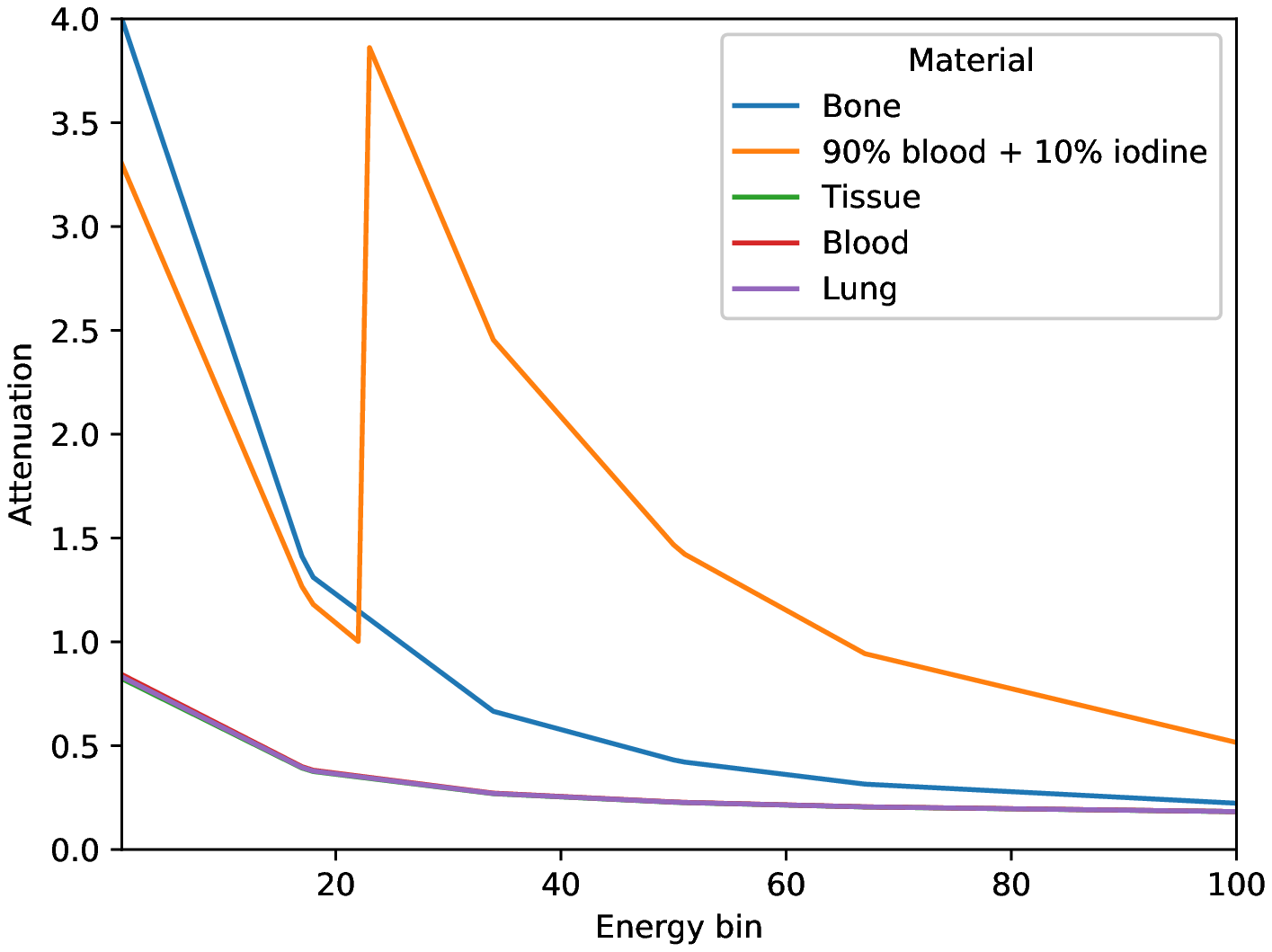} \quad & \includegraphics[width=0.45\textwidth,clip]{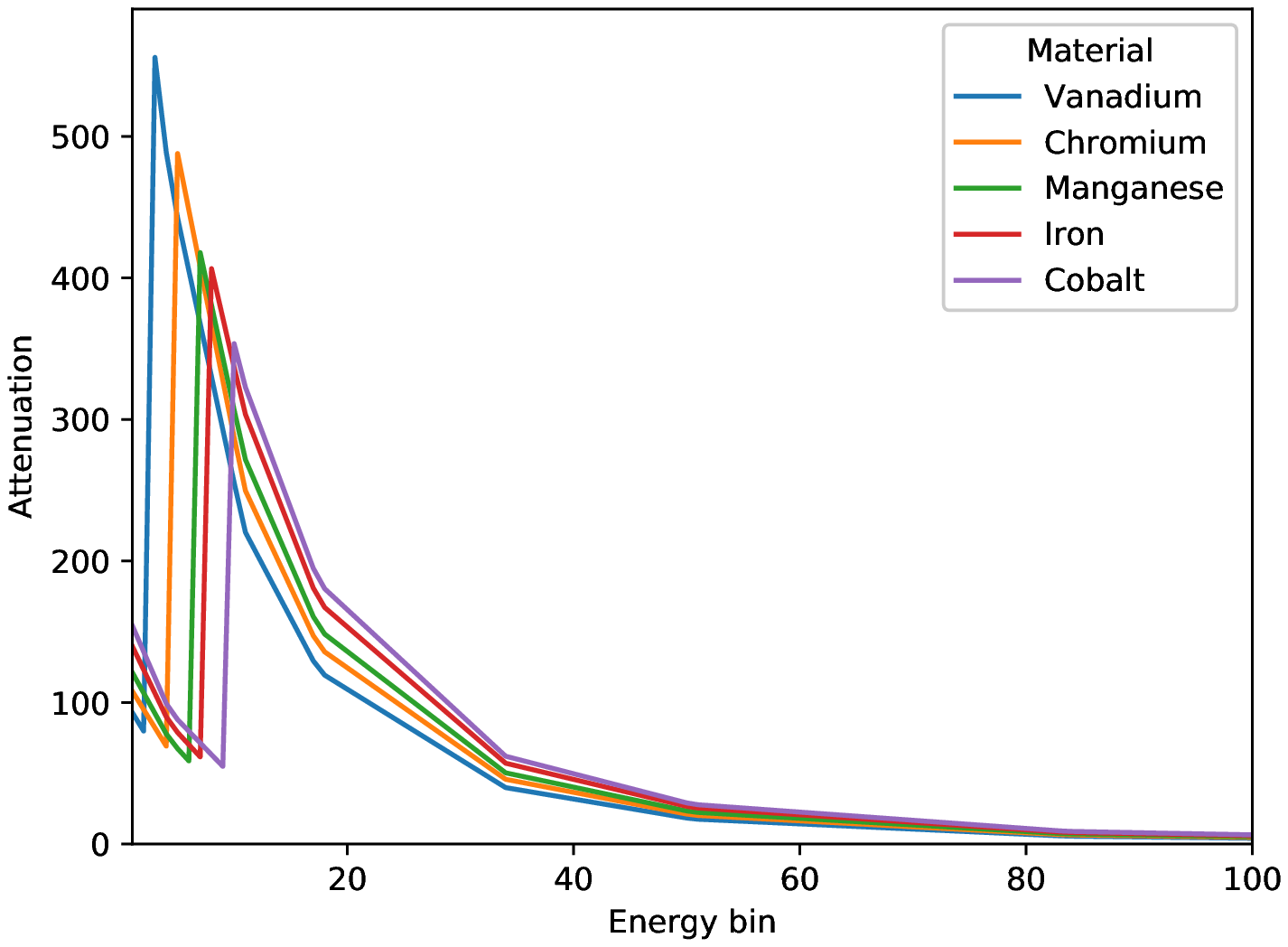} \\
        (a) Thorax - 100 spectral bins & (b) Shepp-Logan - 100 spectral bins \\[2ex]
        \includegraphics[width=0.45\textwidth,clip]{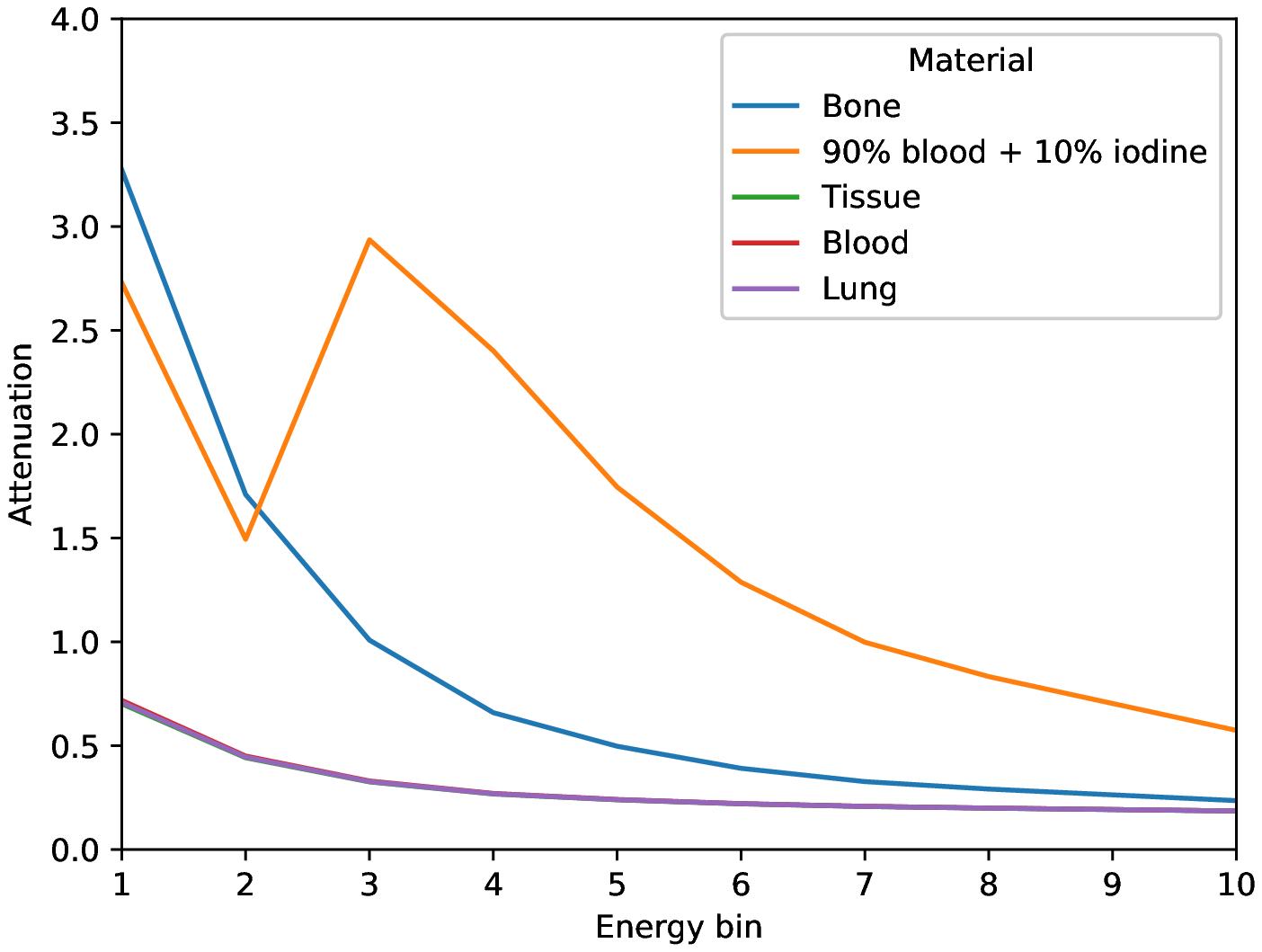} \quad & \includegraphics[width=0.45\textwidth,clip]{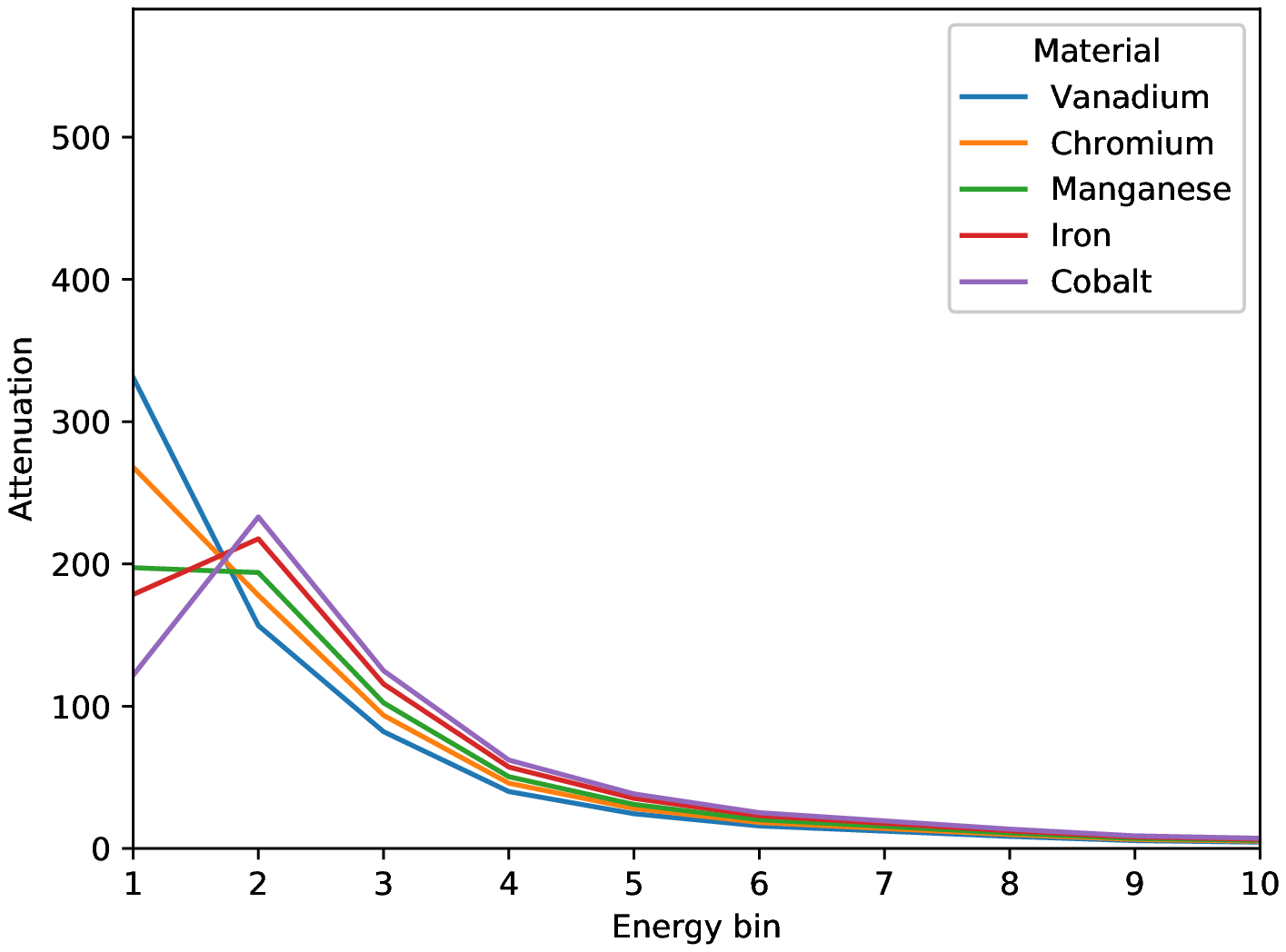} \\
        (c) Thorax - 10 spectral bins & (d) Shepp-Logan - 10 spectral bins
    \end{tabular}
    \caption{Attenuation spectra of different materials in the Thorax phantom (left) and the Shepp-Logan phantom right (right), partitioned over 100 energy bins (top) and 10 energy bins (bottom) over the same energy range.}
    \label{fig:Exp:LimSpecRes_Spectra}
\end{figure}

In this section, we investigate the performance of the algorithm when each spectral bin spans a wider energy range. This problem is also called \textit{limited spectral resolution}. To this end, we have simulated data with 10 spectral bins instead of 100 bins for the Thorax and Shepp-Logan phantoms over the same energy range as before. The selected number of energy bins is in the same order as the number of bins resulting from the use of multi-spectral X-ray photon counting detectors, rather than with the hundreds of energy bins of hyperspectral X-ray detectors. This results in a coarser energy resolution for both the source spectra and the attenuation spectra, and the dictionary has been updated accordingly. Apart from the different energy bins, we apply the same settings and configuration as the experiments in Section~\ref{sec:ExpSetup}.

To illustrate the spectral differences, Figure~\ref{fig:Exp:LimSpecRes_Spectra} shows the attenuation spectra for the materials in the phantoms with the two different energy resolutions. The soft materials, bone and iodine present in the Thorax phantom are easily identifiable since their attenuation spectra are very different from each other in the entire spectrum (\ie, $20-80$~keV).
The materials in the Thorax phantom will still be separable with decrease in the spectral resolution. In the Shepp-Logan phantom, however, all included materials have K-edges very close to each other. Hence, differences resulting from the K-edge of the materials in the Shepp-Logan phantom largely disappear. This is also reflected in the performance of various algorithms in Table~\ref{tab:Exp:comparisonLimSpecRes}, with considerably lower SSIM and PSNR compared to the full spectral resolution experiments for the Shepp-Logan phantom in Table~\ref{tab:Exp:comparison}. On the other hand, for the Thorax phantom the results are similar. Hence, we conclude that the successful application of the proposed algorithm to spectral measurements acquired in bins with larger spectral ranges depends heavily on the complexity of the problem (\ie, separability of the materials in the chosen energy range based on spectral responses in the bins).

\subsection{Noisy measurement patterns}

\begin{figure}[!b]
    \centering
    \begin{tabular}{cccc}
         \includegraphics[width=0.2\textwidth]{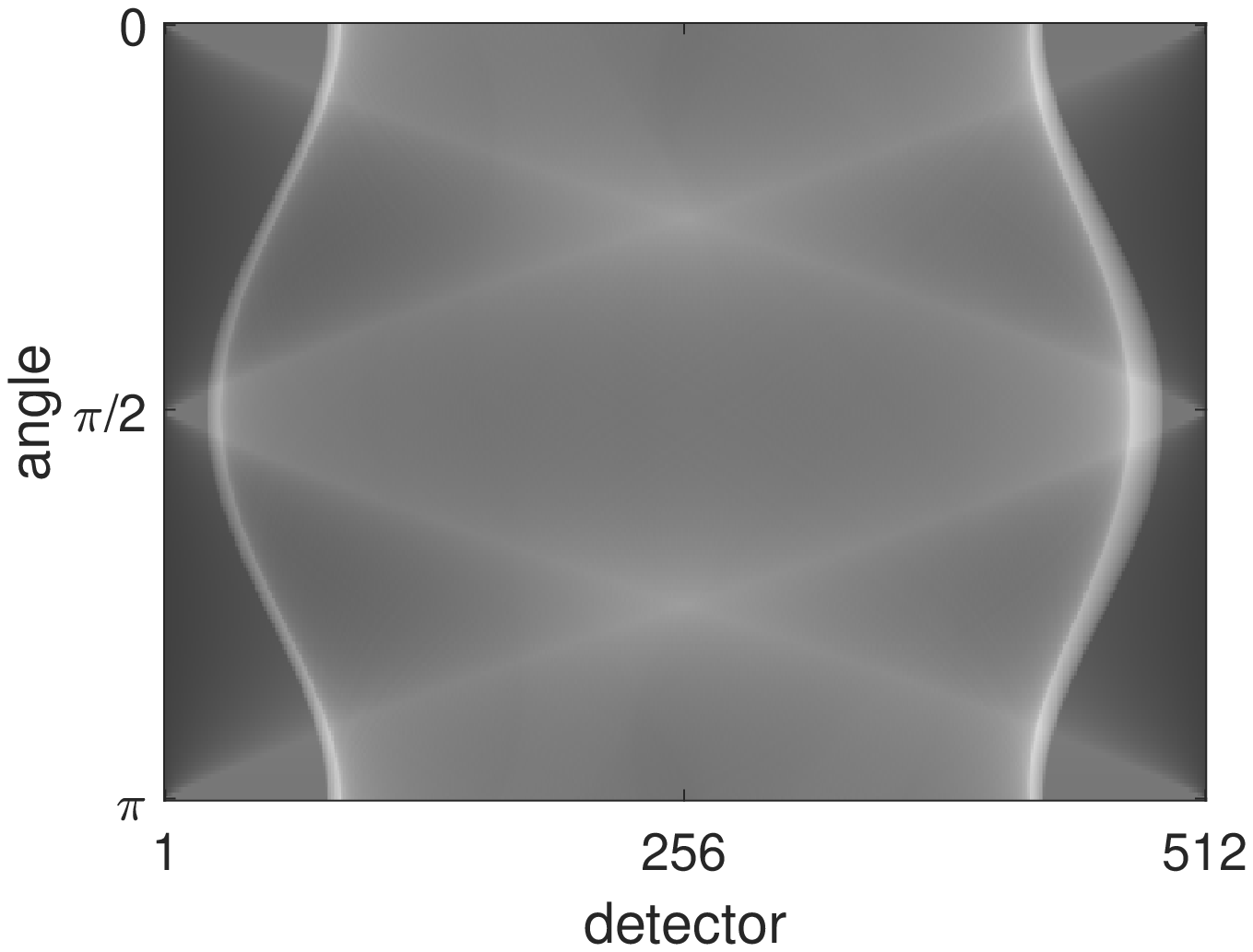} & \includegraphics[width=0.2\textwidth]{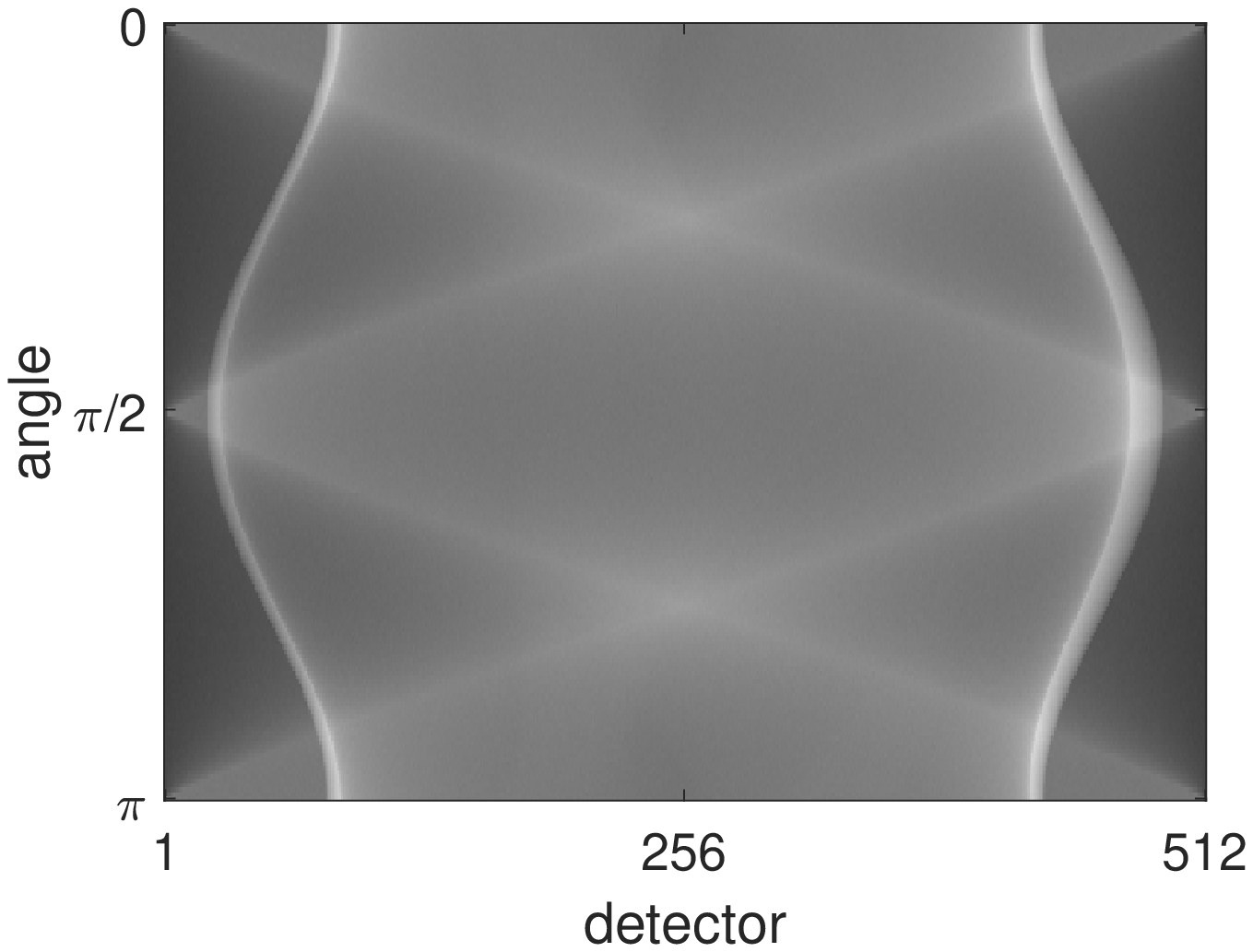} & \includegraphics[width=0.2\textwidth]{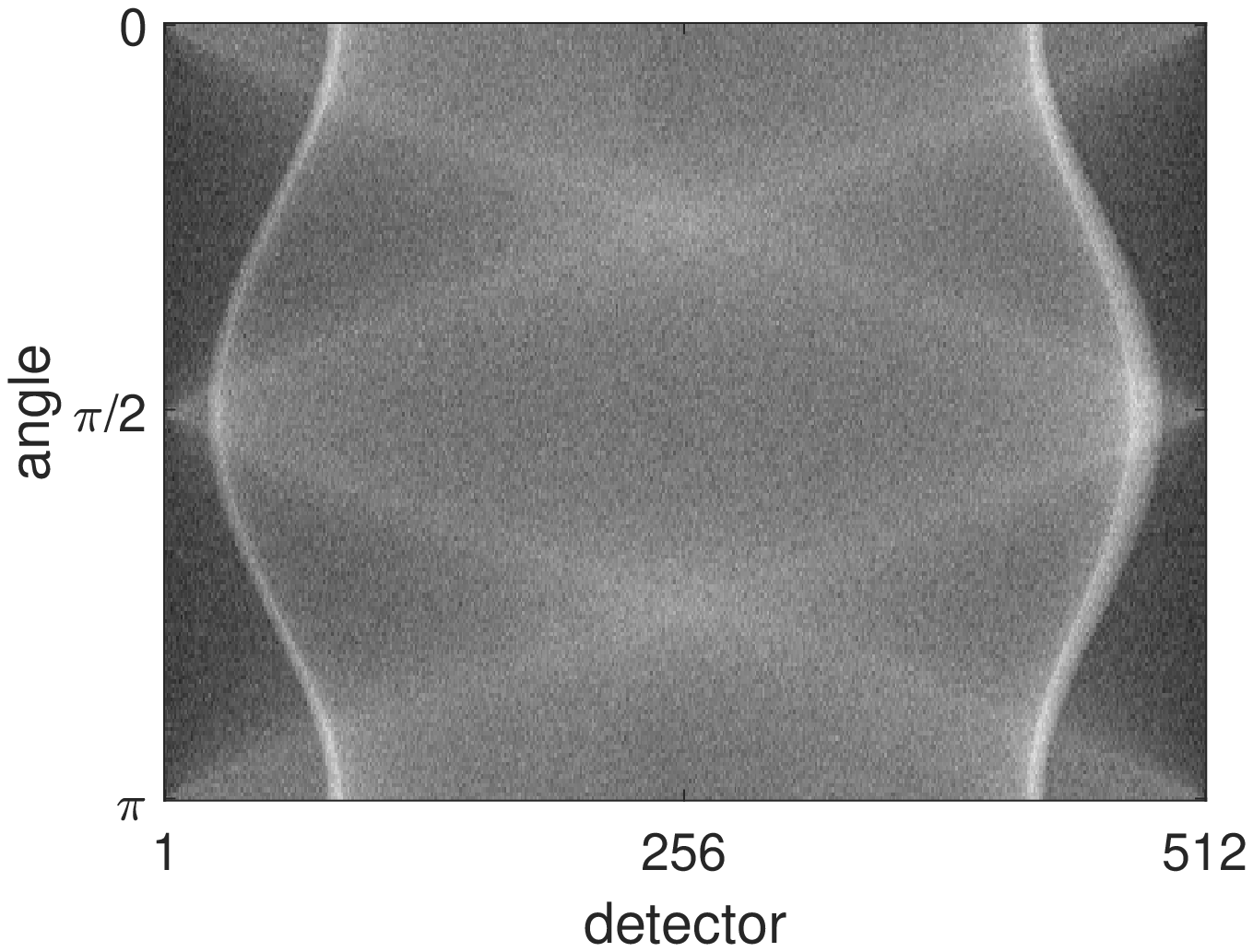} & \includegraphics[width=0.2\textwidth]{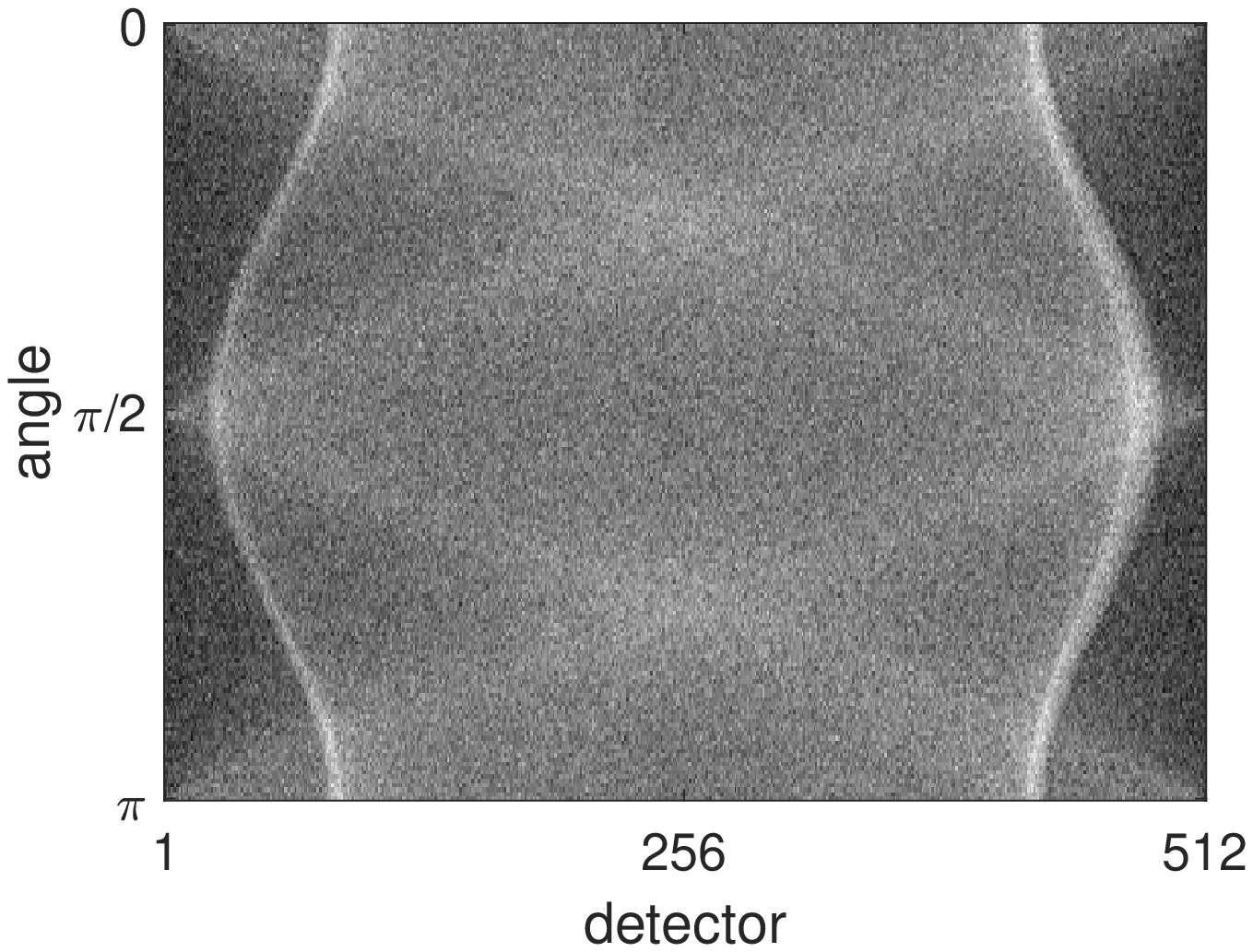} \\
         \includegraphics[width=0.2\textwidth]{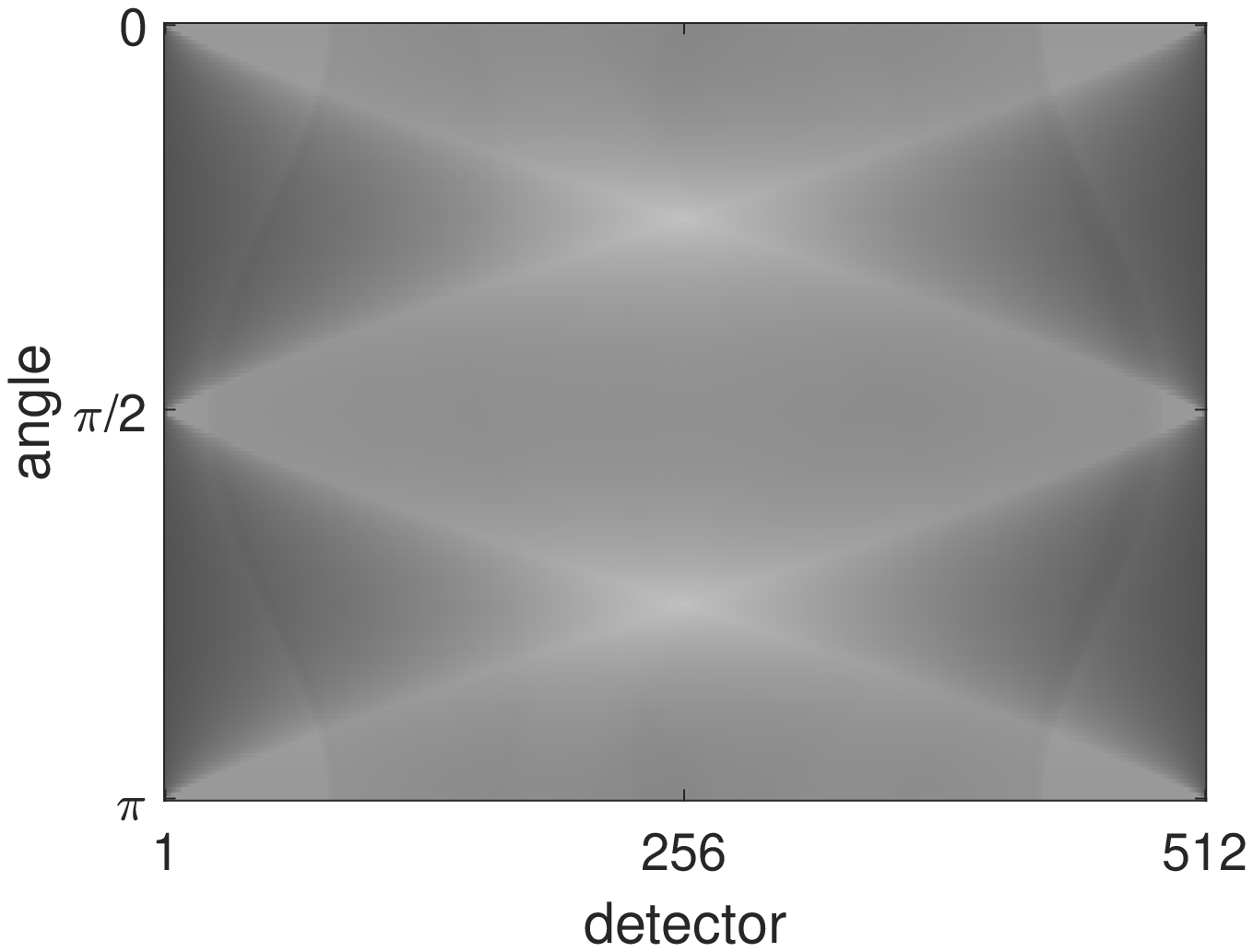} & \includegraphics[width=0.2\textwidth]{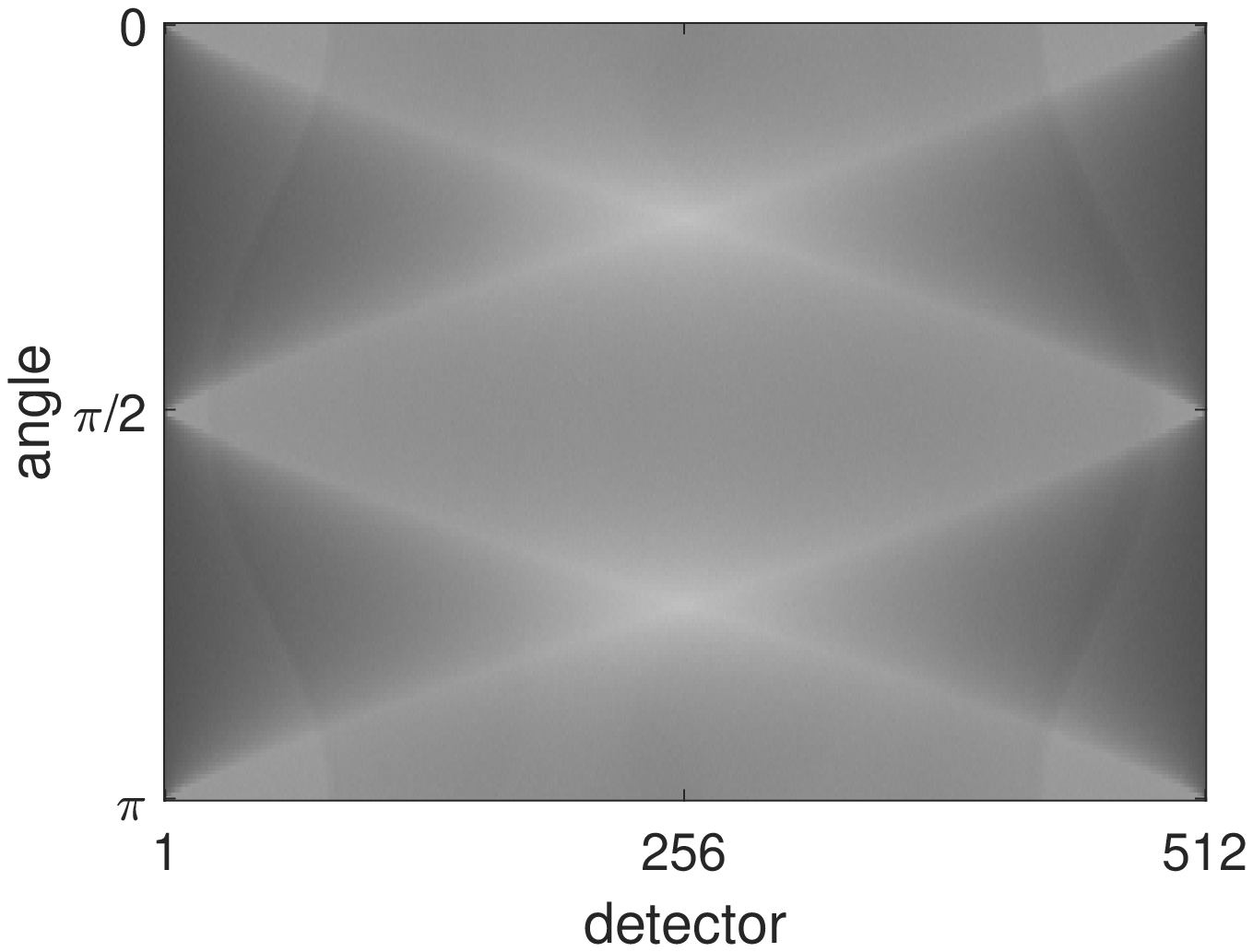} & \includegraphics[width=0.2\textwidth]{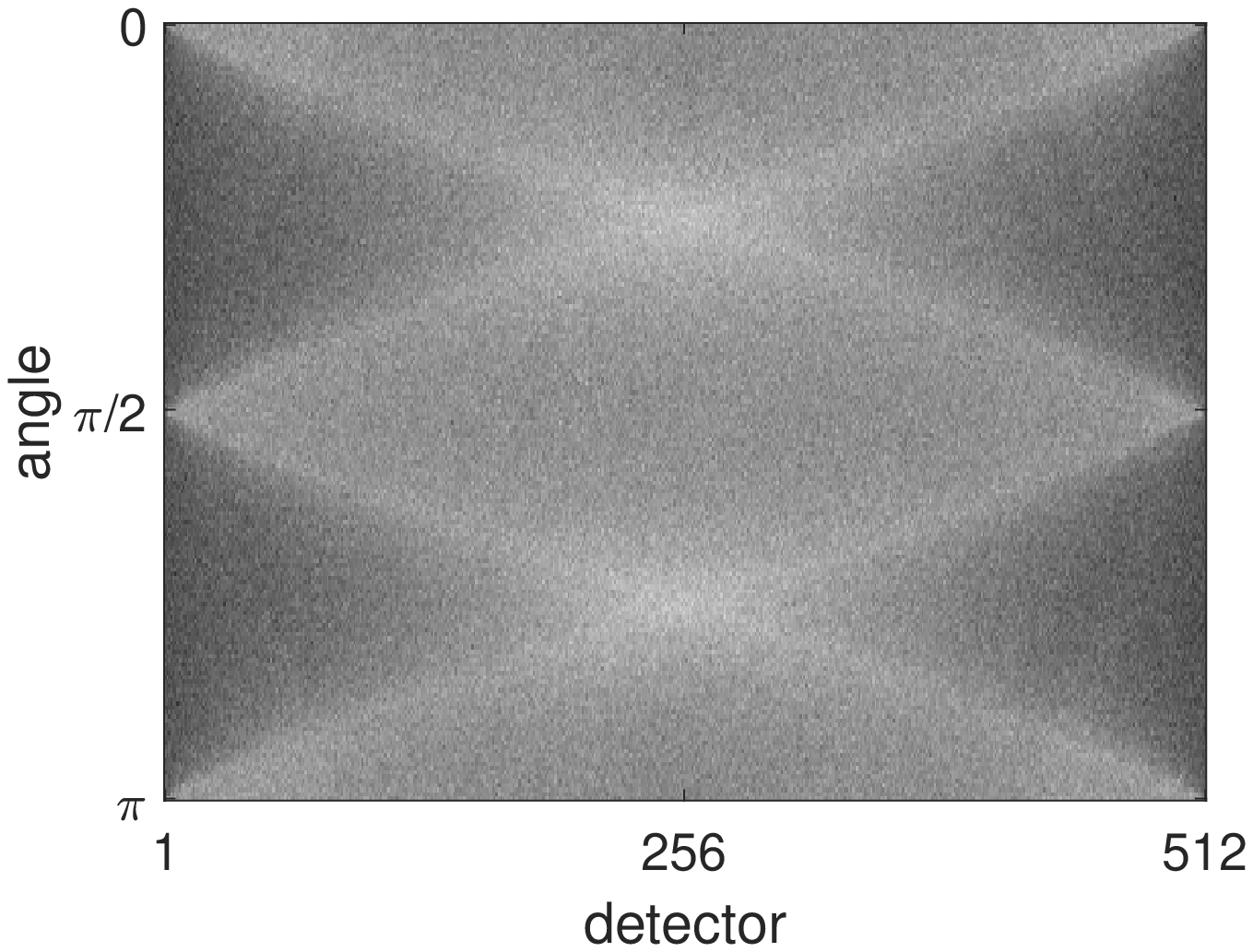} & \includegraphics[width=0.2\textwidth]{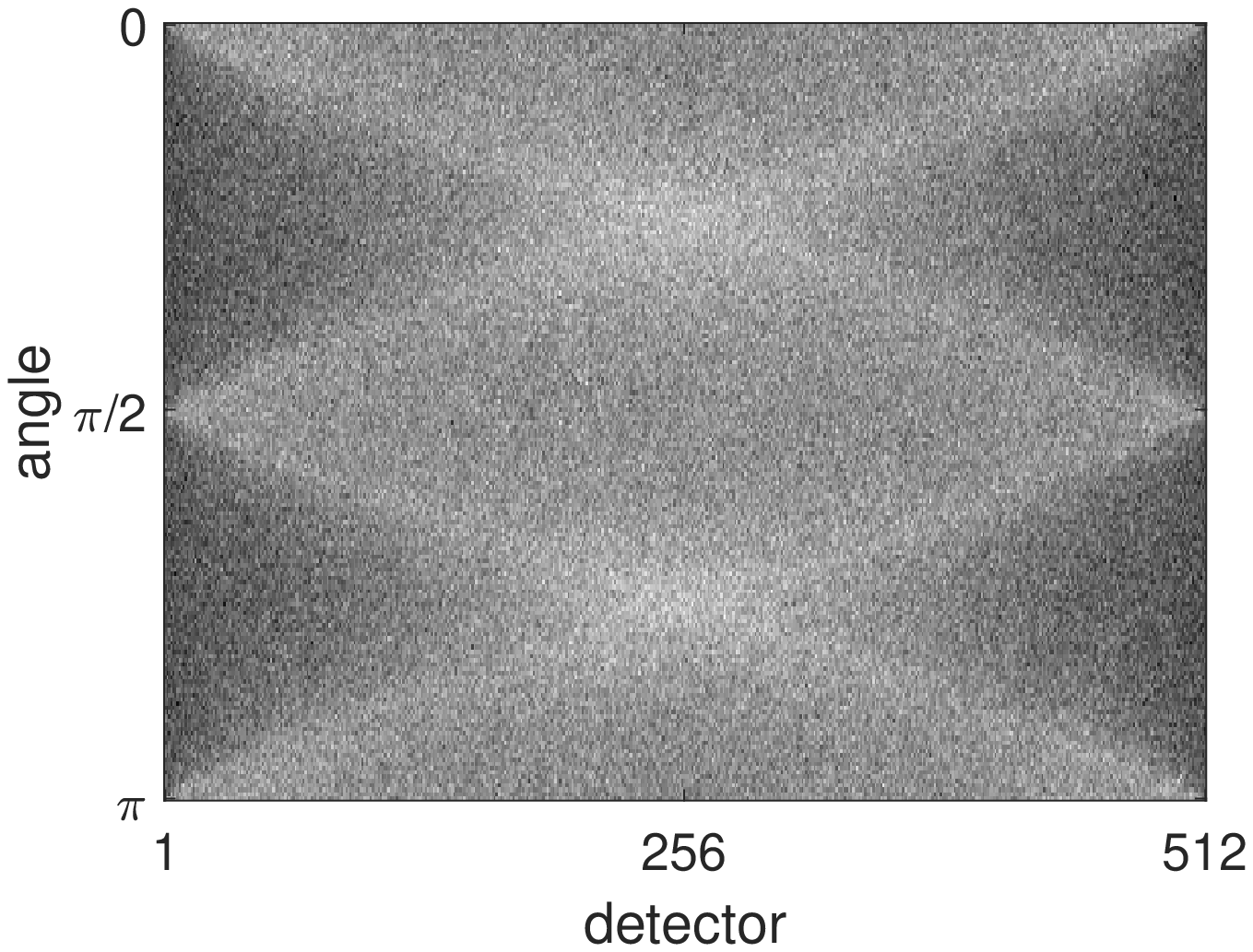} \\
        (a) Clean & (b) $1\%$ noise & (c) $10 \%$ noise & (d) $20 \%$ noise
    \end{tabular}
    \caption{Clean and noisy sinograms (\ie, tomographic projections) of Shepp-Logan phantom for energy of $5.6$~keV (top-row) and $7.7$~keV (bottom-row).}
    \label{fig:Exp:noisyProj}
\end{figure}

We consider three different noise levels to check the robustness of ADJUST against noise. In particular, we corrupt the spectral tomographic measurements with additive Gaussian noise of strength $\{1, 10, 20\}\%$ followed by a Poisson noise with intensity corresponding to the source spectrum. In all three cases, $180$ angular projections in the $[0, \pi)$ range are acquired in 100 spectral bins. The effect of noise on the spectral tomographic projections is demonstrated in Figure~\ref{fig:Exp:noisyProj}. In table~\ref{tab:Exp:comparisonNoisy}, we list the performance measures of ADJUST on three different phantoms. We observe a steady decrease in PSNR and SSIM values for the Shepp-Logan and Thorax phantom.  However, the PSNR and SSIM do not suffer from high noise levels (\ie\ $20 \%$ noise) for the Disks phantom. Since the Disks phantom consists of low-rank shapes of small sizes, they can be retrieved approximately well from noisy measurements. However, the performance of ADJUST on noisy datasets may not be extended for complex shapes, as seen from the numerical studies on the Thorax and Shepp-Logan phantom.  Hence, these numerical experiments demonstrate that ADJUST is stable against a moderate amount of noise, and may not be reliable against the high noise. 

\begin{table}[!t]
    \centering
    \renewcommand{\arraystretch}{1.2}
    \begin{tabular}{c|c|P{6em}|P{6em}|P{6em}}
        \toprule
        {\bf Phantom} & & {\bf $1\%$ noise} & {\bf $10 \%$ noise} & {\bf $20 \%$ noise}  \\ \midrule
        \multirow{3}*{\bf Shepp-Logan} & MSE & 0.0032 & 0.0067 & 0.0187 \\
        & PSNR & 25.68 & 22.53 & 18.68 \\
        & SSIM & 0.9738 &  0.9012 & 0.7076 \\ \midrule
        \multirow{3}*{\bf Disks} & MSE & 0.0001 & 0.0034 & 0.0003  \\ 
        & PSNR & 39.20 & 32.44 & 35.53  \\
        & SSIM & 0.9989 & 0.9779 & 0.9785  \\ \midrule
        \multirow{3}*{\bf Thorax} & MSE & 0.0061 & 0.0129 & 0.0123 \\ 
        & PSNR & 29.70  & 25.93 & 23.33  \\
        & SSIM & 0.8716 & 0.8474 & 0.8464 \\ \bottomrule
    \end{tabular}
    \caption{Reconstruction error in terms of MSE, PSNR, SSIM for various phantoms with three different noise levels.}
    \label{tab:Exp:comparisonNoisy}
\end{table}

\subsection{Experiment on micro-CT data}
\label{sec:RealData}

\begin{figure}[H]
    \centering
    \begin{tabular}{c c}
        \includegraphics[width=0.4\textwidth]{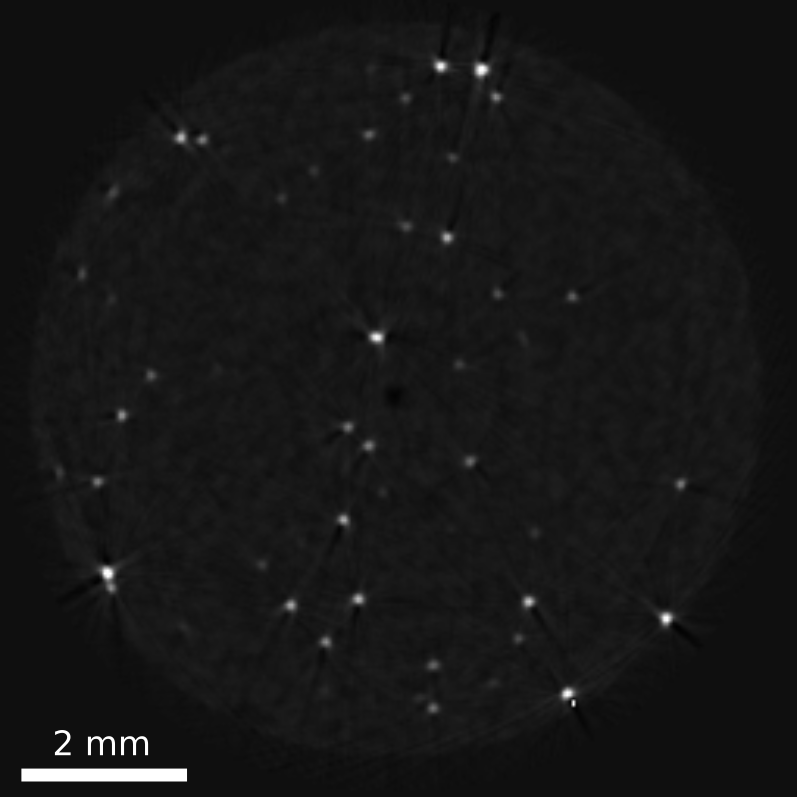} & \includegraphics[width=0.4\textwidth]{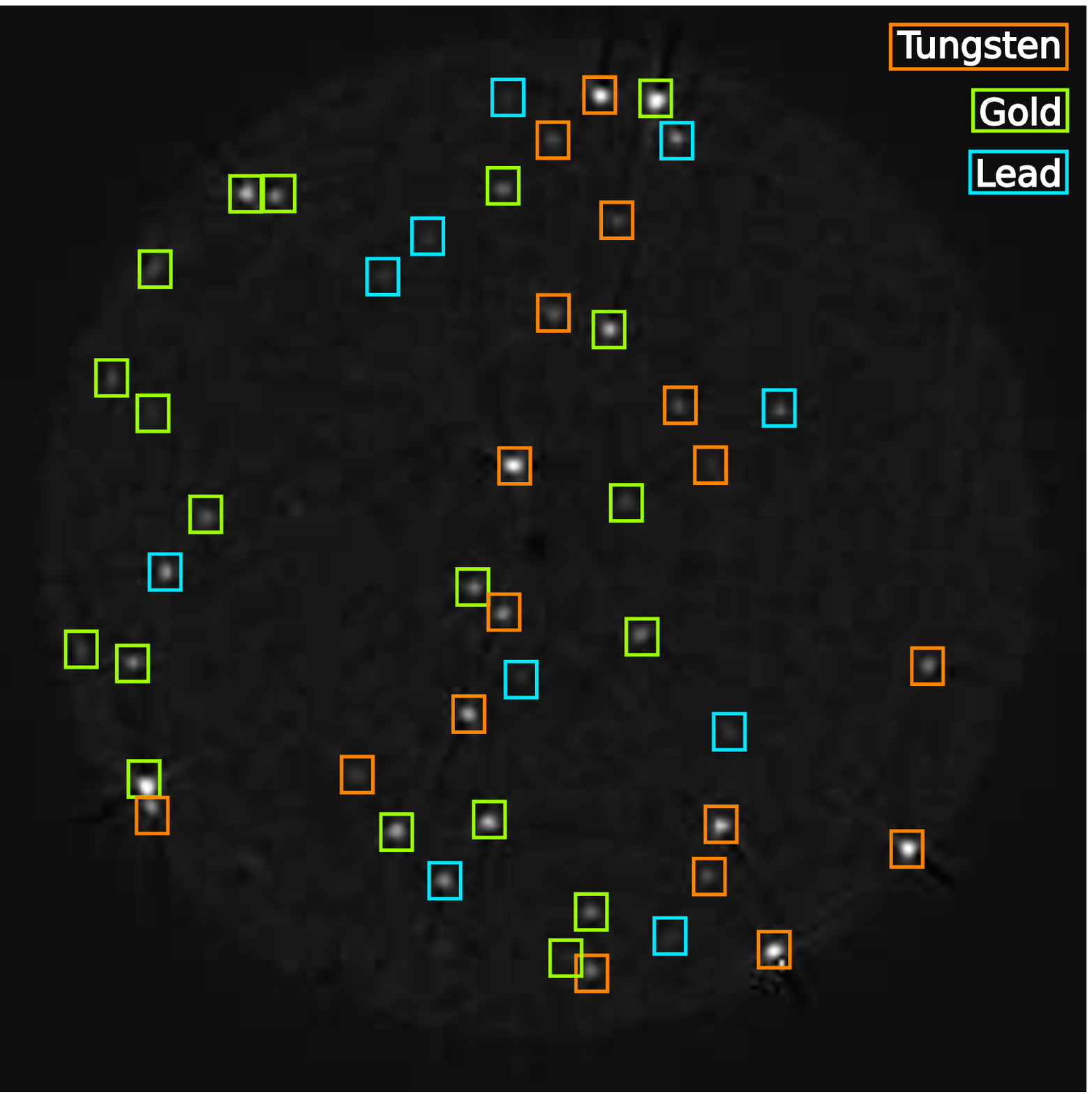} \\
        (a) & (b)
    \end{tabular}
    \caption{CT reconstruction of the 7th energy bin (\textbf{a}) of the spectral micro-CT dataset, in which the locations of the tungsten (orange), gold (green) and lead (blue) particles are highlighted (\textbf{b}).}
    \label{fig:Exp:Particlelocations}
\end{figure}

\begin{figure}[H]
    \centering
    \includegraphics[width=0.5\textwidth]{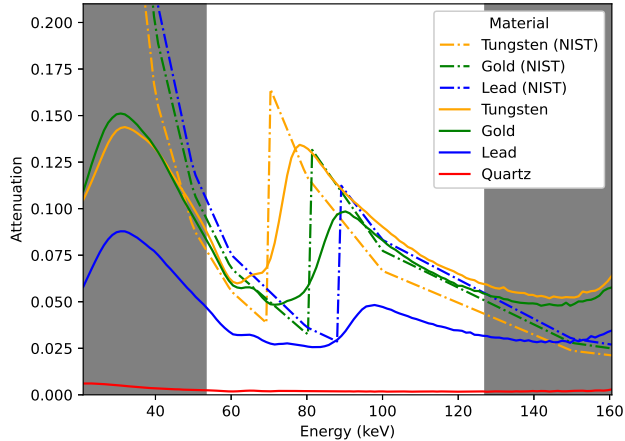}
    \caption{Attenuation spectra of the materials in the sample, showing both the spectra from NIST (scaled by 0.015) and the spectra extracted from the channel-wise CT reconstructions, including quartz. The spectral range is from 20.35 keV to 161.15 keV (white region), with the bins having a spectral resolution of 1.1 keV.}
    \label{fig:Exp:RealSpectra}
\end{figure}

In this subsection, we test the performance of ADJUST on a publicly available X-ray microtomography data set~\cite{SittnerGodinhoData} generated by a conventional laboratory-based CT scanner equipped with a photon-counting line detector (TESCAN Polydet) containing a semiconductor crystal (CdTe)~\cite{SittnerGodinho}. The particle mixture sample contains pure gold, tungsten and lead, along with quartz. For a total of 128 energy bins with a spectral range of ca. 20 to 160~keV, tomographic projections are acquired on 255 equally-spaced detector pixels for 600 angles in $[0, 2 \pi)$. We perform a reconstruction using the simultaneous iterative reconstruction algorithm (SIRT) \cite{KakSlaney} for each channel and determine the location of each material manually (refer to Figure~\ref{fig:Exp:Particlelocations}). We plot the gold, lead, and tungsten spectrum obtained through channel-wise reconstructions along with the corresponding NIST spectra in Figure~\ref{fig:Exp:RealSpectra}.   

\begin{figure}[!b]
    \centering
    \begin{tabular}{cccc}
        {\bf RU} & {\bf UR} & {\bf cJoint} & {\bf ADJUST} \\
        \begin{tikzpicture}[spy using outlines={circle,col1,magnification=4,size=1.5cm, connect spies}]
        \node {\pgfimage[interpolate=true,height=3.5cm]{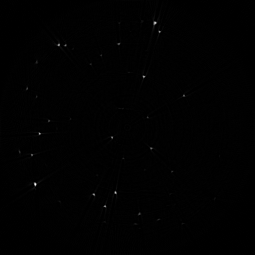}};
        \spy on (0.32,1.35) in node [] at (1.4,1.25);
        \end{tikzpicture}
        & 
        \begin{tikzpicture}[spy using outlines={circle,col1,magnification=4,size=1.5cm, connect spies}]
        \node {\pgfimage[interpolate=true,height=3.5cm]{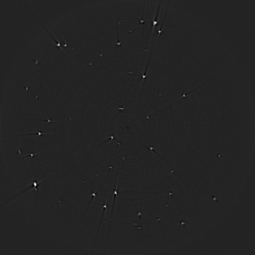}};
        \spy on (0.32,1.35) in node [] at (1.4,1.25);
        \end{tikzpicture}
        & 
        \begin{tikzpicture}[spy using outlines={circle,col1,magnification=4,size=1.5cm, connect spies}]
        \node {\pgfimage[interpolate=true,height=3.5cm]{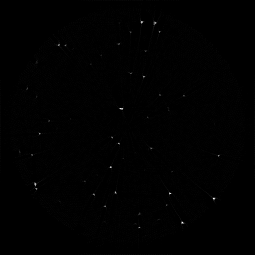}};
        \spy on (0.32,1.35) in node [] at (1.4,1.25);
        \end{tikzpicture}
        &
        \begin{tikzpicture}[spy using outlines={circle,col1,magnification=4,size=1.5cm, connect spies}]
        \node {\pgfimage[interpolate=true,height=3.5cm]{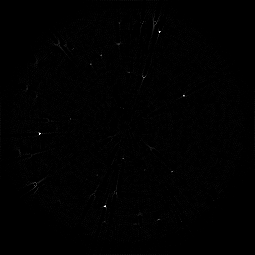}};
        \spy on (0.32,1.35) in node [] at (1.4,1.25);
        \end{tikzpicture} \\
         \begin{tikzpicture}[spy using outlines={circle,col2,magnification=4,size=1.5cm, connect spies}]
        \node {\pgfimage[interpolate=true,height=3.5cm]{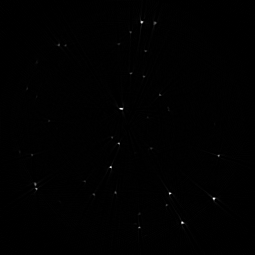}};
        \spy on (0.32,1.35) in node [] at (1.4,1.25);
        \end{tikzpicture}
        & 
        \begin{tikzpicture}[spy using outlines={circle,col2,magnification=4,size=1.5cm, connect spies}]
        \node {\pgfimage[interpolate=true,height=3.5cm]{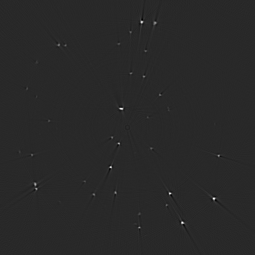}};
        \spy on (0.32,1.35) in node [] at (1.4,1.25);
        \end{tikzpicture}
        & 
        \begin{tikzpicture}[spy using outlines={circle,col2,magnification=4,size=1.5cm, connect spies}]
        \node {\pgfimage[interpolate=true,height=3.5cm]{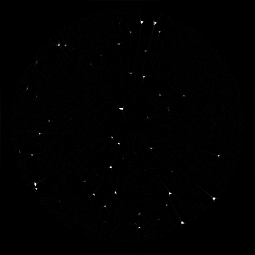}};
        \spy on (0.32,1.35) in node [] at (1.4,1.25);
        \end{tikzpicture}
        &
        \begin{tikzpicture}[spy using outlines={circle,col2,magnification=4,size=1.5cm, connect spies}]
        \node {\pgfimage[interpolate=true,height=3.5cm]{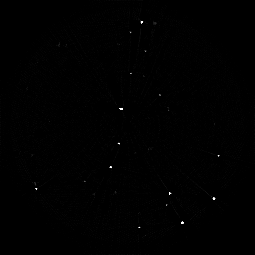}};
        \spy on (0.32,1.35) in node [] at (1.4,1.25);
        \end{tikzpicture} \\
         \begin{tikzpicture}[spy using outlines={circle,col3,magnification=4,size=1.5cm, connect spies}]
        \node {\pgfimage[interpolate=true,height=3.5cm]{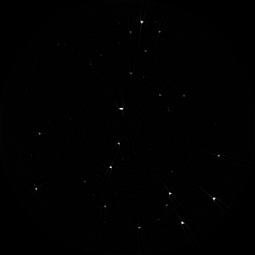}};
        \spy on (0.32,1.35) in node [] at (1.4,1.25);
        \end{tikzpicture}
        & 
        \begin{tikzpicture}[spy using outlines={circle,col3,magnification=4,size=1.5cm, connect spies}]
        \node {\pgfimage[interpolate=true,height=3.5cm]{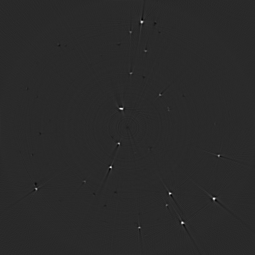}};
        \spy on (0.32,1.35) in node [] at (1.4,1.25);
        \end{tikzpicture}
        & 
        \begin{tikzpicture}[spy using outlines={circle,col3,magnification=4,size=1.5cm, connect spies}]
        \node {\pgfimage[interpolate=true,height=3.5cm]{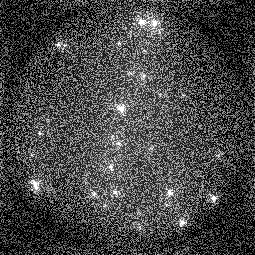}};
        \spy on (0.32,1.35) in node [] at (1.4,1.25);
        \end{tikzpicture}
        &
        \begin{tikzpicture}[spy using outlines={circle,col3,magnification=4,size=1.5cm, connect spies}]
        \node {\pgfimage[interpolate=true,height=3.5cm]{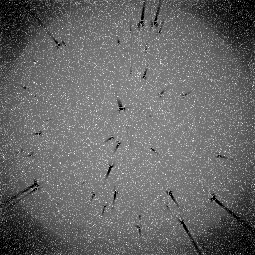}};
        \spy on (0.32,1.35) in node [] at (1.4,1.25);
        \end{tikzpicture} \\
         \begin{tikzpicture}[spy using outlines={circle,col4,magnification=4,size=1.5cm, connect spies}]
        \node {\pgfimage[interpolate=true,height=3.5cm]{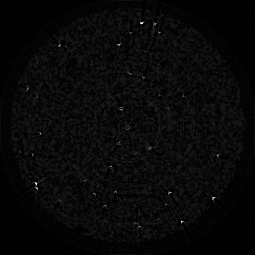}};
        \spy on (0.32,1.35) in node [] at (1.4,1.25);
        \end{tikzpicture}
        & 
        \begin{tikzpicture}[spy using outlines={circle,col4,magnification=4,size=1.5cm, connect spies}]
        \node {\pgfimage[interpolate=true,height=3.5cm]{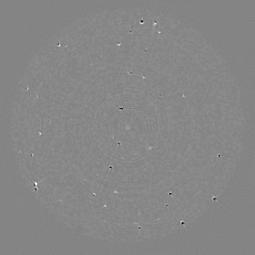}};
        \spy on (0.32,1.35) in node [] at (1.4,1.25);
        \end{tikzpicture}
        & 
        \begin{tikzpicture}[spy using outlines={circle,col4,magnification=4,size=1.5cm, connect spies}]
        \node {\pgfimage[interpolate=true,height=3.5cm]{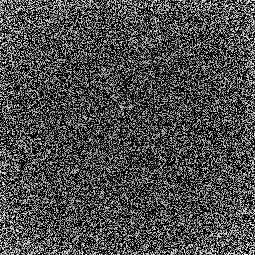}};
        \spy on (0.32,1.35) in node [] at (1.4,1.25);
        \end{tikzpicture}
        &
        \begin{tikzpicture}[spy using outlines={circle,col4,magnification=4,size=1.5cm, connect spies}]
        \node {\pgfimage[interpolate=true,height=3.5cm]{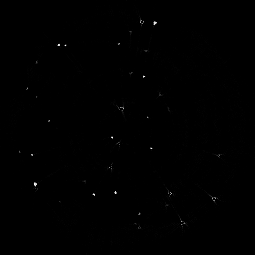}};
        \spy on (0.32,1.35) in node [] at (1.4,1.25);
        \end{tikzpicture} \\
        \includegraphics[width=0.23\textwidth,clip]{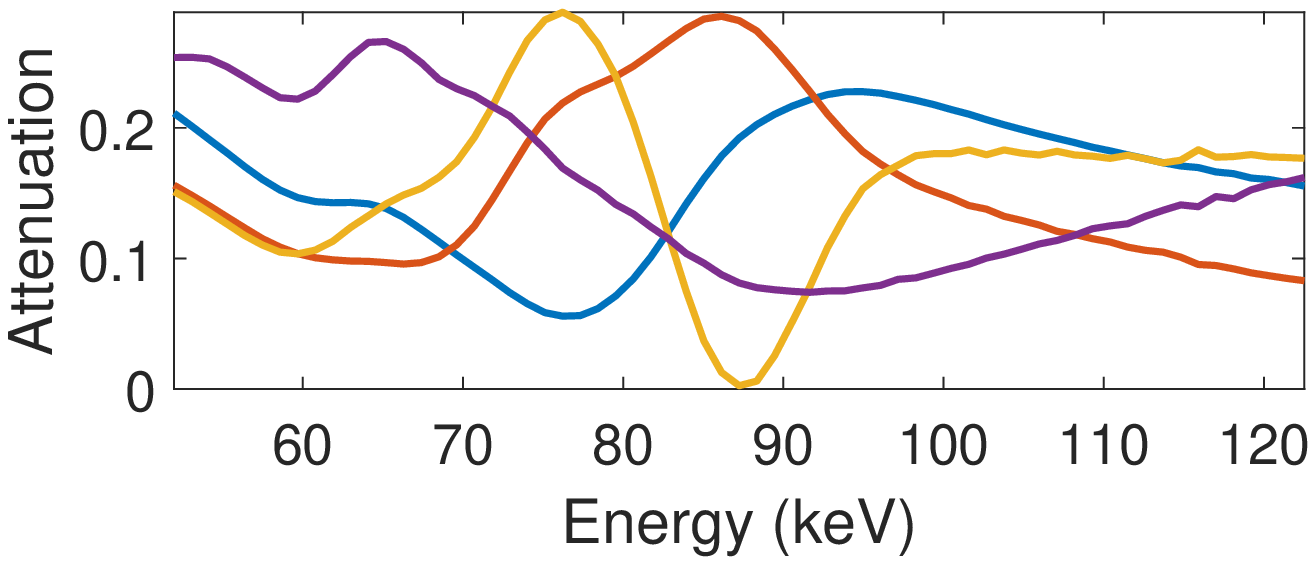} & \includegraphics[width=0.23\textwidth,clip]{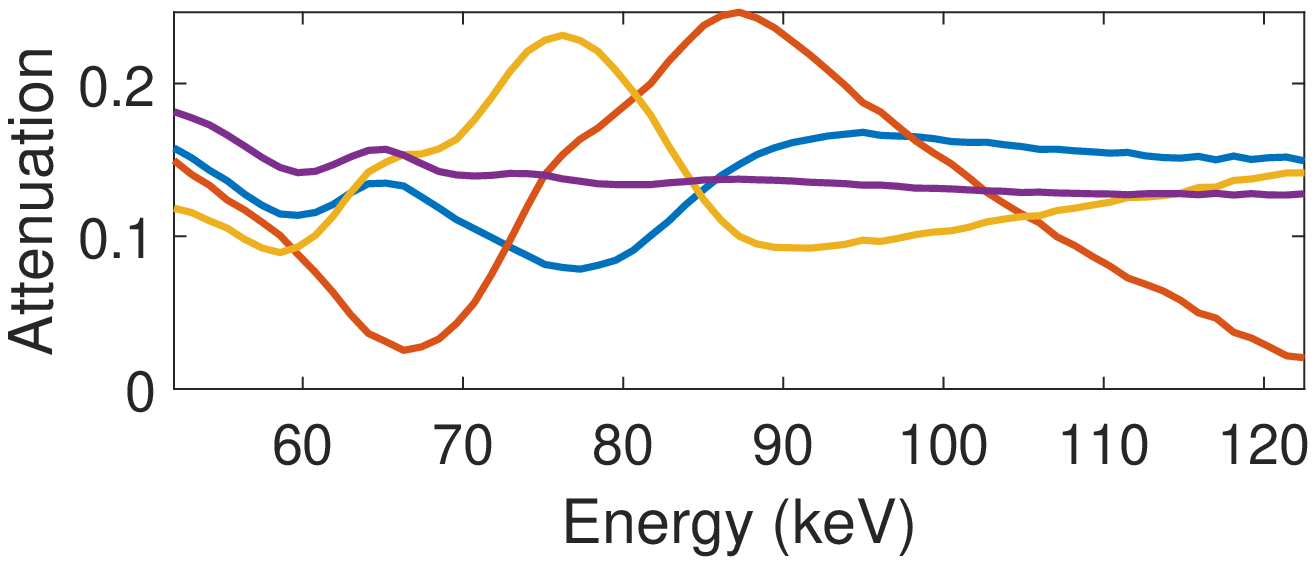} & \includegraphics[width=0.23\textwidth,clip]{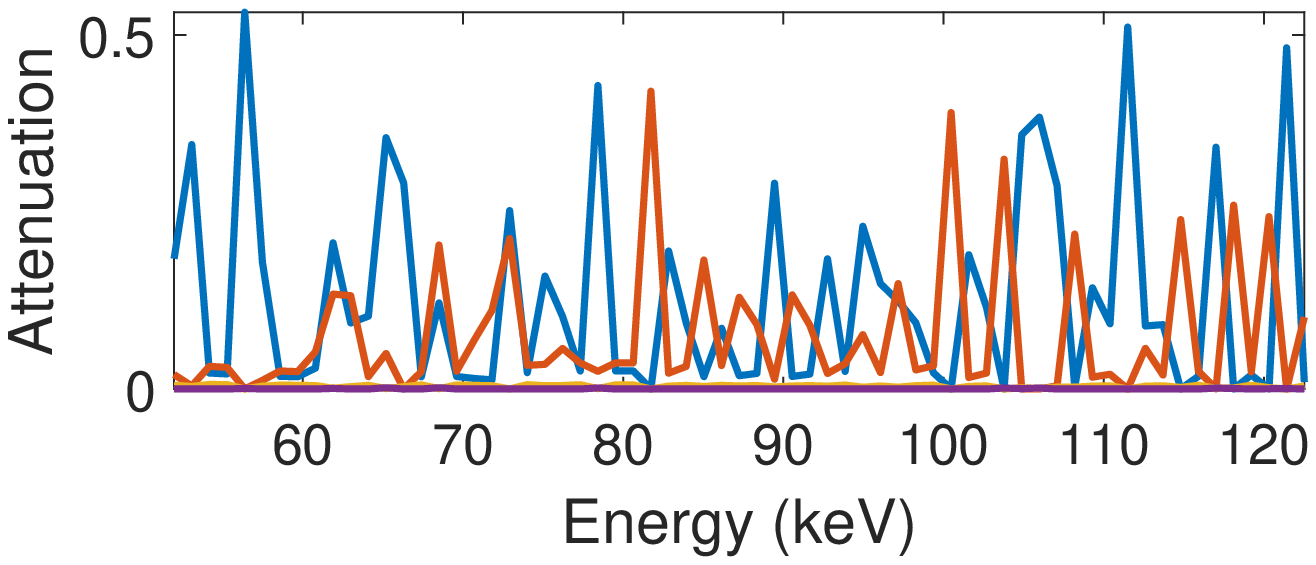} & \includegraphics[width=0.23\textwidth,clip]{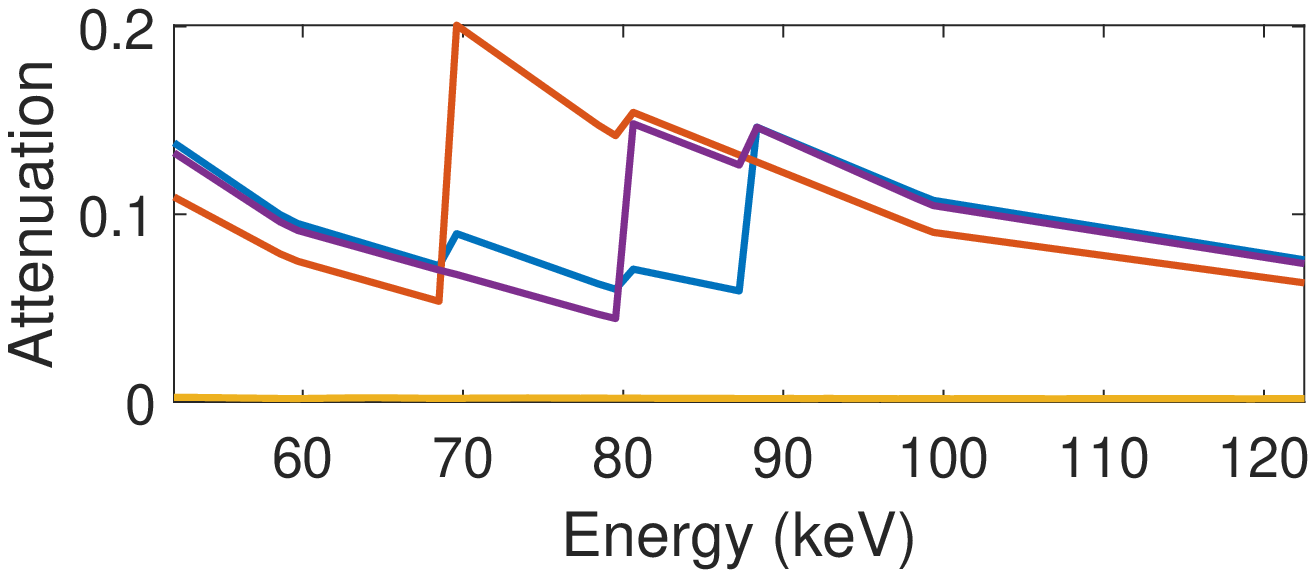} \\
    \end{tabular}
    \caption{Reconstruction results of methods RU, UR, cJoint, and ADJUST on microtomography dataset of particle mixtures. The zoomed sections demonstrate the separation capabilities of ADJUST compared to RU, UR, and cJoint on particles of three different materials (tungsten, gold and lead). We match the color of zoomed sections with the spectral plots for improved readability.}
    \label{fig:Exp:RealData:Results}
\end{figure}

\begin{figure}[H]
    \centering
    \begin{small}
    \renewcommand{\arraystretch}{1}
    \begin{tabular}{c c c}
        \includegraphics[width=0.31\textwidth,clip]{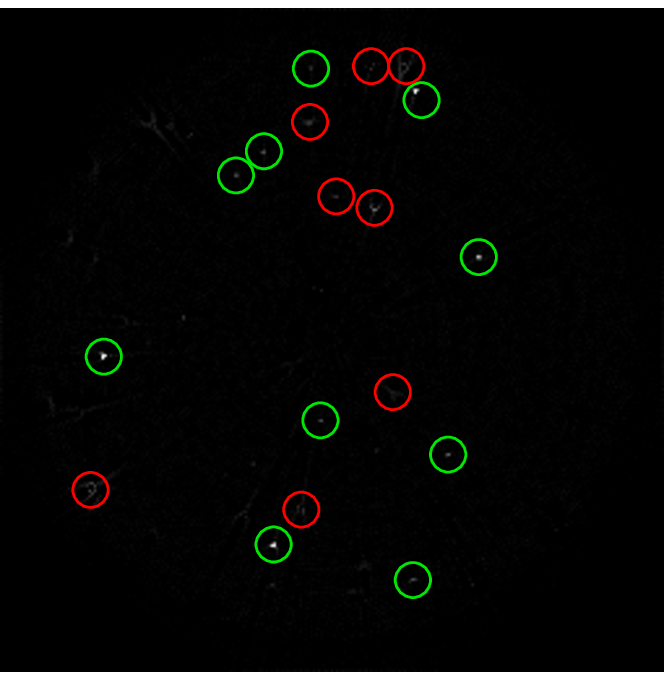} & \includegraphics[width=0.31\textwidth,clip]{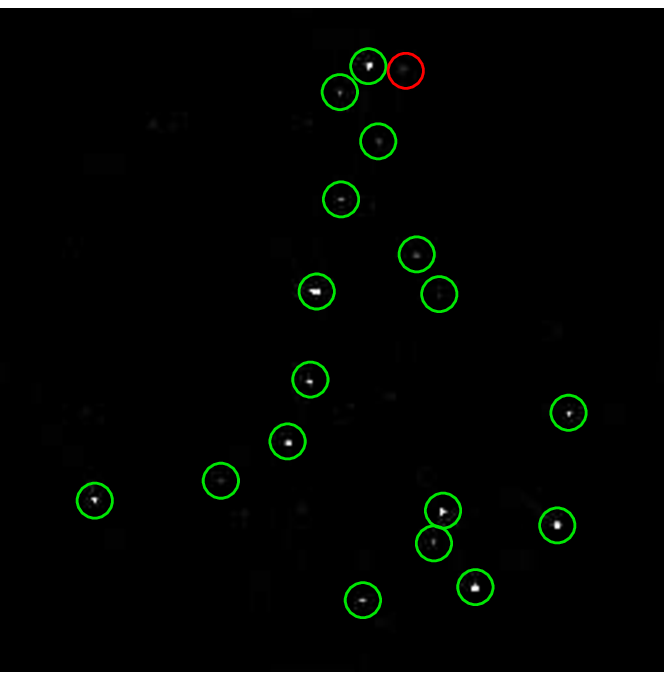} & \includegraphics[width=0.31\textwidth,clip]{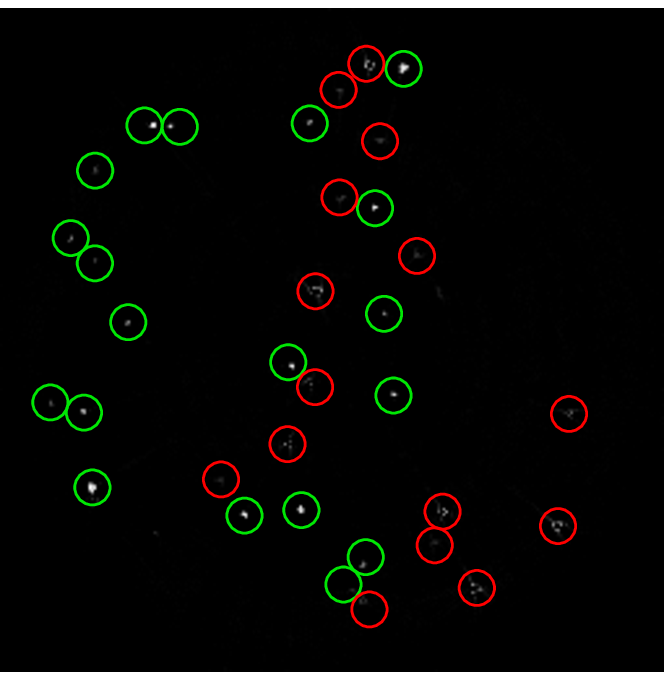} \\
        (a) Lead & (b) Tungsten & (c) Gold
    \end{tabular}
    \end{small}
    \caption{Detailed reconstruction results of ADJUST on X-ray microtomography dataset of particle mixtures. Reconstructed material maps for lead, tungsten and gold are shown, and green circles indicate correctly identified particles. None of the maps contain false negatives, and the remaining identified particles are false positives (denoted by red circles).}
    \label{fig:Exp:RealData:ResultsADJUST}
\end{figure}

For a fair comparison, we apply RU, UR, and cJoint along with ADJUST on this microtomography spectral dataset. We use the same settings as mentioned in Section~\ref{sec:NumericalImplementation}. For ADJUST, we use the spectral dictionary with four materials (\ie, gold, lead, tungsten and quartz) with their spectrum obtained from NIST (the spectrum of quartz is obtained from the reconstructed channels). Since it is evident from Figure~\ref{fig:Exp:RealSpectra} that the NIST spectra with appropriate scaling match closely to the spectra in the sample in the range of [53.35,127.05]~keV, we reduce the spectral range of the dataset from 53.35 to 127.05 keV (amounting to 64 spectral bins in total). The reconstruction results are demonstrated in Figure~\ref{fig:Exp:RealData:Results}. From these results, it is clear that RU, UR, and cJoint cannot precisely reconstruct the spectral signatures of materials. These results, furthermore, suggest the need of stronger spectral regularization to separate the materials. Out of all the spatial maps reconstructed with these classical methods, only the spatial map of tungsten (red) produced by the RU method approximately matches the expected material map (shown in Figure~\ref{fig:Exp:Particlelocations}). On the other hand, the spatial maps recovered from ADJUST match almost precisely with the expected material maps. Figure~\ref{fig:Exp:RealData:ResultsADJUST} shows that all particles are identified and that there are no false negatives. False positives are mostly small and faint, and many of these occur at the borders of other particles, mainly between gold and tungsten. The reconstruction images are also shown in high resolution in Appendix~\ref{sec:VisualResultsReal}. We note that the NIST spectra do not overlap entirely with the spectra recovered through channel-wise reconstruction. Therefore, the spectral dictionary must be calibrated for the spectral detector setup for various materials. We expect improvement in reconstruction results of ADJUST with a calibrated spectral dictionary. 

\section{Conclusions \& Discussion}
\label{sec:Conclusions}

Spectral imaging is an emerging topic in X-ray tomography since it adds an additional dimension to the measurements, which can be exploited to retrieve the material composition of the object of interest. Recently, joint approaches (also known as one-step methods) have emerged as a promising technique for solving the spectral imaging problem by incorporating all the prior information in a single step. These joint approaches reduce the ill-posedness of the spectral imaging problem. However, the spectral signatures of many materials are very similar, making the joint approaches likely to fail when many materials are involved. To tackle the problems with the joint approach, we propose the ADJUST framework that integrates the structure of spectral signatures by creating a dictionary of all the known materials and uses this to jointly reconstruct and carry out material decomposition in a single step. Since the resulting formulation is a bi-convex optimization problem, we propose an Alternating Accelerated Proximal Minimization (AAPM) scheme to find a solution. Through numerical experiments, we show that ADJUST performs better over practical methods as well as state-of-the-art joint approaches on various simulated phantoms. \\

Obtaining projections from all directions requires high experimental time. Moreover, X-ray machines may not allow for sampling in all directions. Hence, practical methods do not help determine the material composition of structurally more complicated objects (either because of intricate structures or a wider variety of materials) with most X-ray configurations. However, ADJUST is robust against limited tomography measurement patterns on phantoms that are more complicated. \\

A natural question is to check if the utilization of a spectral dictionary (\ie\ representing $\mF$ as $\mR \mT$) in two-step methods can produce optimal results. For example, in UR method, we can first decompose spectral measurements $\mY = \mZ \mR \mT$, where $\mZ$ incorporates the projections per material. Later, the spatial maps can be extracted by solving $\mW \! \mA = \mZ$. However, in the decomposition step, we allow for unrealistic projections due to lack of knowledge of tomography operator $\mW$ and spatial properties of $\mA$, especially when dealing with limited measurements. Moreover, allowing unrealistic projections leads to unrealistic materials since simplex constraints on $\mR$ give rise to convex combination of (many) dictionary elements, and not of only a single element. A similar argument holds for RU method combined with the spectral dictionary. These unrealistic solutions are, however, penalized by proper \textit{spatio-spectral regularization} in the ADJUST framework. Hence, ADJUST will always perform better than spectral dictionary versions of two-step methods even when complete measurements are available. \\

There are some limitations to the ADJUST framework. It can only separate hard materials from each other and separate hard materials from soft materials. Moreover, if the spectral signature of the material is not present in the dictionary, or if it can not be composed as a linear combination of the elements from the dictionary, ADJUST will fail in recovering that material. Although we tested ADJUST against moderate Gaussian and Poisson noise, it is not straightforward to assume that ADJUST will behave stably against real (extremely) noisy datasets that are common in energy-dispersive X-ray tomography. \\

The experiments conducted on the micro-CT dataset suggest the potential of ADJUST for application to experimental spectral CT data. Apart from the noise level on experimental data, the successful application of ADJUST depends on several factors. As shown in the experiments with the limited spectral resolution, the number and range of the energy bins, combined with the problem's complexity, determines the degree of possible material decomposition. If the spectral resolution is too small such that materials cannot be distinguished anymore, ADJUST will not yield proper material decompositions (and neither will other methods). Moreover, for successful reconstruction and decomposition, there is a limit to the number of (differentiable) materials that can be included in the dictionary for a given spectral resolution. \\

For an uncalibrated spectral detector, the spectral dictionary can be measured, if it is not available through the manufacturer. Additionally, when working with real data, an interesting consideration for future work would be to estimate the source spectrum along with the spatial material maps and their signatures. Although our framework is based on the assumption that the spectral tomographic measurements consist of additive white noise, we can extend it to tackle Poisson noise by replacing the least-squares loss by Kullback-Leibler function. However, the bi-convexity can no longer be guaranteed and the solution obtained through AAPM may not be partially optimal. We leave this extension for future work. \\

\section*{Acknowledgements}
The authors acknowledge financial support from the Netherlands Organisation for Scientific Research (NWO), project number 639.073.506.

\section*{Code and data availability}
The source code of ADJUST, along with the RU, UR, and cJoint algorithms, are available on \jumplink{https://github.com/mzeegers/ADJUST}{https://github.com/mzeegers/ADJUST}. These MATLAB codes make use of open-source toolboxes, in particular the ASTRA toolbox\cite{AarlePalenstijn,AarlePalenstijn2}, Spot Operator toolbox\cite{BergFriedlander}, MinConf optimization package\cite{SchmidtBerg}. The micro-CT spectral dataset is available at \jumplink{https://rodare.hzdr.de/record/1627}{https://rodare.hzdr.de/record/1627}. The scripts for testing these mentioned algorithms are also made open-source in the ADJUST Github repository.

\section*{Conflict of interest}
The authors declare no conflict of interest.

\newpage

\appendices

\numberwithin{equation}{section}

\label{sec:Proofs}

\section{Proof of Theorem~\ref{thm:ProxR}} \label{sec:ProofsThm2}
Since the convex set $\mathcal{C} = \left\lbrace \mX \in \R^{M \times D} \, | \, \mX \geq 0, \, \mX \vone \leq \vone, \mX^T \vone \leq \vone \right\rbrace$ is composed of convex sets $\mathcal{C}_1 = \left\lbrace \mX \in \R^{M \times D} \, | \, \mX \geq 0, \, \mX \vone \leq \vone \right\rbrace$ and $\mathcal{C}_2 = \left\lbrace \mX \in \R^{M \times D} \, | \, \mX \geq 0, \, \mX^T \vone \leq \vone \right\rbrace$, the indicator function $\delta_{\mathcal{C}}$ can be expressed as
\[
    \delta_{\mathcal{C}}(\mX) = \delta_{\mathcal{C}_1}(\mX) + \delta_{\mathcal{C}_2} (\mX).
\]
Hence, the projection onto set $\mathcal{C}$ amounts to solving the following minimization problem
\[
    \proj_{\mathcal{C}}(\mZ) = \argmin_{\mX} \left\lbrace \frac{1}{2} \| \mX - \mZ \|_F^2 + \delta_{\mathcal{C}_1}(\mX) + \delta_{\mathcal{C}_2} (\mX) \right\rbrace.
\]
Since the cost function is the composition of two indicator functions, we can redefine a minimization problem by introducing a new slack variable $\mY$:
\[
     \underset{\mX, \mY}{\mbox{minimize}} \left\lbrace \frac{1}{2} \| \mX - \mZ \|_F^2 + \delta_{\mathcal{C}_1}(\mX) + \delta_{\mathcal{C}_2} (\mY) + \frac{1}{2} \| \mX - \mY \|_F^2 \right\rbrace,
\]
where we have penalized the slack variable $\mY$ to stay close to the original variable $\mX$ using quadratic term. The optimal point of this minimization problem must satisfy the following fixed point equation:
\begin{align*}
    \mX - \mZ + \partial \delta_{\mathcal{C}_1}(\mX) + \mX - \mY & \in \mzero, \\
    \partial \delta_{\mathcal{C}_2}(\mY) + \mY - \mX & \in \mzero, 
\end{align*}
where $\partial f$ denotes the sub-gradient of the function $f$. Hence, the fixed point iteration scheme to find the optimal point leads to
\begin{align*}
    \left(\mI + (1/2) \partial \delta_{\mathcal{C}_1} \right)\mX_{t+1} &= \frac{\mZ + \mY_{t}}{2}, \\
    \left(\mI + \partial \delta_{\mathcal{C}_2} \right) \mY_{t+1} &= \mX_{t+1},
\end{align*}
for $t = 1, \dots, T$ with setting $\mY_0$ to an arbitrary vector. Since the operation $\left(\mI + \alpha \partial \delta_{\mathcal{C}} \right)^{-1}$ with $\alpha > 0$ is equivalent to the definition of proximal operator, we can compactly rewrite the iteration scheme as
\begin{align*}
    \mY_{t+1} &= \left(\mI + \partial \delta_{\mathcal{C}_2} \right)^{-1} \left( \left(\mI + (1/2) \partial \delta_{\mathcal{C}_1} \right)^{-1} \left( \frac{\mZ + \mY_{t}}{2} \right) \right), \\
    &= \proj_{\mathcal{C}_2} \left( \proj_{\mathcal{C}_1} \left( \frac{\mZ + \mY_{t}}{2} \right) \right).
\end{align*}

\section{Bi-convexity of ADJUST and partial optimality}
\label{sec:Biconvexity}

In this section, we show that the optimization problem~(\ref{eq:ADJUST}) is bi-convex. We start with the definitions related to bi-convexity.
\begin{defn}[Bi-convex set] \label{def:biconvex_set}
    A set $\setB \subset \setX \times \setY$ is bi-convex on $\setX \times \setY$ if $\setB_x = \{y \in \setY: (x,y) \in \setB \}$ is convex for every $x\in \setX$ and $\setB_y = \{x \in \setX: (x,y) \in \setB\}$ is convex for every $y \in \setY$.
\end{defn}
\begin{defn}[Bi-convex function] \label{def:biconvex_function}
    A function $\mathcal{F}: \setB \to \mathbb{R}$ on a bi-convex set $\setB \subseteq \setX \times \setY$ is bi-convex if and only if for every fixed $y$, the function $\mathcal{F}(x,\cdot): \setB_x \to \R$ is convex on $\setB_x$, and for every fixed $x$, the function $\mathcal{F}(\cdot, y): \setB_y \to \R$ is convex on $\setB_y$.
\end{defn}
\begin{defn}[Bi-convex optimization problem] \label{def:biconvex_opt}
    A minimization problem of the form 
    \[ 
        \underset{x,y}{\mbox{minimize}} \quad \mathcal{F}(x,y) \quad \mbox{subject to} \quad x,y \in \setB
    \]
    is bi-convex if the set $\setB$ is bi-convex on $\setX \times \setY$ and the objective function $\mathcal{F}$ is bi-convex on $\setB$.
\end{defn}

Therefore, to show bi-convexity of problem~(\ref{eq:ADJUST}), we need to show that the constraint set $\mathcal{C}_A \times \mathcal{C}_R$ is bi-convex on $\R^{N \times M} \times \R^{M \times D}$, and the function $\mathcal{J}: \R^{N \times M} \times \R^{M \times D} \to \R $ is a bi-convex function.

\begin{lem}\label{lem:convex_set}
    The set $\setB \triangleq \mathcal{C}_A \times \mathcal{C}_R$ is bi-convex on $\R^{N \times M} \times \R^{M \times D}$.
\end{lem}
\begin{proof}
Since the set $\setB$ is partitioned into two independent sets $\mathcal{C}_A$ and $\mathcal{C}_R$, we only need to show that these sets are convex. The set 
\[
 \mathcal{C}_A = \bigg\lbrace \mX \in \R^{N \times M} \, | \, x_{ij} \geq 0, \, \sum_{j=1}^{M} x_{ij} \leq 1 \bigg\rbrace
\]
is a convex set on $\R^{N \times M}$ since it is an intersection of non-negative orthant ($x_{ij} \geq 0$) with $N$ number of hyperplanes ($\sum\nolimits_{j=1}^{M} x_{ij} \leq 1 $) (see 2.2.4 of \cite{BoydVanden}). Similarly, the set 
\[
 \mathcal{C}_R = \bigg\lbrace \mX \in \R^{M \times D} \, | \, x_{ij} \geq 0, \, \sum_{j=1}^D x_{ij} = 1, \, \sum_{i=1}^M x_{ij} \leq 1\bigg\rbrace,
\]
is a convex set on $\R^{M \times D}$ because it is an intersection of non-negative orthant ($x_{ij} \geq 0$) with $M$ number of hyperplanes ($\sum\nolimits_{j=1}^D x_{ij} = 1$) and $D$ number of halfspaces ($\sum\nolimits_{i=1}^M x_{ij} \leq 1$). Hence, from definition~\ref{def:biconvex_set}, the set $\setB = \mathcal{C}_A \times \mathcal{C}_R$ is a bi-convex set on $\R^{N \times M} \times \R^{M \times D}$. 
\end{proof}

\begin{lem}\label{lem:convex_fn}
    The function $\mathcal{J}(\mA,\mR) = \frac{1}{2} \| \mY - \mW \mA \mR \mT \|_F^2$ is bi-convex.
\end{lem}
\begin{proof}
First, we rewrite the function in the form
\begin{align*}
    \mathcal{J}(\mA,\mR) &= \frac{1}{2} \| \mY - \mW \mA \mR \mT \|_F^2, \\
    &= \frac{1}{2} \Tr((\mY - \mW \mA \mR \mT)(\mY - \mW \mA \mR \mT)^T) \qquad \qquad \|\mX\|_F^2 = \Tr(\mX\mX^T), \\
    &=  \frac{1}{2}\underbrace{\Tr \left( \mT^T \mR^T \mA^T \mW^T \mW \mA \mR \mT \right)}_{\mathcal{P}(\mA,\mR)} - \underbrace{\Tr \left(\mY^T \mW \mA \mR \mT \right)}_{\mathcal{Q}(\mA,\mR)} + \frac{1}{2} \| \mY \|_F^2.
\end{align*}
Hence, to show that $\mathcal{J}(\mA,\mR)$ is bi-convex, we need to show that $\mathcal{P}(\mA,\mR)$ and $\mathcal{Q}(\mA,\mR)$ are bi-convex. 

We first show the bi-convexity of $\mathcal{Q}(\mA,\mR)$. To do so, fix $\overline{\mA} \in \mathcal{C}_A$. Now, let $\mR_1, \mR_2 \in \mathcal{C}_R$ and $\lambda \in (0,1)$. Then we have
\begin{align*}
    \lambda \mathcal{Q}(\overline{\mA},\mR_1) + (1-\lambda) \mathcal{Q}(\overline{\mA},\mR_2)
    &= \lambda \Tr(\mY^T\mW \overline{\mA} \mR_1 \mT) + (1-\lambda)\Tr(\mY^T\mW \overline{\mA} \mR_2 \mT) \\
    &= \Tr(\lambda \mY^T\mW \overline{\mA} \mR_1 \mT) + \Tr((1-\lambda)\mY^T\mW \overline{\mA} \mR_2 \mT)\\
    &= \Tr(\lambda \mY^T\mW \overline{\mA} \mR_1 \mT + (1-\lambda)\mY^T\mW \overline{\mA} \mR_2 \mT)\\
    &= \Tr(\mY^T\mW \overline{\mA} (\lambda \mR_1 + (1-\lambda)\mR_2) \mT)\\
    &= \mathcal{Q}(\overline{\mA},\lambda \mR_1 + (1-\lambda)\mR_2)
\end{align*}
Hence, $\mathcal{Q}(\overline{\mA},\mR)$ is a convex function over $\R^{M \times D}$ for every $\mA \in \mathcal{C}_A$. Similarly, fixing $\overline{\mR} \in \mathcal{C}_R$ and using an analogous deduction as above shows that 
\begin{equation*}
    \lambda \mathcal{Q}(\mA_1,\overline{\mR}) + (1-\lambda) \mathcal{Q}(\mA_2,\overline{\mR})
    = \mathcal{Q}(\lambda \mA_1 + (1-\lambda)\mA_2,\overline{\mR})
\end{equation*}
for every $\mA_1, \mA_2 \in \mathcal{C}_A$ and $\lambda \in (0,1)$. Hence, $\mathcal{Q}(\mA,\overline{\mR})$ is a convex function over $\R^{N \times M}$ for every $\mR \in \mathcal{C}_R$. This shows that $\mathcal{Q}(\mA,\mR)$ is bi-convex.\\

Next, we show the bi-convexity of $\mathcal{P}(\mA,\mR)$. Thus, fix $\overline{\mA} \in \mathcal{C}_A$. Now, to show that $\mathcal{P}(\overline{\mA},\mR)$ is convex, we use the first-order condition (see 3.1.4 of \cite{BoydVanden}). Let $\mQ = \overline{\mA}^T \mW^T \mW \overline{\mA}$ and $\mP = \mT \mT^T$. The first-order condition states that $\forall \, \mR_1, \mR_2 \in \R^{M \times D}$, we need
\begin{align*}
    \mathcal{P}(\overline{\mA},\mR_2) &\geq \mathcal{P}(\overline{\mA},\mR_1) + \Tr \left( (\mR_2 - \mR_1)^T \nabla_{\mR_1} \mathcal{P}(\overline{\mA},\mR_1)  \right) \\
    \Tr(\mR_2^T \mQ \mR_2 \mP ) &\geq \Tr(\mR_1^T \mQ \mR_1 \mP ) + 2 \Tr \left( (\mR_2 - \mR_1)^T \overline{\mA}^T \mW^T \mW \overline{\mA} \mR_1 \mT \mT^T \right) \\
    \Tr(\mR_2^T \mQ \mR_2 \mP ) &\geq \Tr(\mR_1^T \mQ \mR_1 \mP ) + 2 \Tr \left( (\mR_2 - \mR_1)^T \mQ \mR_1\mP \right)
\end{align*}
To arrive at this condition, let us consider
\begin{align*}
    \Tr \left( (\mR_1 - \mR_2)^T \mQ (\mR_1 - \mR_2) \mP \right)
    &= \Tr (\mR_1^T \mQ \mR_1 \mP) + \Tr (\mR_2^T \mQ \mR_2 \mP) - \Tr (\mR_1^T \mQ \mR_2 \mP) - \Tr(\mR_2^T \mQ \mR_1 \mP) \\
    &= \Tr (\mR_1^T \mQ \mR_1 \mP) + \Tr (\mR_2^T \mQ \mR_2 \mP) - \Tr (\mR_1^T \mQ \mR_2 \mP) - \Tr(\mR_2^T \mQ^T \mR_1 \mP^T) \\
    &= \Tr (\mR_1^T \mQ \mR_1 \mP) + \Tr (\mR_2^T \mQ \mR_2 \mP) - 2 \Tr (\mR_1^T \mQ \mR_2 \mP).
\end{align*}
Since $\mQ$ and $\mP$ are positive semi-definite matrices, we have $\Tr \left( (\mR_1 - \mR_2)^T \mQ (\mR_1 - \mR_2) \mP \right) \geq 0$. Hence, we get
\[
    \Tr (\mR_1^T \mQ \mR_1 \mP) + \Tr (\mR_2^T \mQ \mR_2 \mP) \geq 2 \Tr (\mR_1^T \mQ \mR_2 \mP),
\]
which proves the first-order condition. Similarly, we can show that  $\mathcal{P}(\mA,\overline{\mR})$ is a convex function over $\R^{N \times M}$ for fixed $\overline{\mR} \in \mathcal{C}_R$. Hence, $\mathcal{P}(\mA,\mR)$ is a bi-convex function.

Since $\mathcal{P}(\mA,\mR)$ and $\mathcal{Q}(\mA,\mR)$ are bi-convex functions, their linear combination is also a bi-convex function \cite{GorskiPfeuffer}. Hence, we prove that $\mathcal{J}(\mA,\mR)$ is bi-convex.
\end{proof}

\begin{cor}
    The optimization problem~(\ref{eq:ADJUST}) is bi-convex.
\end{cor}
\begin{proof}
    Since the cost function $\mathcal{J}(\mA,\mR) = \frac{1}{2}\| \mY - \mW \mA \mR \mT \|_F^2$ is bi-convex (Lemma~\ref{lem:convex_fn}) and $\mathcal{C}_A \times \mathcal{C}_R$ is a bi-convex set (Lemma~\ref{lem:convex_set}), the optimization problem
    \[
        \mbox{minimize} \quad \mathcal{J}(\mA,\mR)  \quad \mbox{subject to} \quad \mA \in \mathcal{C}_A, \, \mR \in \mathcal{C}_R
    \]
    is bi-convex (from definition~\ref{def:biconvex_opt}).
\end{proof}

Bi-convex optimization problems may have a large number of local minima as they are global optimization problems in general \cite{GorskiPfeuffer}. Since we are interested in finding a stationary point of~(\ref{eq:ADJUST}), we define the notion of partial optimality.
\begin{defn}[Partial optimality]
    Let $\mathcal{F}: \setX \times \setY \mapsto \R$ be a given function and let $(x^\star, y^\star) \in \setX \times \setY$. Then, $(x^\star, y^\star)$ is called a partial optimum of $\mathcal{F}$ on $\setX \times \setY$, if
    \[
        \mathcal{F}(x^\star, y^\star) \leq \mathcal{F}(x, y^\star) \quad \forall \, x \in \setX \quad \mbox{and} \quad \mathcal{F}(x^\star, y^\star) \leq \mathcal{F}(x^\star, y) \quad \forall \, y \in \setY.
    \]
\end{defn}
It is easy to show that a partial optimum $z^\star = (x^\star,y^\star)$ is also a stationary point of $\mathcal{F}$ in $\setX \times \setY$ if $\mathcal{F}$ is differentiable at $z^\star$. Also, the converse is true \cite{GorskiPfeuffer}. Finally, the following theorem (adapted from \cite{WendellHurter}) connects the local optimality (\ie, stationary points) to the partial optimality:
\begin{thm}
    Let $(\mA^\star, \mR^\star) \in \mathcal{C}_A \times \mathcal{C}_R$ be a partial optimum of $\mathcal{J}(\mA, \mR) = \frac{1}{2} \| \mY - \mW \mA \mR \mT \|_F^2$. Furthermore, let $\setU (\mR^\star)$ denote the set of all optimal solutions to~(\ref{eq:ADJUST}) with $\mR = \mR^\star$ and let $\setV (\mA^\star)$ be the set of optimal solutions to~(\ref{eq:ADJUST}) with $\mA = \mA^\star$. If $(\mA^\star, \mR^\star)$ is a local optimal solution to~(\ref{eq:ADJUST}), then it necessarily holds that
    \[
        \mathcal{J}(\mA^\star, \mR^\star) \leq \mathcal{J}(\mA, \mR) \quad \forall \, \mA \in \setU(\mR^\star), \, \mR \in \setV(\mA^\star).
    \]
\end{thm}
This theorem implies that the natural solution of any alternating minimization algorithm will lead to a partial optimal solution. The proof of the theorem can be found in \cite{WendellHurter}.

\section{Derivation of AAPM}
\label{sec:AAPM}

First, we rephrase the original ADJUST problem in the following form:\begin{align*}
    \underset{\mA, \mR}{\mbox{minimize}} \quad & \mathcal{J}(\mA,\mR) + \delta_{\mathcal{C}_{A}}(\mA) + \delta_{\mathcal{C}_{R}}(\mR), \\
    \mbox{subject to} \quad & \mW \! \mA \mR \mT = \mY,
\end{align*}
where $\delta_{\mathcal{C}}$ is an extended value function for the constraint set $\mathcal{C}$ that is $0$ when constraint is satisfied and $\infty$ otherwise. Here, we have introduced the constraints on the misfit between simulated and true measurements in the linear form. The Lagrangian for this optimization problem reads
\begin{align}
    \mathcal{L}(\mA,\mR,\mU) &=  \mathcal{J}(\mA,\mR)   + \delta_{\mathcal{C}_{A}}(\mA) + \delta_{\mathcal{C}_{R}}(\mR) +  \left\langle \mU,  \mY - \mW \! \mA \mR \mT \right\rangle  \\[1ex]
    & = \underbrace{\mathcal{J}(\mA,\mR) + \left\langle \mU, \mY - \mW \! \mA \mR \mT \right\rangle \vphantom{\delta_{\mathcal{C}_{R}}}}_{\triangleq \widetilde{\mathcal{J}}(\mA, \mR, \mU)} + \delta_{\mathcal{C}_{A}}(\mA) + \delta_{\mathcal{C}_{R}}(\mR)
    \label{eq:Lagrangian}
\end{align}
where $\mU \in \R^{J \times C}$ is a Lagrange multiplier for constraint $\mW \mA \mR \mT = \mY$. The Lagrange multiplier $\mU$ can also be thought of as a running-sum-of-error as it captures the misfit between the true measurements and simulated measurements. The goal is to find a saddle point of this Lagrangian, since the saddle point will give the optimal solution to~(\ref{eq:ADJUST}). The saddle point of the Lagrangian is given by
\[
    \left(\mA^\star, \mR^\star, \mU^\star \right) = \argmax_{\mU} \argmin_{\mA, \mR} \, \mathcal{L}(\mA,\mR,\mU).
\]
It is important to note that the Lagrangian is non-differentiable due to the presence of $\delta_{\mathcal{C}_{A}}$ and $\delta_{\mathcal{C}_{R}}$. Since the min-max problem can not be solved using a simple gradient-based iterative scheme due to non-differentiability of the Lagrangian, we need to make use of proximal alternating iterative algorithm. To derive such scheme, we approximate the Lagrangian~(\ref{eq:Lagrangian}) near point $\left(\mA_{k},\mR_{k}, \mU_{k} \right)$ using the Taylor series for the differentiable function $\widetilde{\mathcal{J}}(\mA,\mR,\mU)$. This approximation reads
\begin{align}
    \begin{split} 
    \mathcal{L}(\mA,\mR,\mU) & \approx \widetilde{\mathcal{L}}(\mA,\mR,\mU | \mA_k,\mR_k,\mU_k) \\
    &= \widetilde{\mathcal{J}}(\mA_{k},\mR_{k},\mU_{k}) +  \\
    & \quad \big\langle \nabla_{\mR}\widetilde{\mathcal{J}}(\mA_k, \mR_k, \mU_k), \mR - \mR_k  \big\rangle + 1/(2 \alpha) \| \mR - \mR_{k} \|_F^2 \, + \\
    & \quad \big\langle \nabla_{\mA}\widetilde{\mathcal{J}}(\mA_k, \mR_k, \mU_k), \mA - \mA_k \big\rangle + 1/(2 \beta) \| \mA - \mA_{k} \|_F^2  \, + \\
    & \quad \delta_{\mathcal{C}_{A}}(\mA) + \delta_{\mathcal{C}_{R}}(\mR) ,
    \end{split}
    \label{eq:ApproxLagrangian}
\end{align}
where $\alpha$ and $\beta$ are the Lipschitz constants of the partial gradients of $\widetilde{\mathcal{J}}(\mA,\mR,\mU)$ with respect to $\mA$ and $\mR$ respectively. This approximation leads to the following alternating scheme where we minimize with respect to the primal variables $\mA$ and $\mR$, and maximize with respect to the dual variable $\mU$:
\begin{align*}
    \mR_{k+1} &= \argmin_{\mR} \widetilde{\mathcal{L}}(\mA,\mR,\mU | \mA_k,\mR_k,\mU_k) \\
    \mA_{k+1} &= \argmin_{\mA} \widetilde{\mathcal{L}}(\mA,\mR,\mU | \mA_k,\mR_{k+1},\mU_k) \\
    \mU_{k+1} &= \mU_{k} + \rho \left( \mW \! \mA_{k+1} \mR_{k+1} \mT - \mY \right)
\end{align*}
with $k=0,\dots,K$, and $\rho > 0$ is the acceleration parameter. This alternating scheme requires initial values of $\mR$ and $\mA$, while the initial value of $\mU$ can be set to $\mathbf{0}$. We update the dual variable $\mU$ using the linearized ascent, a standard technique used by many alternating methods, \eg, alternating direction method of multipliers~\cite{BoydParikh}. Since the approximate Lagrangian~(\ref{eq:ApproxLagrangian}) is composed of quadratic term and non-smooth terms for $\mA$ and $\mR$, we can express the iterates using proximal operations. To derive $\mR$, we use the identity $\| \mX+\mY \|_F^2 = \|\mX\|_F^2 + \|\mY\|_F^2 + 2 \big\langle \mX, \mY \big\rangle$, or equivalently, $\big\langle \mX, \mY \big\rangle + \frac{1}{2}\|\mY\|_F^2 = \frac{1}{2}\| \mX+\mY \|_F^2 - \frac{1}{2}\|\mX\|_F^2$. The derivation is now as follows:
\begin{align}
    \begin{split} 
    \mR_{k+1} &= \argmin_{\mR} \widetilde{\mathcal{L}}(\mA,\mR,\mU | \mA_k,\mR_k,\mU_k), \\
    &= \argmin_{\mR} \left\lbrace \big\langle \nabla_{\mR}\widetilde{\mathcal{J}}(\mA_k, \mR_k, \mU_k), \mR - \mR_k  \big\rangle + \frac{1}{2 \alpha} \| \mR - \mR_{k} \|_F^2 + \delta_{\mathcal{C}_{R}}(\mR)  \right\rbrace, \\
    &= \argmin_{\mR} \bigg\lbrace \frac{1}{\alpha} \big\langle \alpha \nabla_{\mR}\widetilde{\mathcal{J}}(\mA_k, \mR_k, \mU_k), \mR - \mR_k  \big\rangle + \frac{1}{2 \alpha} \| \mR - \mR_{k} \|_F^2 + \delta_{\mathcal{C}_{R}}(\mR)  \bigg\rbrace, \\
    &= \argmin_{\mR} \Bigg\lbrace \frac{1}{\alpha}  \underbrace{\left( \big\langle \alpha \nabla_{\mR}\widetilde{\mathcal{J}}(\mA_k, \mR_k, \mU_k), \mR - \mR_k  \big\rangle + \frac{1}{2} \| \mR - \mR_{k} \|_F^2\right)}_{\text{applying the identity with } \mX \triangleq \alpha \nabla_{\mR}\widetilde{\mathcal{J}}(\mA_k, \mR_k, \mU_k), \, \mY \triangleq \mR - \mR_k} + \delta_{\mathcal{C}_{R}}(\mR)  \Bigg\rbrace, \\
    &= \argmin_{\mR} \Bigg\lbrace  \frac{1}{2\alpha}\| \alpha \nabla_{\mR}\widetilde{\mathcal{J}}(\mA_k, \mR_k, \mU_k) + \mR - \mR_k\|_F^2\\
    & \qquad \qquad \underbrace{- \frac{1}{2}\|\alpha \nabla_{\mR}\widetilde{\mathcal{J}}(\mA_k, \mR_k, \mU_k)\|_F^2}_{\text{independent of $\mR$}} + \delta_{\mathcal{C}_{R}}(\mR)  \Bigg\rbrace, \\
    &= \argmin_{\mR} \left\lbrace \frac{1}{2 \alpha} \| \mR - \mR_{k} + \alpha \nabla_{\mR}\widetilde{\mathcal{J}}(\mA_k, \mR_k, \mU_k)\|_F^2 + \delta_{\mathcal{C}_{R}}(\mR)  \right\rbrace, \\
    &=  \prox_{\delta_{\mathcal{C}_{R}}} \left( \mR_{k} - \alpha \nabla_{\mR} \widetilde{\mathcal{J}}(\mA_k, \mR_k, \mU_k) \right),
    \end{split}
    \label{eq:updateRderivation}
\end{align}
where the proximal for a function $f: \R^n \mapsto \R$ reads
\[
    \prox_{\gamma f} (\vz) = \argmin_{\vx \in \R^n} \left\lbrace \frac{1}{2 \gamma} \| \vx - \vz \|_2^2 + f(\vx ) \right\rbrace 
\]
with $\gamma > 0$. The proximal operator allows us to work with non-differentiable functions. Moreover, proximal operators for many functions have explicit expressions, making it a very computationally-friendly tool. The proximal operator for $\delta_{\mathcal{C}}$ with $\mathcal{C} \subset \R^n$ takes the following form:
\[
    \prox_{\delta_{\mathcal{C}}}(\vz)
    = \argmin_{\vx \in \R^n} \left\lbrace \frac{1}{2}\|\vx - \vz \|_2^2 + \delta_{\mathcal{C}}(\vx) \right\rbrace
\]
Indeed, the proximal operator of a $\delta_{\mathcal{C}}$ is just an orthogonal projection of a vector onto the set $\mathcal{C}$. If the set $\mathcal{C}$ is convex, the proximal point is unique. Similar to~(\ref{eq:updateRderivation}), we can explicitly write down the update of $\mA$ in terms of the proximal operator.

\section{Gradient Computations} 
\label{sec:AppendixGradientComputations}

Here we show how the gradients are computed at the final comments in Section~\ref{sec:NumericalOptimization}. We only show the derivation of $\nabla_{\mA} \widetilde{\mathcal{J}} ( \mA, \mR, \mU )$ since the derivation of $\nabla_{\mR} \widetilde{\mathcal{J}} ( \mA, \mR, \mU )$ is very similar.
\begin{align*}
    \nabla_{\mA} \widetilde{\mathcal{J}} ( \mA, \mR, \mU )
    &= \nabla_{\mA} \left( \frac{1}{2} \| \mY - \mW \! \mA \mR \mT \|_F^2 + \big\langle \mU, \mY - \mW\mA\mR\mT \big\rangle \right) \\
    &= \nabla_{\mA} \left( \frac{1}{2} \| \mY - \mW \! \mA \mR \mT \|_F^2 \right) + \nabla_{\mA} \big\langle \mU, \mY - \mW\mA\mR\mT \big\rangle  \\
    &\stackrel{*}{=} \frac{1}{2} \nabla_{\mA} \left( \|\mY\|_F^2 + \| \mW \mA \mR \mT \|_F^2 - 2 \Tr \left( \mY^T \mW \mA \mR \mT \right)  \right) \\
    &\ \ \ + \nabla_{\mA} \Tr \left(\mU^T(\mY - \mW\mA\mR\mT) \right) & \qquad \big\langle \mX, \mY \big\rangle = \Tr(\mA^T\mB)\\
    &= \frac{1}{2} \nabla_{\mA} \left( \| \mW \mA \mR \mT \|_F^2 \right) - \nabla_{\mA} \left( \Tr \left( \mY^T \mW \mA \mR \mT \right) \right) \\
    &\ \ \ + \nabla_{\mA} \Tr(\mU^T\mY) - \nabla_{\mA} \Tr(\mU^T\mW\mA\mR\mT)  \\
    &= \frac{1}{2} \nabla_{\mA} \left( \Tr \left( \mT^T \mR^T \mA^T \mW^T \mW \mA \mR \mT \right) \right) - \mW^T \mY \mT^T \mR^T \\
    &\ \ \ - (\mU\mT\mW)^T(\mR\mT)^T& \qquad \frac{\partial}{\partial \mX} \Tr \left( \mA \mX \mB \right) = \mA^T \mB^T \\
    &= \mW^T \left( \mW \mA \mR \mT \right)\mT^T \mR^T  - \mW^T \left( \mY \right) \mT^T \mR^T  - \mW^T\mU\mT^T\mR \\
    &= \mW^T \left( \mW \mA \mR \mT - \mY \right) \mT^T \mR^T - \mW^T\mU\mT^T\mR.
\end{align*}
In the third step (*), we use the following identity:
\begin{align*}
    \|\mX - \mY\|_F^2 &= \Tr \left( (\mX - \mY)^T (\mX - \mY) \right) \\
    &= \Tr \left( (\mX^T - \mY^T) (\mX - \mY) \right) \\
    &= \Tr \left( \mX^T \mX - \mY^T \mX - \mX^T \mY + \mY^T \mY \right) \\
    &= \Tr \left( \mX^T \mX \right) - \Tr \left( \mY^T \mX \right) - \Tr \left( \mX^T \mY \right) + \Tr \left( \mY^T \mY \right) \\
    &= \|\mX\|_F^2 + \|\mY\|_F^2 - 2 \Tr(\mY^T \mX) & \quad \left( \mY^T \mX \right)^T = \mX^T \mY
 \end{align*}

\section{Dictionary matrix} \label{sec:Dictionary}

In this section, we list the 42 materials that are used in the dictionary matrix $\mT$ for the Disks and Shepp-Logan phantoms. The spectra are retrieved from the National Institute for Standards and Technology (NIST)~\cite{NIST, HubbellSeltzer}. \\

\begin{center}
\begin{tabular}{|c|c|c|}
\hline
\begin{tabular}{@{}c@{}}Mat. \\ no. \end{tabular}  & \begin{tabular}{@{}c@{}}Material \\ name \end{tabular}  & \begin{tabular}{@{}c@{}}At. \\ no. \end{tabular} \\ \hline
23 & Vanadium & 23 \\
24 & Chromium & 24 \\
25 & Manganese & 25 \\
26 & Iron & 26 \\
27 & Cobalt & 27 \\
28 & Nickel & 28 \\
29 & Copper & 29 \\
30 & Zinc & 30 \\ 
31 & Gallium & 31 \\ 
32 & Germanium & 32 \\
33 & Arsenic & 33 \\
34 & Selenium & 34 \\
35 & Bromine & 35 \\
36 & Krypton& 36 \\
37 & Rubidium & 37 \\
38 & Strontium & 38 \\
39 & Yttrium & 39 \\
40 & Zirconium & 40 \\ 
41 & Niobium& 41 \\
42 & Molybdenum & 42 \\
43 & Technetium & 43 \\ \hline
\end{tabular} \qquad
\begin{tabular}{|c|c|c|}
\hline
\begin{tabular}{@{}c@{}}Mat. \\ no. \end{tabular}  & \begin{tabular}{@{}c@{}}Material \\ name \end{tabular}  & \begin{tabular}{@{}c@{}}At. \\ no. \end{tabular} \\ \hline
44 & Ruthenium & 44 \\
45 & Rhodium & 45 \\
46 & Palladium & 46 \\
47 & Silver & 47 \\
48 & Cadmium & 48 \\
49 & Indium & 49 \\
50 & Tin & 50 \\
51 & Antimony & 51 \\
52 & Tellurium & 52 \\
53 & Iodine & 53 \\
54 & Xenon & 54 \\
55 & Cesium & 55 \\
56 & Barium & 56 \\
57 & Lanthanum & 57 \\
58 & Cerium & 58 \\
59 & Praseodymium & 59 \\
60 & Neodymium & 60 \\
61 & Promethium & 61 \\
62 & Samarium & 62 \\
63 & Terbium & 63 \\
64 & Gadolinium & 64 \\ \hline
\end{tabular}
\end{center}

\begin{figure}[H]
    \centering
    \includegraphics[width=0.75 \textwidth]{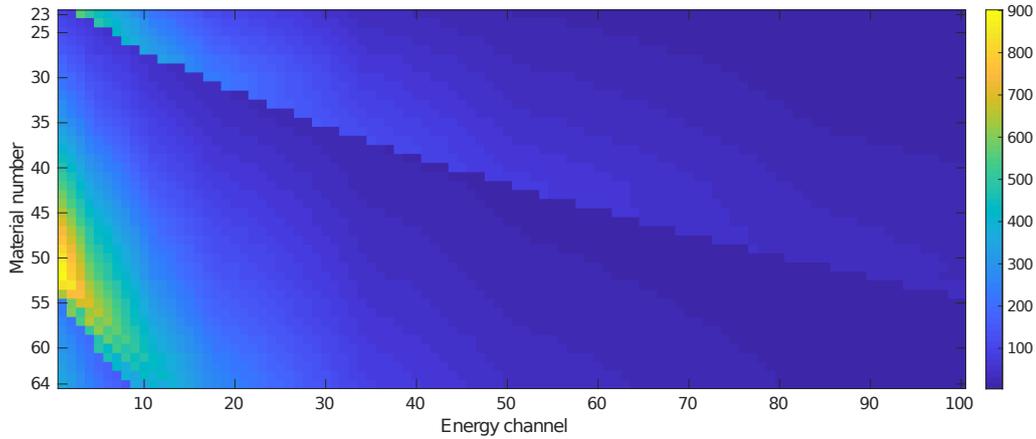}
    \caption{Dictionary matrix $\mT$: Attenuation values over 100 spectral channels for 42 materials, with energies ranging from $20$KeV to $119$KeV.}
    \label{fig:dict_matrix}
\end{figure}

\begin{figure}[H]
    \centering
    \includegraphics[width=0.5 \textwidth]{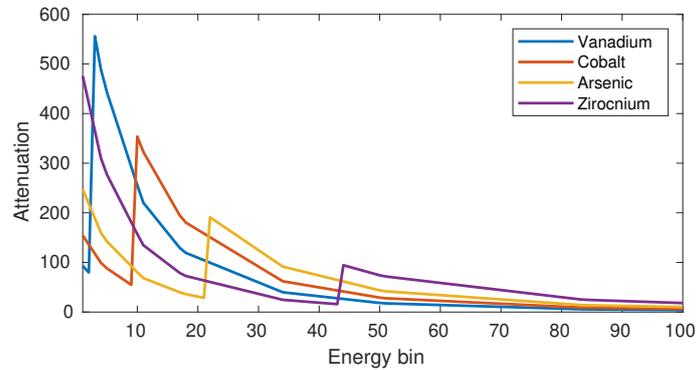}
    \caption{Attenuation values over 100 spectral channels for four materials from the dictionary matrix $\mT$, with energies ranging from $5$keV to $35$keV.}
    \label{fig:Example_Spectra}
\end{figure}

We plot the attenuation spectra for all dictionary elements for each bin within the selected range in Figure~\ref{fig:dict_matrix}. Additionally, Figure~\ref{fig:Example_Spectra} shows the spectra for a few selected materials. All of these materials have a K-edge in the considered spectral range.

\newpage
\section{Performance measures}
\label{sec:PerformanceMeasuresExt}

To assess the quality of the reconstructions that ADJUST (and the comparison methods) generates, we compare the reconstructions with the ground truth. Since for the UR, RU, cJoint and ADJUST methods the best matching reconstruction of a certain channel in the ground truth may be located in a different channel in the material map matrix, a matching that minimizes the total error over the channels needs to be carried out. Let $\mA^{\text{GT}} \in \R^{N\times M}$ be the matrix containing the ground truth material maps and $\mA^{\text{rec}} \in \R^{N\times M}$ be the reconstructed material map. We compute a matrix $\mA^{\text{error}}$ containing the mutual errors between channels of $\mA^{\text{GT}}$ and $\mA^{\text{rec}}$, defined by
\begin{equation*}
    A_{ij}^{\text{error}} = \left\|(\mA_{ki}^{\text{rec}})_{i \leq k \leq N} - (\mA_{kj}^{\text{GT}})_{1 \leq k \leq N}\right\|_2
\end{equation*}
Given this error matrix, we use an iterative greedy approach to match the channels of the $\mA^{\text{GT}}$ and $\mA^{\text{rec}}$ matrices based on their mutual channel errors. We repeatedly compute the minimum of the error matrix and remove the possibility to match the corresponding channels. To do so, let $\mathcal{M}_0^{\text{GT}} = \mathcal{M}$, $\mathcal{M}_0^{\text{rec}} = \mathcal{M}$ and $\mathcal{M}_0^{\text{match}} = \emptyset$. In each iteration $1 \leq l \leq M$, we compute
\begin{equation*}
    (i_l, j_l) = \argmin_{i \in \mathcal{M}_l^{\text{rec}} \atop j \in \mathcal{M}_l^{\text{GT}}} A_{ij}
\end{equation*}
and define $\mathcal{M}_{l+1}^{\text{rec}} = \mathcal{M}_{l}^{\text{rec}} \backslash \{i_l\}$, $\mathcal{M}_{l+1}^{\text{GT}} = \mathcal{M}_{l}^{\text{GT}} \backslash \{j_l\}$ and $\mathcal{M}_{l+1}^{\text{match}} = \mathcal{M}_{l}^{\text{match}} \cup \{(i_l,j_l\}$. Given the final channel-matching represented by $\mathcal{M}_{M}^{\text{match}}$, we compute the following three error metrics for each $(i,j) \in \mathcal{M}_M^{\text{match}}$:

\begin{itemize}[noitemsep]
    \item \textit{Mean squared error} (MSE) for each matched material pair:
        \begin{equation*}
            \text{MSE}(i,j) = \left\|(\mA_{ki}^{\text{rec}})_{i \leq k \leq N} - (\mA_{kj}^{\text{GT}})_{1 \leq k \leq N}\right\|_2^2
        \end{equation*}
    \item \textit{Peak signal-to-noise ratio} (PSNR) for each matched material pair:
        \begin{equation*}
            \text{PSNR}(i,j) = 10 \log_{10} \left( \left(\max_k{(\mA_{kj}^{\text{GT}})_{1 \leq k \leq N}}\right)^2 \slash \left\|(\mA_{ki}^{\text{rec}})_{i \leq k \leq N} - (\mA_{kj}^{\text{GT}})_{1 \leq k \leq N}\right\|_2^2 \right)
        \end{equation*}
    \item \textit{Structural similarity index} (SSIM) for each matched  material pair:
        \begin{equation*}
            \text{SSIM}(i,j) = \left( \left(2 \mu_{i} \mu_j + C_1)(2\sigma_{ij} + C_2\right) \slash \left( \mu_i^2 + \mu_j^2 + C_1 )(\sigma_i^2 + \sigma_j^2 + C_2) \right)\right)
        \end{equation*}
        with $\mu_{i}, \mu_{j}$ and $\sigma_i, \sigma_j$ being the means and the standard deviations of the matrices $(\mA_{ki}^{\text{rec}})_{i \leq k \leq N}$ and $(\mA_{kj}^{\text{GT}})_{1 \leq k \leq N}$ respectively, with $\sigma_{ij}$ being the cross-correlation between these two matrices, and with $C_1 = (0.01L)^2$, $C_2 = (0.03L)^2$ and $L = 1$.
\end{itemize}
The averages of the MSE, PSNR and SSIM over all materials are then given by:\begin{align*}
    \text{MSE}_{\text{avg}} &= \sum_{(i,j) \in \mathcal{M}_M^{\text{match}}} \text{MSE}(i,j)/ M, \\
    \text{PSNR}_{\text{avg}} &= \sum_{(i,j) \in \mathcal{M}_M^{\text{match}}} \text{PSNR}(i,j)/M , \\
    \text{SSIM}_{\text{avg}} &= \sum_{(i,j) \in \mathcal{M}_M^{\text{match}}} \text{SSIM}(i,j)/M .
\end{align*}


\section{Numerical Studies: Comparison of methods}
\label{sec:NumStudComparison}
\begin{figure}[H]
    \centering
    \renewcommand{\arraystretch}{1.0}
    \begin{tabular}{c | cccc}
        {\bf True} & {\bf RU} & {\bf UR} & {\bf cJoint} & {\bf ADJUST} \\
        \includegraphics[width=0.15\textwidth,cfbox=col1 1pt 0pt]{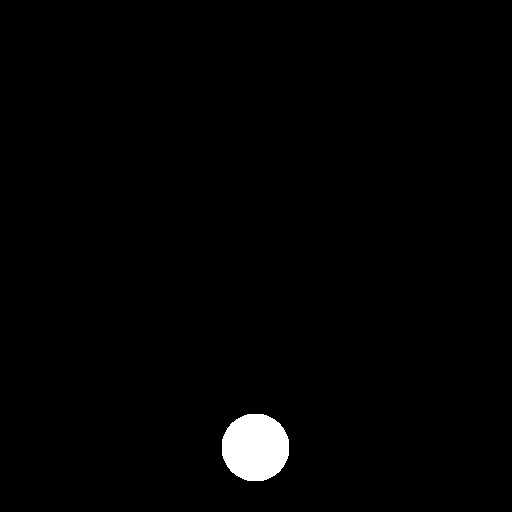} \quad & \includegraphics[width=0.15\textwidth,cfbox=col1 1pt 0pt]{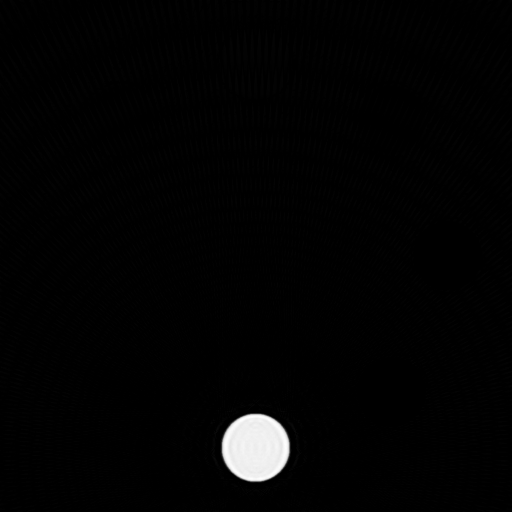} & \includegraphics[width=0.15\textwidth,cfbox=col1 1pt 0pt]{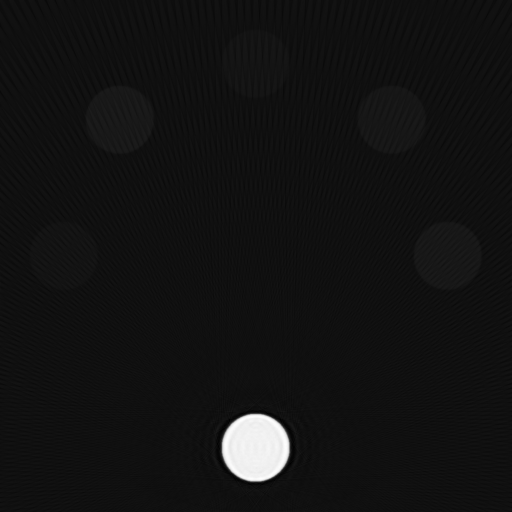} & \includegraphics[width=0.15\textwidth,cfbox=col1 1pt 0pt]{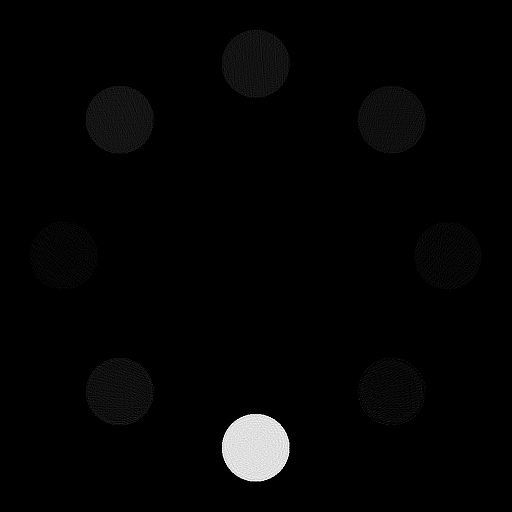} & \includegraphics[width=0.15\textwidth,cfbox=col1 1pt 0pt]{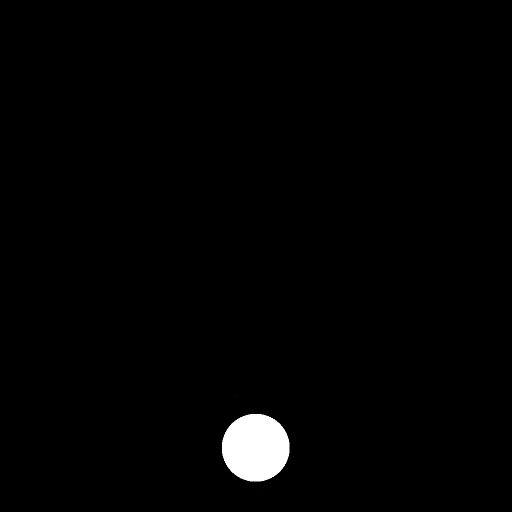} \\
        \includegraphics[width=0.15\textwidth,cfbox=col2 1pt 0pt]{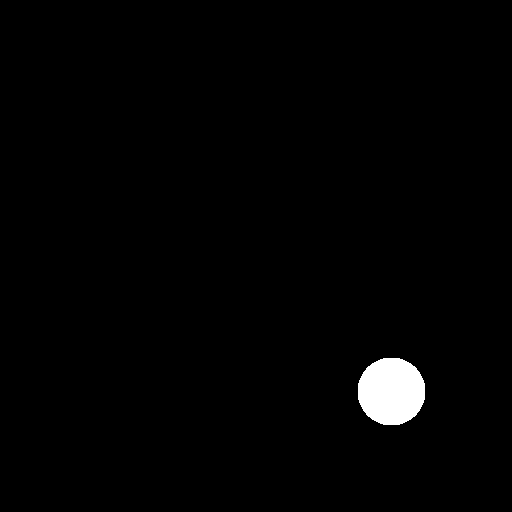} \quad & \includegraphics[width=0.15\textwidth,cfbox=col2 1pt 0pt]{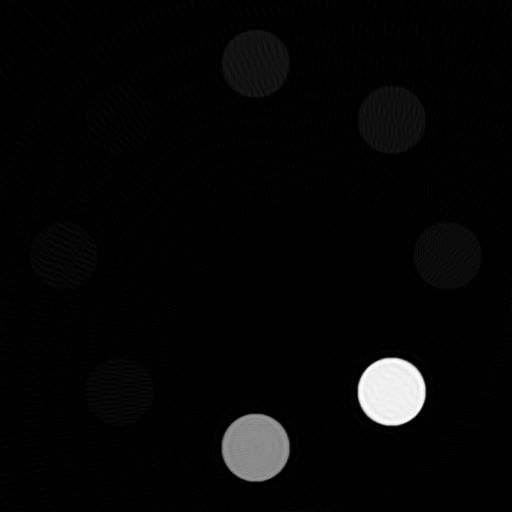} & \includegraphics[width=0.15\textwidth,cfbox=col2 1pt 0pt]{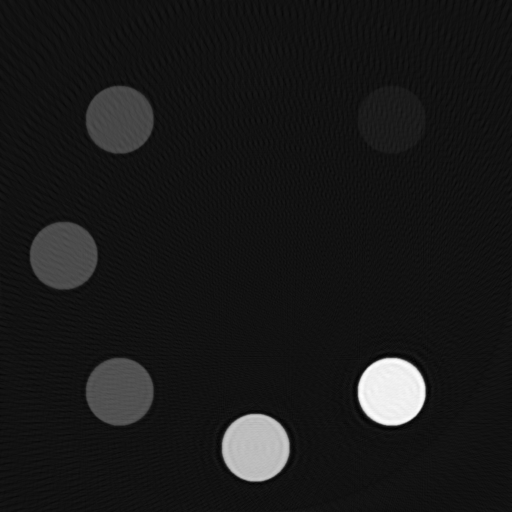} & \includegraphics[width=0.15\textwidth,cfbox=col2 1pt 0pt]{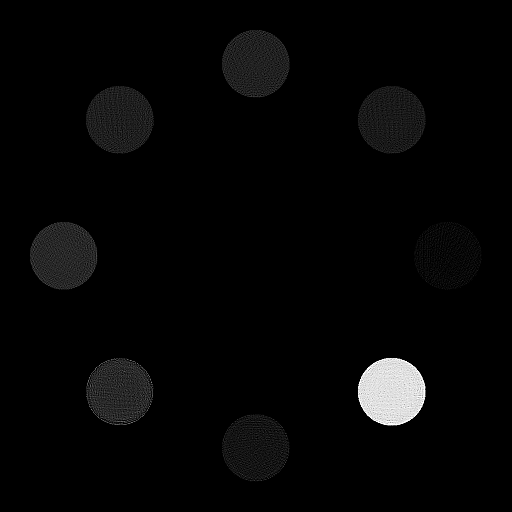} & \includegraphics[width=0.15\textwidth,cfbox=col2 1pt 0pt]{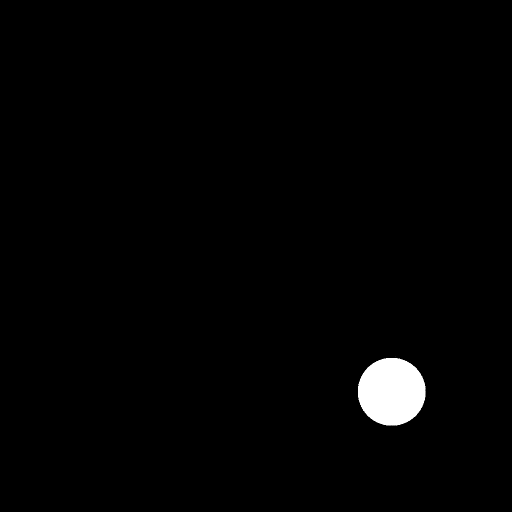} \\
        \includegraphics[width=0.15\textwidth,cfbox=col3 1pt 0pt]{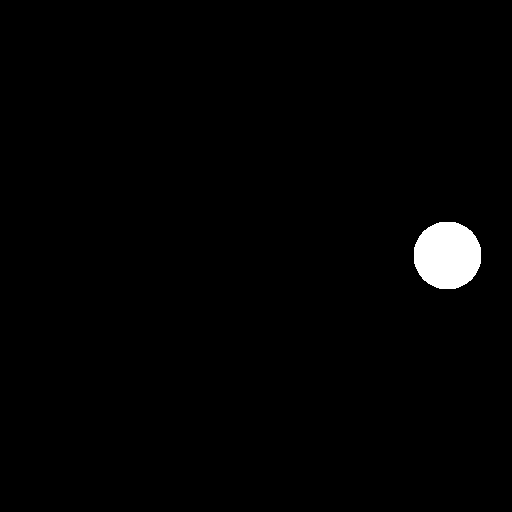} \quad & \includegraphics[width=0.15\textwidth,cfbox=col3 1pt 0pt]{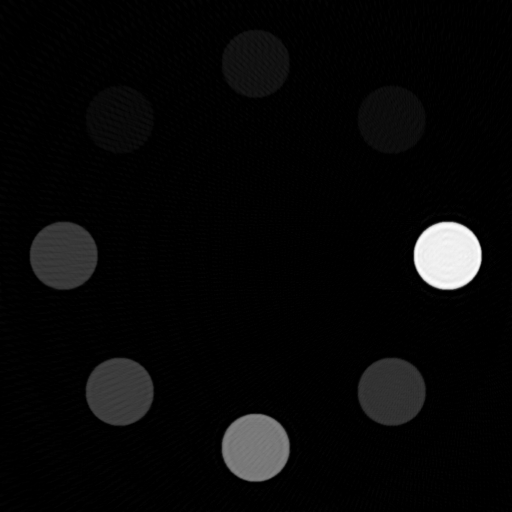} & \includegraphics[width=0.15\textwidth,cfbox=col3 1pt 0pt]{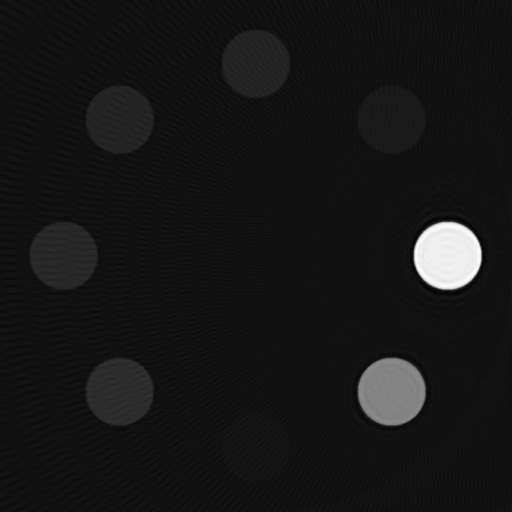} & \includegraphics[width=0.15\textwidth,cfbox=col3 1pt 0pt]{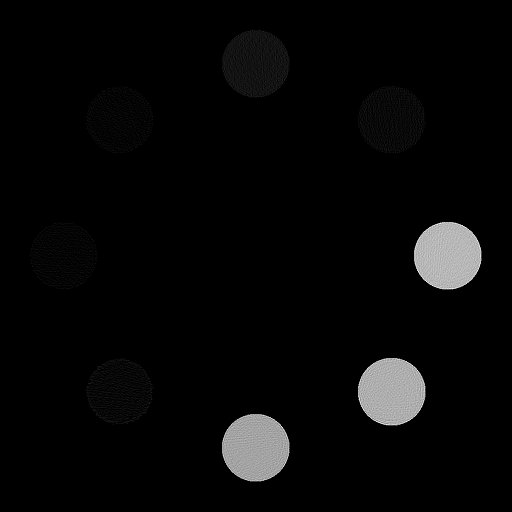} & \includegraphics[width=0.15\textwidth,cfbox=col3 1pt 0pt]{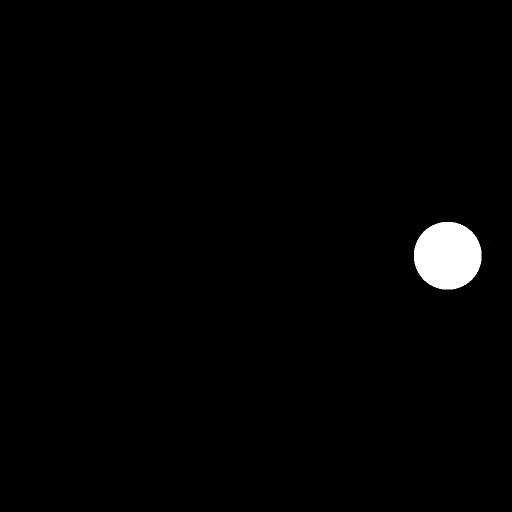} \\
        \includegraphics[width=0.15\textwidth,cfbox=col4 1pt 0pt]{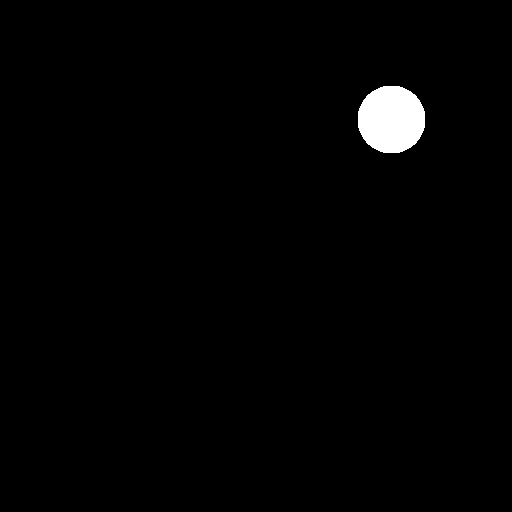} \quad & \includegraphics[width=0.15\textwidth,cfbox=col4 1pt 0pt]{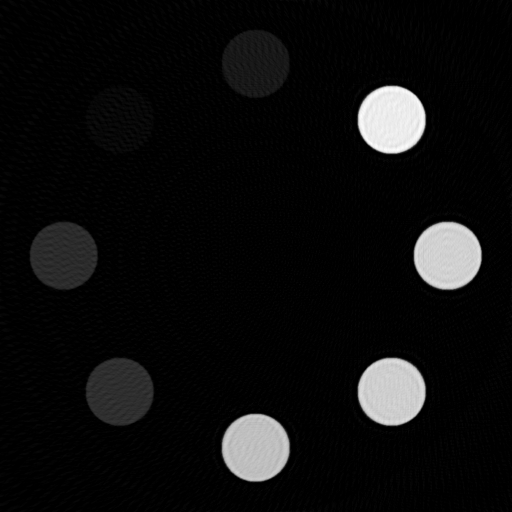} & \includegraphics[width=0.15\textwidth,cfbox=col4 1pt 0pt]{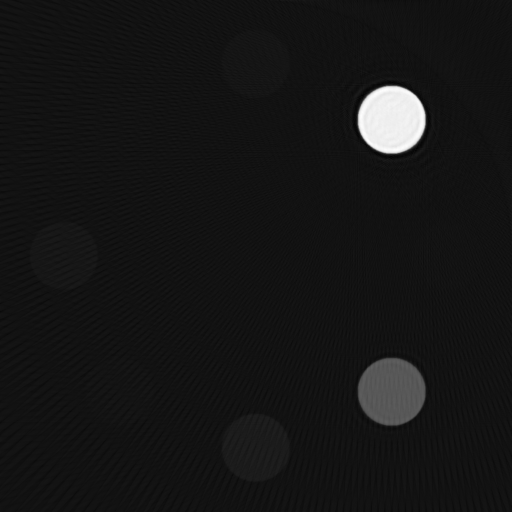} & \includegraphics[width=0.15\textwidth,cfbox=col4 1pt 0pt]{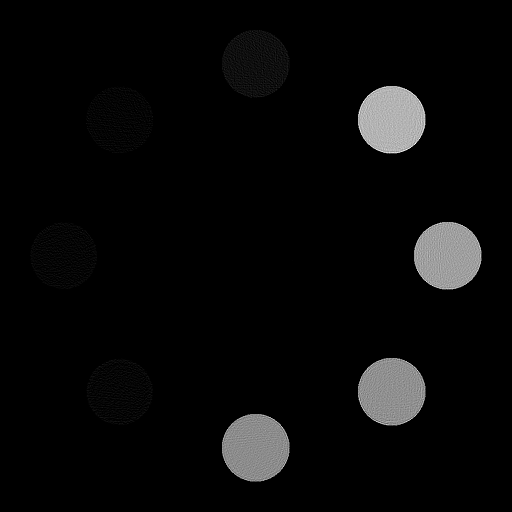} & \includegraphics[width=0.15\textwidth,cfbox=col4 1pt 0pt]{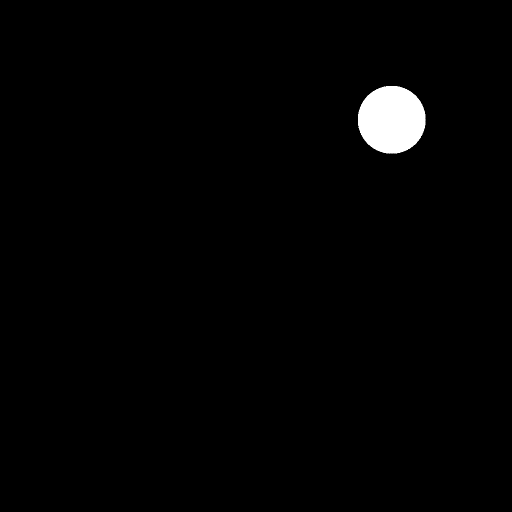} \\
        \includegraphics[width=0.15\textwidth,cfbox=col5 1pt 0pt]{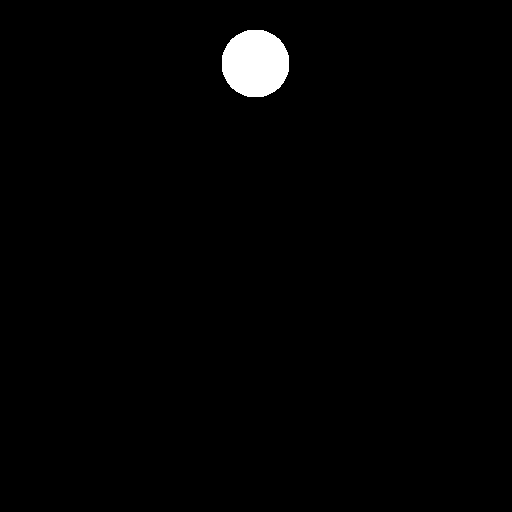} \quad & \includegraphics[width=0.15\textwidth,cfbox=col5 1pt 0pt]{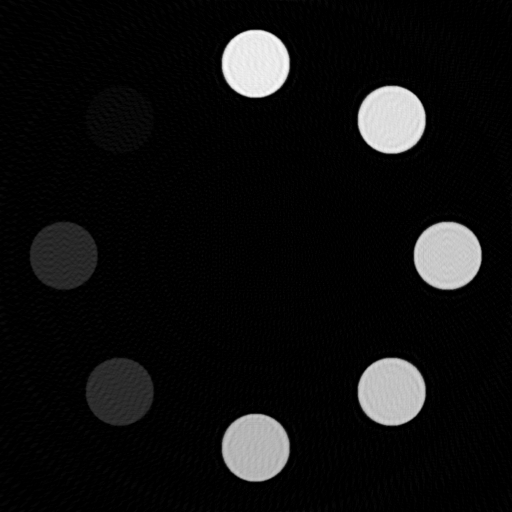} & \includegraphics[width=0.15\textwidth,cfbox=col5 1pt 0pt]{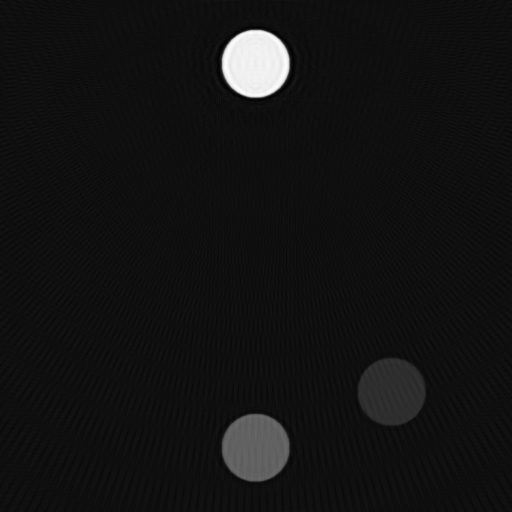} & \includegraphics[width=0.15\textwidth,cfbox=col5 1pt 0pt]{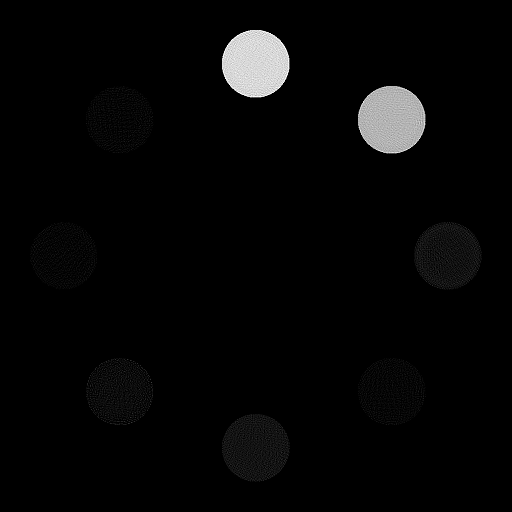} & \includegraphics[width=0.15\textwidth,cfbox=col5 1pt 0pt]{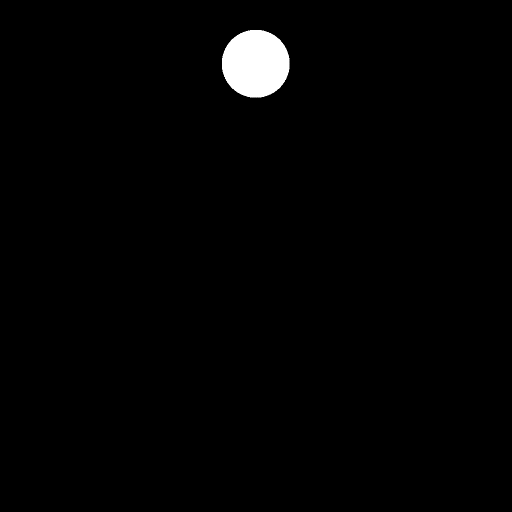} \\
        \includegraphics[width=0.15\textwidth,cfbox=col6 1pt 0pt]{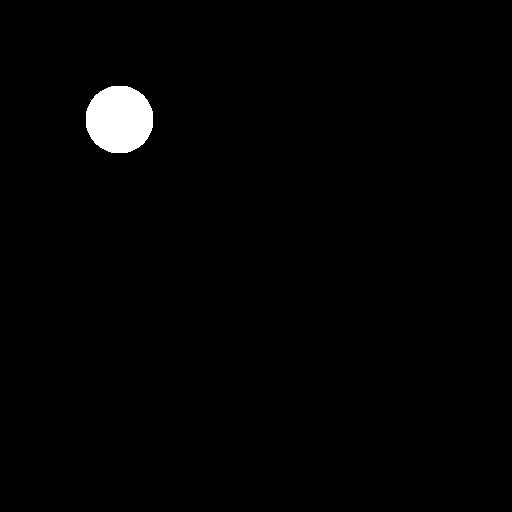} \quad & \includegraphics[width=0.15\textwidth,cfbox=col6 1pt 0pt]{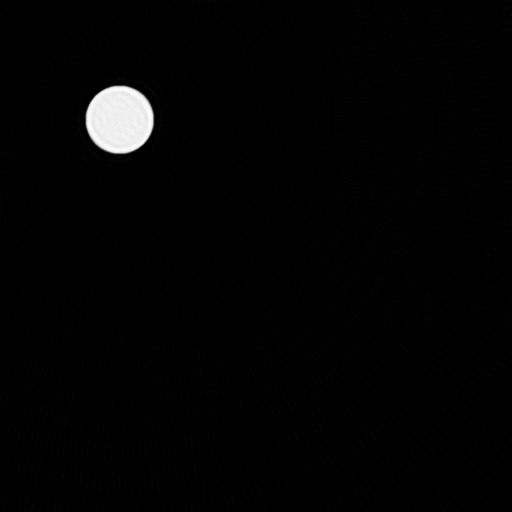} & \includegraphics[width=0.15\textwidth,cfbox=col6 1pt 0pt]{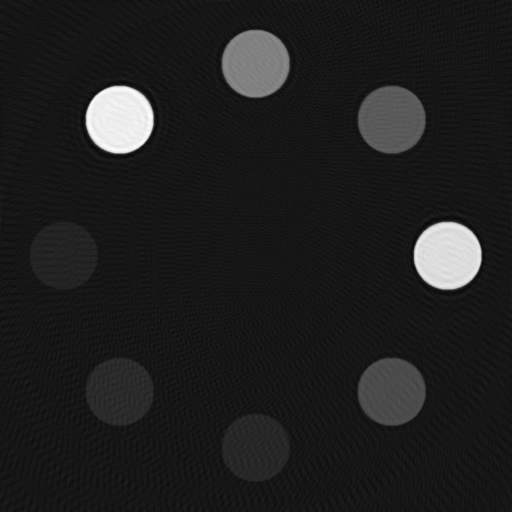} & \includegraphics[width=0.15\textwidth,cfbox=col6 1pt 0pt]{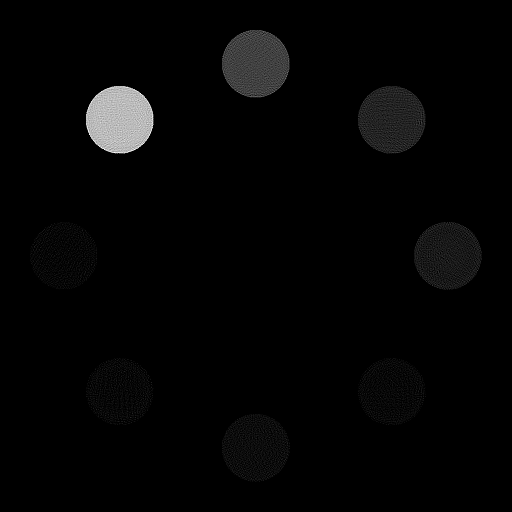} & \includegraphics[width=0.15\textwidth,cfbox=col6 1pt 0pt]{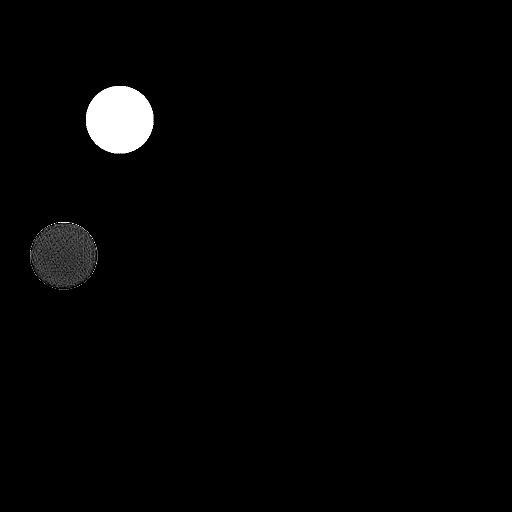} \\
        \includegraphics[width=0.15\textwidth,cfbox=col7 1pt 0pt]{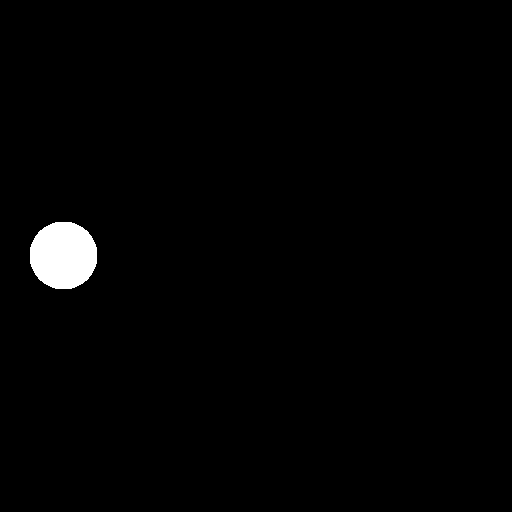} \quad & \includegraphics[width=0.15\textwidth,cfbox=col7 1pt 0pt]{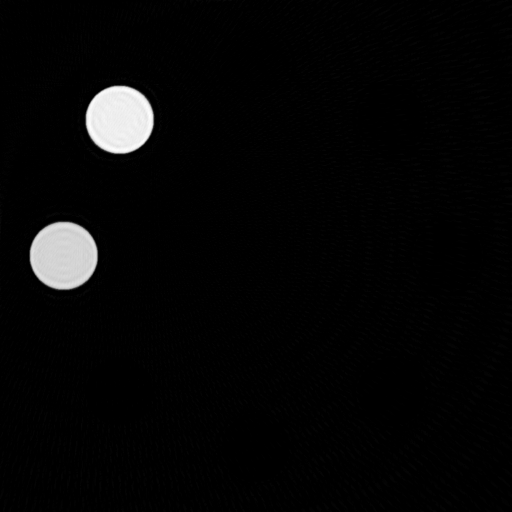} & \includegraphics[width=0.15\textwidth,cfbox=col7 1pt 0pt]{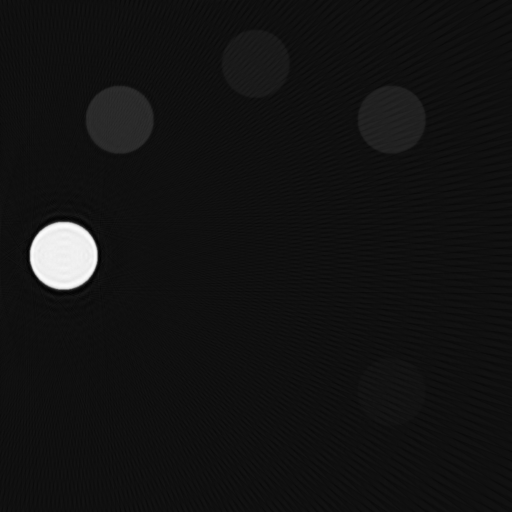} & \includegraphics[width=0.15\textwidth,cfbox=col7 1pt 0pt]{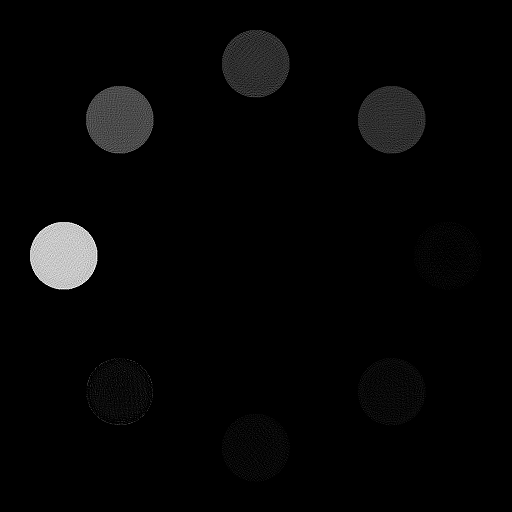} & \includegraphics[width=0.15\textwidth,cfbox=col7 1pt 0pt]{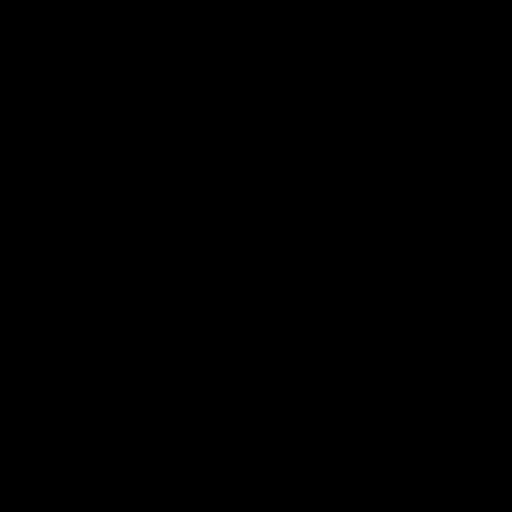} \\
        \includegraphics[width=0.15\textwidth,cfbox=col8 1pt 0pt]{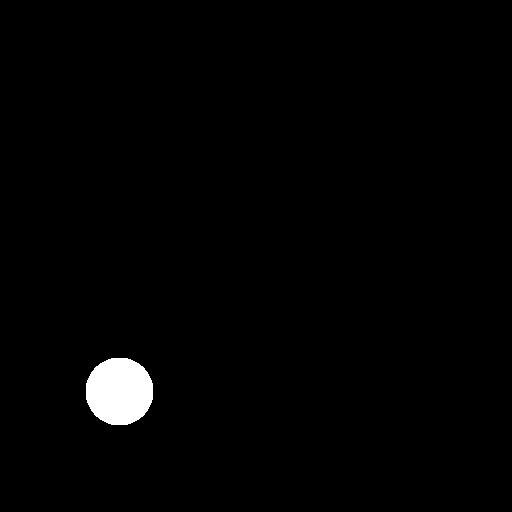} \quad & \includegraphics[width=0.15\textwidth,cfbox=col8 1pt 0pt]{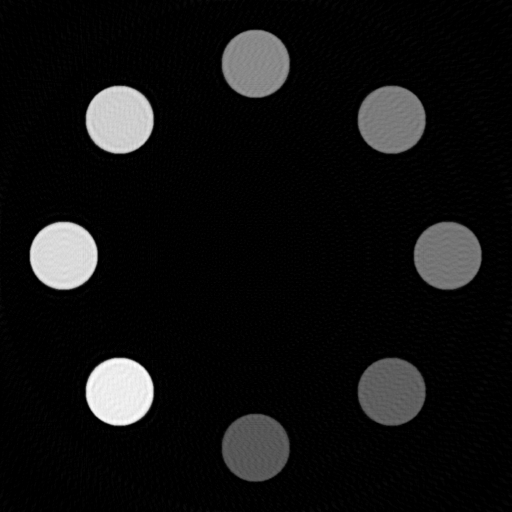} & \includegraphics[width=0.15\textwidth,cfbox=col8 1pt 0pt]{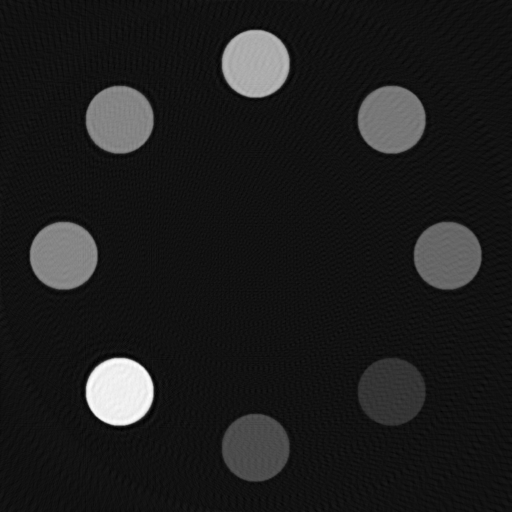} & \includegraphics[width=0.15\textwidth,cfbox=col8 1pt 0pt]{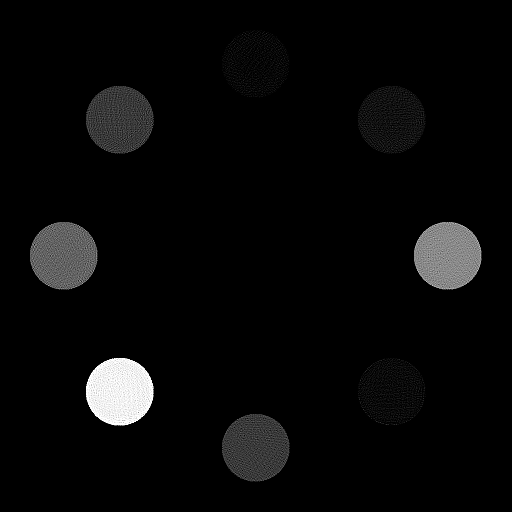} & \includegraphics[width=0.15\textwidth,cfbox=col8 1pt 0pt]{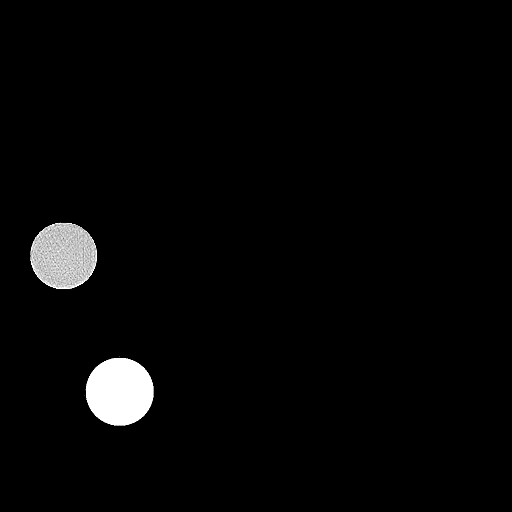} \\
        \includegraphics[width=0.15\textwidth]{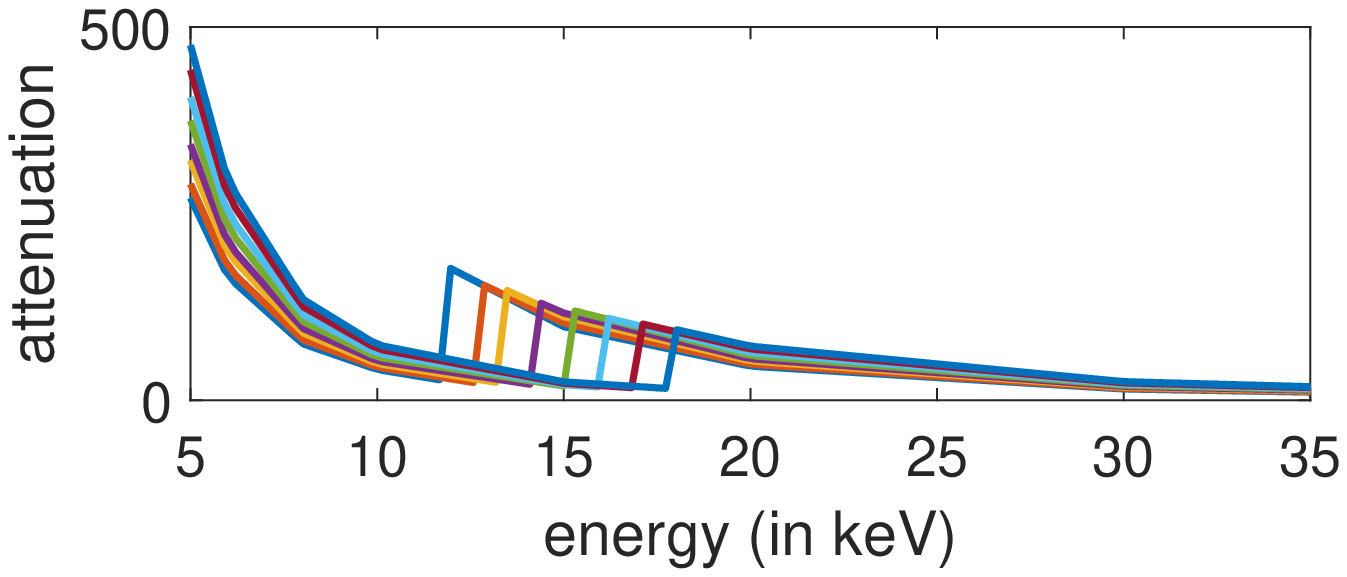} \quad & 
        \includegraphics[width=0.15\textwidth]{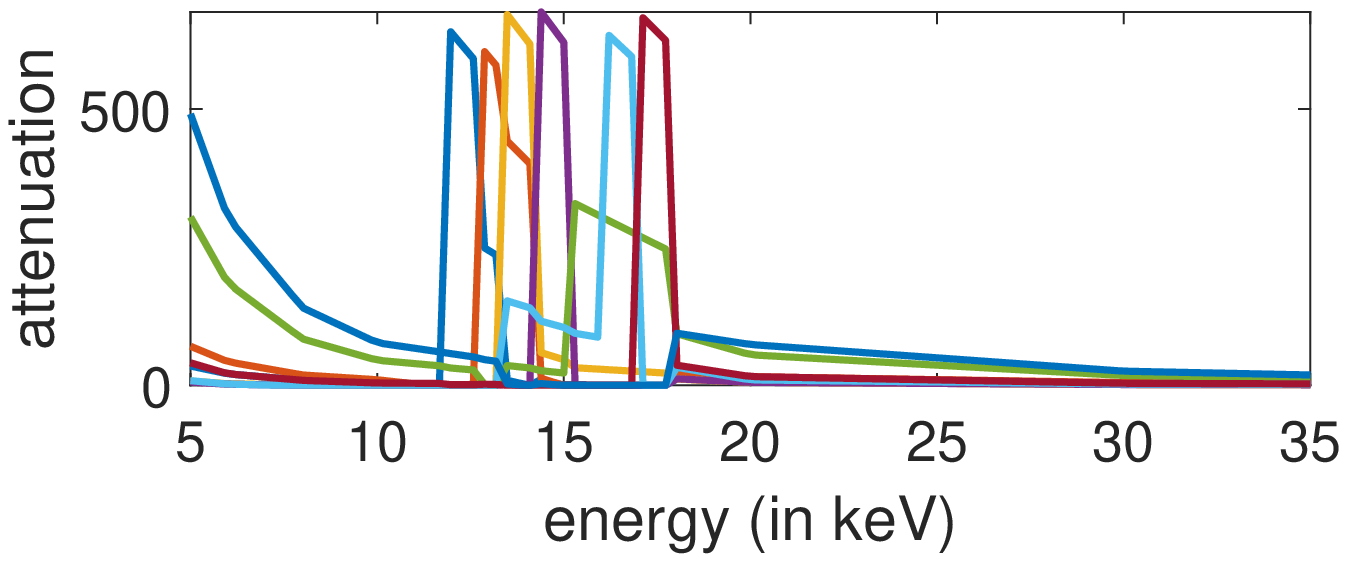} &
        \includegraphics[width=0.15\textwidth]{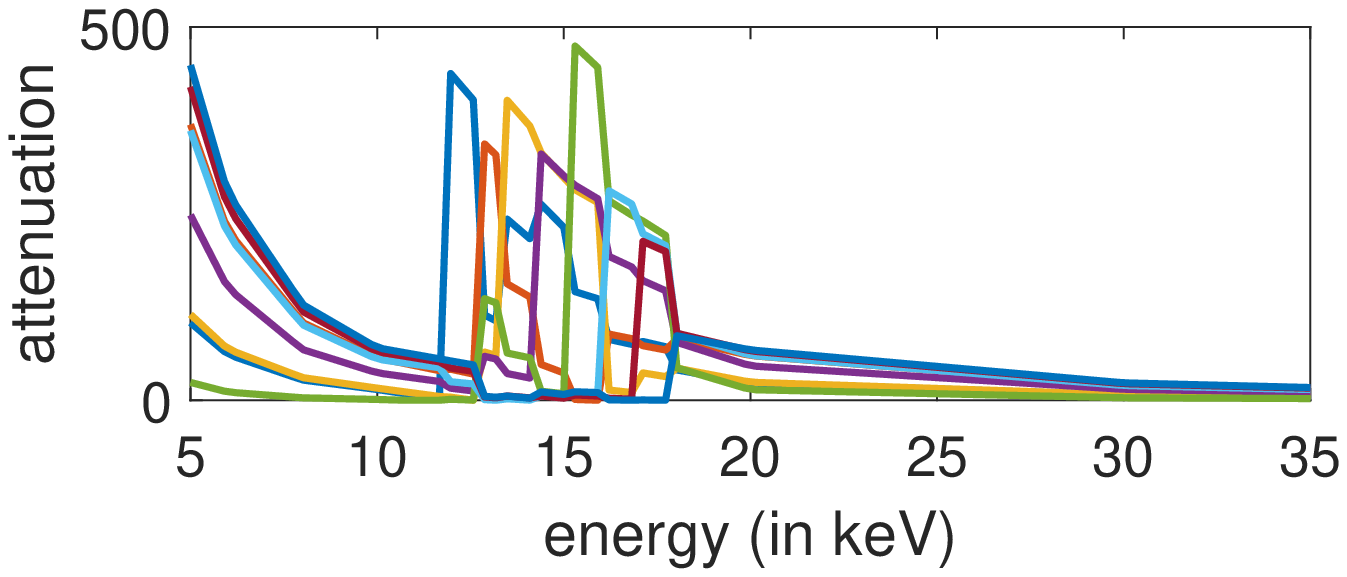} &
        \includegraphics[width=0.15\textwidth]{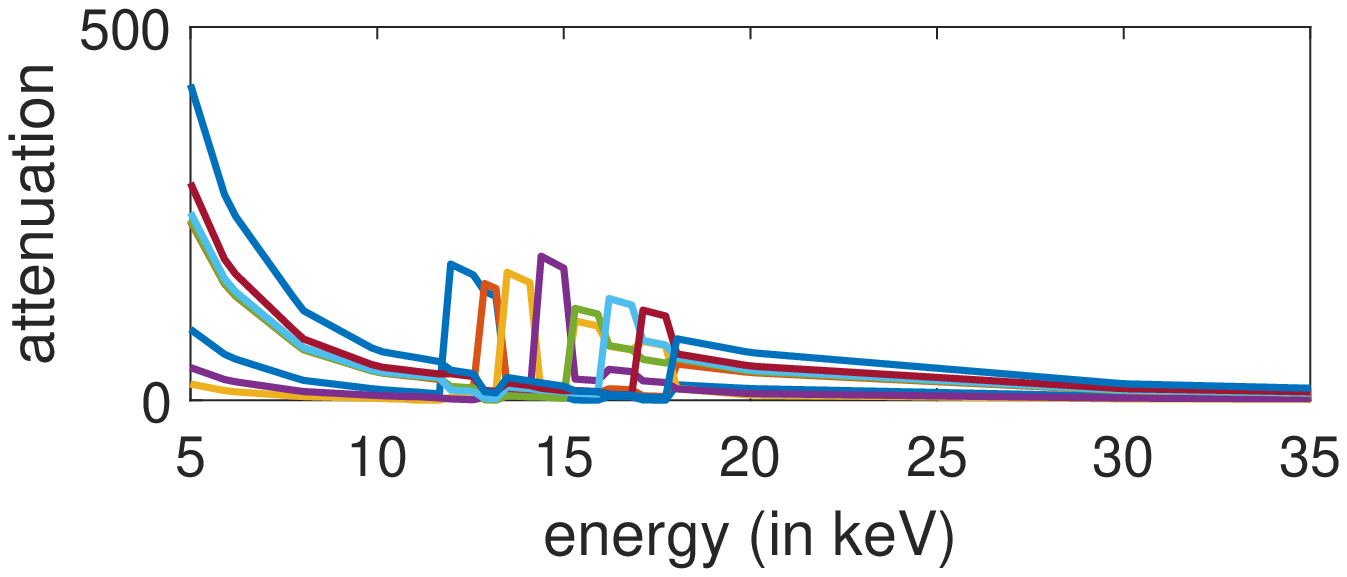} &
        \includegraphics[width=0.15\textwidth]{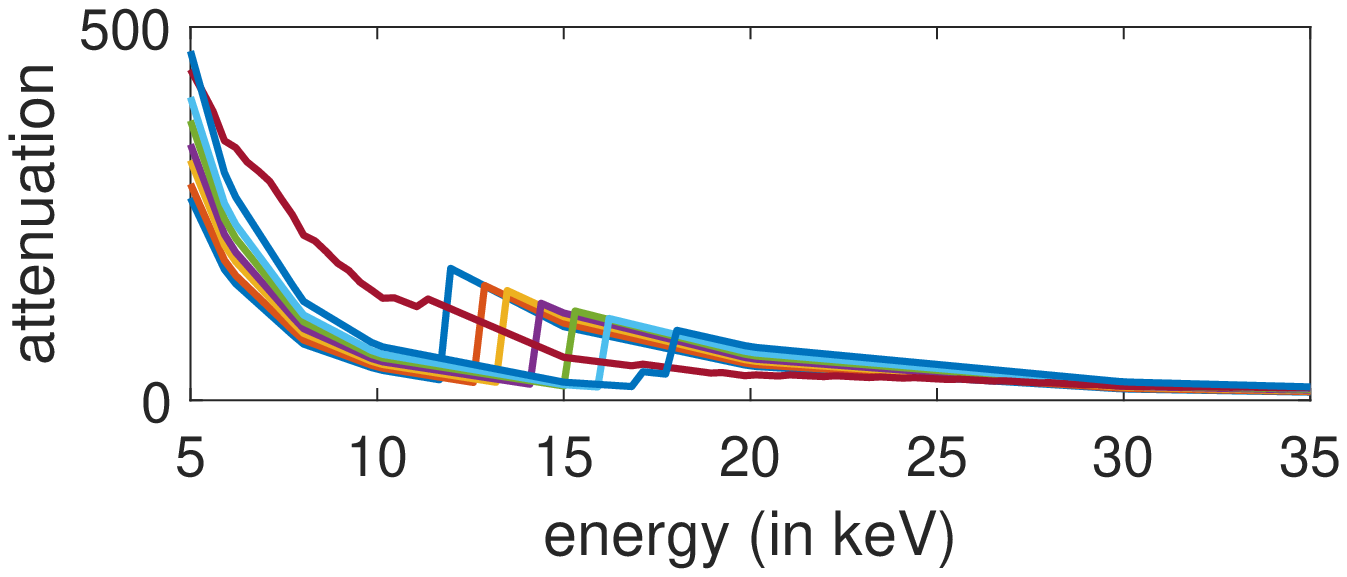}\\ 
    \end{tabular}
    \caption{Visual comparison of ADJUST with RU, UR, and cJoint method on the Disks phantom. We only show the reconstructions of all disks here for the comparison. Moreover, we match the colors of the bounding box for material maps with the (recovered) spectral signatures of the materials (shown in the bottom row).}
    \label{fig:Exp:comparison}
\end{figure}

\begin{figure}[t]
\centering
\begin{tabular}{cc}
    \rotatebox[origin=l]{90}{\hphantom{0ex} Shepp-Logan \hphantom{0ex}} & \includegraphics[width=0.9\textwidth]{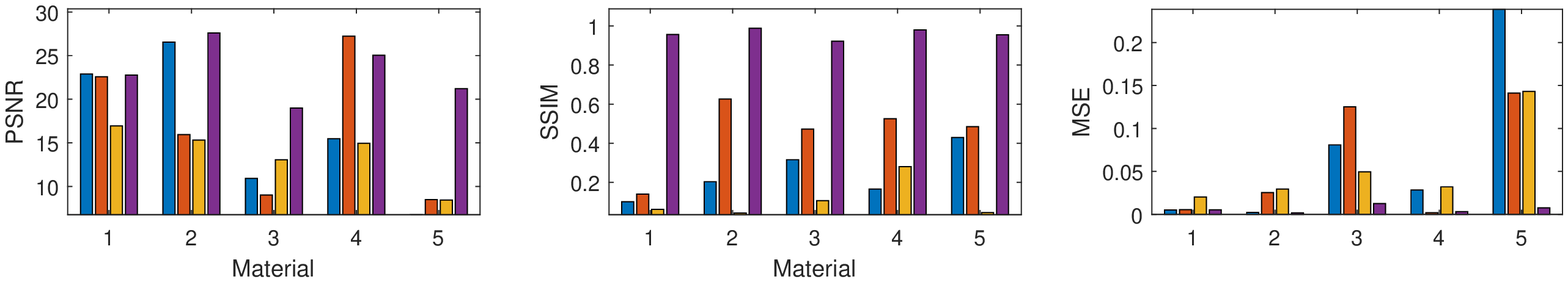} \\[3ex]
    \rotatebox[origin=l]{90}{\hphantom{3ex} Thorax \hspace{-5ex}} & \includegraphics[width=0.9\textwidth]{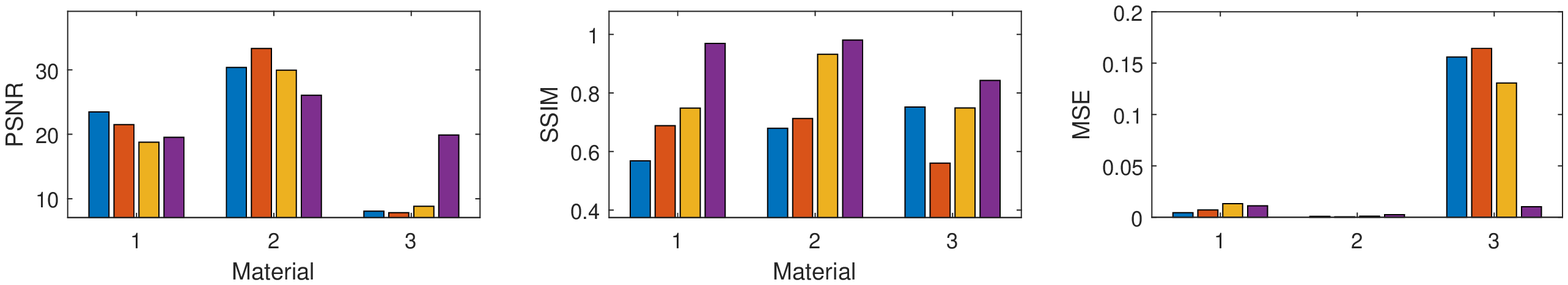}  \\[3ex]
    \rotatebox[origin=l]{90}{\hphantom{4ex} Disks  } & \includegraphics[width=0.9\textwidth]{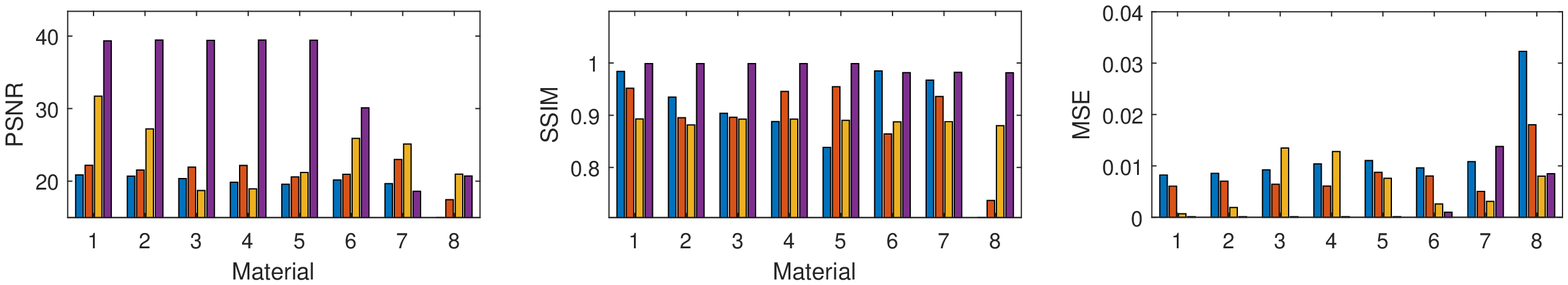} \\[3ex]
    & \includegraphics[width=0.4\textwidth]{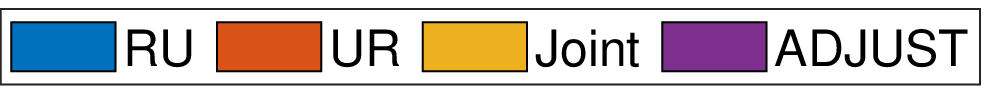} 
\end{tabular}
\caption{Performance plots showing PSNR (left column), SSIM (middle column), MSE (right column) of the reconstructed materials against the ground truth for various numerical algorithms on the phantoms.}
\label{fig:Exp:comparison_graphs}
\end{figure}

As stated in the main paper, we have compared ADJUST with RU, UR, and cJoint on three numerical phantoms, mainly the Shepp-Logan phantom, the Disks phantom, and the Thorax phantom. Figure~\ref{fig:Exp:comparison} shows the reconstruction results (\ie\ reconstructed spatial maps and the spectra of materials) of these methods on Disks phantom. Moreover, we also plot the performance measures of these methods per material in Figure~\ref{fig:Exp:comparison_graphs}. 

\section{Numerical Studies: Limited Measurement Patterns}
\label{sec:NumStudLimited}

\begin{figure}[H]
    \centering
    \setlength{\tabcolsep}{4pt}
    \begin{tabular}{p{0.12\textwidth} cccccc}
    \centered{\bf GT} &
    \centered{\includegraphics[height=0.06\textheight,cfbox=col1 1pt 0pt]{SL_GT_1.png}} & \centered{\includegraphics[height=0.06\textheight,cfbox=col2 1pt 0pt]{SL_GT_2.png}} & \centered{\includegraphics[height=0.06\textheight,cfbox=col3 1pt 0pt]{SL_GT_3.png}} & \centered{\includegraphics[height=0.06\textheight,cfbox=col4 1pt 0pt]{SL_GT_4.png}} &
    \centered{\includegraphics[height=0.06\textheight,cfbox=col5 1pt 0pt]{SL_GT_5.png}} &
    \centered{\includegraphics[height=0.06\textheight]{SL_GT_f.eps}} \\ 
    {\bf Sparse angles} &
    \centered{\includegraphics[height=0.06\textheight,cfbox=col1 1pt 0pt]{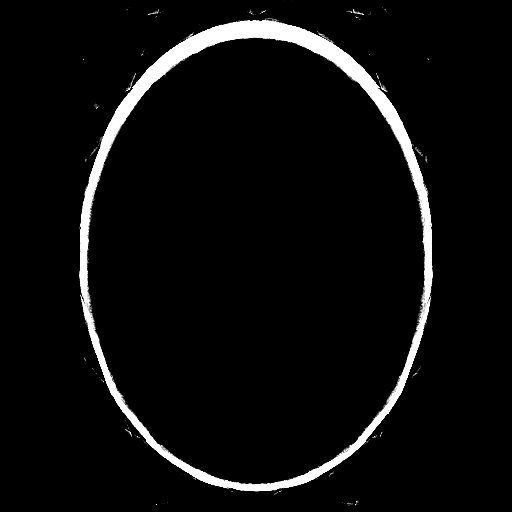}} & \centered{\includegraphics[height=0.06\textheight,cfbox=col2 1pt 0pt]{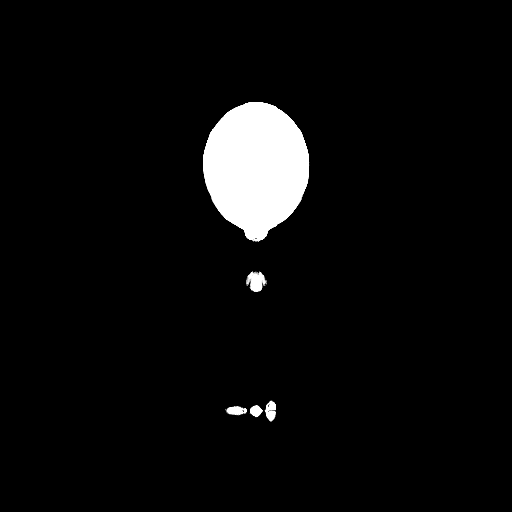}} & \centered{\includegraphics[height=0.06\textheight,cfbox=col3 1pt 0pt]{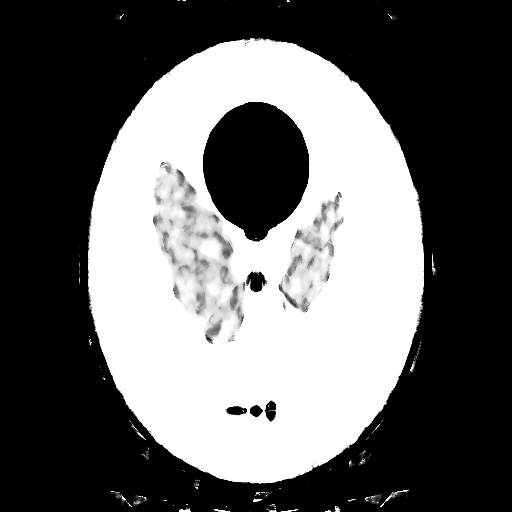}} & \centered{\includegraphics[height=0.06\textheight,cfbox=col4 1pt 0pt]{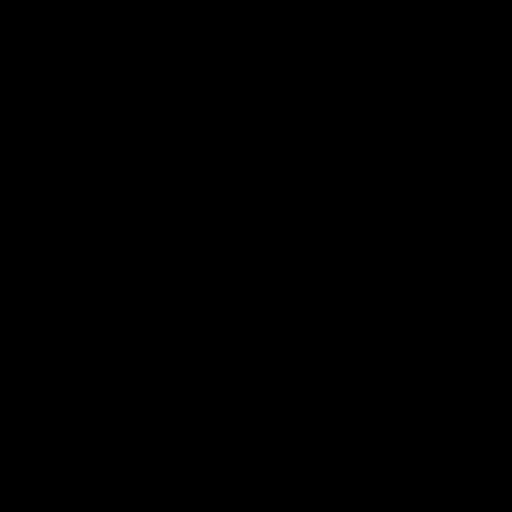}} &
    \centered{\includegraphics[height=0.06\textheight,cfbox=col5 1pt 0pt]{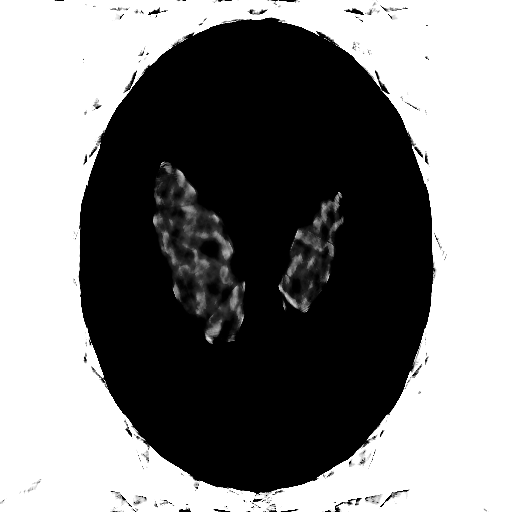}} &
    \centered{\includegraphics[height=0.06\textheight]{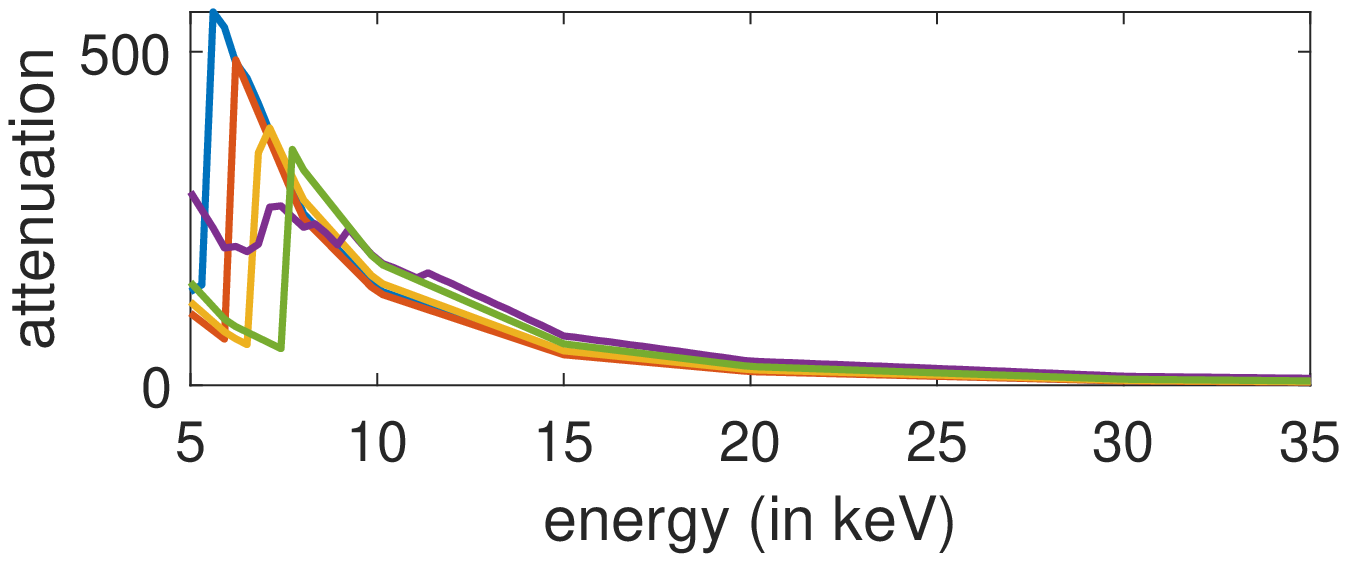}} \\ 
    {\bf Limited view} &
    \centered{\includegraphics[height=0.06\textheight,cfbox=col1 1pt 0pt]{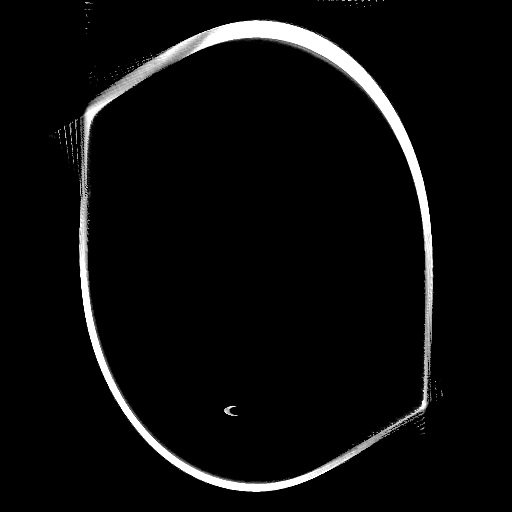}} & \centered{\includegraphics[height=0.06\textheight,cfbox=col2 1pt 0pt]{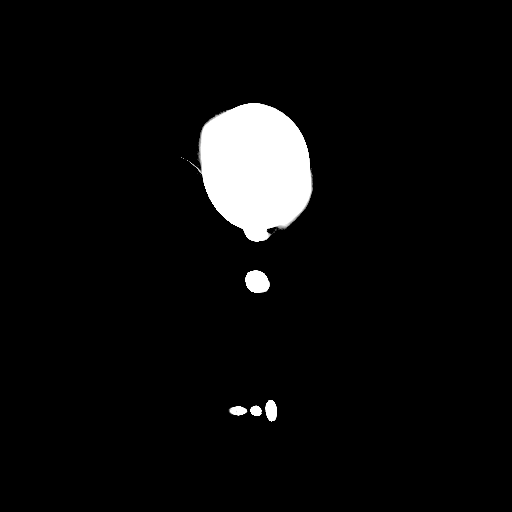}} & \centered{\includegraphics[height=0.06\textheight,cfbox=col3 1pt 0pt]{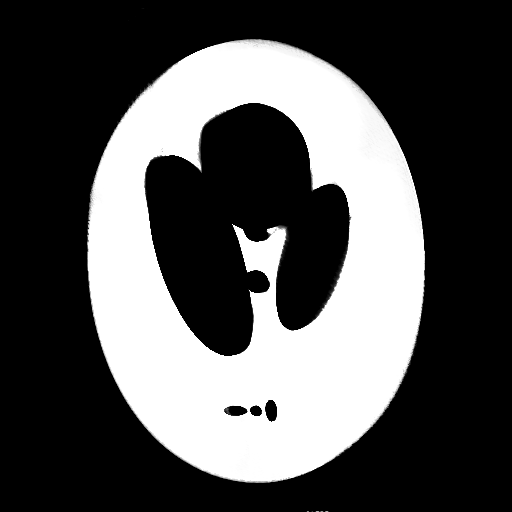}} & \centered{\includegraphics[height=0.06\textheight,cfbox=col4 1pt 0pt]{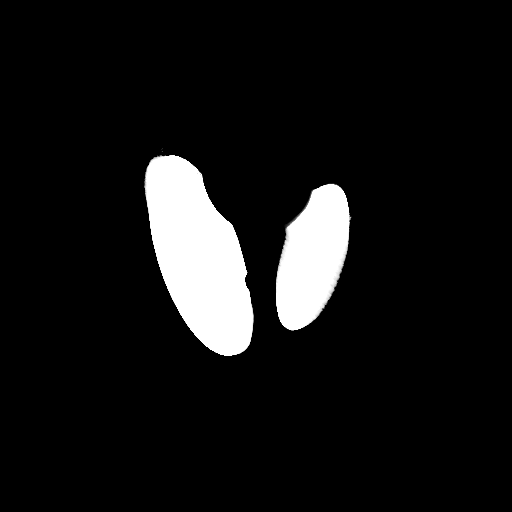}} &
    \centered{\includegraphics[height=0.06\textheight,cfbox=col5 1pt 0pt]{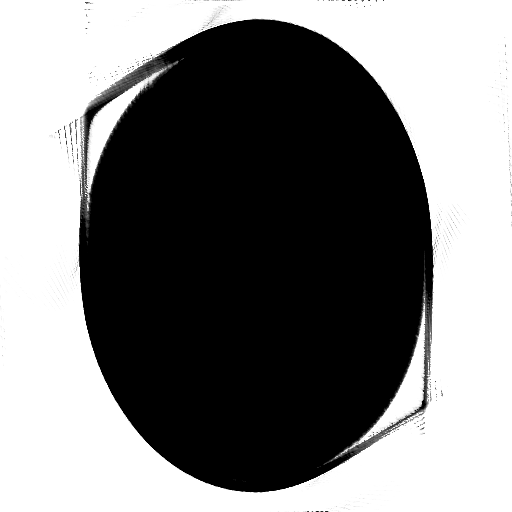}} &
    \centered{\includegraphics[height=0.06\textheight]{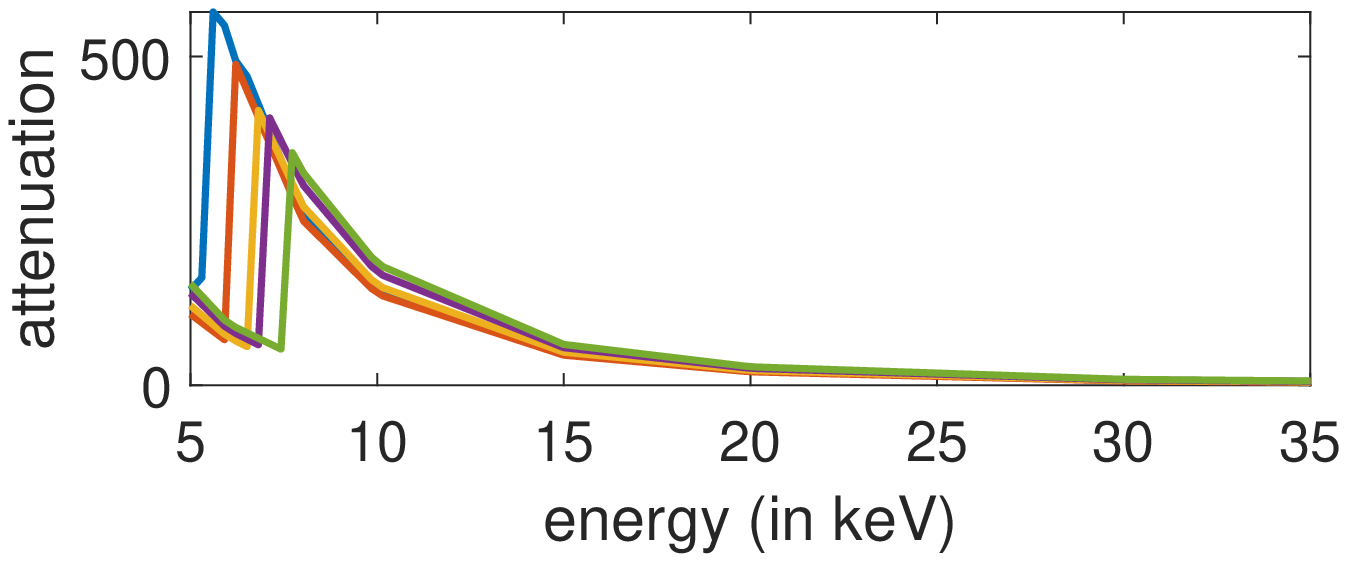}} \\ 
    {\bf Sparse channels} &
    \centered{\includegraphics[height=0.06\textheight,cfbox=col1 1pt 0pt]{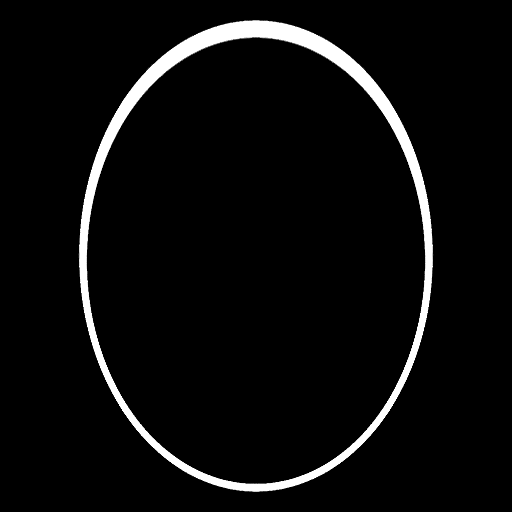}} & \centered{\includegraphics[height=0.06\textheight,cfbox=col2 1pt 0pt]{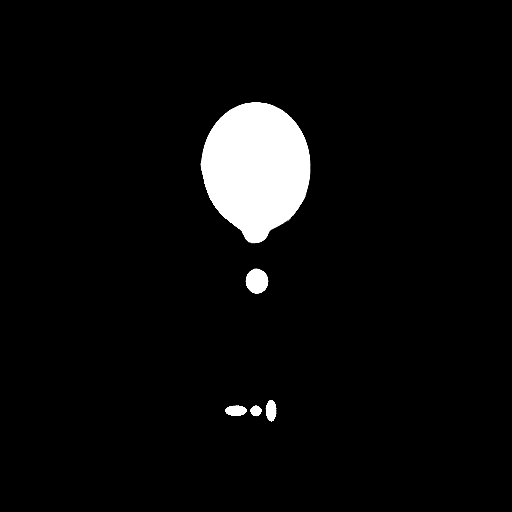}} & \centered{\includegraphics[height=0.06\textheight,cfbox=col3 1pt 0pt]{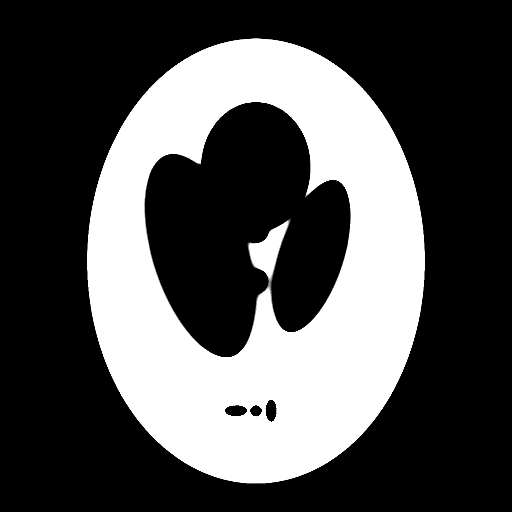}} & \centered{\includegraphics[height=0.06\textheight,cfbox=col4 1pt 0pt]{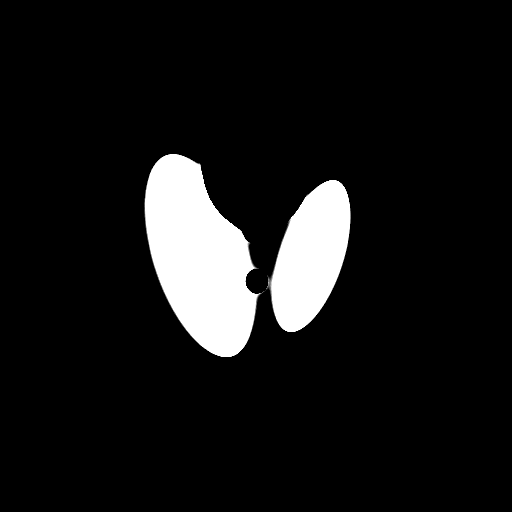}} &
    \centered{\includegraphics[height=0.06\textheight,cfbox=col5 1pt 0pt]{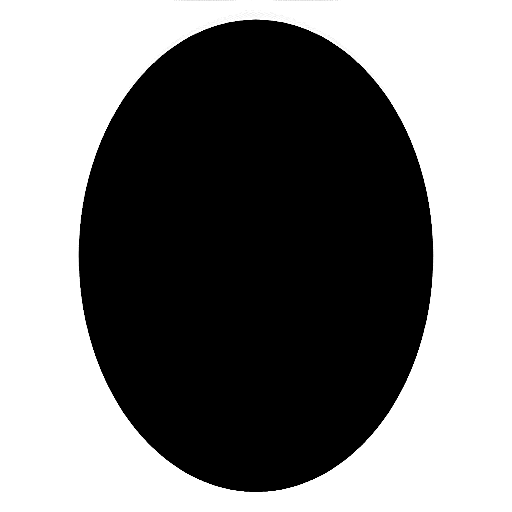}} &
    \centered{\includegraphics[height=0.06\textheight]{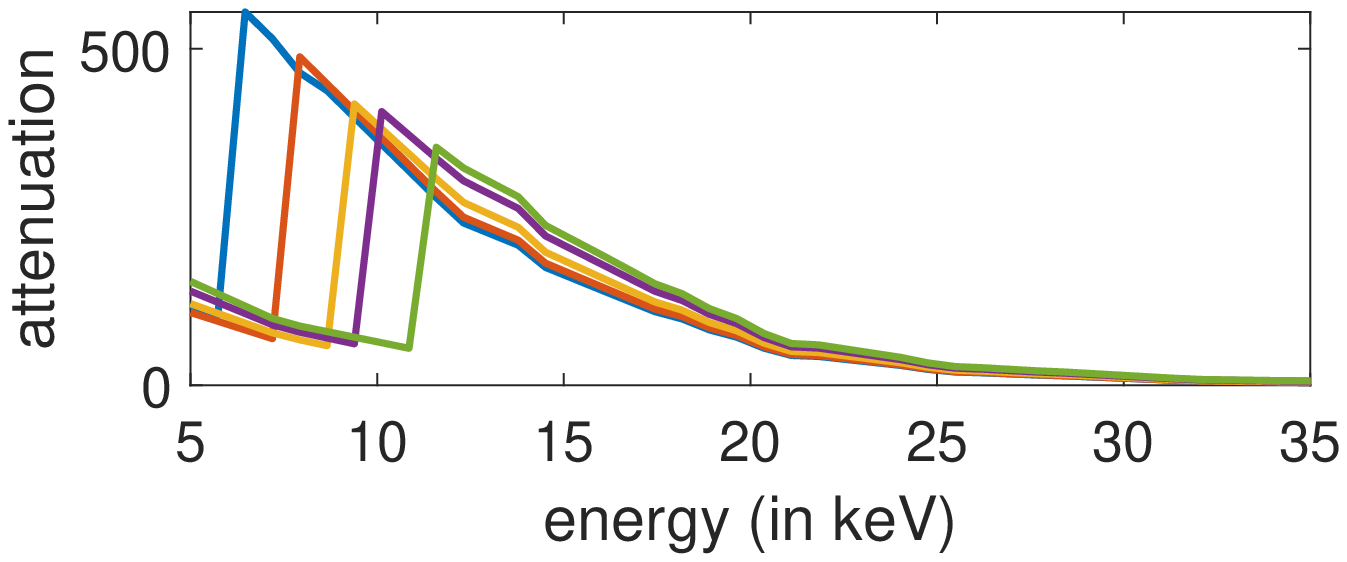}} \\ 
    \end{tabular}
    \caption{Results of ADJUST with sparse-angle data, limited view data and sparse spectral channels on Shepp-Logan phantom. The colors of the bounding box of material maps are matched with the spectral signatures shown in the rightmost corner.}
    \label{fig:Exp:LimitedSL}
\end{figure}

\begin{figure}[H]
    \centering
    \setlength{\tabcolsep}{4pt}
    \begin{tabular}{p{0.12\textwidth} cccccc}
    \centered{\bf GT} &
    \centered{\includegraphics[height=0.06\textheight,cfbox=col1 1pt 0pt]{disk_GT_1.png}} & \centered{\includegraphics[height=0.06\textheight,cfbox=col2 1pt 0pt]{disk_GT_2.png}} & \centered{\includegraphics[height=0.06\textheight,cfbox=col3 1pt 0pt]{disk_GT_3.png}} & \centered{\includegraphics[height=0.06\textheight,cfbox=col4 1pt 0pt]{disk_GT_4.png}} &
    \centered{\includegraphics[height=0.06\textheight,cfbox=col5 1pt 0pt]{disk_GT_5.png}} &
    \centered{\includegraphics[height=0.06\textheight]{disk_GT_f.eps}} \\ 
    {\bf Sparse angles} &
    \centered{\includegraphics[height=0.06\textheight,cfbox=col1 1pt 0pt]{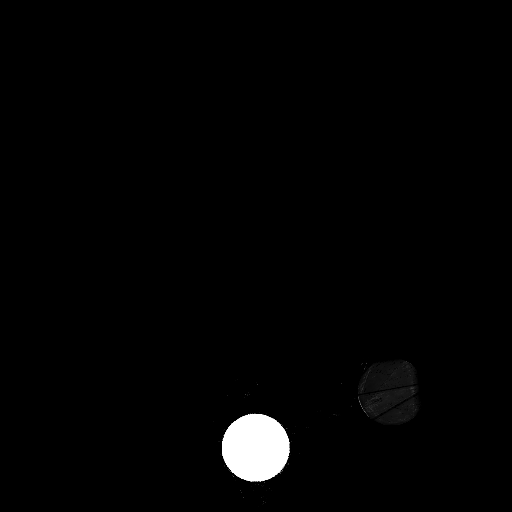}} & \centered{\includegraphics[height=0.06\textheight,cfbox=col2 1pt 0pt]{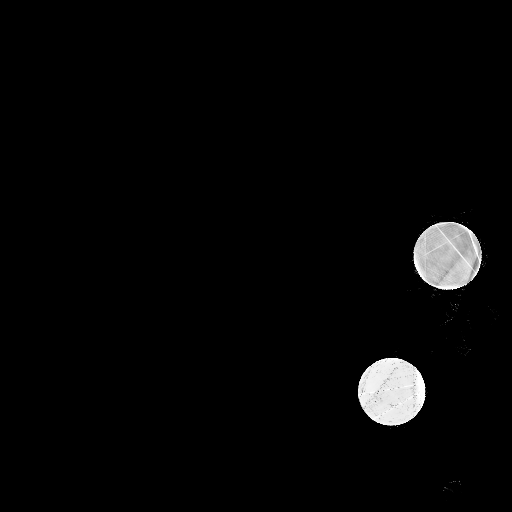}} & \centered{\includegraphics[height=0.06\textheight,cfbox=col3 1pt 0pt]{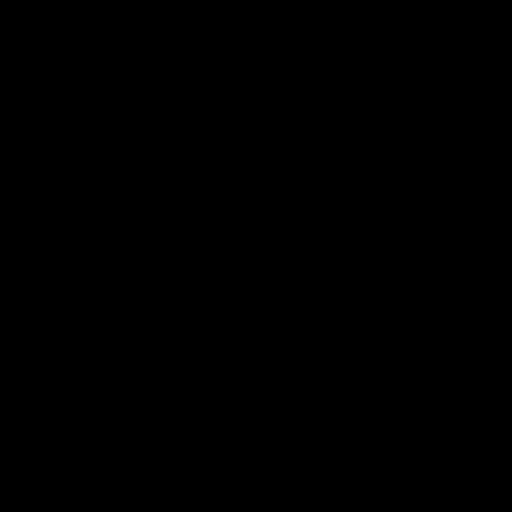}} & \centered{\includegraphics[height=0.06\textheight,cfbox=col4 1pt 0pt]{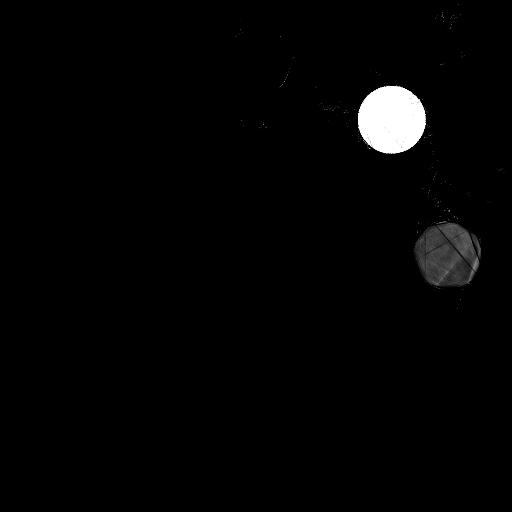}} &
    \centered{\includegraphics[height=0.06\textheight,cfbox=col5 1pt 0pt]{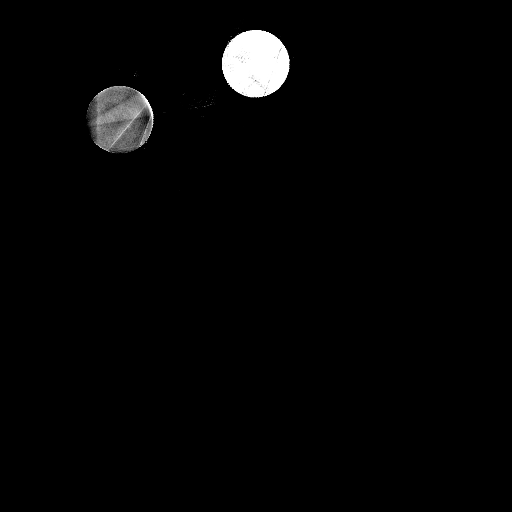}} &
    \centered{\includegraphics[height=0.06\textheight]{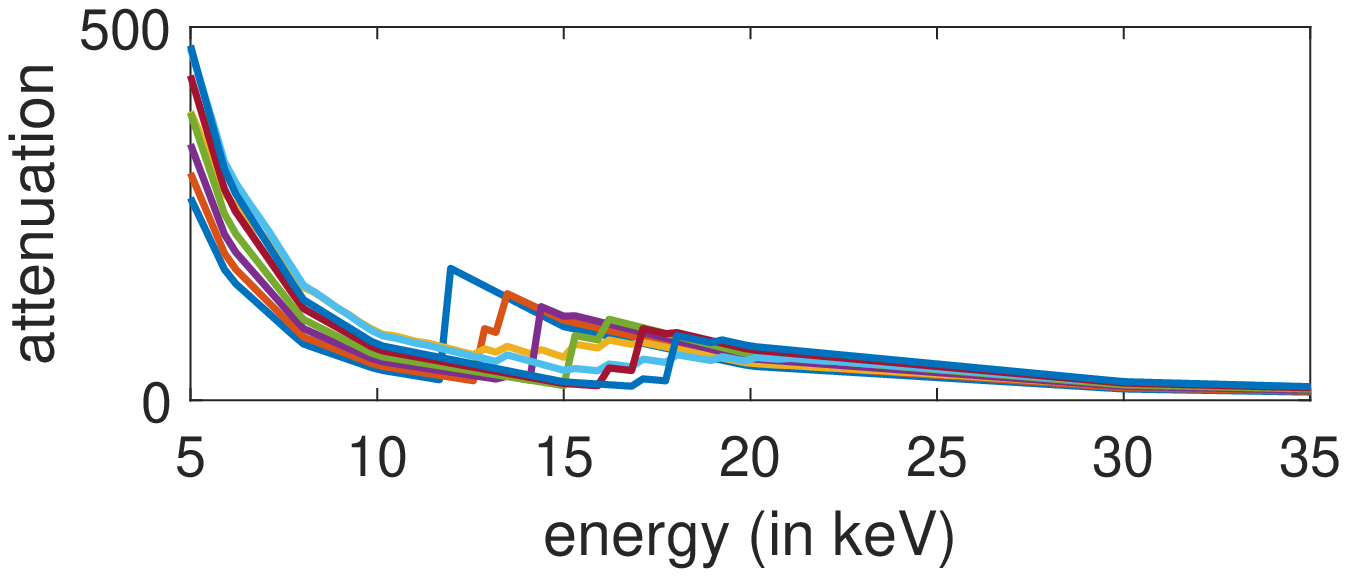}} \\ 
    {\bf Limited view} &
    \centered{\includegraphics[height=0.06\textheight,cfbox=col1 1pt 0pt]{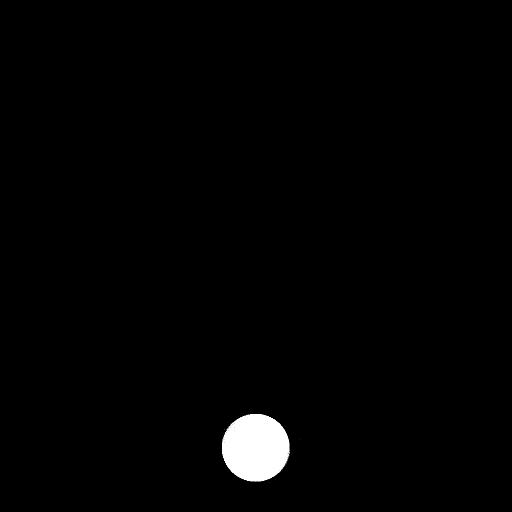}} & \centered{\includegraphics[height=0.06\textheight,cfbox=col2 1pt 0pt]{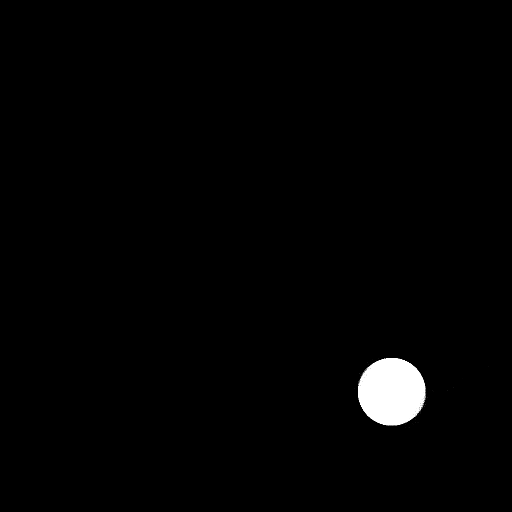}} & \centered{\includegraphics[height=0.06\textheight,cfbox=col3 1pt 0pt]{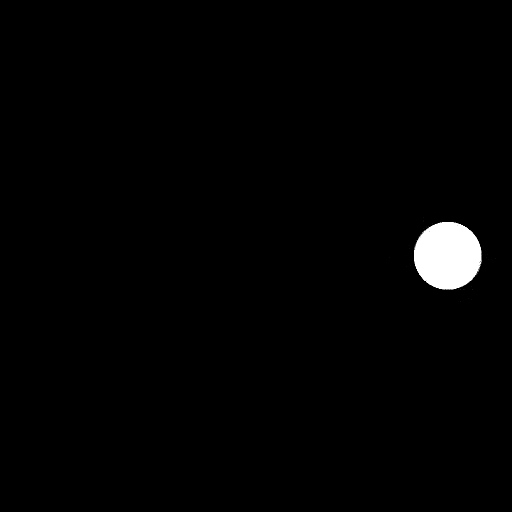}} & \centered{\includegraphics[height=0.06\textheight,cfbox=col4 1pt 0pt]{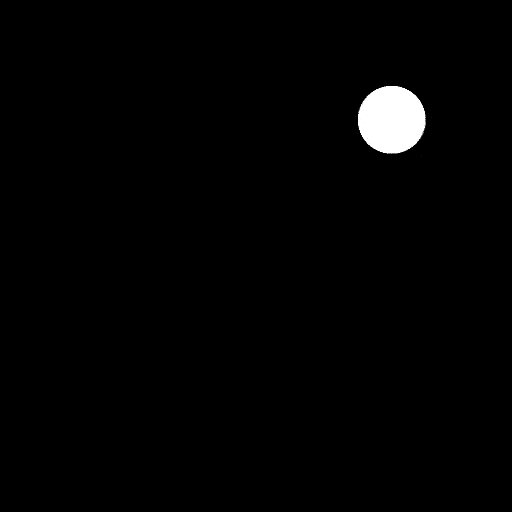}} &
    \centered{\includegraphics[height=0.06\textheight,cfbox=col5 1pt 0pt]{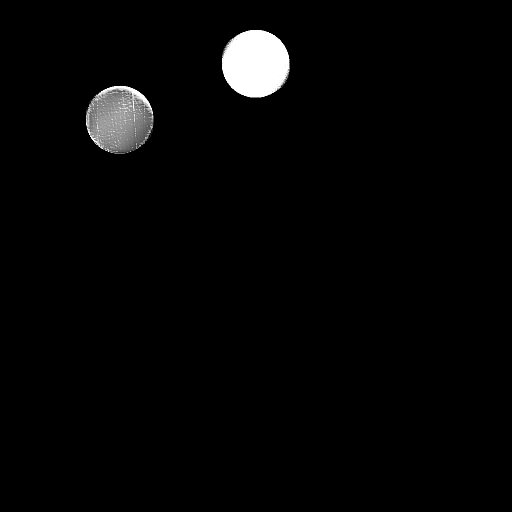}} &
    \centered{\includegraphics[height=0.06\textheight]{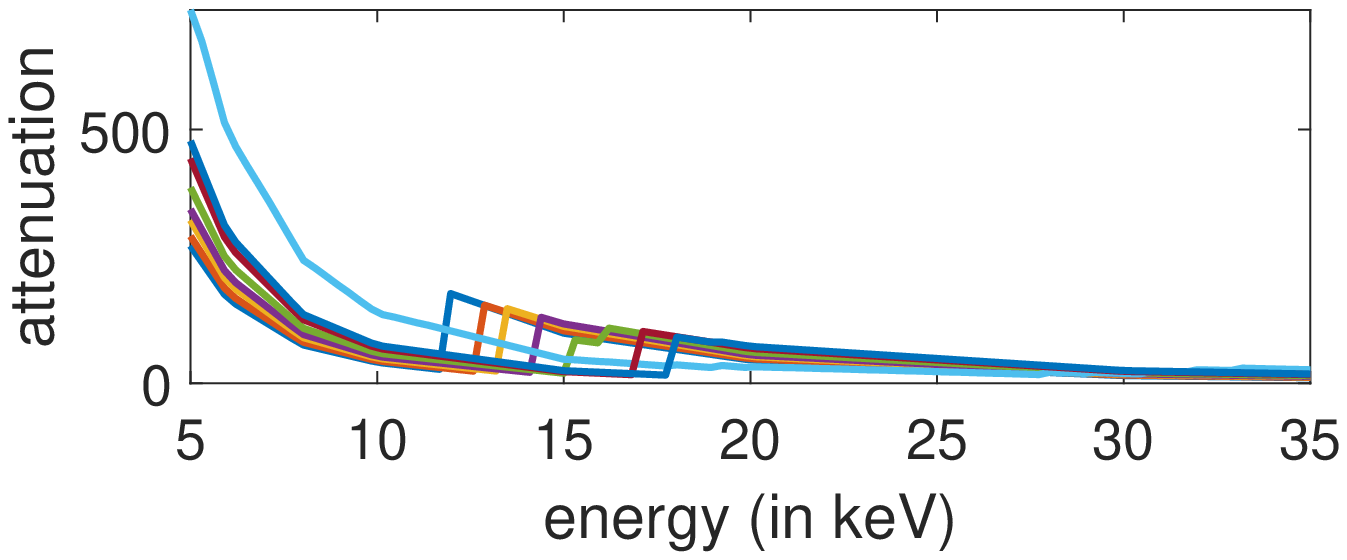}} \\ 
    {\bf Sparse channels} &
    \centered{\includegraphics[height=0.06\textheight,cfbox=col1 1pt 0pt]{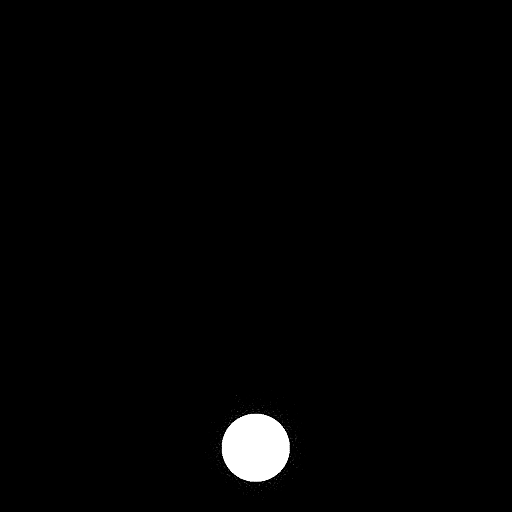}} & \centered{\includegraphics[height=0.06\textheight,cfbox=col2 1pt 0pt]{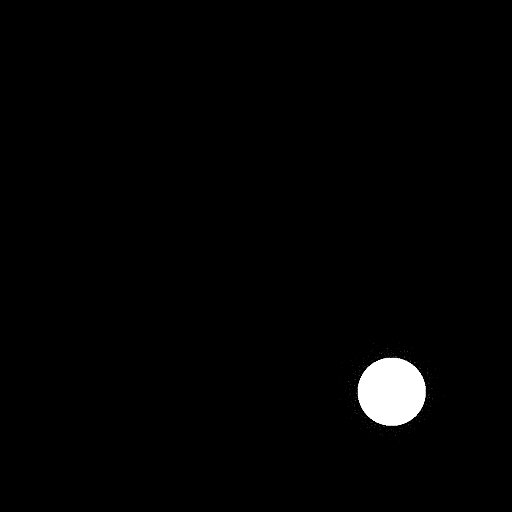}} & \centered{\includegraphics[height=0.06\textheight,cfbox=col3 1pt 0pt]{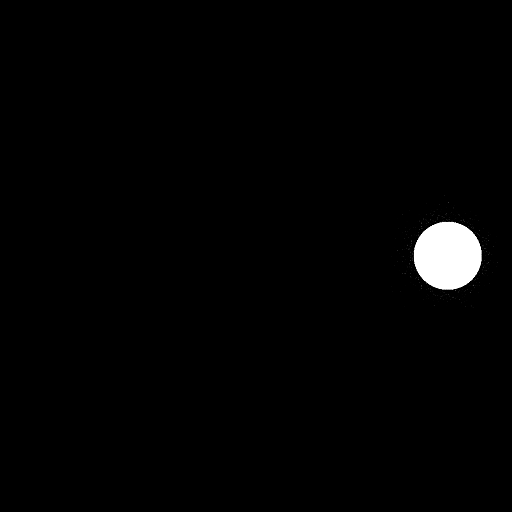}} & \centered{\includegraphics[height=0.06\textheight,cfbox=col4 1pt 0pt]{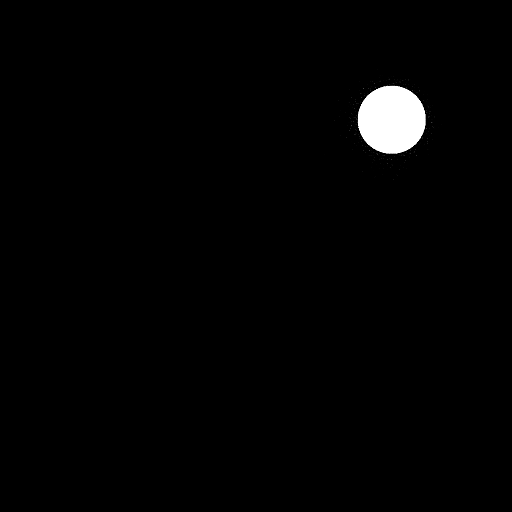}} &
    \centered{\includegraphics[height=0.06\textheight,cfbox=col5 1pt 0pt]{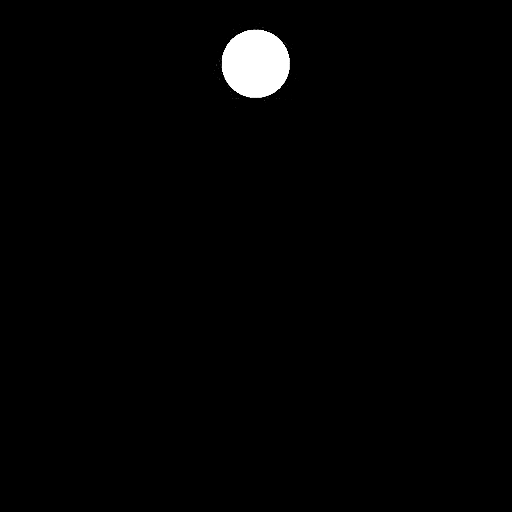}} &
    \centered{\includegraphics[height=0.06\textheight]{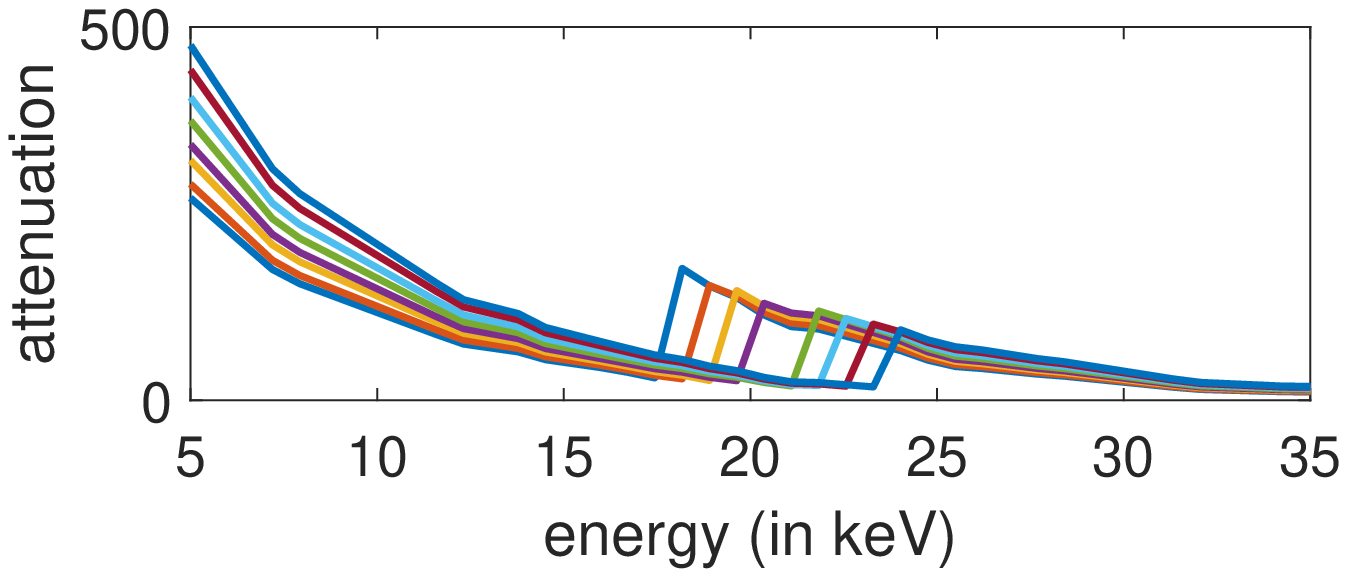}} \\ 
    \end{tabular}
    \caption{Results of ADJUST with sparse-angle data, limited view data and sparse spectral channels on Disks phantom. We only show material maps of first five materials. The colors of the bounding box of material maps are matched with the spectral signatures shown in the rightmost corner.}
    \label{fig:Exp:LimitedDisk}
\end{figure}

We consider three different types of limited measurement patterns: (\emph{i}) \textit{Sparse-angle}: tomographic projections from $10$ equidistant angles in the range of $0$ to $\pi$ for 100 spectral channels,  (\emph{ii}) \textit{Limited-view}: 60 equidistant projection angles in the limited range of $[0,2\pi/3]$ for 100 spectral channels, (\emph{iii}) \textit{Sparse channels}: 60 equidistant angles between $0$ and $\pi$, but with only 30 spectral channels. We test ADJUST on the two numerical spectral phantoms, \ie, the Shepp-Logan phantom and the Disks phantom. Figure~\ref{fig:Exp:LimitedSL} and~\ref{fig:Exp:LimitedDisk} demonstrate the reconstructions of ADJUST for all three limited measurement patterns on these two phantoms. 

\section{Numerical Studies: Mixed Material Phantom}
\label{sec:NumStudMixed}

\begin{figure}[H]
    \centering
    \renewcommand{\arraystretch}{1.0}
    \begin{tabular}{c | cccc}
        {\bf True} & {\bf RU} & {\bf UR} & {\bf cJoint} & {\bf ADJUST} \\
        \includegraphics[width=0.11\textwidth,cfbox=col1 1pt 0pt]{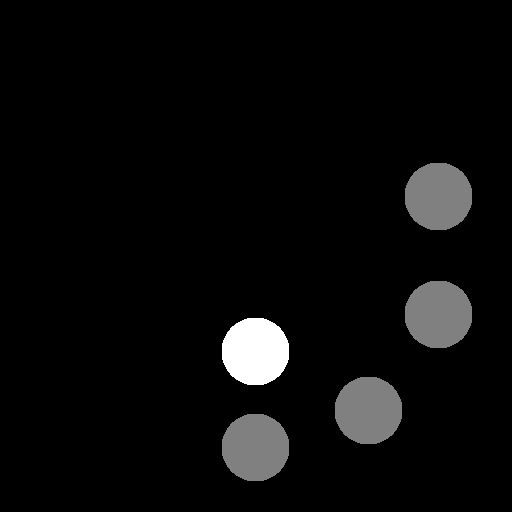} \quad & \includegraphics[width=0.11\textwidth,cfbox=col1 1pt 0pt]{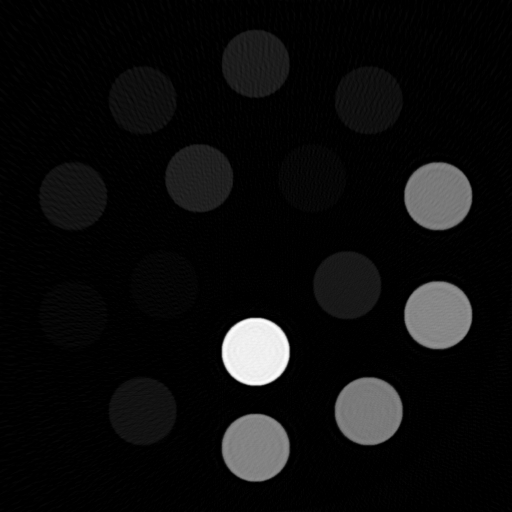} & \includegraphics[width=0.11\textwidth,cfbox=col1 1pt 0pt]{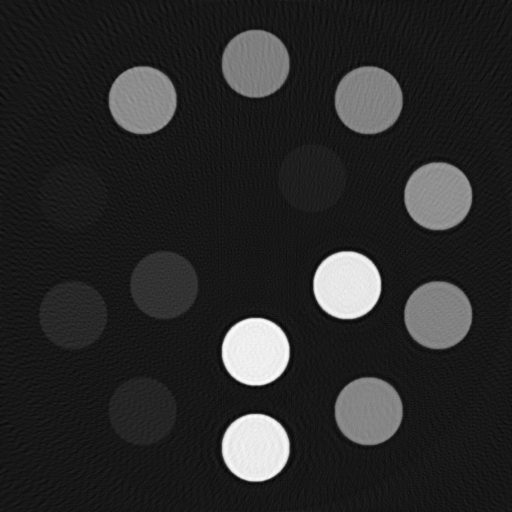} & \includegraphics[width=0.11\textwidth,cfbox=col1 1pt 0pt]{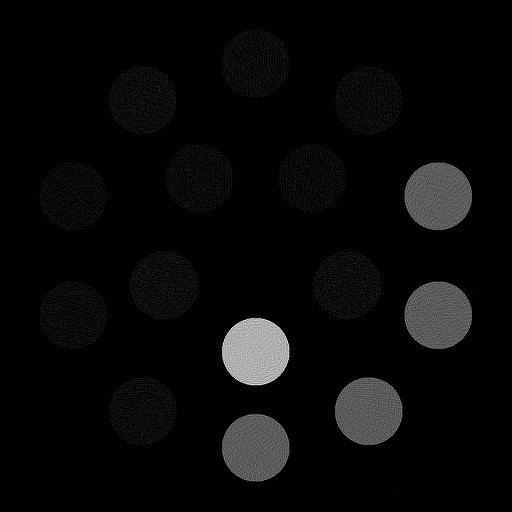} & \includegraphics[width=0.11\textwidth,cfbox=col1 1pt 0pt]{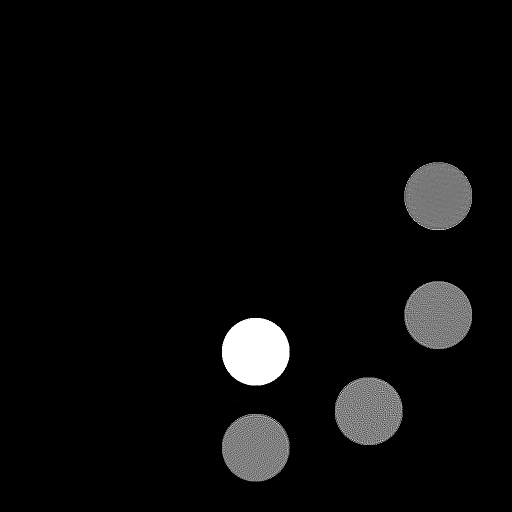} \\
        \includegraphics[width=0.11\textwidth,cfbox=col2 1pt 0pt]{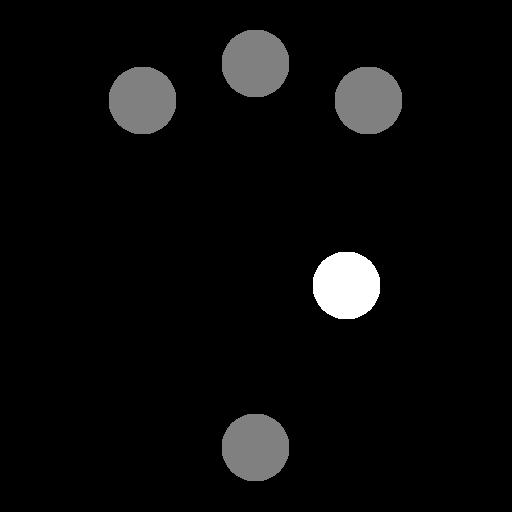} \quad & \includegraphics[width=0.11\textwidth,cfbox=col2 1pt 0pt]{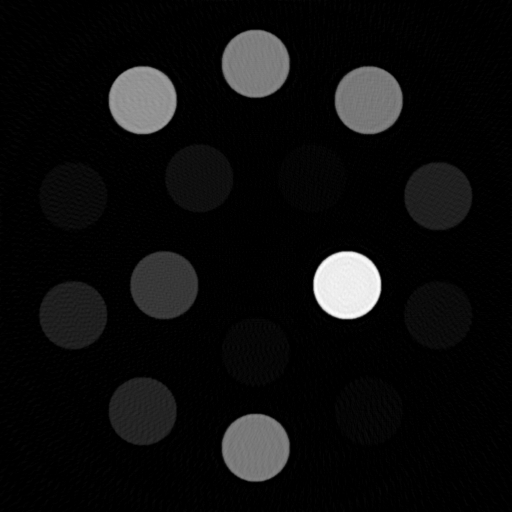} & \includegraphics[width=0.11\textwidth,cfbox=col2 1pt 0pt]{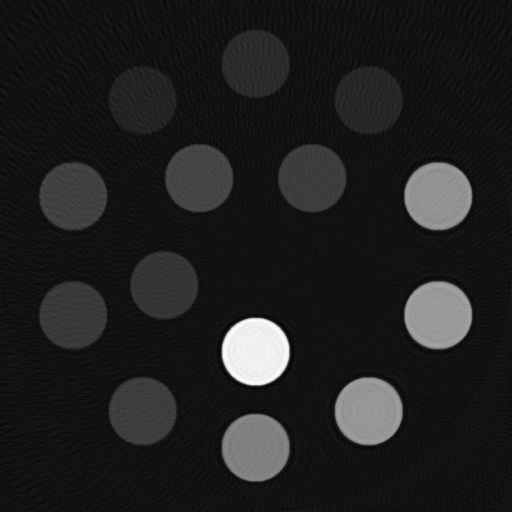} & \includegraphics[width=0.11\textwidth,cfbox=col2 1pt 0pt]{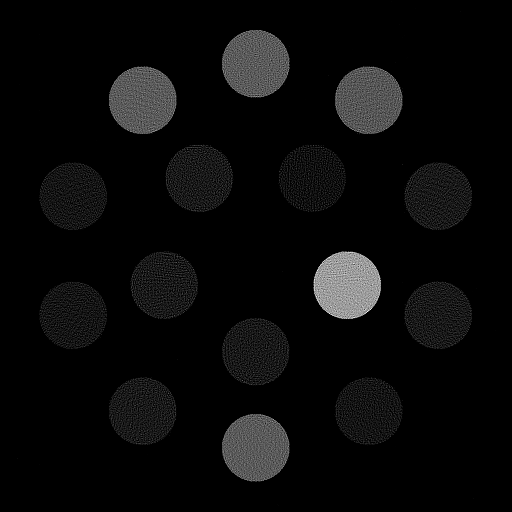} & \includegraphics[width=0.11\textwidth,cfbox=col2 1pt 0pt]{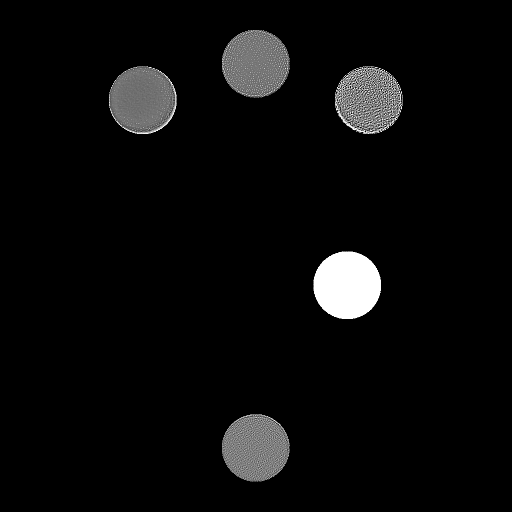} \\
        \includegraphics[width=0.11\textwidth,cfbox=col3 1pt 0pt]{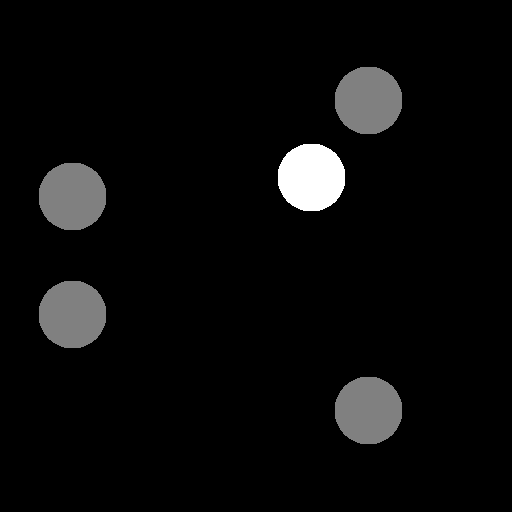} \quad & \includegraphics[width=0.11\textwidth,cfbox=col3 1pt 0pt]{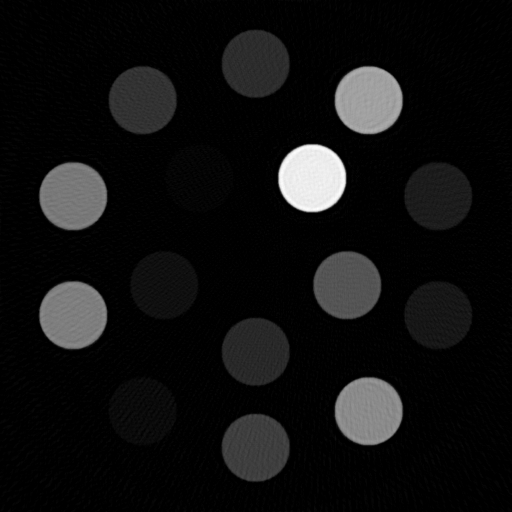} & \includegraphics[width=0.11\textwidth,cfbox=col3 1pt 0pt]{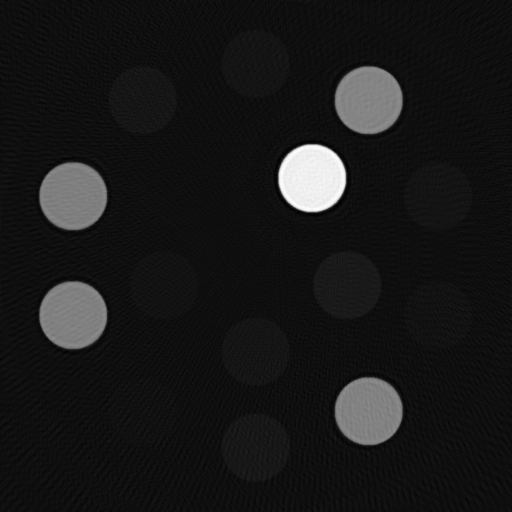} & \includegraphics[width=0.11\textwidth,cfbox=col3 1pt 0pt]{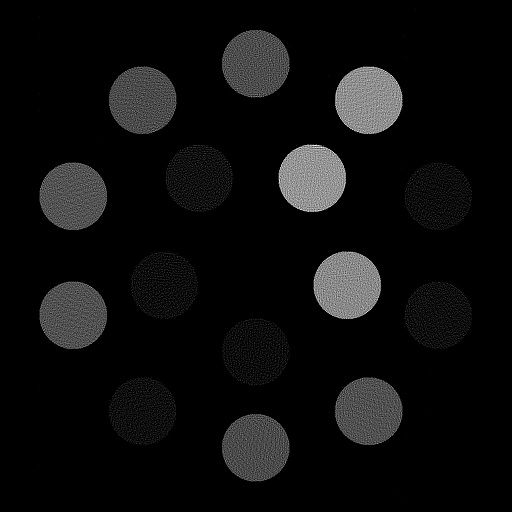} & \includegraphics[width=0.11\textwidth,cfbox=col3 1pt 0pt]{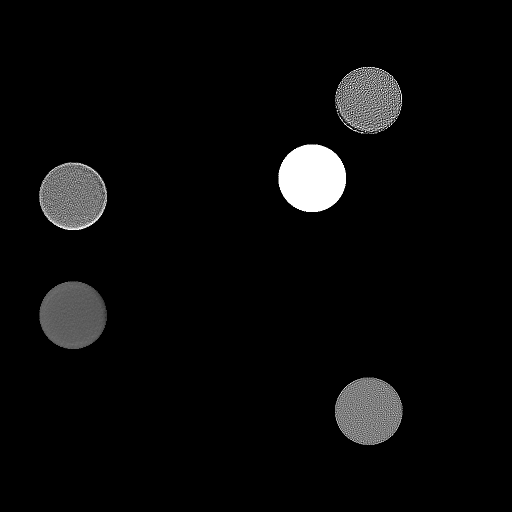} \\
        \includegraphics[width=0.11\textwidth,cfbox=col4 1pt 0pt]{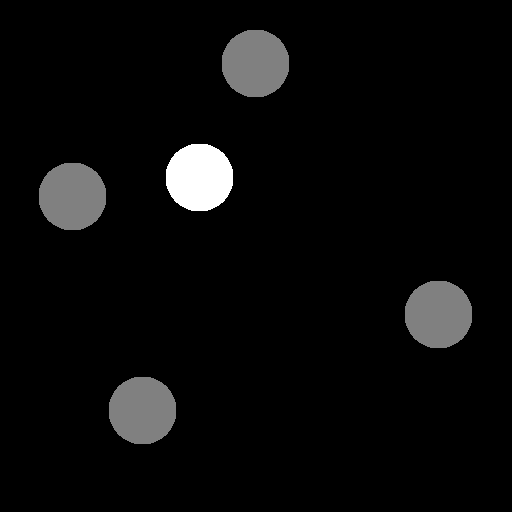} \quad & \includegraphics[width=0.11\textwidth,cfbox=col4 1pt 0pt]{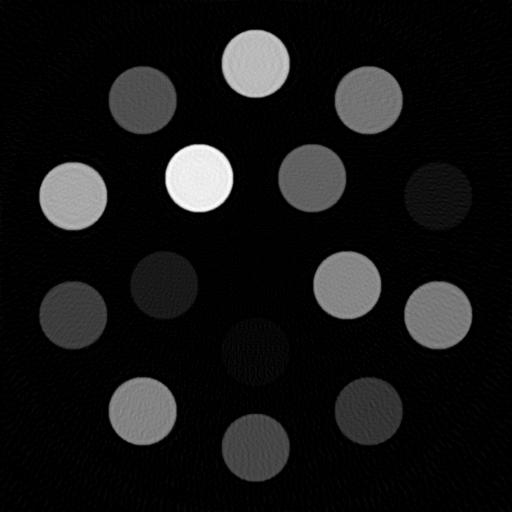} & \includegraphics[width=0.11\textwidth,cfbox=col4 1pt 0pt]{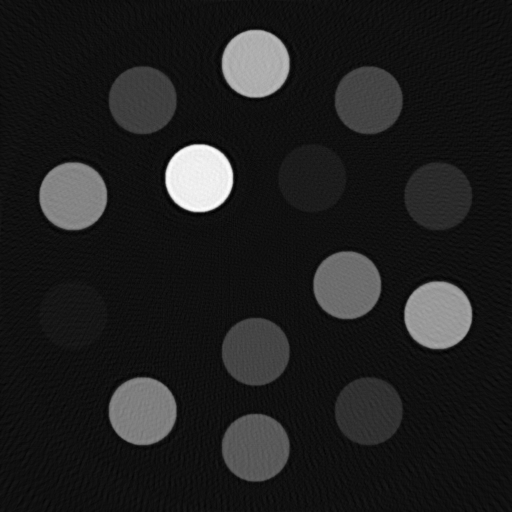} & \includegraphics[width=0.11\textwidth,cfbox=col4 1pt 0pt]{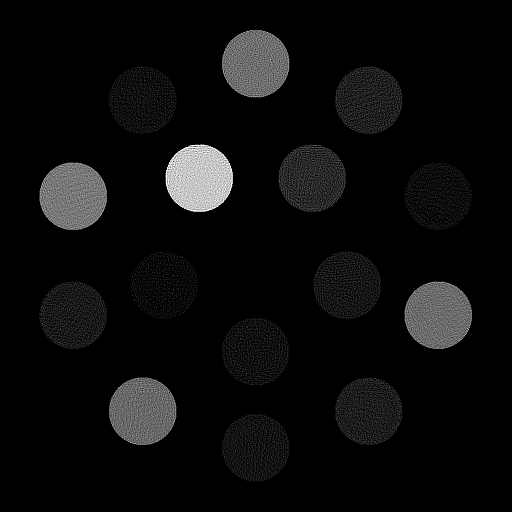} & \includegraphics[width=0.11\textwidth,cfbox=col4 1pt 0pt]{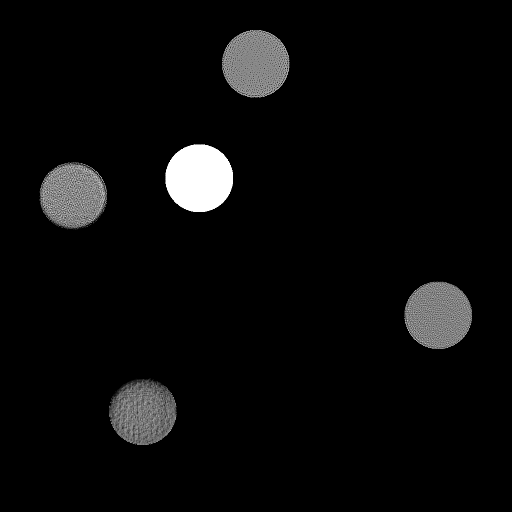} \\
        \includegraphics[width=0.11\textwidth,cfbox=col5 1pt 0pt]{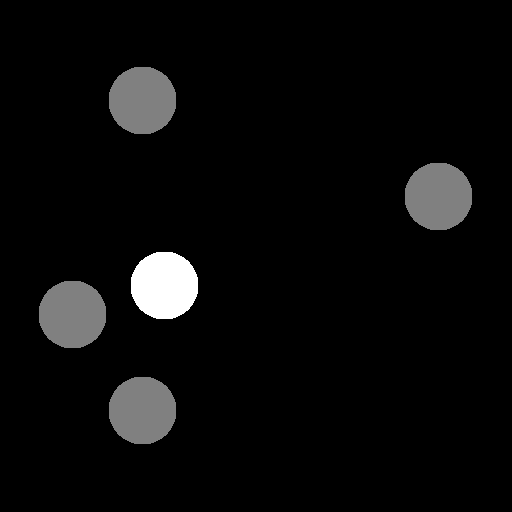} \quad & \includegraphics[width=0.11\textwidth,cfbox=col5 1pt 0pt]{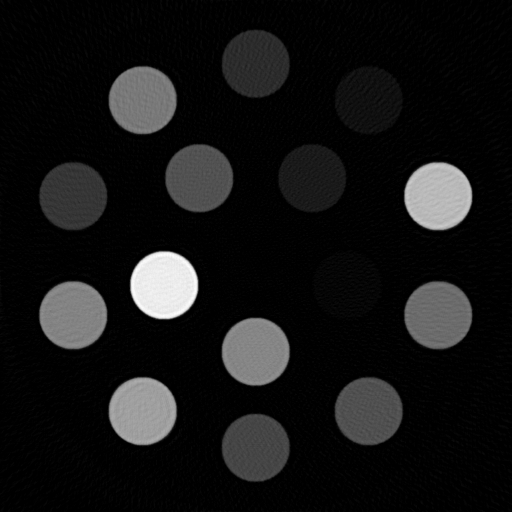} & \includegraphics[width=0.11\textwidth,cfbox=col5 1pt 0pt]{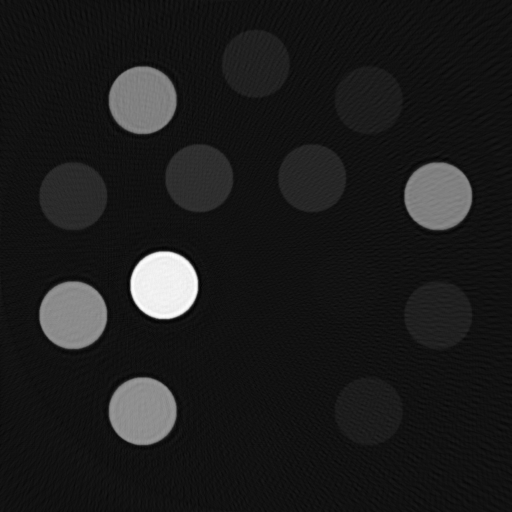} & \includegraphics[width=0.11\textwidth,cfbox=col5 1pt 0pt]{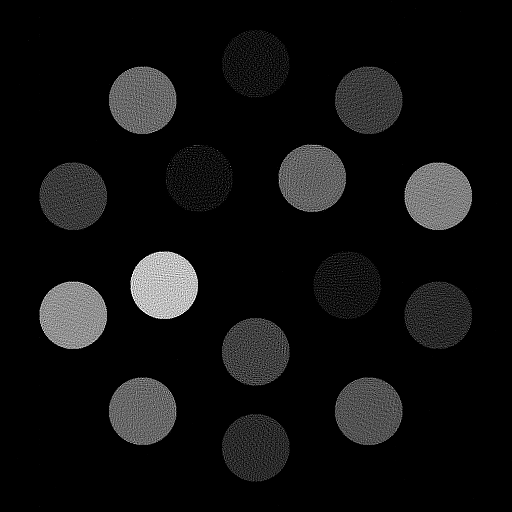} & \includegraphics[width=0.11\textwidth,cfbox=col5 1pt 0pt]{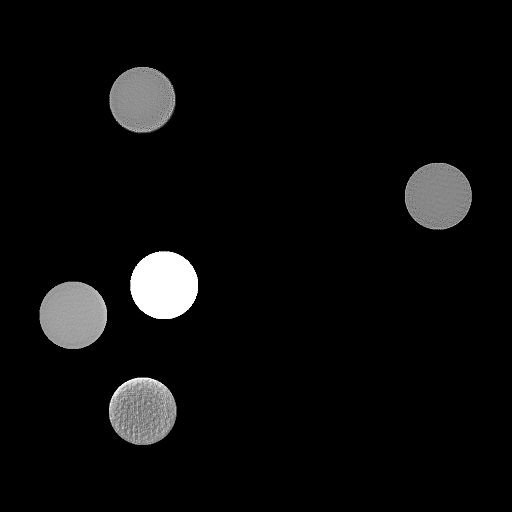} \\
        \includegraphics[width=0.11\textwidth]{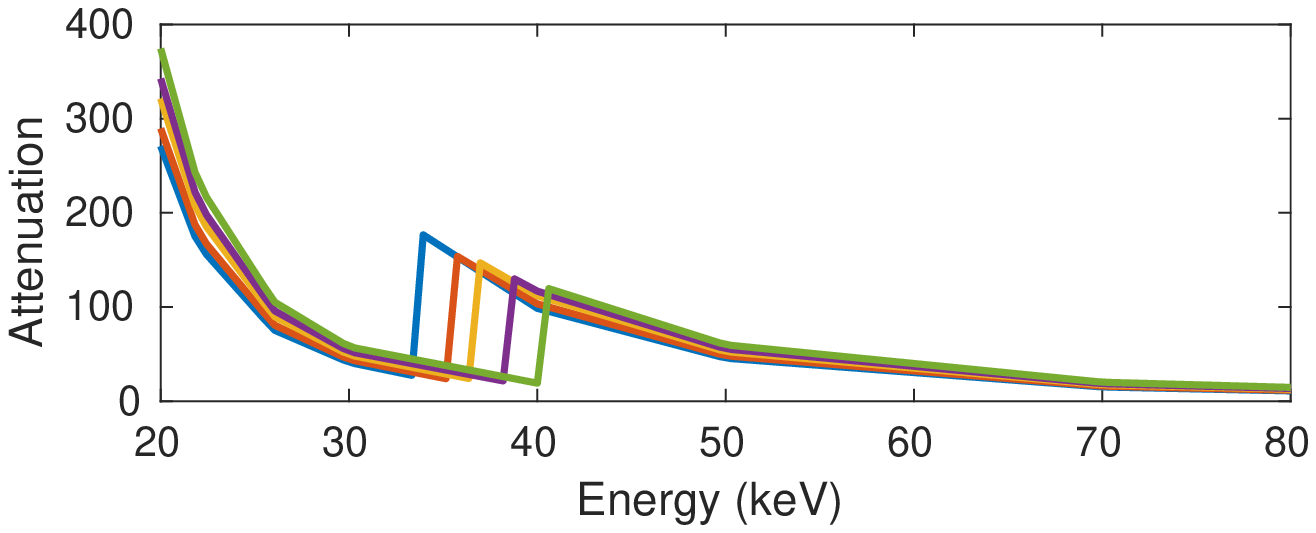} \quad & 
        \includegraphics[width=0.11\textwidth]{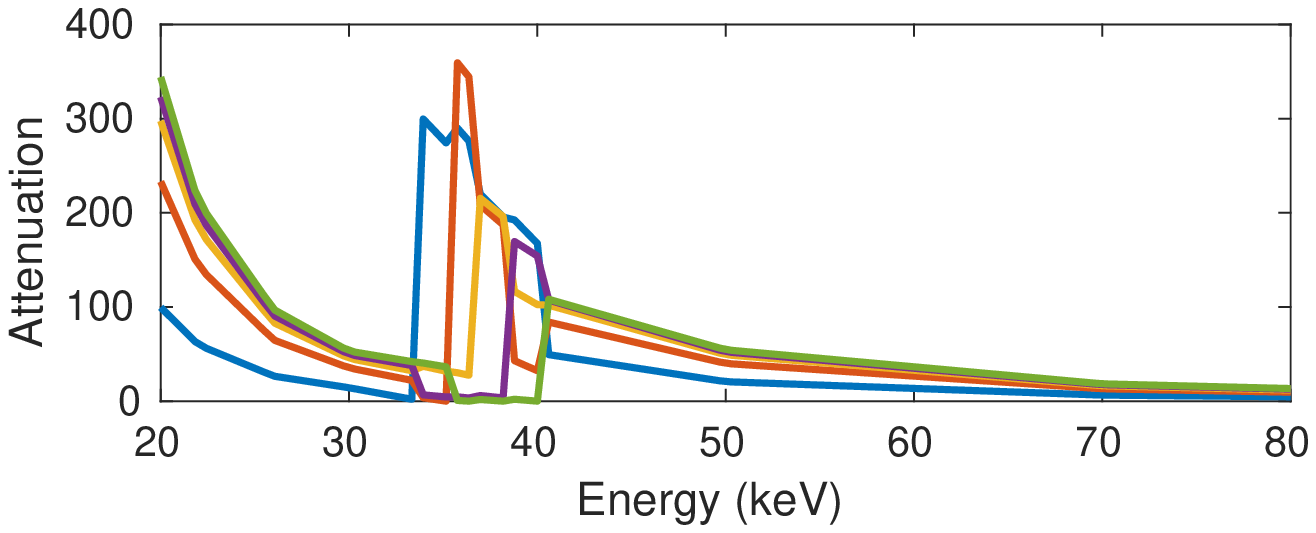} &
        \includegraphics[width=0.11\textwidth]{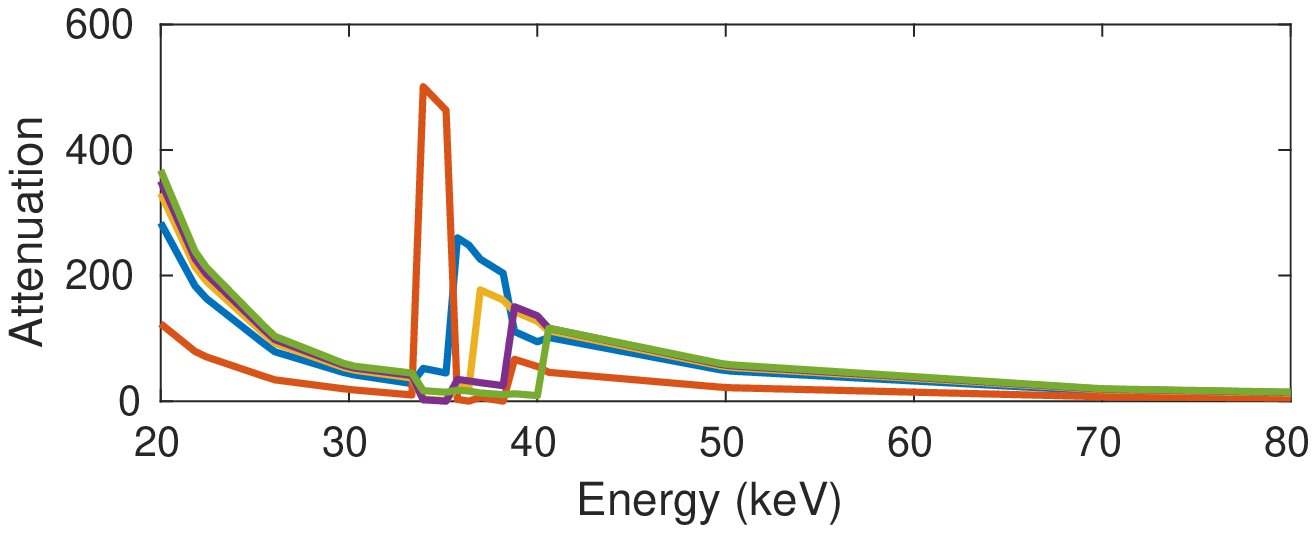} &
        \includegraphics[width=0.11\textwidth]{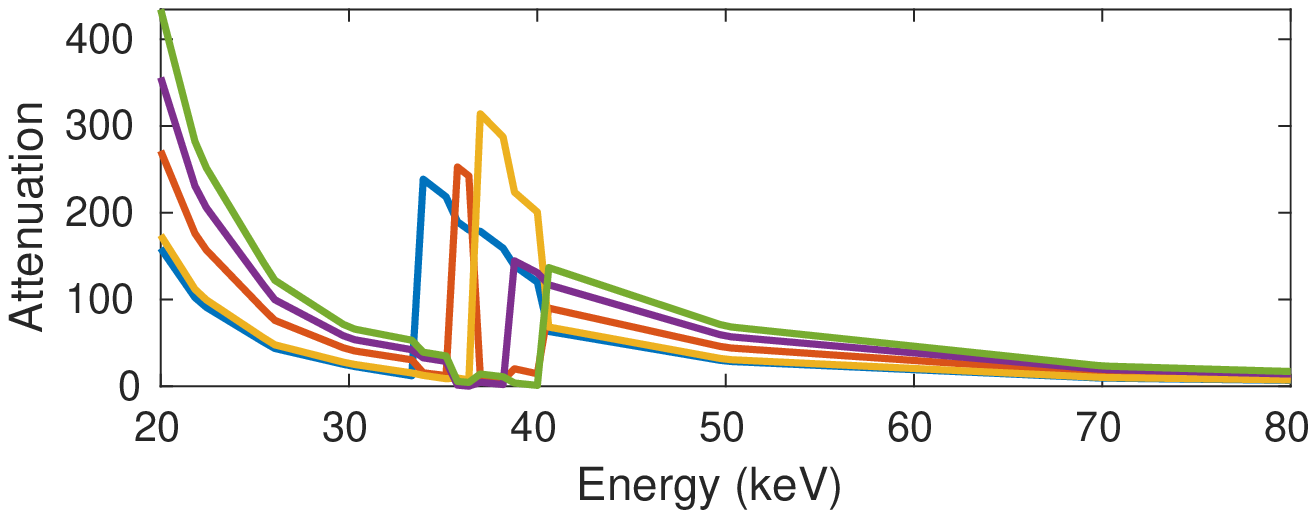} &
        \includegraphics[width=0.11\textwidth]{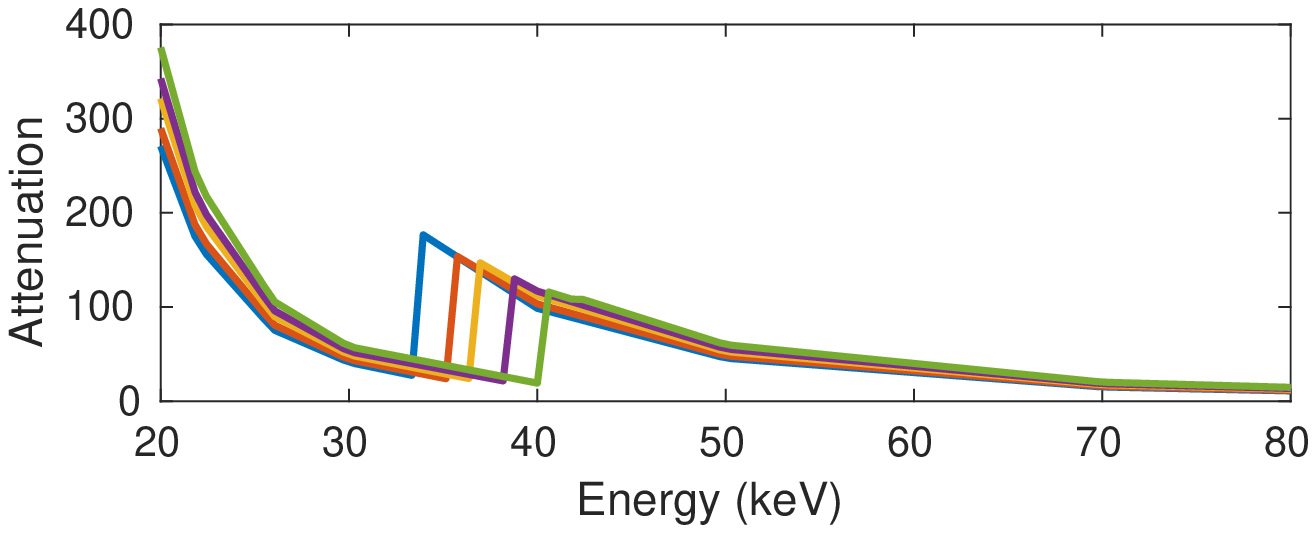}\\ 
    \end{tabular}
    \caption{Comparison of various methods for spectral CT for a mixed-material Disks phantom. The materials contained in this phantom are arsenic (top row), selenium, bromine, krypton and rubidium (bottom row).}
    \label{fig:Exp:Mixed:comparison}
\end{figure}

We consider the Mixed Disks phantom, which consists of solid disks in an inner circle and mixed disks on an outer circle. All material mixtures are present on the outer circle. With $M=5$ disks on the inner circle, this amounts to $10$ mixed disks on the outer circle. The materials are the same as the first $5$ selected materials in the Disks phantom. The ADJUST method with $2000$ iterations is compared with RU, UR, and cJoint. The results of this experiment are shown in Figure~\ref{fig:Exp:Mixed:comparison}, with the results for each material on a separate row. We see that the RU, UR, and cJoint methods are not capable of fully separating the mixtures, and retrieving the disks on the inner circle. On the other hand, ADJUST nearly perfectly reconstructs the disks on the inner circle and the mixture disks on the outer circle.

\section{Numerical Studies: 3D Phantom}
\label{sec:NumStud3D}

\begin{figure}[H]
    \centering
    \begin{tabular}{c c c c c}
        3D half-section & XY slice & YZ slice & XZ slice & Spectral profile\\
        \includegraphics[width=0.15\textwidth]{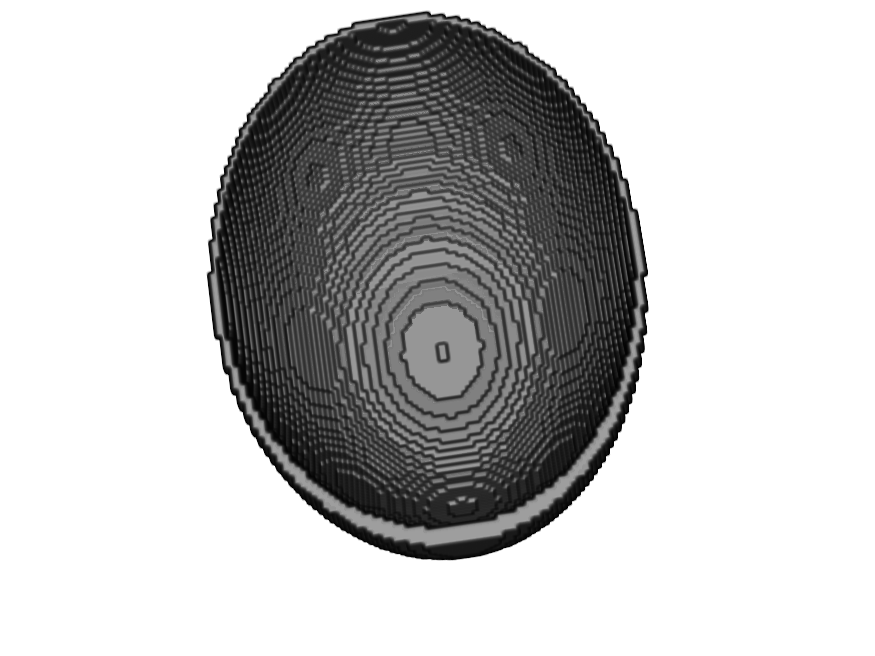} & 
        \includegraphics[width=0.15\textwidth]{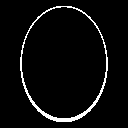} & \includegraphics[width=0.15\textwidth]{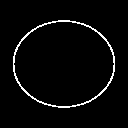} & 
        \includegraphics[width=0.15\textwidth]{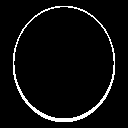} & \includegraphics[width=0.2\textwidth]{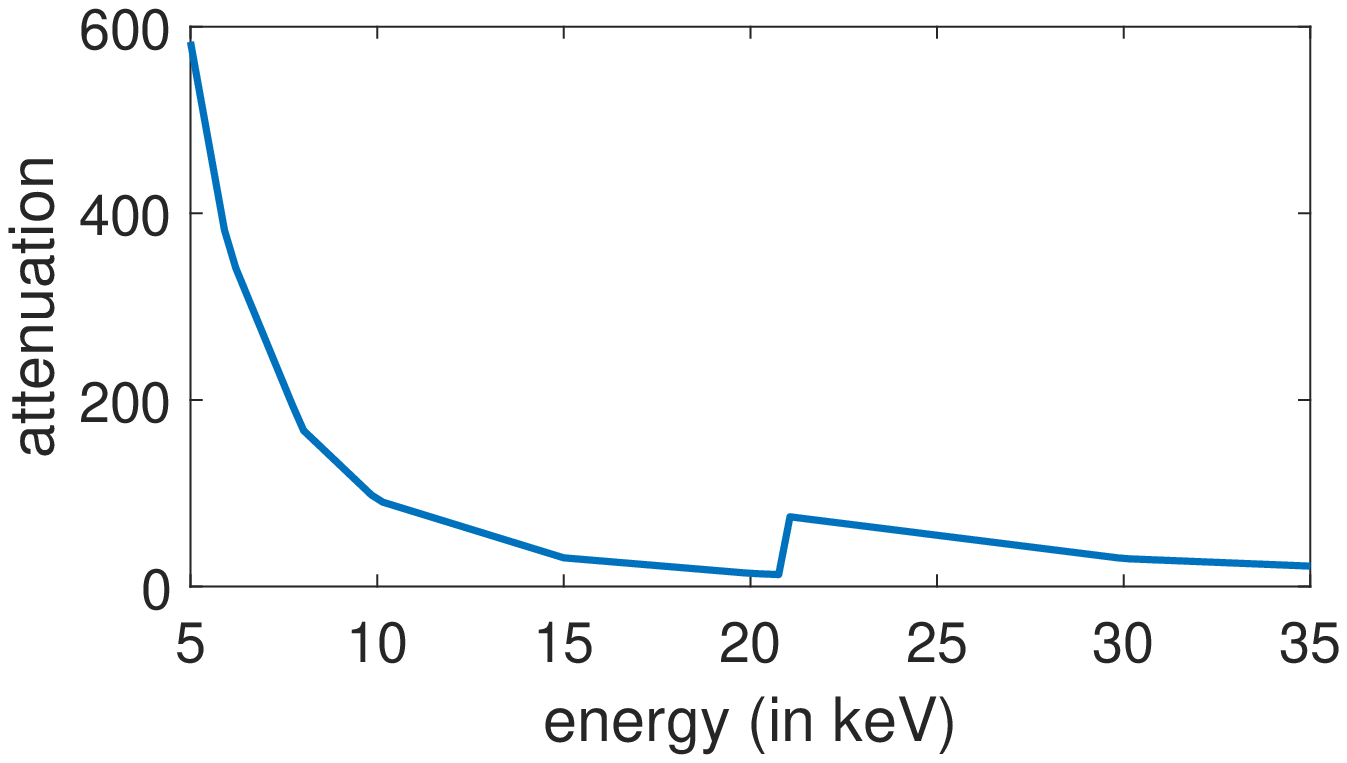}  \\
        \includegraphics[width=0.15\textwidth]{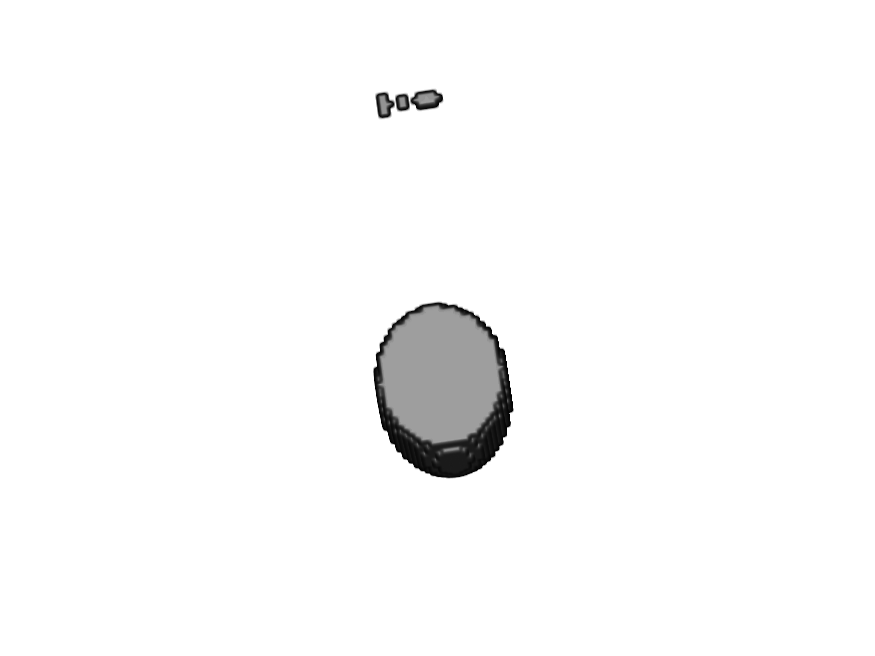} & 
        \includegraphics[width=0.15\textwidth]{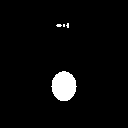} & \includegraphics[width=0.15\textwidth]{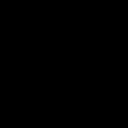} & 
        \includegraphics[width=0.15\textwidth]{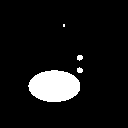} & \includegraphics[width=0.2\textwidth]{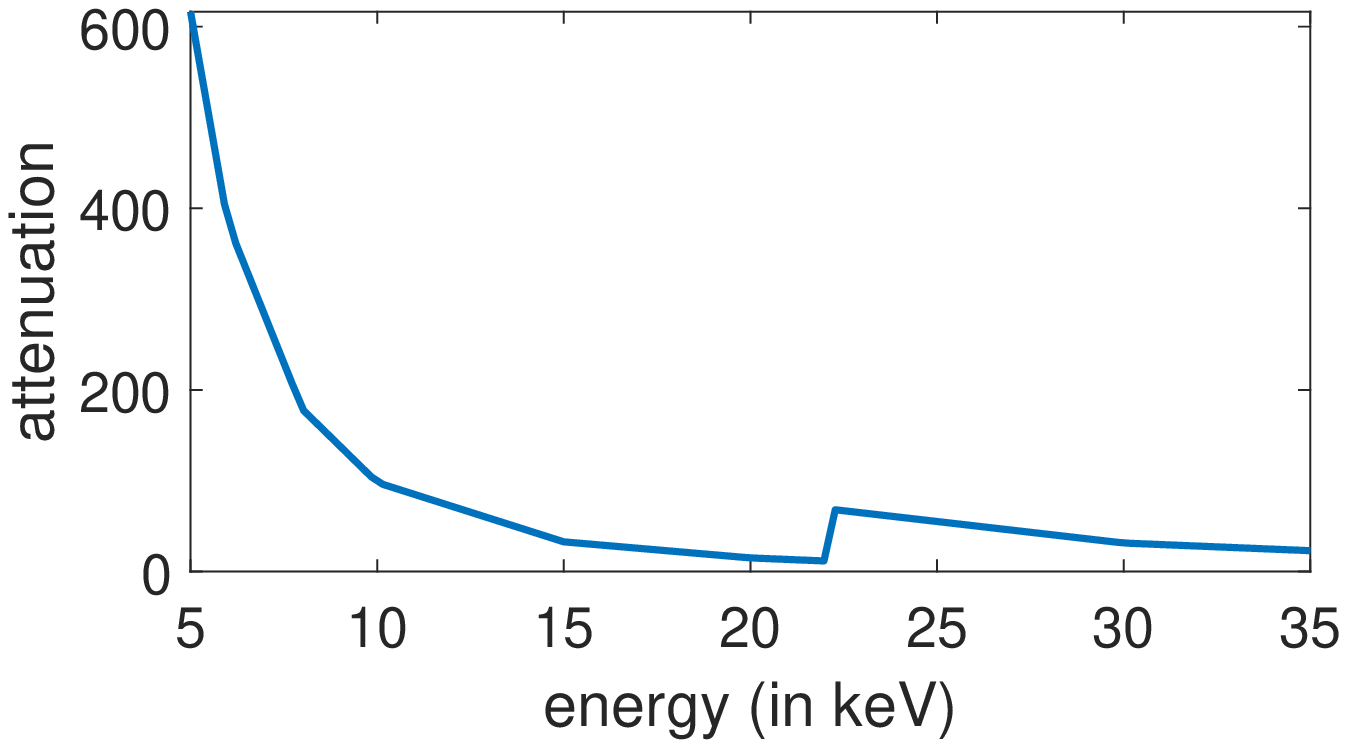} \\
        \includegraphics[width=0.15\textwidth]{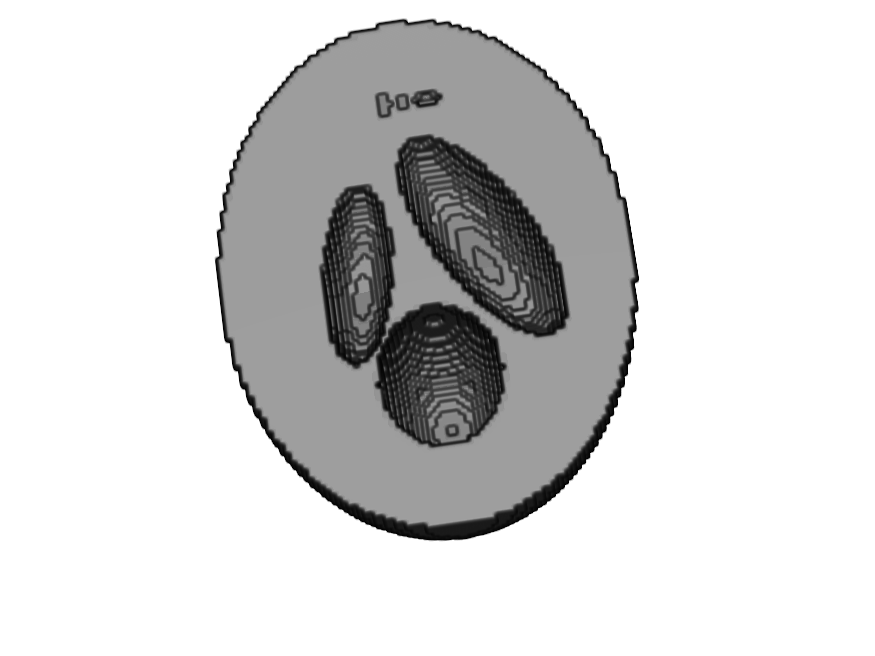} & 
        \includegraphics[width=0.15\textwidth]{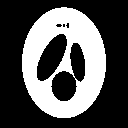} & \includegraphics[width=0.15\textwidth]{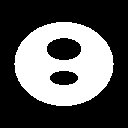} & 
        \includegraphics[width=0.15\textwidth]{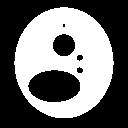} & \includegraphics[width=0.2\textwidth]{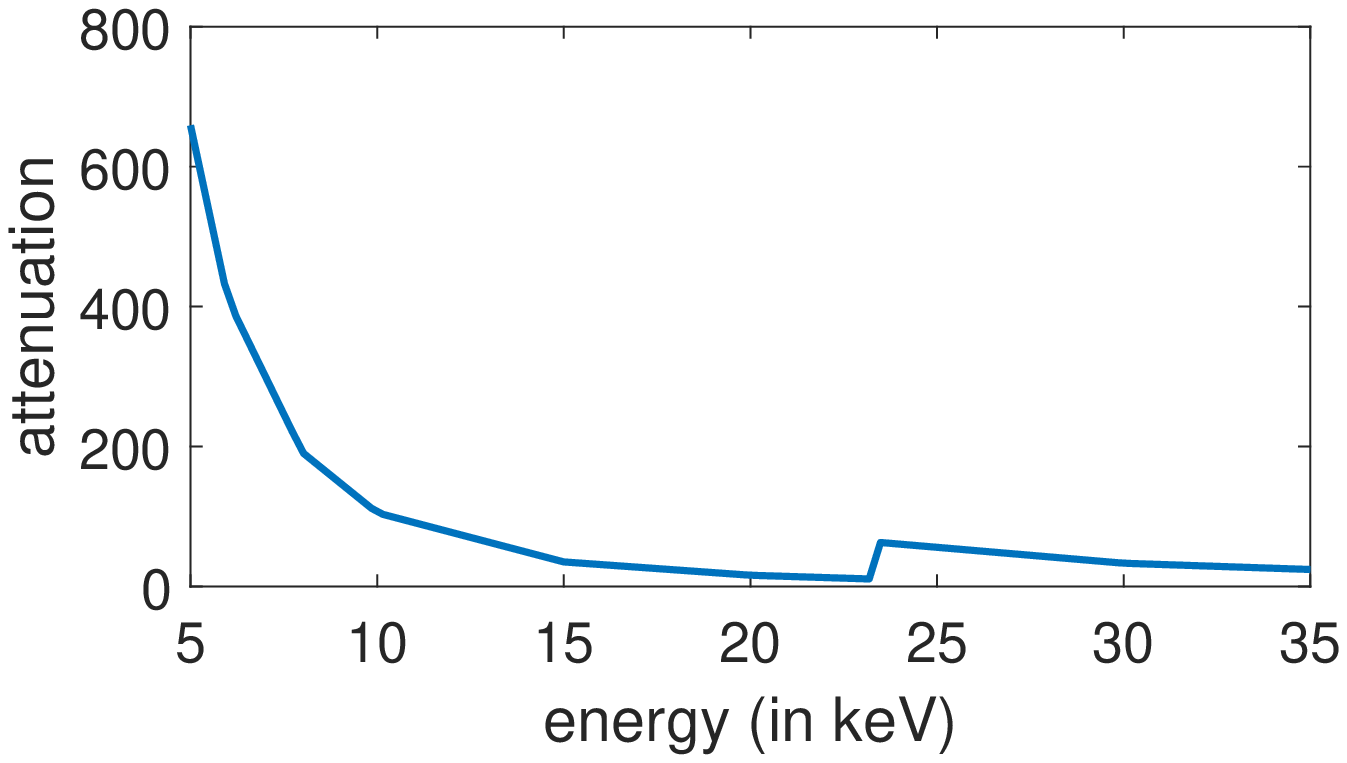} \\
    \end{tabular}
    \caption{The true material compositions of the 3D Shepp-Logan phantom. The materials contained in this phantom are vanadium (top row), chromium (middle row), and manganese (bottom row).}
    \label{fig:Exp:3D:GT}
\end{figure}

\begin{figure}[H]
    \centering
    \begin{tabular}{c c c c c}
        3D half-section & XY slice & YZ slice & XZ slice & Spectral profile \\
        \includegraphics[width=0.15\textwidth]{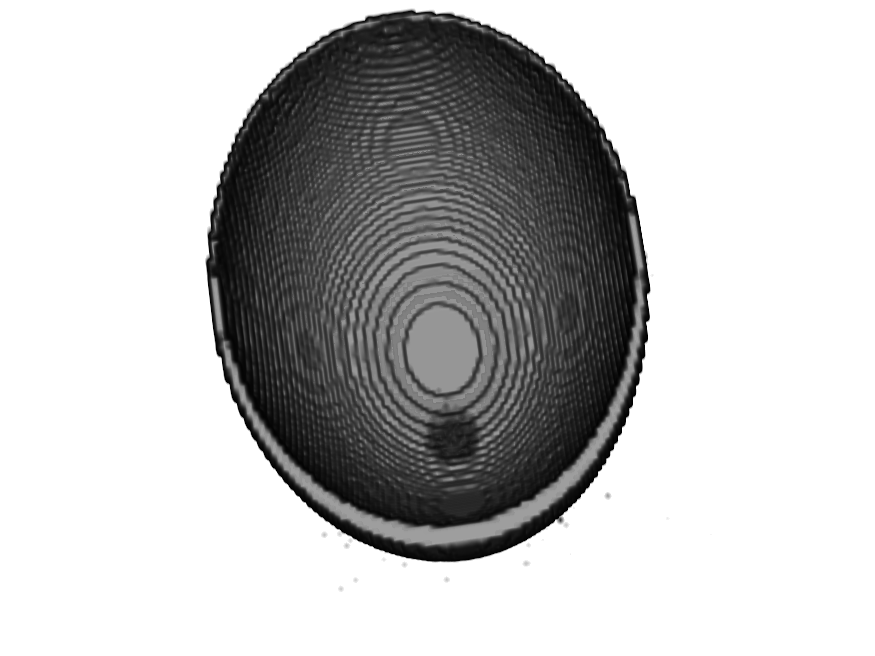} & 
        \includegraphics[width=0.15\textwidth]{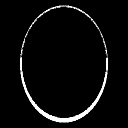} & \includegraphics[width=0.15\textwidth]{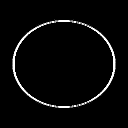} & 
        \includegraphics[width=0.15\textwidth]{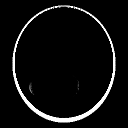} &
        \includegraphics[width=0.2\textwidth]{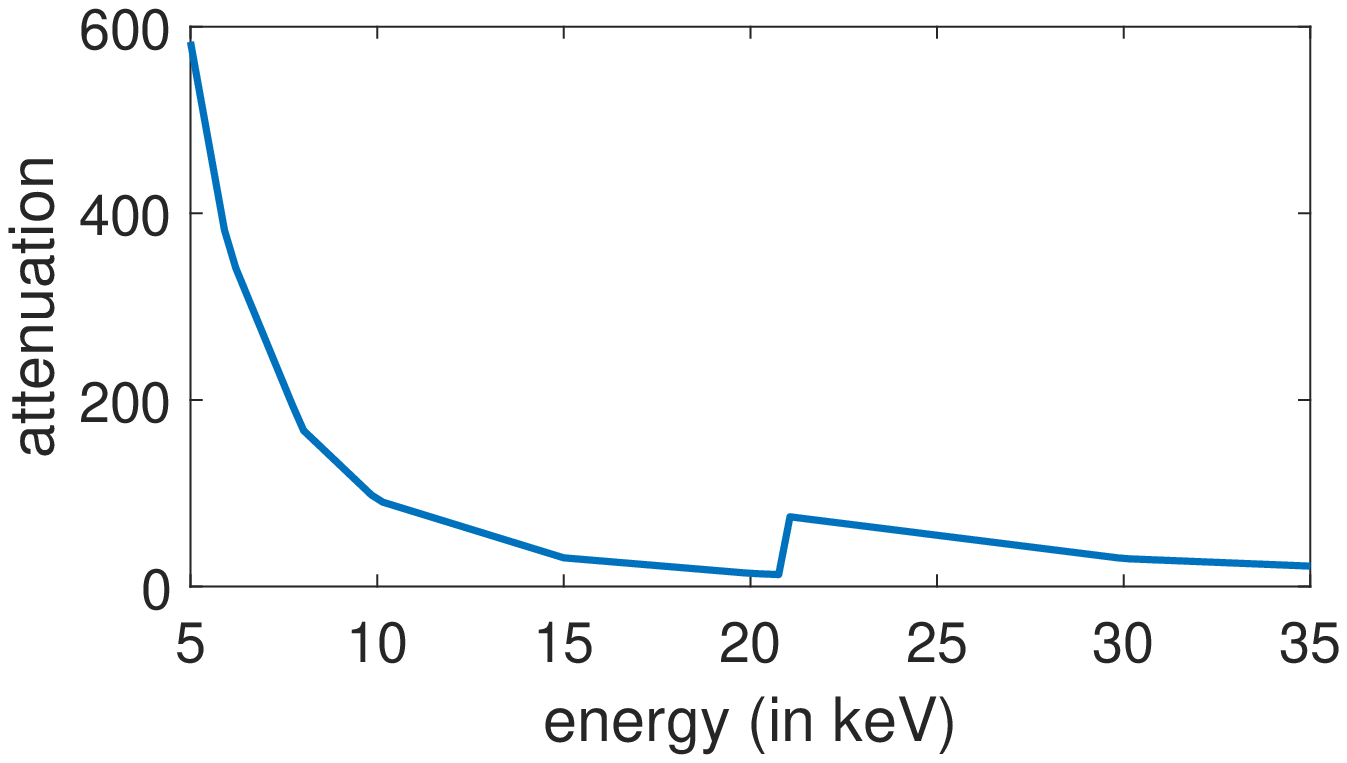} \\
        \includegraphics[width=0.15\textwidth]{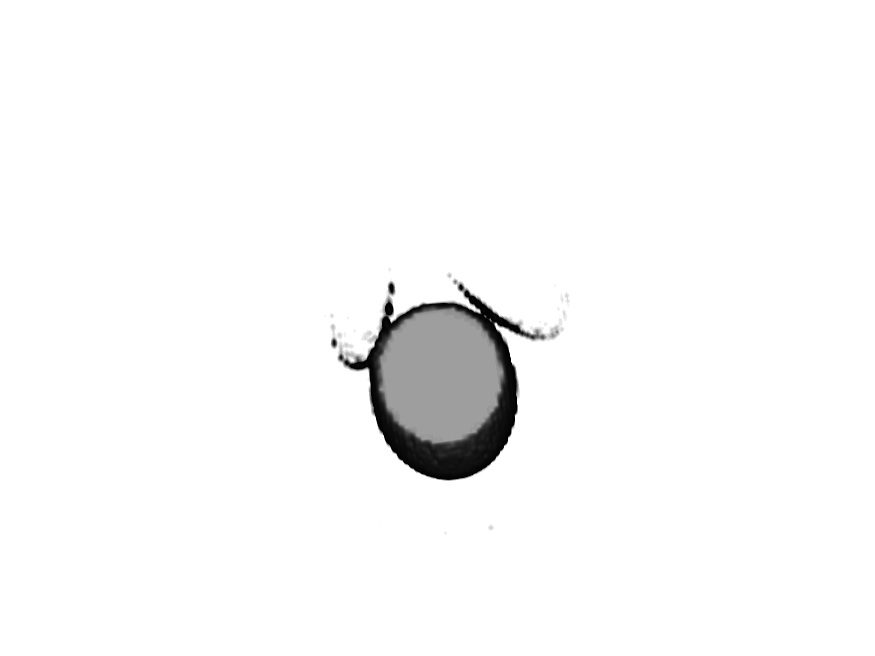} & 
        \includegraphics[width=0.15\textwidth]{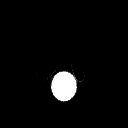} & \includegraphics[width=0.15\textwidth]{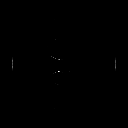} & 
        \includegraphics[width=0.15\textwidth]{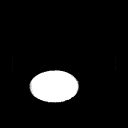} & \includegraphics[width=0.2\textwidth]{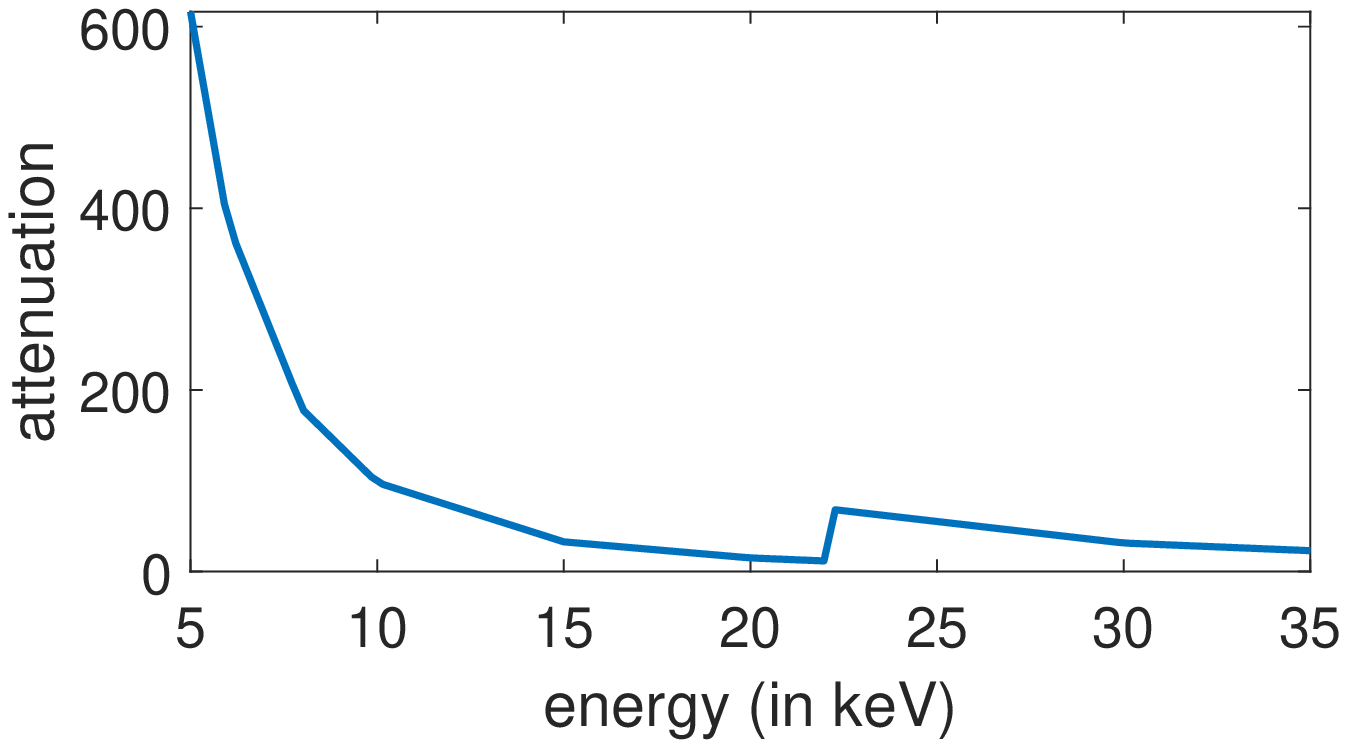} \\
        \includegraphics[width=0.15\textwidth]{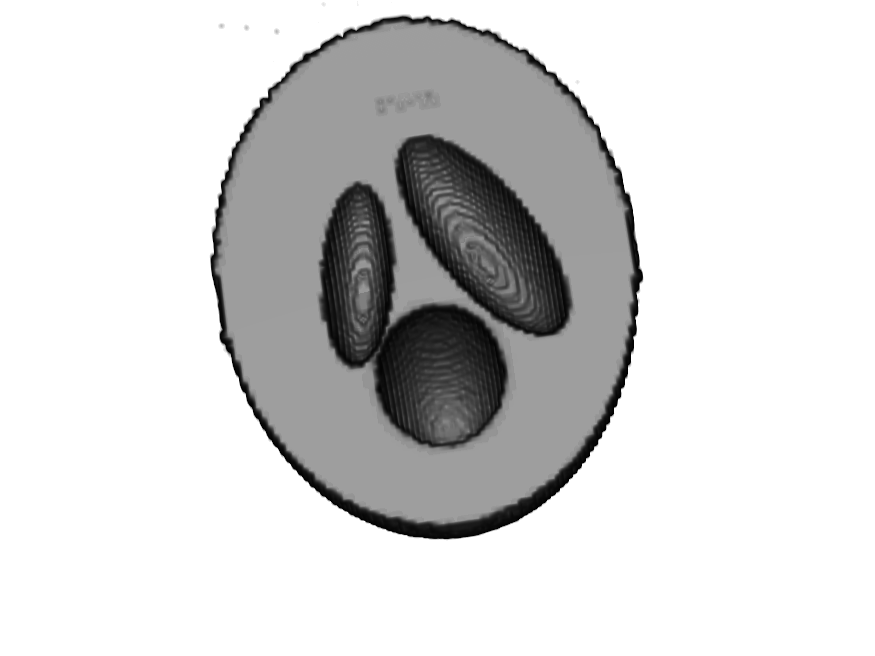} & 
        \includegraphics[width=0.15\textwidth]{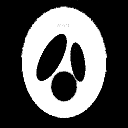} & \includegraphics[width=0.15\textwidth]{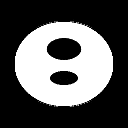} & 
        \includegraphics[width=0.15\textwidth]{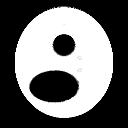} & \includegraphics[width=0.2\textwidth]{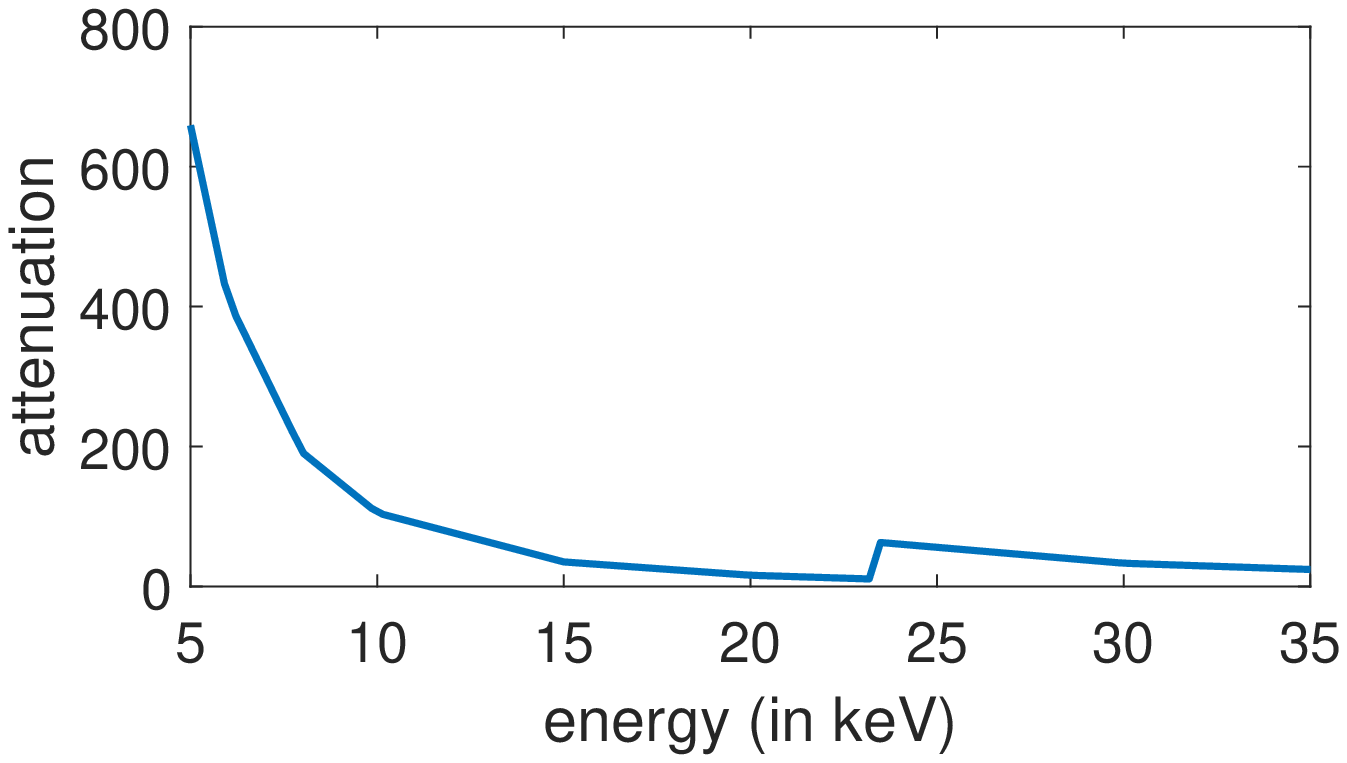}   \\
    \end{tabular}
    \caption{The reconstructed material composition of the 3D Shepp-Logan phantom from ADJUST algorithm. The materials contained in this phantom are vanadium (top row), chromium (middle row), and manganese (bottom row).}
    \label{fig:Exp:3D:ADJUST}
\end{figure}

We also apply the ADJUST algorithm to the 3D Shepp-Logan phantom to show the ability to reconstruct a 3D phantom. This 3D phantom is four times as large as the 2D Shepp-Logan phantom. The phantom is discretized on a grid of $128 \times 128 \times 128$ voxels. The phantom is shown in Figure~\ref{fig:Exp:3D:GT}. We consider $60$ equidistant projection angles in the range of $[0, \pi]$ with a parallel-beam acquisition geometry. We show the visual results of the 3D material decomposition in Figure~\ref{fig:Exp:3D:ADJUST}. The average MSE is $0.0029$, the average PSNR is $26.67$ and the average SSIM is $0.9763$, indicating that the 3D reconstructions are almost accurate.

\newpage

\section{Visual results on spectral micro-CT dataset} 
\label{sec:VisualResultsReal}

\begin{figure}[H]
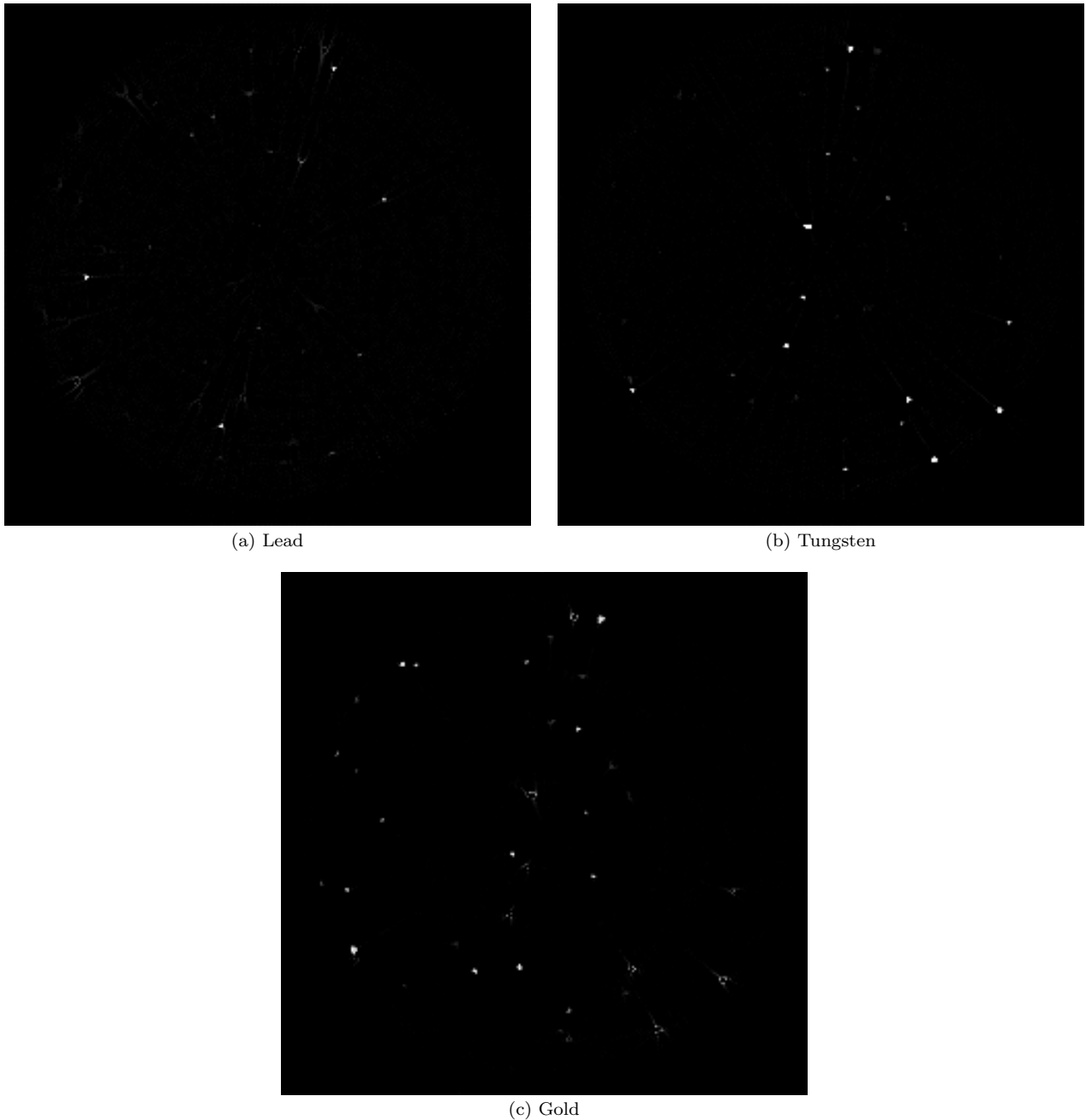

    \centering
    \begin{small}
    \renewcommand{\arraystretch}{1}
    \begin{tabular}{c c c}
        \includegraphics[width=0.45\textwidth,clip]{RD_ADJUST_1.png} & \includegraphics[width=0.45\textwidth,clip]{RD_ADJUST_2.png} \\ 
        (a) Lead & (b) Tungsten \\
        \\
        \multicolumn{2}{c}{\includegraphics[width=0.45\textwidth,clip]{RD_ADJUST_4.png}} \\
        \multicolumn{2}{c}{(c) Gold}
    \end{tabular}
    \end{small}
    \caption{Detailed reconstruction results of ADJUST on microtomography dataset of particle mixtures. Reconstructed material maps for (a) lead, (b) tungsten and (c) gold are shown.}
    \label{fig:Exp:RealData:ResultsADJUSTRaw}
\end{figure}

In this section, we present the reconstructed material maps of ADJUST at high resolution (without true or false positive indicators) on the spectral micro-CT dataset as described in Section~\ref{sec:RealData}. The reconstructions per material are given in Figure~\ref{fig:Exp:RealData:ResultsADJUSTRaw}.

\end{document}